\documentclass[11pt,twoside]{extreport}
\usepackage{xspace}

\usepackage{lipsum}

\input{sty/uhead.sty}
\input{sty/boxit.sty}
\input{sty/icthesis.sty}

\usepackage[compact]{titlesec}
\usepackage{multicol}
\usepackage{acronym}
\usepackage{extsizes} 
\usepackage{layout}
\usepackage{listings}
\usepackage[pdftex]{graphicx}
\usepackage{verbatim}
\usepackage{booktabs}
\usepackage{tabularx}
\usepackage{moresize}
\usepackage{subcaption}
\usepackage[T1]{fontenc}
\usepackage{amssymb}
\usepackage{nameref}
\usepackage{textcomp}
\usepackage{makecell}
\usepackage{dcolumn}
\usepackage{hyphenat}
\usepackage{amsmath}
\usepackage{float}
\usepackage{wrapfig}

\usepackage{booktabs}
\usepackage{multirow}
\usepackage{adjustbox}
\usepackage{colortbl}

\usepackage{tikz}
\usetikzlibrary{patterns}
\usetikzlibrary{decorations.pathreplacing}

\usepackage{mdframed}
\newtheorem{definitionx}{Definition}
\newenvironment{definition}
  {\begin{mdframed}\begin{definitionx}}
  {\end{definitionx}\end{mdframed}}

\usepackage[ngerman,russian,english]{babel}
\usepackage[utf8]{inputenc}
\usepackage{csquotes}

\usepackage[style=numeric,sortcites=true,maxcitenames=1,maxbibnames=100,defernumbers=true]{biblatex}
\AtEveryBibitem{%
  \clearfield{issn}
  \clearfield{doi}
  \clearfield{url}
  \clearfield{biburl}
  \clearfield{timestamp}
  \clearname{editor}
}

\usepackage{todonotes}
\presetkeys{todonotes}{inline,backgroundcolor=yellow,caption={}}{}

\usepackage[shortlabels]{enumitem}
\setlist[enumerate]{nosep}
\setlist[itemize]{nosep}

\newcommand{\normallinespacing}{\renewcommand{\baselinestretch}{1.1} \normalsize}
\newcommand{\toclinespacing}{\renewcommand{\baselinestretch}{0.6} \normalsize}
\newcommand{\acronymslinespacing}{\renewcommand{\baselinestretch}{1.1} \normalsize}
\newcommand{\newdoublepage}{\pagestyle{plain}\cleardoublepage\pagestyle{uheadings}}

\usepackage{minitoc}

\usepackage[a-2a]{pdfx} 
\pdfcatalog{/MarkInfo <</Marked true>>}  

\hypersetup{
  pdftitle    = {A Layered Architecture for Log Analysis in Complex IT Systems},
  pdfsubject  = {A Layered Architecture for Log Analysis in Complex IT Systems},
  pdfauthor   = {Thorsten Wittkopp},
  pdfkeywords = {Dissertation, Log Analysis, Anomaly Detection, Root Cause Analysis, AIOps},
  pdfcreator  = {pdfx},
  pdfproducer = {LaTeX with hyperref},
  pdflang     = {en},
  bookmarksopenlevel = {section},
  colorlinks = {true},
  linkcolor = {black}, 
  urlcolor  = {black}, 
  citecolor = {black} 
}

\usepackage[a4paper]{geometry}

\makeatletter
\renewcommand*\l@section{\@dottedtocline{1}{0.55cm}{0.9cm}}
\renewcommand*\l@subsection{\@dottedtocline{2}{1.45cm}{1.2cm}}
\renewcommand*\l@subsubsection{\@dottedtocline{3}{2.65cm}{1.5m}}
\renewcommand*\l@paragraph{\@dottedtocline{4}{4.15cm}{1.8cm}}
\renewcommand*\l@subparagraph{\@dottedtocline{5}{5.95cm}{2.1cm}}
\makeatother

\newcommand\YAMLcolonstyle{\scriptsize\ttfamily\color{black}\bfseries}
\newcommand\YAMLkeystyle{\scriptsize\ttfamily\color{black}\bfseries}
\newcommand\YAMLvaluestyle{\scriptsize\ttfamily\color{black}\mdseries}

\makeatletter

\newcommand\language@yaml{yaml}

\expandafter\expandafter\expandafter\lstdefinelanguage
\expandafter{\language@yaml}
{
  keepspaces=true,
  keywords={true,false,null,y,n},
  keywordstyle=\color{darkgray}\bfseries,
  basicstyle=\YAMLkeystyle,                                 
  sensitive=false,
  comment=[l]{\#},
  morecomment=[s]{/*}{*/},
  commentstyle=\color{purple}\ttfamily,
  stringstyle=\YAMLvaluestyle\ttfamily,
  moredelim=[l][\color{orange}]{\&},
  moredelim=[l][\color{magenta}]{*},
  moredelim=**[il][\YAMLcolonstyle{:}\YAMLvaluestyle]{:},   
  morestring=[b]',
  morestring=[b]",
  literate =    {---}{{\ProcessThreeDashes}}3
                {>}{{\textcolor{red}\textgreater}}1     
                {|}{{\textcolor{red}\textbar}}1 
                {\ -\ }{{\mdseries\ -\ }}3,
}

\lst@AddToHook{EveryLine}{\ifx\lst@language\language@yaml\YAMLkeystyle\fi}
\makeatother

\newcommand\ProcessThreeDashes{\llap{\color{cyan}\mdseries-{-}-}}

\newfloat{mylisting}{thp}{lop}
\floatname{mylisting}{Listing}
\newfloat{submylisting}{thp}{lop}
\floatname{submylisting}{Listing}

\graphicspath{ {figures/} }
\DeclareGraphicsExtensions{.pdf,.jpeg,.jpg,.png,.svg}

\def\maketitle{
    \setlength{\parindent}{0cm}
    \newpage
    \linespread{1}
    \begin{center}
        \vspace*{1cm}
        \huge
        {\bf A Layered Architecture for Log Analysis in Complex IT Systems\par }
        
        \normalsize
        \vspace{2cm}\par
        vorgelegt von\\
        Thorsten Wittkopp, M.Sc.\\
        \vspace{1cm}\par
        an der Fakultät IV – Elektrotechnik und Informatik\\
        der Technischen Universität Berlin\\
        zur Erlangung des akademischen Grades\\
        \vspace{8mm}\par
        Doktor der Ingenieurwissenschaften\\
        - Dr.-Ing. -

        \vspace{8mm}\par
        genehmigte Dissertation

    \end{center}

    Promotionsausschuss:\\
    \\
    \begin{tabular}{@{}ll}
    Vorsitzender: & Prof. Dr. Matthias Böhm \\
    Gutachter: & Prof. Dr. Odej Kao \\
    Gutachter: & Prof. Dr. Roberto Natella \\
    Gutachter: & Prof. Dr. David Bermbach
                
    \end{tabular}
    \vspace{1cm}\par
    
    Tag der wissenschaftlichen Aussprache: 21.~Oktober 2024
    \vspace{1cm}\par
    
    {\centering Berlin 2024\par }
    \cleardoublepage
}

\begin{document}

\pagenumbering{roman}
\pagestyle{empty}
\normallinespacing
\maketitle
\setlength{\parindent}{0pt}
\pagestyle{plain}
\normallinespacing

\begin{acknowledgements}
First, I would like to express my deep gratitude to my advisor, Prof. Odej Kao, for welcoming me into his research group and for granting me trust and freedom. 
His support, encouragement, and patience have been instrumental in my journey, allowing me to explore my own research path. 
I am also very grateful to Prof. David Bermbach and Prof. Roberto Natella for their willingness to review this thesis.

Being part of the Distributed and Operating Systems research group has been a privilege, enabling me to meet and collaborate with a brilliant array of researchers. 
Our numerous discussions and shared meals have not only enriched my professional experience, but also developed friendships. 
To all of you, thank you for your encouragement, humor, and insights that have been a constant source of strength, making my days productive and enjoyable.
I am deeply grateful to Dominik Scheinert, who has not only proofread parts of this thesis but has also been a great friend outside of our academic endeavors.
Special thanks go to Alexander Acker for encouraging and inspiring me to pursue a PhD in the first place.
I extend my deep thanks to Jana Bechstein for her tremendous organizational and administrative support throughout this journey.

This work could not have been accomplished without the constant support of my family and friends, who always believed in me.
I am deeply grateful to my parents for their unwavering support and encouragement throughout my PhD journey.

Thank you.

\end{acknowledgements}

\cleardoublepage
\begin{abstract}

In the rapidly evolving landscape of \ac{IT}, the stability and reliability of IT systems and services are important because they underpin numerous aspects of modern life.
However, their increasing complexity poses significant challenges for \ac{DevOps} teams, who are responsible for their implementation and maintenance. 
Log analysis, a core component of Artificial Intelligence for IT Operations (AIOps), plays an essential role by serving as a major source for investigating the complex behaviors and failures of IT systems.
Therefore, this dissertation addresses the critical need for effective log analysis in complex IT systems by introducing a three-layered architecture designed to enhance the capabilities of DevOps teams in failure resolution.

The first layer, \ac{LI}, focuses on autonomous labeling and anomaly classification to provide the groundwork for the next layers. 
We developed a method that accurately labels log data autonomously, facilitating supervised model training and the precise evaluation of anomaly detection methods.
In addition, we created a taxonomy to classify anomalies into three different categories, guaranteeing the selection of a suitable anomaly detection method.

Within the second layer, \ac{AD}, we identify behaviors of IT systems that deviate from the norm.
Therefore, we propose a flexible \ac{AD} method adaptable to various training scenarios: Unsupervised, weakly supervised, or supervised. 
Evaluations on public and industry data sets demonstrate that our method achieves F1-Scores ranging from 0.98 to 1.0 in different training scenarios, ensuring a reliable anomaly detection.

The third layer addresses \ac{RCA}.
With our developed \ac{RCA} method we can identify a minimal set of log lines that describe a failure along with its origin and the sequence of events that led to it.
By balancing training data and identifying the primary services involved, our \ac{RCA} method consistently identifies 90-98\% of root cause log lines within the top 10 candidates, providing precise and actionable insights for failure mitigation.

Our research answers the overarching question of how log analysis methods can be designed and optimized to help DevOps teams resolve failures efficiently. 
By integrating these three layers, our architecture equips DevOps teams with the necessary methods to enhance IT system reliability. 

\end{abstract}

\cleardoublepage
\begin{otherlanguage}{ngerman}
\begin{abstract}

In der sich schnell entwickelnden Landschaft der Informationstechnologie (IT) sind die Stabilität und Zuverlässigkeit von IT-Systemen und -Diensten von großer Bedeutung. Diese Systeme unterstützen zahlreiche Aspekte des modernen Lebens, aber ihre zunehmende Komplexität stellt DevOps-Teams, die für ihre Wartung verantwortlich sind, vor große Herausforderungen. 
Die Log-Analyse, eine Kernkomponente der Artificial Intelligence for IT-Operations (AIOps), spielt eine wesentliche Rolle, da sie als eine der Hauptquellen für die Untersuchung von IT-Systemen dient und als Grundlage für eine Fehleruntersuchung benutzt werden kann.
Diese Dissertation befasst sich daher mit der Log-Analyse in komplexen IT-Systemen, indem sie eine dreischichtige Architektur vorstellt, die DevOps-Teams bei der Fehleranalyse und Fehlerbehebung unterstützt.

Die erste Ebene, Log Investigation, konzentriert sich auf das automatisierte Labeln von Datensätzen und die Klassifikation von Anomalien. Dafür haben wir einerseits eine Methode entwickelt, die Anomalien selbstständig labelt und anderseits eine Taxonomie zur Klassifikation von Anomalien erstellt. 
Somit gewährleistet diese Ebene eine Auswertung von Anomalieerkennungsmethoden durch die Bereitstellung nahezu perfekt gelabelter Datensätze sowie eine zielgerichtete Auswahl von Anomalieerkennungsmethoden durch die Bereitstellung von verschiedenen Anomalienarten in den Log-Daten.

Auf der zweiten Ebene, Anomaly Detection, beschreiben wir eine allgemeine Anomalieerkennungsmethode, die sich an verschiedene Trainingsbedingungen anpassen lässt: unbeaufsichtigt, schwach überwacht oder überwacht. Dabei zeigen unsere Auswertungen auf öffentlichen und industriellen Datensätzen, dass unsere Methode in verschiedenen Trainingsszenarien F1-Werte von 0,98 bis 1,0 erreichen kann und somit eine zuverlässige Anomalieerkennung gewährleistet.

Die dritte Ebene befasst sich mit der Ursachenanalyse, wobei irrelevante Anomalien herausgefiltert werden.  
Dafür erstellen wir automatisch ausbalancierte Trainingsdaten, um unsere Root Cause Analyse Methode zu trainieren. Im Anschluss analysieren wir, welche Services an dem Fehler beteiligt sind und präsentieren die entsprechenden anomalen Log-Zeilen dem DevOps Team.
Dabei befinden sich 90-98\% der präsentierten Log-Zeilen innerhalb der Top-10-Kandidaten und liefern präzise Erkenntnisse zur Fehlerbehebung.

Im Ergebnis beantwortet diese Forschungsarbeit die übergreifende Frage, wie eine Log-Analyse so gestaltet und optimiert werden kann, dass sie DevOps-Teams genügend Details liefert, damit diese Fehler im System beheben können. 
Durch die Integration dieser drei Ebenen: Log Investigation, Anomaly Detection und Root Cause Analyse, stattet unsere Architektur DevOps-Teams mit den notwendigen Werkzeugen aus, um die Zuverlässigkeit und Leistung von IT-Systemen zu verbessern. 

\end{abstract}
\end{otherlanguage}
\newdoublepage
\toclinespacing
\dominitoc
\tableofcontents
\newdoublepage
\listoftables 
\newdoublepage
\listoffigures 
\newdoublepage

\chapter*{Abbreviations}
\acronymslinespacing

\begin{multicols}{2}
\begin{acronym}[MCDM]
    \acro{AI}{Artificial Intelligence}
    \acro{AIOps}{IT Operations with Artificial Intelligence}
    \acro{AD}{Anomaly Detection}
    \acro{BERT}{Bidirectional Encoder Representations from Transformers}
    \acro{DL}{Deep Learning}
    \acro{DoS}{Denial of Service}
    \acro{DevOps}{Development and Operation}
    \acro{FFN}{Feedforward Neural Network}
    \acro{FTW}{Failure Time Window}
    \acro{IT}{Information Technology}
    \acro{ITW}{Investigation Time Windows}
    \acro{LI}{Log Investigation}
    \acro{LSTM}{Long Short Term Memory}
    \acro{ML}{Machine Learning}
    \acro{NLP}{Natural Language Processing}
    \acro{NN}{Neural Network}
    \acro{PCA}{Principal Component Analysis}
    \acro{RCA}{Root Cause Analysis}
    \acro{RCC}{Root Cause Candidate}
    \acro{SVM}{Support Vector Machine}
    \acro{TCN}{Temporal Convolutional Networks}
    \acro{TEA}{Transformer Encoder Architecture}   
\end{acronym}
\end{multicols}

\newpage

\cleardoublepage
\pagenumbering{arabic}
\normallinespacing

\cleardoublepage

\chapter{Introduction}
\minitoc
\label{ch:intro}

In the rapidly evolving landscape of \ac{IT}, the stability and reliability of IT systems and services is critical~\cite{zio2009reliability,avizienis2004basic,gray1986computers}. 
IT systems and services are integral to almost every aspect of modern daily life, from communication and commerce to transportation and healthcare~\cite{atzori2010internet,akter2022transforming,del2023big}. 
We rely on these systems for seamless connectivity, efficient operations, and instant access to information across diverse platforms and devices. 
Whether managing personal finances through online banking, navigating using GPS applications, or accessing entertainment via streaming services, IT systems underpin the functionality and convenience we experience daily~\cite{kitsios2021digital,azzaakiyyah2023impact}.
To handle all these responsibilities, IT systems and services are becoming increasingly complex~\cite{zhang2010cloud,santos2017analyzing,newkirk2008rapid}.

DevOps teams play a critical role in creating, deploying, and maintaining these systems, yet they often face unforeseen challenges during runtime and execution~\cite{ko2008debugging}. 
Unexpected failures can disrupt operations, leading to significant downtime and financial loss~\cite{natella2016assessing}.
Consequently, significant research has been dedicated to the operation and maintenance of complex IT systems since the development of the field of computer science~\cite{simon1977organization,senapathi2018devops}. 
Due to rapid innovations and new business opportunities, IT systems and services become more complex.
This leads to an increase in data volumes and complexity, which intensifies the challenges of operating and maintaining these systems and services, limiting the success achieved~\cite{zhang2010cloud,bass2015devops,santos2017analyzing,shen2020evolving}.
Operators of IT services need assistance maintaining control of this complexity to ensure dependability, stability, and serviceability~\cite{hamooni2016logmine}.

With the invention of \ac{AI}, a new field known as \ac{AIOps} emerged, that help DevOps teams address various challenges in this domain~\cite{dang2019aiops,bogatinovski2021artificial,notaro2020systematic,becker2020towards}.
AIOps has many domains, one of which is log analysis~\cite{zhaoxue2021survey}.
To keep track of how the system works, developers incorporate log statements throughout the software development process. 
These logs capture and record the execution of the source code, providing a detailed chronological record of the system’s behavior and usage patterns~\cite{gulenko2016system,miranskyy2016operational}. 
Through logging, developers can track various operational metrics, including system usage, user interactions, and error occurrences.
These logs serve as a primary source of information that reflects how the system is executed in real life, providing invaluable insights into its performance and potential issues.

The logs generated by IT systems are voluminous and complex, potentially containing billions of lines~\cite{oliner2012advances}. 
These logs capture a wide range of events, from routine operations to unexpected errors and attacks~\cite{yen2013beehive}, offering a comprehensive view of how the system behaves under various conditions. 
The sheer volume of log data generated can be overwhelming.
Logs record every detail of the system's execution, including user interactions, system processes, and error messages. 
Manually analyzing these logs is impractical due to their size and the intricate patterns they contain~\cite{xu2009detecting}.

The task of manual log analysis is not only time-consuming, but also resource-inten\-sive~\cite{beschastnikh2011mining}. 
Companies must invest in highly skilled experts capable of interpreting complex and voluminous data. 
This leads to significant financial costs, as the hiring and retention of skilled professionals is expensive~\cite{shengqi2010mining}.
These experts, often referred to as system administrators, data scientists, or DevOps engineers, are required to sift through immense data sets to identify anomalies and potential issues. However, even with their expertise, the sheer scale and complexity of the logs make it virtually impossible to perform a thorough analysis manually~\cite{jiang2011efficient}. 
Their time could be better utilized on more strategic tasks that drive innovation and business growth.
Inefficiencies in manual log analysis can lead to prolonged system downtimes, delayed responses to critical issues, and an overall increase in operational costs~\cite{fu2009execution}. 
For many organizations, this inefficiency in analyzing failures results in substantial financial losses, in terms of delayed problem resolution~\cite{kreps2011kafka}.

Failures can be subtle and multifaceted, often buried within layers of seemingly unrelated log.
Consequently, substantial research is focused on log \acf{AD}~\cite{guo2021logbert, du2017deeplog, nedelkoski2020self, zawawy2010log, wittkopp2021loglab, korzeniowski2022landscape, lu2017log, wittkopp2020a2log, wittkopp2022pull, li2020swisslog, meng2019loganomaly}. 
However, some of these approaches require a significant amount of labeled data for training or parameter tuning, or they are tailored to specific IT systems or training strategies. 
Thus, a research gap we address in this thesis involves the necessary groundwork and preparatory steps for effective anomaly detection models and their evaluation. 
Furthermore, we focus on generalizing anomaly detection models by allowing them to handle different training data.

In addition, there is a second field in the log analysis, namely \acf{RCA}. 
However, there are relatively few studies that focus on \ac{RCA} of log data~\cite{notaro2023logrule,lu2017log,lu2019ladra,zawawy2010log}, highlighting a significant research gap in this area.
\ac{RCA} for logs involves identifying and analyzing the context and the relationship of specific events that triggered the failure. 
These events are documented in the logs~\cite{fu2012logmaster}, allowing the team to understand the origin of the failure and ultimately resolve it.

Although individual methods are very useful for anomaly detection and root cause analysis, they are often underutilized due to several inherent limitations. 
Developing and deploying these methods is complex and resource intensive, requiring substantial expertise and customization for each specific IT system~\cite{sipos2014log,he2016experience}. 
They tend to be system-specific, limiting their applicability to different platforms and environments~\cite{zhaoxue2021survey}. 
In addition, focusing on a single aspect of log analysis can lead to fragmented insights, making it difficult for companies to gain a comprehensive understanding of their IT operations. To address these issues, companies need holistic and standardized approaches that integrate multiple facets of log analysis into a cohesive framework. Such an integrated approach can provide more actionable insights and improve the overall efficiency and effectiveness of IT system and service management.

The need for automated, accurate, and efficient log analysis methods that work hand in hand is more critical than ever. 
What is required are not just standalone methods for the log analysis layers \ac{AD} and \ac{RCA}, but tools, processes, and methods to effectively develop and evaluate these methods. 
However, a crucial preliminary layer that precedes \ac{AD} and \ac{RCA}, is underrepresented in the current research and therefore introduced in this thesis as \ac{LI}.
With this holistic architecture, DevOps teams can then proficiently analyze the system's performance, improving the overall reliability and serviceability, and ensure greater stability in future updates.


\section{Problem Definition}
\label{sec:intro:problem_definition}
The entire log analysis process in AIOps is not just a single method, it should include several methods that together support the DevOps team in the best possible way.
Therefore, the research question is formulated as follows:


\begin{center}
    \textit{\textbf{"How can log analysis methods be designed and optimized to help DevOps teams resolve failures efficiently?"}}
\end{center}

We have identified three primary layers in log analysis for complex IT systems and services. 
Using these three layers, DevOps teams should gain sufficient insight into their systems to address issues in future updates. 
The first layer is \acf{LI}, which aims to gather enough information about the logs to determine and evaluate the most suitable anomaly detection method for the system. 
The second layer is \acf{AD}, where it is important to choose an anomaly detection method that can be trained with existing training data.
Finally, in the third layer \acf{RCA}, the anomalies need to be filtered to highlight those that most accurately represent the source and action of a failure. 
Consequently, we have developed a detailed research question for each layer.

\begin{itemize}
    \item \textit{\textbf{RQ 1.1:} How can an anomaly taxonomy and labeling support DevOps teams in the deployment and evaluation of appropriate anomaly detection methods?}\\
    
    \item \textit{\textbf{RQ 1.2:} How should an anomaly detection method be designed to identify anomalies in any system without depending on specific training data?}\\
    
    \item \textit{\textbf{RQ 1.3:} Is it possible to identify a set of log lines that describe the root cause leading to a failure, so that DevOps teams can remediate the failure?}
\end{itemize}

\section{Contributions}
\label{sec:intro:contributions}

To address these challenges, this thesis introduces a three-layered log analysis architecture. 
This layered log analysis architecture provides an integrated approach to log management, addressing the entire log data analysis lifecycle. It is designed to be adaptable and applicable across various IT systems and services, including the analysis of log data gathered from field tests, enabling DevOps teams to efficiently handle the complexities of modern IT environments.
The modular nature of the architecture allows each layer to be executed independently, providing flexibility and scalability. 
By automating the log analysis process, our methods reduce the time and effort required for manual analysis, offering substantial improvements in efficiency and cost-effectiveness. 
The layered architecture provides DevOps teams with methods to identify failures and understand their root causes to implement corrective updates.
The primary contributions for each of our layers are described in the following.

\begin{itemize}
    \item \textbf{Log Investigation} 
    The first layer focuses on log investigation, which involves classification of anomalies and autonomous labeling of logs. This layer makes the following contributions: This layer provides a systematic approach to classifying different types of anomalies in log data. By categorizing anomalies, DevOps teams can select the most suitable detection methods, improving the accuracy of anomaly detection. Furthermore, an autonomous labeling method is introduced, facilitating the training and evaluation of anomaly detection methods. This reduces the dependency on manually labeled data, which is often scarce and expensive to obtain.
    \\

\item \textbf{Anomaly Detection} 
    The second layer is dedicated to anomaly detection, aiming to identify states in the system that deviate from normal behavior. The contribution include: The development of an adaptable anomaly detection method that can be trained unsupervised, weakly supervised, or supervised. The method is designed to be robust and provides reliable detection of anomalies in different IT systems and services under different conditions.
    \\

\item \textbf{Root Cause Analysis} 
    The third layer is root cause analysis, which is the filtering of detected anomalies to pinpoint those that directly indicate the underlying causes of failures. The contribution in this layer includes a method to filter anomalies effectively, identifying the most relevant to system failures. This helps DevOps teams focus on critical issues, reducing the time needed to troubleshoot and resolve problems.
\end{itemize}

The work in this thesis builds upon the following peer-reviewed publications:

\begin{refsection}
\nocite{*}
\newrefcontext[sorting=ydnt]
\printbibliography[keyword={contribution}, heading=none, resetnumbers=true]
\end{refsection}

Furthermore, parts of this thesis builds on the following patent: \textit{LOG ANOMALY DETECTION MODEL TRAINING METHOD, APPARATUS AND DEVICE}, with the identification number: \textit{WO2022227388}.

\section{Thesis Outline}
The rest of this thesis is structured as follows:

\textbf{Chapter 2} provides an overview of AIOps, its components, and its importance in IT operations. 
It discusses the nature of log data and log data processing and feature extraction techniques. 
Furthermore, it covers various learning strategies for machine learning models in log analysis and provide concepts that are used in our methods.

\textbf{Chapter 3} reviews the existing literature related to AIOps, anomaly detection, and root cause analysis. 
It explores various methodologies and technologies that have been proposed and implemented in these areas, providing a comprehensive understanding of the current state of research. 

\textbf{Chapter 4} presents the layered architecture for log analysis. 
It describes how the different layers and methods interact with each other to form an architecture. 
The chapter outlines the primary objectives of each layer, setting the foundation for the detailed descriptions of each method that follow in subsequent chapters.

\textbf{Chapter 5} focuses on the first layer, \acl{LI}. 
It introduces a method for classifying anomalies, providing the basis to select an appropriate anomaly detection method. 
The chapter also presents an autonomous log labeling method, which is crucial for training and evaluating anomaly detection models.

\textbf{Chapter 6} details the second layer, which involves the design and implementation of an anomaly detection method. 
It describes the mathematical foundations of the method and how it can be trained with different types of training data. The chapter also covers the inference process for differently trained models.

\textbf{Chapter 7} addresses the third layer, \acl{RCA}. 
It explains our method developed for filtering anomalies to identify a set of root cause log lines to understand system failures to enhance the troubleshooting capabilities of DevOps teams.

\textbf{Chapter 8} provides a comprehensive evaluation of each method proposed in the previous chapters. 
It presents experimental results and performance metrics that demonstrate the effectiveness and efficiency of our architecture. 
The chapter also discusses how the results from one layer of the architecture can be beneficial to the next layer.
In addition, we answer our three subresearch questions in this chapter.

\textbf{Chapter 9} concludes the thesis by summarizing the key findings and contributions of the research and answers our main research question. 
The chapter also suggests potential directions for future research and improvements, aiming to further advance the field of AIOps and log analysis.

\cleardoublepage
\chapter{Background}
\minitoc
\label{ch:background}

This chapter provides background information for this thesis, laying the groundwork for log analysis and different training strategies. 
Hence, providing the information needed to understand our methods in the three different layers of our log analysis architecture.

Initially, we present information about AIOps in~\autoref{sec:background:aiops}. 
Next, we discuss log data and log data processing techniques in~\autoref{sec:background:log_data}. 
We then cover common machine learning strategies along with their advantages and disadvantages in~\autoref{sec:background:learning_strategies}.
Finally, we introduce general concepts in~\autoref{sec:background:concepts} that are applied in several of our methods.

\section{AIOps: IT Operations Through Artificial Intelligence}
\label{sec:background:aiops}

In the ever-evolving landscape of modern \ac{IT} environments, where the complexity and scale of large computer systems reach unprecedented heights, the challenges of effective development, maintenance, and operation are difficult~\cite{notaro2020systematic,rosendo2018improve}. 
Traditional approaches to \ac{IT} operations are often strained to cope with the dynamic and intricate nature of contemporary IT infrastructure~\cite{notaro2021survey}. 
In response to this pressing need, Artificial Intelligence for IT Operations emerges as a new paradigm poised to transform the way we perceive, manage, and optimize IT systems and services~\cite{gulenko2020ai,notaro2020systematic,andenmatten2019aiops}. 

As organizations grapple with the intricacies of large-scale computer systems, the demand for automation and intelligent analysis becomes increasingly apparent~\cite{bermbach2021future}. Traditional IT operations, often constrained by manual intervention and reactive approaches, do not address the dynamic nature of cloud or IoT environments. \ac{AIOps}, at its core, is a response to this need for proactive, data-driven, and intelligent IT operations management. Hence, it consists of a convergence of processes for artificial intelligence, automation, and IT operations. 
AIOps is fundamentally driven by the vast amount of data continuously generated by IT environments.  
Thereby, AIOps systems rely on the three pillars of observability: metrics, traces, and logs~\cite{sridharan2018distributed}.
Logs capture detailed records of events and transactions generated by log statements embedded in the source code, metrics provide quantitative measures of system performance, and traces offer insights into the flow and dependencies of processes across distributed systems. To effectively analyze and derive actionable insights from these voluminous and heterogeneous data, it is stored and managed in big data systems. These systems are designed to handle the scale, complexity, and diversity of monitoring data, enabling AIOps platforms to perform advanced analytics and machine learning. 
Using advanced analytics and machine learning, AIOps platforms can detect anomalies and root causes of failures in the IT system to then propose or implement countermeasures.

\begin{figure}[h]
\centering
\includegraphics[width=0.8\columnwidth]{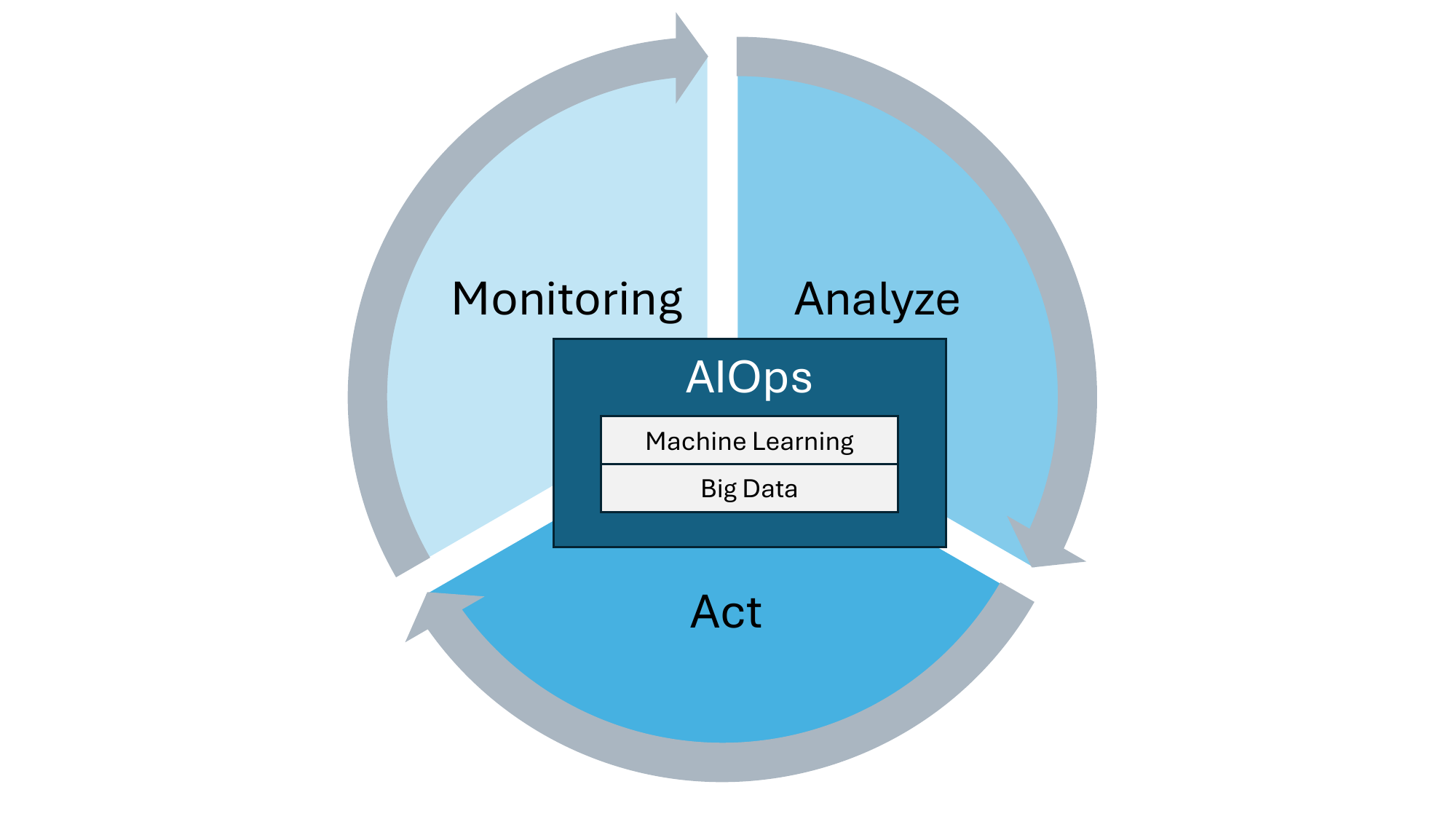}
\caption{The AIOps lifecycle inspired by Gartner}
\label{fig:background:AIOps}
\end{figure}

\autoref{fig:background:AIOps} inspired by Gartner\footnote{AIOps Platforms: Gartner: https://blogs.gartner.com/andrew-lerner/2017/08/09/aiops-platforms/. Revisited January 2024.} illustrates the core of Artificial Intelligence for IT Operations.
Hence, they present AIOps as a circular flow, symbolizing its life cycle.

The initial stage of the \ac{AIOps} cycle is dedicated to monitoring. 
This involves the systematic collection and analysis of large amounts of data generated by IT infrastructure, applications, and services. 
In this phase, various data sources contribute to the creation of a comprehensive understanding of the IT environment. 
Following the monitoring phase, the diagram transitions inward to the analysis process. 
Here, IT professionals and stakeholders interpret the monitoring data to gain valuable insight into performance, health, and potential issues within the IT system or service. 
The Act phase presents informed decision making based on the data analyzed. 
This phase involves the utilization of automation technologies to respond quickly and intelligently to the insights gained from the analyzing stage based on the monitored information and data. 
Automation facilitates the execution of predefined actions or the initiation of dynamic responses to optimize IT operations, mitigate risks, and improve overall system performance. 

Nestled at the core of the \ac{AIOps} cycle are two main concepts: Machine Learning and Big Data. These embody the technologies that enable the effective functioning of AIOps systems.
Big Data technologies handle the vast volumes of data generated by monitoring applications, ensuring scalability, storage, and efficient processing.
Hence, big data technologies cover the vast amount of data and build the foundation for machine learning algorithms that are trained on the monitoring data.
Machine learning algorithms analyze historical and real-time data, discerning patterns, anomalies, and trends to provide predictive insights. 
In summary, the Gartner \ac{AIOps} diagram encapsulates the iterative flow of Monitoring, Engage, and Act.


\subsection{Anomaly Detection}
\label{subsec:background:anomaly_detection}



Anomaly detection in AIOps has its roots in the early days of IT operations, when simple threshold-based alerts were used to identify deviations from expected system behavior. As IT environments grew in complexity, these rudimentary methods proved inadequate. The advent of machine learning and statistical techniques in the late 20th century marked a significant evolution, enabling more sophisticated approaches to anomaly detection~\cite{nishu2022anomaly}. Using historical data and learning normal behavior patterns, these techniques can identify subtle and complex anomalies that traditional methods might overlook. 

In AIOps, anomaly detection is a main component that is used to proactively identify unusual patterns or behaviors within IT systems that can indicate potential issues such as performance degradation, security breaches, or system failures. 
By detecting anomalies in real time, AIOps solutions allow quick intervention, minimize downtime, and maintain optimal performance of IT services~\cite{al2021review,rijal2022aiops}. This capability is essential for managing the dynamic nature of modern IT environments, where early detection and resolution of anomalies can significantly improve operational resilience and efficiency.

The financial implications of IT system downtimes are substantial for companies, often resulting in significant revenue losses, decreased customer trust, and operational inefficiencies~\cite{franke2011optimal}. In today’s fast-paced digital landscape, even a few minutes of downtime can disrupt critical business operations, leading to cascading effects across the organization. This underscores the importance of anomaly detection within AIOps~\cite{diaz2023joint}. By identifying anomalies early, before they escalate into full-blown outages, companies can take proactive measures to mitigate potential disruptions. 
The ability to foresee and prevent downtime not only safeguards revenue streams but also enhances customer satisfaction and trust. 

Anomalies in IT systems can manifest in various forms in monitored metrics, logs, and traces, each indicating potential problems that require attention. In monitored metrics, an anomaly might be an unexpected spike in CPU usage or memory consumption, which could signal a malfunctioning application or a potential cyber attack. For example, a sudden increase in network latency could indicate a network bottleneck or a \ac{DoS} attack. In logs, anomalies might appear as error messages or unusual parameters in normal log lines.  
In traces, anomalies are often found in the form of unusual request patterns or delays in service response times. For example, an unexpected increase in the duration of a transaction trace might indicate a performance bottleneck. 

Given that this thesis concentrates on log analysis for AIOps, we now present a brief example to illustrate what anomalies in log data might look like:

\begin{lstlisting}[basicstyle=\ttfamily\scriptsize, breaklines=true]
2023-06-21 14:30:00 ERROR Unable to connect to database server at 192.168.1.100:5432
2023-06-21 14:31:00 INFO Enable connection to null
\end{lstlisting}

The first log line is marked with a severity level \textit{Error}, indicating that the system cannot connect to the database server, which is clearly an anomaly. The second log line has a severity level of \textit{Info} and states that it enables a connection to null. Typically, this log message includes an IP address, but due to current system conditions, the IP address is null, making this anomaly less obvious.
Different types of log anomalies will be presented in~\autoref{sec:data:classifying_anomalies}.

Consequently, investing in robust anomaly detection mechanisms within AIOps is a strategic move for companies aiming to maintain high availability and reliability of their IT systems and services, thereby ensuring business continuity and competitive advantage.

\subsection{Root Cause Analysis}
\label{subsec:background:rca}


Root cause analysis is a systematic process to identify the underlying causes of problems or incidents in IT systems~\cite{wu2020microrca}. By determining the root cause, IT teams can address fundamental problems rather than treating symptoms, leading to more effective and lasting solutions. RCA involves collecting and analyzing data from the monitoring systems, such as logs, metrics, and traces, to analyze the origin of a failure and understand how it propagated through the system~\cite{zhang2021cloudrca}. This process is crucial in complex IT environments, where multiple interdependent components can obscure the true source of an issue~\cite{larsson2007real}.

The history of root cause analysis dates back to early industrial processes, where it was used to improve manufacturing quality and reliability. In the context of IT, RCA evolved alongside the development of more complex and distributed systems. Initially, RCA relied on manual inspection and correlation of logs and metrics, which was time-consuming and error-prone~\cite{diaz2023joint}. With the advent of automated monitoring tools and more sophisticated data analysis techniques in the late 20th century, RCA became more efficient and accurate. The integration of machine learning and big data analytics into IT operations further revolutionized RCA, enabling the handling of large amounts of monitoring data and the analysis of failure causes.

Root cause analysis supports AIOps by improving the ability to diagnose and resolve failures quickly and accurately. In AIOps, RCA leverages advanced analytics and machine learning to automate the identification of root causes~\cite{zhang2021cloudrca}. This automation significantly reduces the time and effort required for troubleshooting, allowing IT teams to focus on preventing future incidents and optimizing system performance. 
Furthermore, the continuous learning aspect of AIOps ensures that the system improves over time, becoming more adept at identifying root causes of failures.

By applying root cause analysis, DevOps teams can analyze unexpected runtime failures and gain insight into their underlying causes. With this knowledge, they can implement solutions in future updates to prevent similar failures from recurring. This proactive approach not only enhances the reliability and performance of the system, but also provides significant financial benefits for companies. By reducing downtime and minimizing the need for costly and reactive fixes, companies can reduce operational costs and avoid revenue losses associated with service interruptions.

Examples of root cause analysis in AIOps can be seen across different types of monitoring data. In log data, RCA can involve correlating error messages between different services to identify a faulty component that triggered a cascade of failures. For instance, repeated database connection errors in application logs could point to an overloaded or misconfigured database server. In metrics data, RCA might involve analyzing performance degradation patterns to pinpoint a specific resource bottleneck, such as a sudden increase in CPU usage linked to a particular service update. In trace data, RCA could involve mapping the flow of transactions to identify a delay in one microservice that causes a ripple effect throughout the system~\cite{wu2020microrca}. For example, a significant increase in response time for a payment processing microservice might be traced back to a recent change in its configuration or a dependency on a slow external API. 

As we focus on log data, we now provide some log lines that show the root cause of a failure, which are multiple log lines that show the propagation through the system until it emerges to a failure. 

\begin{lstlisting}[basicstyle=\ttfamily\scriptsize, breaklines=true]
2023-06-21 16:00:00 INFO Android device initiated system update
2023-06-21 16:05:00 INFO WiFi module turned off for system update
2023-06-21 16:10:00 INFO WiFi module turned on after system update
2023-06-21 16:12:00 INFO WiFi scanning for available networks
2023-06-21 16:15:00 INFO Connected to primary WiFi network 'HomeWiFi'
2023-06-21 16:20:00 ERROR Failed to connect to smart TV 'Living Room TV'
2023-06-21 16:22:00 WARNING Reattempting connection to smart TV.
2023-06-21 16:25:00 ERROR Unable to establish connection to smart TV 'Living Room TV'.
2023-06-21 16:30:00 INFO Reconnected to primary WiFi network 'HomeWiFi'
2023-06-21 16:38:00 ERROR Manual connection to smart TV 'Living Room TV' failed
\end{lstlisting}

In this situation, an Android device was initially connected to a smart TV but then lost connection, which constitutes failure. The underlying cause of this failure is that the Android device initiated an update and turned off the WiFi module. Upon completion of the update, the module was switched on again and connected to home WiFi instead of the smart TV. Because the device is now connected to the home WiFi, it cannot reconnect to the smart TV, leading to an error that indicates that the connection to the smart TV has been lost.

By effectively applying RCA to different data sources from monitoring systems, AIOps can ensure precise problem resolution while maintaining optimal system performance and reliability.

\section{Log Data}
\label{sec:background:log_data}

In the complex landscape of modern \ac{IT} systems and services, the generation, collection, and analysis of log data is a critical cornerstone in understanding the dynamics and performance of these complex environments~\cite{miranskyy2016operational}. 
Log data, often referred to as event or audit trails, encapsulates a chronological record of activities, events, and transactions within an IT system~\cite{he2016experience}. 
This section provides a summary of log data, explaining their nature and the techniques used to process them for subsequent analysis.

Log data is generated from the execution of the source code of an IT system which was developed by numerous developers.
These developers write complex source code and embed logging statements. 
During runtime, these logging statements capture and record the current state of variables and other relevant data, generating log lines that are written to log files~\cite{bogatinovski2022qulog}.

The volume of log data generated can be enormous, often comprising millions of log lines, and analyzing them falls into the vast domain of big data due to its sheer volume, velocity, and variety. 

They serve as a critical source of information for understanding system behavior and forensics, offering a window into the internal workings of IT systems.
Log data analysis can reveal patterns, anomalies, and trends, providing indispensable information for troubleshooting, anomaly detection, and root cause analysis. 
The ability to extract actionable insights from log data is essential for maintaining operational integrity and ensuring the dependability of IT systems.

\subsection{Log Data Characteristics}
\label{subsec:background:log_data:characteristics}

In this section, we describe the appearance of raw and unstructured log data and discuss their attributes and features.
The log file documents the executions of the software and is created by log instructions (e.g. \texttt{printf()} or \texttt{log.info()}). 
Each log instruction results in a single log line, so that the complete log is a sequence of lines $\mathcal{L} = \{l_1, l_2, \ldots, l_n\}$. 
The raw and unstructured log files appear as follows:

\begin{lstlisting}[basicstyle=\ttfamily\scriptsize, breaklines=true]
2024-05-24 14:23:01 INFO auth PID: 123 user johndoe successfully logged in
2024-05-24 14:24:45 ERROR auth PID: user:johndoe set_new_password:None
2024-05-24 14:25:30 WARN db PID: 567 Database connection is slow, response time: 30ms
2024-05-24 14:26:15 DEBUG cache PID: 876 Cache hit for key 'session_ABC123'
2024-05-24 14:27:00 INFO payment PID: 678 Payment processed successfully for ID 987
2024-05-24 14:27:45 ERROR auth PID: 123 Password reset failed for user 'janedoe'
2024-05-24 14:28:30 WARN db PID: 567 Disk space is running low: 10% remaining
2024-05-24 14:29:15 INFO web PID: 432 Rendering homepage for user 'johndoe'
2024-05-24 14:30:00 INFO cache PID: 876 Cache cleared for key 'session_XYZ789'
2024-05-24 14:30:45 ERROR payment PID: 678 Payment processing failed
\end{lstlisting}

The above example illustrates ten example log lines. 
Each log line $l_i$ is composed of \textit{meta-information}, such as timestamp, severity level, service name, and Process ID (PID), which provide context about the logged event itself, followed by the log message itself, also called \textit{content}.
To access the content we write $c_i$.
The content or message contains dynamic information specific to each event.  
The following notation applies to the entire thesis. 
When we speak of a log line, we mean the entire log line, including the log message. 
In the case of the log message or content, we only mean the human-written section.

For instance, in the log line \texttt{2024-05-24 14:23:01 INFO auth PID: 123 user johndoe successfully logged in}, the timestamp is \texttt{2024-05-24 14:23:01}, the severity level is \texttt{INFO}, the service is \texttt{auth}, and the PID is \texttt{123}, which are the \textit{meta-information}.
The content \texttt{user johndoe successfully logged in} is free text, written by the developer and consists of a static and a variable part~\cite{he2017drain, wittkopp2021taxonomy}. 

\subsection{Log Data Processing Techniques}
\label{subsec:background:log_data:processing}

The continuous inflow of log events from diverse sources contributes to complexity and scale. 
The sheer volume of data, coupled with its unstructured nature, demands advanced processing techniques. 
Therefore, this section presents log data processing techniques that cover tokenization, embedding, and creating templates.

\subsubsection{Tokenization}
In recent years, advances in tokenization methods have been fueled by deep learning techniques, particularly transformer models like BERT~\cite{devlin2018bert} and GPT, which have demonstrated remarkable performance across various \ac{NLP} benchmarks. 
Tokenization is a fundamental process in \ac{NLP}, which involves breaking down text into smaller units, or tokens, such as words, phrases, or characters, for analysis~\cite{a2018fast,khurana23natural}. 

As the content $c_i$ of a log line $l_i$ is the only part that is freely written text, tokenization is typically applied solely to it.
The smallest indecomposable unit within a log content $c_i$ is a token. 
Consequently, each log content $c_i$ can be interpreted as a sequence $c_i = (w_j: w_j \in V, j = 1,2,\ldots, s_i)$ of tokens, where $w_j$ is the $j$-th token in $c_i$, $V$ is a set of all known tokens, which constitutes the \emph{vocabulary}, and $s_i$ denotes the total number of tokens in $c_i$. 
Thus, tokenization translates a text into a sequence of tokens, creating a well-formed vocabulary.
This process splits written text into segments (for example, words, word stems, or characters).
The amount of tokens $c_i$ for each log message and the structure of each token can vary, depending on the concrete tokenization method. 

As an illustration, we tokenize the first two log lines from the provided log example: \\
\texttt{user johndoe successfully logged in} and \\
\texttt{user:johndoe set\_new\_password:None}
with different tokenizers.

\begin{table}[h]
  \centering
  \begin{tabular}{|c|c|c|c|}
    \hline
    split characters & token sequences & $s_i$ & \makecell{$|V|$} \\ 
    
    \hline
    \texttt{blank space} & 
    \makecell{(user, johndoe, successfully, logged, in) \\ (user:johndoe, set\_new\_password:None)} & 
    \makecell{5\\2} & 
    7 \\ 
    
    \hline
    \texttt{blank space :}  &
    \makecell{(user, johndoe, successfully, logged, in) \\ (user, johndoe, set\_new\_password, None)} &
    \makecell{5\\4} & 
    7\\ 
    
    \hline
    \texttt{blank space : \_}  & 
    \makecell{(user, johndoe, successfully, logged, in) \\ (user, johndoe, set, new, password, None)} &
    \makecell{5\\6} & 
    9 \\
    
    \hline
  \end{tabular}
  \caption{Influence of the splitting character on the vocabulary and the token sequence.}
  \label{tab:background:log_data:processing:splitting}
\end{table}

This~\autoref{tab:background:log_data:processing:splitting} effectively illustrates how the choice of splitting characters can significantly impact the token sequence and the size of the vocabulary.

\subsubsection{Embeddings} 
As tokens themselves are elements of the vocabulary $V$ and therefore consist of characters or strings, they cannot be passed directly into a \ac{NN}. 
This is where embeddings come into play by computing a numerical representation of each token such that a machine learning model can process it.
Embeddings are vectors that represent a token in a $d$ dimensional space.
Moreover, tokens alone do not provide any information about their similarities or differences, but vectors can capture these relationships.
Embeddings are trainable units adapted during the model training process to represent the meaning of the original token.

\begin{figure}[h]
\centering
\includegraphics[width=0.3\columnwidth]{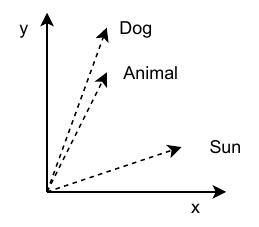}
\caption{Example of embedding vectors in a two dimensional space.}
\label{fig:background:embeddings}
\end{figure}

Figure~\ref{fig:background:embeddings} examines the trained embeddings, where \textit{Dog} and \textit{Animal} are more close to each other, whereas \textit{Sun} is mapped to another point.
In an n-dimensional space, distinct words with different meanings can be mapped to various points, illustrating the relationships or similarities among them.
Moreover, embeddings can be derived not just from single words but also from complete sentences. This is significant for log analysis, as an embedded log line can be represented by a single vector and mapped to a point in a n-dimensional space.


\subsubsection{Templates}
Another way to process and standardize logs is to create templates.
Log templating is a technique used to create structured formats for log messages to ensure consistency and facilitate easier analysis. This process involves mining templates that capture the common structure of log messages while using placeholders for variable elements~\cite{he2017drain}. By employing templates, logs become standardized, making them more readable and easier to process automatically.

The process of creating log templates starts with parsing raw log messages into their constituent parts, called static and variable parts~\cite{nedelkoski2020self,zhu2019tools,he2017drain}. 

\begin{lstlisting}[basicstyle=\ttfamily\scriptsize, breaklines=true]
server started at 15:05:37
user johndoe successfully logged in
user sandra successfully logged in
server started at 18:17:12
\end{lstlisting}

For example, the four log lines can be translated into the following templates:

\begin{lstlisting}[basicstyle=\ttfamily\scriptsize, breaklines=true]
server started at <*>
user <*> successfully logged in
\end{lstlisting}

A log message such as \textit{server started at 15:05:37} can be broken down into static text and variable elements, also called attributes, resulting in a template like \textit{server started at <*>}. 
Here, "<*>" acts as a placeholder for any attribute that can appear at that position, such as a specific time or other changing data.

Templating ensures that log messages follow a consistent format, which is crucial for several reasons. Firstly, it improves readability by maintaining a uniform structure across all logs, making it easier for developers and system administrators to quickly understand the log data. Secondly, it facilitates automated parsing and analysis. 
Moreover, structured logs are invaluable for debugging and troubleshooting. 

Log templating also plays a critical role in monitoring and alert systems. Templates allow for precise monitoring rules and alerts based on specific log patterns, helping to quickly identify and respond to potential issues~\cite{he2021survey}. 
In addition, templating supports data analysis by providing a structured format that can be easily ingested by analytical tools. This structured data can then be used to generate insights, such as usage patterns and error trends, enabling data-driven decision making.

\section{Learning Strategies}
\label{sec:background:learning_strategies}
This section provides background information on different learning strategies for statistical, machine learning models, and neural networks.

In modern data science, the transition from creating functions to building models to creating models from data, also known as data-driven models, indicates a significant change. 
Historically, functions were carefully designed based on theoretical premises or empirical data, with the goal of capturing the fundamental relationships within a system.
Nevertheless, this method frequently proved inadequate when confronted with intricate, real-world data marked by its large scale, diversity, and rapid change.
The rise of machine learning and statistical modeling utilizes data-driven methods to build models that can infer from observed trends, adjusting dynamically to new data and unforeseen situations. 
By learning directly from data, these models offer a more adaptive and scalable approach, capable of discerning intricate patterns that traditional function-based approaches might overlook.

There are multiple strategies for training those data driven models to enable them to learn from the data. 
The need for diverse learning strategies comes from the varied nature of the data and the objectives we aim to achieve. 
Data sets are not always labeled or categorized; they frequently display subtleties, ambiguities, or inconsistencies that complicate direct modeling. Supervised learning techniques, which depend on labeled examples to discern patterns, perform well in situations where there are clear separations between classes or outcomes.
In contrast, unsupervised learning methods explore the unknown domain of unlabeled data, aiming to discover latent patterns or clusters without any predefined labels. 
Moreover, weakly supervised approaches navigate the intermediate terrain, where only partial or noisy labels are available, demanding robust methodologies to distill meaningful insights. 

Each strategy embodies a tailored approach to learning from data, offering distinct advantages and trade-offs suited to the complexity and ambiguity inherent in real-world data sets.

\subsection{Unsupervised Learning}
\label{subsec:background:training_strategies:unsupervised}


Unsupervised learning is a type of machine learning in which algorithms are trained on data without predefined labels or outcomes~\cite{dike2018unsupervised}. 
In data-driven models, this learning strategy aims to uncover hidden patterns or structures within the data. 
It enables the segmentation of data into meaningful groups without having known these groups before~\cite{sinaga2020unsupervised}. 

Unsupervised learning encompasses various techniques, each suited for specific tasks and types of data. 
Clustering techniques group data points based on similarity, making them useful for market segmentation, customer categorization, and image compression without knowing the number of groups in advance.
Furthermore, dimensionality reduction techniques are also part of unsupervised learning. Examples are \ac{PCA}~\cite{jolliffe2005principal} and t-distributed stochastic neighbor embedding (t-SNE)~\cite{arora2018analysis}, which reduce the number of variables while preserving essential patterns, aiding in data visualization and noise reduction.

Unsupervised learning is also essential in anomaly detection, where it identifies outliers in the data that significantly deviate from the norm, by training normal behavior and calculating a good representation of normal samples.

In the context of log anomaly detection, unsupervised learning is highly applicable due to its ability to operate without labeled examples of normal and anomalous behavior. 
Unsupervised learning techniques can detect anomalies in log data by identifying patterns and deviations without prior knowledge of specific anomalies. 

The application of unsupervised learning in log anomaly detection offers several advantages. 
It allows for the detection of previously unknown anomalies, making it adaptable to new and evolving failures~\cite{ramponi2020neural}. 
This adaptability is crucial in dynamic IT environments where the nature of anomalies can change rapidly~\cite{qi2020small,baier2019cope}. 

However, the use of unsupervised learning for log anomaly detection also presents challenges. 
One significant challenge is the potential for false positives, where benign deviations are incorrectly flagged as anomalies.


\subsection{Weak-Supervised Learning}
\label{subsec:background:training_strategies:weak_supervised}



Weakly supervised learning is a type of machine learning that leverages imperfect or limited labeled data to train algorithms. 
This approach bridges the gap between fully supervised and unsupervised learning by utilizing labels that are incomplete, noisy, false, or coarse-grained~\cite{zhou2018brief,bekker2020learning}. 
It aims to build models that can make accurate predictions or uncover patterns despite the constraints of suboptimal training data.

Weakly supervised learning encompasses various techniques designed to handle different types of weak supervision. 
Techniques such as semi-supervised learning use a small amount of labeled data combined with a large amount of unlabeled data to improve learning accuracy. Self-training and co-training are common methods in which a model iteratively labels the unlabeled data and retrains itself~\cite{zhu2009introduction}. 
Or, the models use coarse-grained labels, where only the superordinate label is available for several samples of the same class, but no label for each individual sample~\cite{carbonneau2018multiple}.

Weakly supervised learning is also valuable in anomaly detection. 
It utilizes limited labeled or incorrect labeled examples of normal and anomalous~\cite{becker2020towards,liu2002partially}. 
In the context of log anomaly detection, weakly supervised learning is particularly effective because labeled examples of anomalies are often scarce and expensive to obtain~\cite{wittkopp2020decentralized,wen2020time}. 
By leveraging a small set of labeled log entries, weakly supervised learning methods can significantly improve the detection of anomalies in large and complex log datasets~\cite{zhang2019robust}. 
These techniques can refine their models iteratively, improving their ability to detect subtle and evolving patterns of abnormal behavior~\cite{liu2021self,wittkopp2022pull}.

The application of weakly supervised learning in log anomaly detection offers several benefits. 
It reduces the dependency on extensive labeled data sets, making it feasible to deploy in real-world scenarios where labeled data is limited. 
This approach also improves the model's ability to generalize from limited examples, making it adaptable to various types of anomalies~\cite{wittkopp2021loglab}. 

However, weakly supervised learning for log anomaly detection also presents challenges. 
The main challenge is dealing with the noise and inaccuracies in the weak labels, which can lead to suboptimal model performance. 
Furthermore, the iterative nature of some weakly supervised methods can be computationally intensive, requiring careful tuning and validation to maintain accuracy and efficiency~\cite{ren2020not}.

\subsection{Supervised Learning}
\label{subsec:background:training_strategies:supervised}




Supervised learning is a type of machine learning in which algorithms are trained on a labeled data set, meaning each training example is paired with an output label. 
This learning strategy aims to learn a mapping from inputs to outputs, allowing the model to make predictions on unseen data based on the patterns it has learned~\cite{mohri2018foundations}. 
Supervised learning is essential for building models for classification and regression tasks, where the goal is to predict discrete labels or continuous values, respectively.

Supervised learning encompasses various techniques tailored to different types of prediction tasks. 
Classification techniques, such as decision trees, support vector machines, and neural networks, are used when the output is a discrete label. 
These techniques are commonly applied in tasks like image recognition, spam detection, and medical diagnosis~\cite{bishop2006pattern}. 
Regression techniques, such as linear regression, polynomial regression, and support vector regression, are used when the output is a continuous value. These techniques are used to predict stock prices, estimate house values and forecast sales~\cite{hastie2009elements}. 

Supervised learning is also critical in anomaly detection, particularly when labeled examples of normal and anomalous behavior are available. 
By training on data sets that include both types of examples, supervised learning algorithms can learn to distinguish between normal patterns and anomalies effectively. 
This approach is beneficial in scenarios where anomalies have well-defined characteristics that can be captured during the training phase.

In the context of log anomaly detection, supervised learning is highly effective when a sufficiently large and representative set of labeled logs is available. 
These models are then extremely effective for the log data of the underlying IT system or service~\cite{du2017deeplog,wittkopp2020a2log}.

The application of supervised learning in log anomaly detection offers several advantages. 
It provides high accuracy in detecting known anomalies due to the direct learning from labeled examples.
In addition, some supervised models can provide detailed explanations for detected anomalies, facilitating easier interpretation and response~\cite{brown2018recurrent}.

However, the use of supervised learning for log anomaly detection also presents challenges. 
One significant challenge is the dependency on a large and accurately labeled data set, which can be costly and time-consuming to obtain. 
Moreover, supervised models may struggle to detect novel or rare anomalies that were not represented in the training data, limiting their applicability in rapidly changing environments. 
Continuous retraining and validation are necessary to maintain the relevance and performance of the model~\cite{chalapathy2019deep}.

\section{Modeling Concepts}
\label{sec:background:concepts}
After illustrating how our methods interact across three distinct layers, in this section, we introduce two concepts that reappear throughout the subsequent chapters, forming the backbone of our methods for solving complex problems in log analysis. 
First, we discuss a widely used neural network architecture known for its effectiveness in handling natural language in~\autoref{subsec:background:concepts:transformer}. 
Second, we present a training strategy tailored for this architecture, specifically designed for scenarios frequently encountered in real-world applications in~\autoref{subsec:background:concepts:pu_learning}.

\subsection{Transformer Encoder Architecture}
\label{subsec:background:concepts:transformer}
In this section, we present the \ac{TEA}, which is a neural network architecture that is used in the landscape of \ac{NLP} and \ac{ML}. 
Thereby we provide a detailed overview of the Transformer encoder, its mathematical foundations, and its application in translating log lines to a latent space.

The field of NLP and \ac{DL} has seen transformative advances with the introduction of the transformer architecture~\cite{DevlinCLT19} with self-attention by Vaswani et al.~\cite{VaswaniSPUJGKP17}. 
The transformer model has revolutionized the way we approach sequence-to-sequence tasks, including machine translation, text generation, and various other NLP applications by modeling relationships within input sequences through the attention mechanism.

The architecture consists of an encoder-decoder structure, where both components are made up of multiple layers of self-attention and feed-forward neural networks. 
The encoder processes the input sequence to generate context-aware representations in latent space, while the decoder uses these representations to generate the output sequence step by step, conditioned on the previously generated tokens.
In our work, only the encoder component of the architecture is utilized to encapsulate all important information within the latent space.

\begin{figure}[htbp]
\centering
\includegraphics[width=\columnwidth]{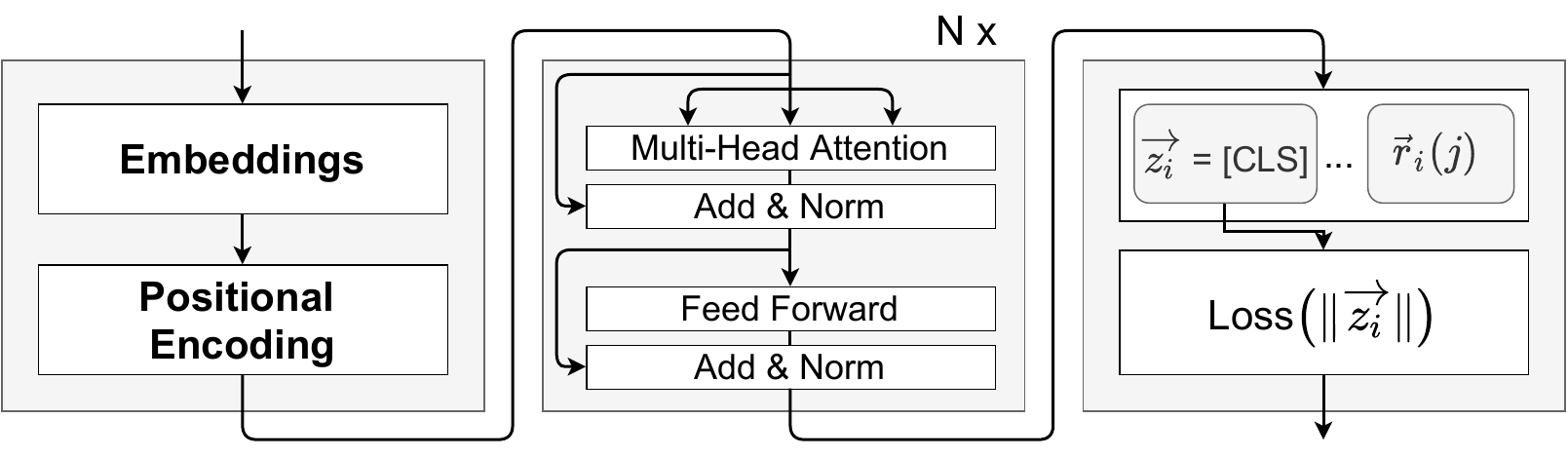}
\caption{Transformer encoder architecture.}
\label{fig:log_analysis_layer:concepts:transformer:transformer_architecture}
\end{figure}

\autoref{fig:log_analysis_layer:concepts:transformer:transformer_architecture} depicts the aforementioned network architecture.
Initially, the embeddings derived from tokenized log messages are used as input (see~\autoref{subsec:background:log_data:processing}), followed by the application of positional encoding to each embedding, due to the fact that the transformer lacks the capability to recognize the sequence of tokens.

Let $\mathcal{L} = \{l_1, l_2, \ldots, l_n\}$ be a sequence of log lines, $t_i$ the tokenized content of each log line $l_i$, and let $\mathbf{E}$ be the embedding matrix. 
The preprocessed token sequence $t_i$ is translated into embeddings using the embedding matrix $\mathbf{E}$.
The embedding matrix $\mathbf{E}$ is a matrix of learned parameters of size $V \times d$, where $V$ is the vocabulary size and $d$ is the embedding dimension. 
The $m$-th row of $\mathbf{E}$ corresponds to the embedding of the token with ID $m$.
The embedding vector for the $j$-th token of the token sequence $t_i$ is given by $\mathbf{e}: \mathbb{N} \rightarrow \mathbb{R}^d$ and written as $\mathbf{E}(t_i^j)$
We write $\vec{e_i}$ for all embeddings of a token sequence $t_i$, which is the embedded log line.

In order to explicitly encode the position of a token in a token sequence $t_i$ in the
calculated embedding $\vec{e_i}(j)$, \emph{positional encoding} is used~\cite{GehringAGYD17,VaswaniSPUJGKP17}. 
The values are sampled using a function $p:\mathbb{R}^d\rightarrow \mathbb{R}^d$ as 
\begin{equation}
    p(j,k):= \begin{cases}
    \sin (\omega_k \cdot j) & \text{if}\quad j=2k\\
    \cos (\omega_k \cdot j) & \text{if}\quad j=2k + 1\\
    \end{cases}
\end{equation}
where 
\begin{equation}
    \omega_k = \frac{1}{10000^{2k/d}}.
\end{equation}
Furthermore, $j$ signals the position of the token and $\omega_k$ determines the frequency of the sine and cosine functions, ensuring that different dimensions use different frequencies.
Moreover, $k$ is an index given by $k = \left\lfloor \frac{j}{2} \right\rfloor$, which represents the position of the dimension in the embedding space.

Then the corresponding embeddings, enrichted with positional encoding, $\vec{e_i}$ are assigned to each token of the token sequence $t_i$ and captured in a prepanding $\mathbf{CLS}$ token by:
\begin{equation}
\mathbf{X} = \mathbf{E}(\text{CLS}) \oplus \mathbf p({E}(t_1)) \oplus \mathbf{E}(t_2) \oplus \cdots \oplus \mathbf p({E}(t_n))
\end{equation}
where $\oplus$ denotes concatenation.

Each log line is embedded into a continuous vector space through all the tokens it contains and their respective embeddings in the embedding matrix $\mathbf{E}$ and their positional encoding $p$. 

The Transformer encoder consists of $N$ identical layers, depicted in the box in the middle of~\autoref{fig:log_analysis_layer:concepts:transformer:transformer_architecture}, each containing two main sub-layers: a multi-head self-attention mechanism and a position-wise fully connected feed-forward network. 
Residual connections and layer normalization are employed around each of these sublayers, called Add \& Norm.

The self-attention mechanism computes attention scores to weigh the importance of different tokens in the sequence. 
For each head, the scaled dot-product attention is calculated as:

\begin{equation}
\text{Attention}(\mathbf{Q}, \mathbf{K}, \mathbf{V}) = \text{softmax}\left(\frac{\mathbf{QK}^T}{\sqrt{d_k}}\right) \mathbf{V}
\end{equation}

where $\mathbf{Q}$, $\mathbf{K}$, and $\mathbf{V}$ are the query, key, and value matrices, respectively, and $d_k$ is the dimensionality of the key vectors.

Multi-head attention allows the model to focus on different parts of the input sequence:
\begin{equation}
\text{MultiHead}(\mathbf{Q}, \mathbf{K}, \mathbf{V}) = \oplus(\text{head}_1, \ldots, \text{head}_h) \mathbf{W}^O
\end{equation}
where 
\begin{equation}
\text{head}_i = \text{Attention}(\mathbf{Q} \mathbf{W}_i^Q, \mathbf{K} \mathbf{W}_i^K, \mathbf{V} \mathbf{W}_i^V)
\end{equation}
and $\mathbf{W}_i^Q$, $\mathbf{W}_i^K$, $\mathbf{W}_i^V$, and $\mathbf{W}^O$ are learned parameter matrices~\cite{VaswaniSPUJGKP17}.

The output of the final encoder layer is the $\mathbf{CLS}$ token embedding, which contains all input information as ambeddings, which will be optimized during training.
Thus, the model is indirectly responsible for training the aforementioned embeddings of the tokens through backpropagation of prediction errors.
Subsequent objective functions will focus solely on computing the loss of the $\mathbf{CLS}$ token. 
We denote the output of the model as $\vec{z_i}$ and use it throughout this thesis.

The transformer model is widely used in NLP due to its superiority over other architectures. 
Consequently, it is also effective in log analysis. 
This network is used in multiple methods within our architecture, making it fundamental to this thesis.

\subsection{Training with Positive and Unknown Data}
\label{subsec:background:concepts:pu_learning}
The second concept, training with positive and unknown data, addresses the crucial need for robustness and adaptability in model training without perfect labeled data, offering an approach to improve model performance in different real-world scenarios.

In the field of neural network training, high quality labeled data is crucial to achieve well-performing models, as their efficacy depends on the comprehensiveness and accuracy of the labeled data sets used during training~\cite{krizhevsky2012imagenet}.
The process of labeling data involves significant human effort and expertise, often making it prohibitively expensive and slow. 
Therefore, acquiring such high-quality labeled data is a challenging, cost-intensive, and time-consuming process~\cite{sun2017revisiting, wen2021time}. 
This bottleneck in data preparation can hinder progress in developing robust machine learning models~\cite{zhu2019tools}.
Moreover, in many practical scenarios, the available data is incomplete or noisy, further complicating the training process~\cite{ratner2020snorkel}. 
Therefore, it is essential to develop training strategies that can take advantage of the available data more efficiently, reducing the dependency on large volumes of high quality labeled data.

In numerous cases, the exact moment on millisecond level of a system failure is unclear but is roughly known from different sources.
On the one hand, modern IT systems and services have monitoring tools that notify DevOps teams about failures, offering a preliminary but useful starting point for log analysis.
On the other hand, applications and services utilized by end users are capable of transmitting crash reports that include timestamps and log data.
The alerts associated with the time points and the log data serve as the main anchor for analysis. 
Using the failure time, it becomes feasible to narrow down to single relevant log lines~\cite{oliner2012advances,bekker2020learning}.
Combining this information, we present a training strategy that is applicable in many log analysis tasks that can deal with inaccurate labeled log data.

We use the approximated failure times to create \acfp{FTW} to then approximate two classes. 
One class with normal data and another class with unknown data that contains abnormal and normal data~\cite{elkan2008learning}. 
The scenario outlined is also called PU learning: Learning from positive and unknown~\cite{liu2002partially,liu2003building}.
Learning from positive and unknown samples is an umbrella term for several weakly supervised binary classification methods that classify unknown samples by learning the positive class and the statistical distributions in the unknown class~\cite{liu2002partially,zhu2009introduction,bekker2020learning}. 

In an PU learning scenario, all log messages that occur during the failure time windows represent the unknown class $\mathcal{U}$, while all other log messages represent the positive class $\mathcal{P}$. 
PU-Learning is a special form of weakly supervised training, where some training samples have inaccurate labels. 
These inaccurate labels come from the circumstance that we cannot assign accurate labels due to imprecision of the \ac{FTW}.
Weak supervision with inaccurate labels is defined as a situation in which the supervision information does not always match the ground truth~\cite{zhou2018brief}.
By design, the preliminary labels derived from the assignment do not have to correspond to the underlying ground truth. 
With this training strategy, we are able to calculate the error of the inaccurate labels and find the real anomalies and root cause log lines in the unknown class.

This strategy for training log analysis models, relies on 2 assumptions:
\begin{enumerate}
  \item The approximated normal class consists of true normal samples.
  \item The characteristics of the real abnormal samples, hidden in the unknown samples, differ from those of the normal samples.
\end{enumerate}


In order to clarify this training strategy within the scope of log analysis, we will now present a formal definition.

\subsubsection{Formal Definition}
Let $\mathcal{L} = \{l_1, l_2, \ldots, l_n\}$ be a sequence of log lines.
Using weak supervision with inaccurate labels, each log line $l_i \in \mathcal{L}$ is assigned a triple $(m_i, \tilde{y}_i, y_i)$, where $m_i$ is the preprocessed log message, $\tilde{y}_i$ the inaccurate label and $y_i$ the ground truth. 
The ground truth $y_i$ is only available in the experiment setup, to evaluate and compare different methods. 
We define $\tilde{y}_i, y_i \in \{0,1\}$, where $0$ is the label for normal and $1$ for abnormal log lines, and form the two disjunct classes: 

\begin{equation}
\mathcal{P} = (l_i\ |\ \forall l_i \in \mathcal{L}, \tilde{y}_i = 0) \\
\mathcal{U} = (l_i\ |\ \forall l_i \in \mathcal{L}, \tilde{y}_i = 1)
\end{equation}

Let $\Phi(x_i, \tilde{y}_i, \Theta) : \mathcal{L} \rightarrow [0 \ldots 1]$ be a function represented by a trainable model, which predicts a label $\widehat{y}_i \in \{0,1\}$ for the input log message $m_i$, by training the inaccurate labels $\tilde{y}_i$. 
The task is to learn the parameters $\Theta$ during training and then predict a label $\widehat{y}_i$ for each sample $m_i$. 
The predictions are defined as $\mathcal{Y} = (\widehat{y}_{i}\ |\ \forall l_i \in \mathcal{L}, \widehat{y}_{i}=\Phi(x_i;\Theta))$, where $\Phi(x_i;\Theta)$ is the model output. 
We write $\widehat{y}_i$ to access the predicted label for each log sample. 
The goal is therefore to train all log messages $m_i$ with inaccurate labels $\tilde{y}_i$, to calculate a score for each log message that indicates how likely these log messages are an anomaly, used for anomaly detection or root cause analysis.

This concept gives us the ability to train models for automated labeling, anomaly detection, and root cause analysis in certain scenarios where we only have an approximate time of failure but no exact labels per log line. 
Together with the transformer network, this is a fundamental technology to make multiple facets of log analysis efficient.

\cleardoublepage

\chapter{Related Work}
\minitoc
\label{ch:related_work}

In this chapter, we provide related work for this thesis.
Thereby we also provide details for all methods in different sections that we use as baselines later on in our evaluation.
We structured this chapter according to our three layers for log analysis.
First, we provide related work for \textit{log investigation} in~\autoref{sec:related_work:log_investigation}.
Second, we present related work for \textit{anomaly detection} in~\autoref{sec:related_work:anomaly_detection}.
Third, we elaborate on some work in \textit{root cause analysis} for log data in~\autoref{sec:related_work:root_cause_analysis}.
Finally, we discuss methods for \textit{PU learning} applied to log data in~\autoref{sec:related_work:pu_learning}, since PU learning is one of our main concepts.

\section{Log Investigation}
\label{sec:related_work:log_investigation}

In this section, we review methods and research articles relevant to log investigation, the first layer of our architecture. 
Since this layer is not common in the log analysis literature, we focus on two main areas: anomaly categorization and text data classification.

\subsection{Categorization of Anomaly Types}
\label{sec:related_work:log_investigation:categorization}
Anomalies, often referred to as outliers, exceptions, or deviations, are patterns in data that do not conform to a well-defined notion of normal behavior and can manifest in different forms~\cite{chandola2009anomaly,steinwart2005classification}.
In the literature, numerous approaches have been proposed to categorize anomalies in different data types based on their characteristics and the context in which they occur. 

A comprehensive survey~\cite{chandola2009anomaly} provides a structured overview of these approaches, grouping anomaly detection techniques into distinct categories based on their underlying methodologies and assumptions.
The authors differentiate between \emph{point anomalies}, where an individual data sample is anomalous in comparison to the rest, \emph{contextual anomalies}, where a data sample is only considered anomalous in specific contexts, and \emph{collective anomalies}, where a single data sample is only anomalous when occurring as part of a collection of related data samples, not individually. 

Similarly, the authors of~\cite{sebestyen2018taxonomy} classify anomalies into three categories: \emph{one-point anomaly}, \emph{contextual anomaly}, and \emph{sequential data anomaly}, defining them in a manner consistent with previous works. Their research emphasizes the importance of a systematic classification framework to aid domain specialists in understanding the nature of anomalies within their specific fields. By categorizing anomalies based on their characteristics and context, this work simplifies the complex landscape of anomaly classification methods.

A review of outlier and anomaly detection in time series data~\cite{blazquez2020review} further distinguishes between \emph{point outliers} and \emph{sequence outliers}. In addition, they introduce the concept of \emph{outlier time series}, where the entire time series can be anomalous and only detectable in multivariate contexts. This comprehensive and structured review presents a taxonomy based on the main aspects that characterize outlier detection techniques in time series, aiding in the identification and application of appropriate methods.

Log messages can also be distinguished into event log messages and state log messages~\cite{NagarajKN12} and can also be written in a distributed manner, which introduces additional challenges. They address these challenges in the context of anomaly classification by using machine learning techniques to analyze log data from large-scale distributed systems.


\subsection{Text-Based Classification}
\label{sec:related_work:log_investigation:text_classification}
Text-based classification involves a variety of methods aimed at categorizing textual data into predefined classes or categories~\cite{minaee2021deep}. These methods are pivotal in natural language processing (NLP) applications such as sentiment analysis, topic categorization, spam detection, and more. For all these classification methods, the challenge lies in identifying optimal structures, architectures, and methodologies for robust text classification~\cite{minaee2021deep,kowsari2019text}. 

The PCA algorithm~\cite{jolliffe2005principal} is for instance often employed for dimensionality reduction right before the actual classification procedure, is frequently applied in text classification pipelines to reduce the dimensionality of feature vectors derived from textual data before feeding them into classification algorithms. By capturing the most significant variance in the data, PCA helps improve computational efficiency and mitigate the curse of dimensionality.

Decision Trees~\cite{quinlan1986induction}, represent a widely adopted method in classification tasks due to their interpretability and effectiveness. Safavian and Landgrebe~\cite{safavian1991survey} highlighted their utility across various domains, while Chen et al.~\cite{chen2004failure} applied them to diagnose failures in large-scale IT services.

Random forests~\cite{ho1995random} are well-suited for text classification tasks due to their ensemble learning approach. They combine multiple decision trees to produce robust classifications. Random forests are proficient at managing high-dimensional data and are especially adept at processing noisy or unbalanced text data sets.

Logistic regression, a classic statistical method discussed in~\cite{hosmer2013applied,dou2018comparative}, continues to be fundamental in text classification. Despite its simplicity, logistic regression models provide interpretable results and can handle both binary and multiclass classification tasks effectively. They are valued for their ability to estimate the probabilities of class membership based on input features and has demonstrated its usefulness for text classification~\cite{genkin2007large}.

\ac{SVM}~\cite{hearst1998support}, have proven to be an effective approach for text classification. This technique, noted for its strong foundation in computational learning theory and impressive real-world results. In~\cite{manevitz2001one}, SVM is assessed primarily for one-class classification in document categorization, utilizing different kernels and data representation techniques like tf-idf and binary vectors on the Reuters data set.

The authors of~\cite{schapire2000boostexter} introduce BoosTexter, an ensemble learning method based on boosting algorithms tailored for multi-class text categorization tasks. BoosTexter represents a significant advancement in text classification, leveraging boosted classifiers to improve accuracy and robustness across various categorization challenges. Their implementation demonstrates superior performance compared to traditional text-categorization algorithms, as evidenced by comprehensive evaluations on benchmark datasets. This work underscores the effectiveness of boosting techniques in enhancing classification accuracy and addressing complex classification scenarios in natural language processing.

The Rocchio algorithm, evaluated in~\cite{sowmya2016large}, demonstrates strong performance in large-scale multi-label text classification tasks. Compared with kNN, Rocchio leverages a relevance feedback mechanism to adjust classification boundaries based on the similarity of documents to prototype vectors representing different classes. This approach has shown superiority in scenarios where labeled training data is abundant but requires efficient computational resources.
In another work~\cite{selvi2017text}, the authors design a pipeline that uses the Rocchio algorithm. 
As individual pipeline components improve on each other, the motivated approach is able to outperform methods such as naive Bayesian algorithms.

These methods exemplify the diversity of approaches available for text-based classification, each offering unique advantages depending on the specific characteristics and requirements of the textual data and the classification task at hand.

\section{Log Anomaly Detection}
\label{sec:related_work:anomaly_detection}

The detection of anomalies is crucial in AIOps to identify deviations from the normal behavior of the IT system that may indicate problems or failures~\cite{dang2019aiops,shen2020evolving}. 
Detecting abnormal events in large-scale systems, indicated by log files, is crucial to creating reliable systems~\cite{xu2009detecting}. 

Therefore, log analysis becomes increasingly important for industry and academia~\cite{qi2020small} and a wide range of different anomaly detection techniques have been developed and discussed in detailed surveys~\cite{chandola2009anomaly, he2016experience}.
Hence, log based anomaly detection has been extensively studied, demonstrating strong performance on stable log data~\cite{chandola2009anomaly}. 
They utilize different forms of log templates and log embeddings to convert logs to a machine-readable format. 
Commonly used anomaly detection methods are support vector machines~\cite{liang2007failure} and principal component analysis~\cite{jolliffe2005principal}. 
In addition, there are rules-based~\cite{breier2015anomaly},
tree-based~\cite{chen2004failure},
statistical~\cite{xu2009detecting},
and clustering-based methods~\cite{lou2010mining, cinque2012event, baseman2016relational, LANDAUER201894}. 
However, recent anomaly detection methods are mainly designed with neural networks~\cite{zhang2019robust,yin2020improving,naseer2018enhanced} and based on encoder architectures~\cite{nicolau2016hybrid,sakurada2014anomaly}. 
In addition, recent methods use the attention mechanism that is often used in encoder architectures~\cite{VaswaniSPUJGKP17}.

This section reviews related work on anomaly detection, focusing on unsupervised, weak-supervised, and supervised methods~\cite{he2016experience}.
Supervised methods are usually more accurate, though they train the anomalies of the specific data set as well~\cite{zhang2019robust, yang2019nlsalog}.
However, not all anomalies can be known in advance. 
Hence, in industrial applications, unsupervised methods are more practical, as anomaly labels are mostly unavailable~\cite{meng2019loganomaly, wittkopp2020decentralized}. 
Weakly supervised methods represent a hybrid approach, where exact labels are not required, but rather more general or approximate labels are utilized~\cite{zhou2018brief}.


\subsection{Unsupervised Methods}
\label{sec:related_work:anomaly_detection:unsupervised}

Unsupervised anomaly detection methods are particularly valuable because they can learn the underlying patterns of log data without the need for labeled examples. 
In this subsection, various methods for the unsupervised detection of anomalies are presented, most of which are based on neural networks, since they have the ability to model complex relationships in the data.
We explain their architectures, advantages, and use cases.


One of the most known works in the domain of log-based anomaly detection is DeepLog, a deep neural network model utilizing \ac{LSTM} to identify abnormal sequences of log messages~\cite{du2017deeplog}. DeepLog models a system log as a natural language sequence, allowing it to automatically learn log patterns from normal execution and detect anomalies when log patterns deviate from the model trained on normal log data. For this, log templates are generated with Drain~\cite{he2017drain}, and sequences of templates are formed as model inputs. Thereby it constructs system execution workflow models for diagnosis purposes. The model provides a ranked output with probabilities for the next template in a given sequence, and the anomaly detection is based on whether the next template has a high probability or not. 
Additionally, DeepLog can be incrementally updated in an online fashion to adapt to new log patterns over time, and it constructs workflows from the underlying system log, enabling effective diagnosis and root cause analysis of detected anomalies. Extensive experimental evaluations over large log datasets have shown that DeepLog outperforms other existing log-based anomaly detection methods based on traditional data mining methodologies~\cite{lu2018detecting}.

LogAnomaly~\cite{meng2019loganomaly} is similar to DeepLog and predicts the next log message in a sequence of log messages. However, instead of utilizing sequences of log templates, LogAnomaly employs sequences of log embeddings to improve prediction effectiveness. It introduces the representation method template2vec, a novel and effective approach to extract the semantic information hidden in log templates. This method, combined with LSTM networks into an end-to-end framework, enables LogAnomaly to detect both sequential and quantitative log anomalies simultaneously, a capability not achieved by previous works. Additionally, LogAnomaly can avoid false alarms caused by newly appearing log templates between periodic model retrainings. Evaluations on two public production log datasets demonstrate that LogAnomaly outperforms existing log-based anomaly detection methods.

Another approach is Logsy~\cite{nedelkoski2020logsy}, which incorporates an attention mechanism with an encoder architecture to calculate log embeddings. These embeddings of normal log messages are condensed into a centroid using a hyperspherical loss function, while embeddings of abnormal log messages are pushed away. The anomaly detection task is then based on the distance to the centroid. Evaluations on publicly available datasets show that Logsy improves the F1-score by 0.25 compared to previous methods, demonstrating its effectiveness.

In~\cite{FarzadG20}, the authors present an unsupervised anomaly detection method combining isolation forests and autoencoder networks to analyze log messages in cloud and software systems. The autoencoders handle training and feature extraction, while the Isolation Forest predicts positive samples. Evaluations on BGL, Openstack, and Thunderbird datasets show that this approach effectively reduces false positives and outperforms other models.

In~\cite{YuanALYL020}, the authors introduce ADA, an unsupervised online anomaly detection framework using LSTM networks. ADA adapts to new log patterns in real-time and includes a dynamic threshold algorithm to improve detection accuracy. An adaptive model selection strategy ensures efficient resource use and the system guides the storage of abnormal data to reduce costs. Evaluations on the Los Alamos National Laboratory dataset show that ADA achieves a high F1-score of ~95\%, is 97 times faster than existing methods and incurs low storage costs.

However, all the methods presented above use a manually set decision boundary for anomaly detection. However, optimization requires knowledge of the anomalies, which limits these methods.
In contrast to the methods based on neural networks previously discussed, the following methods focus on anomaly detection in a more classical way.

Invariant Miners~\cite{lou2010mining} retrieve structured logs using log parsing, further group log messages according to log parameter relationships, and eventually mine program invariants from the established groups in an automated fashion. These mined invariants, which reveal the inherent linear characteristics of program workflows, are then used to detect anomalies in logs. Experiments on Hadoop demonstrate that this technique effectively detects execution anomalies, providing high accuracy and intuitive insights into the detected problems.

LogCluster~\cite{LinZLZC16} is a clustering-based method designed to perform anomaly detection in large-scale online service systems. It uses log vectorization and aggregative hierarchical clustering to group logs, followed by the extraction of representative log sequences from these clusters. Additionally, LogCluster utilizes a knowledge base to check if log sequences have occurred before, allowing engineers to focus on a small number of previously unseen log sequences. This significantly reduces the number of logs that need to be examined and improves identification accuracy. Experiments on two Hadoop-based applications and two large-scale Microsoft online service systems demonstrate that LogCluster is effective.

\subsection{Weak-Supervised Methods}
\label{sec:related_work:anomaly_detection:weak_supervised}
Weakly supervised methods encompass various techniques such as semi-supervised learning, self-supervised learning, or learning from inaccurate labels.
These methods utilize falsely labeled or partially unlabeled data in logs, significantly reducing the dependency on large annotated data sets~\cite{zhou2018brief}. 
For example, semi-supervised learning algorithms train models using a small set of labeled data along with a large amount of unlabeled data to improve performance~\cite{chapelle2009semi}. 
Self-supervised learning takes advantage of the inherent structure of the data to generate supervisory signals, avoiding the need for manual labeling~\cite{alexey2016discriminative}.
Learning from inaccurate labels is a situation where the supervision information does not always match the ground truth~\cite{zhou2018brief}.
In this subsection, we introduce several weakly supervised methods used for anomaly detection within logs.

Yang et al.~\cite{yang2021semi} propose PLELog, a semi-supervised approach for log-based anomaly detection that combines an attention mechanism with a gated recurrent unit (GRU) neural network. Log messages are transformed into log templates to predict if a sequence of log templates is normal or abnormal. By estimating the labels of training data, PLELog incorporates knowledge of historical anomalies, achieving effective and efficient anomaly detection. Evaluations on public data sets and real-world systems demonstrate the strength of PLELog.

LogClass~\cite{meng2018device} is a method for identifying anomalous logs in network and service management. It combines word representation, PU learning, and a machine learning classifier. Additionally, the authors introduce the Inverse Location Frequency (ILF) method to weight log words for feature construction. This data-driven framework is designed to detect anomalies based on device logs, achieving high performance metrics.

Another paper introduces LogLG~\cite{guo2023loglg}, a weakly supervised method that constructs a log-event graph to capture the structural relationships between log events. By leveraging these relationships, LogLG improves the anomaly detection performance without the need for extensive labeled data. The method integrates graph-based features with machine learning models, achieving high accuracy in detecting anomalies in logs. The evaluation shows that LogLG outperforms traditional log anomaly detection methods by effectively utilizing weak supervision to enhance the detection capability in various log datasets.

Xie et al. propose a deep anomaly scoring network (DASN)~\cite{xie2023weakly} that operates under weak supervision to detect anomalies in logs. The method uses a multi-task learning framework that combines anomaly detection with other related tasks to improve its performance. By applying convolutional neural networks and attention mechanisms, DASN can effectively learn representations from log data and assign anomaly scores to each log entry. The approach demonstrates superior performance compared to existing anomaly detection methods, especially in scenarios with limited labeled data.

This paper introduces KDAlign~\cite{zhao2024weakly}, a framework that enhances weakly supervised anomaly detection by integrating rule-based knowledge with limited labeled data. By converting human expert rules into the knowledge space and aligning them with data using Optimal Transport, KDAlign effectively improves detection accuracy across various anomaly types. This method demonstrates significant performance improvements over state-of-the-art techniques on multiple real-world datasets, showcasing its robustness and adaptability in different scenarios.

This research~\cite{pang2023deep} presents a method that combines representation learning and anomaly scoring in a single pipeline to optimize anomaly detection scores. The approach uses pairwise relation prediction to learn discriminative features from both labeled and unlabeled data. This method significantly reduces overfitting by learning diverse patterns of normality and abnormality, and has shown superior performance compared to traditional unsupervised and semi-supervised techniques.

\subsection{Supervised Methods}
\label{sec:related_work:anomaly_detection:supervised}
Supervised methods for log anomaly detection are crucial in identifying abnormal patterns. 
These methods involve extracting features from log data and applying machine learning algorithms to classify logs as normal or anomalous.
The primary strength of supervised methods lies in their accuracy and reliability, derived from training on labeled examples of normal and anomalous logs~\cite{zhang2019robust,yang2021semi}. This explicit training allows these models to detect subtle and sophisticated anomalies effectively~\cite{ratner2016data}. However, their success heavily relies on the availability of high-quality labeled datasets, which are often challenging and expensive to obtain~\cite{wittkopp2020decentralized,wen2020time}.

This chapter reviews various supervised learning techniques used in log anomaly detection, discussing their methodologies, strengths, and limitations, and highlights significant advancements in the field.

Some text classification techniques have also been applied in the domain of supervised log anomaly detection.

Liang et al.~\cite{liang2007failure} extended SVM's~\cite{hearst1998support} application to anomaly detection, showing its efficacy in identifying anomalous patterns in complex data sets. These evaluations underscore SVM's versatility and effectiveness across different text classification tasks, such as binary log anomaly detection.

Decision Trees~\cite{quinlan1986induction}, are applied by Chen et al.~\cite{chen2004failure} to diagnose failures in large-scale IT services. By training decision trees on runtime properties of requests during user-visible failures, their approach effectively identifies failures with high accuracy. This adaptation involves ranking paths through the tree based on the correlation with failure and merging nodes according to the observed relationship between system components. Such applications underscore the versatility of decision trees in handling complex text based data.

The fact that anomalies are usually few and considerably different is exploited with Isolation Forests~\cite{liu2008isolation}, an ensemble of isolation trees, where anomalies are isolated closer to the root of a tree and thus identified. Isolation Forest isolates anomalies directly rather than profiling normal instances, achieving linear time complexity and minimal memory usage, which is ideal for large and high-dimensional data sets.
Despite its strengths, isolation forests have weaknesses due to random selection of variables during data splitting. A systematic review of the literature~\cite{al2021isolation} has addressed these weaknesses and offered solutions, highlighting the lack of comprehensive reviews of existing Isolation Forest improvements.

In recent years, deep learning solutions have been proposed for the problem of supervised anomaly detection on logs.

LogRobust~\cite{zhang2019robust} addresses the limitations of traditional log-based anomaly detection methods by tackling the instability of log data, which often includes previously unseen log events or sequences due to the evolution of logging statements and processing noise. Unlike existing methods that assume a stable and known set of log events, LogRobust uses an attention-based Bi-LSTM model to capture contextual information in log sequences and automatically learn the importance of different log events. This approach involves extracting semantic information from log events and representing them as semantic vectors, enabling the model to identify and handle unstable log events and sequences effectively. Evaluated using logs from the Hadoop system and an online service system of Microsoft, LogRobust demonstrates accurate and robust performance on real-world, ever-changing log data.

This idea is further developed with SwissLog~\cite{li2020swisslog}, which explicitly leverages semantic and temporal information to handle diverse faults. 
Thereby SwissLog introduces a robust deep learning-based model integrating semantic and temporal embeddings. It employs an advanced log parser and a unified attention-based BiLSTM architecture to detect anomalies caused by changes in log sequence order and time intervals. Experimental results on both real-world and synthetic datasets confirm SwissLog's effectiveness in adapting to evolving log data and detecting diverse faults.

Guo et al. introduce LogBERT~\cite{guo2021logbert}, a self-supervised framework for log anomaly detection based on the \ac{BERT}~\cite{devlin2018bert}. LogBERT leverages the transformer network, which consists of an encoder and decoder incorporating the attention mechanism. Similar to DeepLog, LogBERT predicts a targeted log template in a sequence by utilizing temporal-related log embeddings around the prediction target as inputs. The model is trained with two self-supervised tasks: masked log message prediction and volume of hypersphere minimization. The prediction of the targeted log template employs Cross-Entropy Loss, extended by a hyperspherical loss function to ensure the compactness of the embeddings.

LogBD~\cite{liu2023logbd} employs pretrained models and domain adaptation to enhance log anomaly detection. Using the BERT model, it learns the semantic information of logs and addresses the issue of log instability due to word meanings and log statement updates. LogBD employs Temporal Convolutional Networks \acfp{TCN} to extract common features from different system logs, mapping them to the same hypersphere space to detect anomalies. Experiments on publicly available datasets indicate that LogBD can effectively address log instability and improve cross-system log anomaly detection.

\section{Log Root Cause Analysis}
\label{sec:related_work:root_cause_analysis}
Several log anomaly detection methods were published to identify abnormal log lines.
They focus on marking individual log lines as anomalous but do not take into account the context and causal relationships between logs from multiple services into account, which leads to \ac{RCA}.

Root cause analysis (RCA) is a systematic process that is used to identify the underlying reasons for faults, failures, or problems within a system. 
By pinpointing the root causes rather than simply addressing the symptoms, \ac{RCA} aims to prevent the recurrence of problems and improve the overall reliability and performance of IT systems and services. 
Within log data analysis, the root cause is a set of logs that describe the reason for an occurred failure.
Although the field of log data root cause analysis lacks extensive research, we introduce some methods.

In LogRule~\cite{notaro2023logrule}, the authors leverage structured logs and association rule mining (ARM) to automate RCA. The LogRule algorithm analyzes structured logs to generate a comprehensive list of explanations for specific events, achieving a 0.921 F1-score for diagnosis tasks. Additionally, LogRule computes results 37 times faster than the current state-of-the-art solution, making it a time-efficient, accurate, and interpretable ARM-based RCA algorithm. The evaluation results demonstrate that LogRule effectively enables RCA in complex multidimensional datasets, overcoming the prohibitive execution times of existing algorithms.

The authors of Log-based abnormal task detection and root cause analysis for spark~\cite{lu2017log} propose an offline approach that leverages Spark log files to accurately detect abnormalities and analyze their root causes. The approach consists of four steps: Preprocessing the raw log files to generate structured log data, selecting relevant features related, detecting where and when abnormalities occur based on these features, and analyzing the problems by assigning weighted factors to determine the probability of root causes. The study considers four system resources: CPU, memory, network, and disk, as potential causes of failures. 

The proposed framework in this work~\cite{zawawy2010log} filters logs according to specific analysis goals and diagnostic hypotheses set by the user or an automated process to facilitate root cause analysis. It uses annotated goal trees to model the constraints and conditions under which the system functions. These constraints and conditions are transformed into queries that are applied to a relational database storing the logged data or use Latent Semantic Indexing to identify the most relevant log entries. The results of these queries yield a subset of logged data compliant with the goal tree, which is then used by a diagnostic SAT-solver based algorithm. Experimental results demonstrate that this filtering process effectively reduces the time and complexity of diagnosing issues in multi-tier heterogeneous service-oriented systems, thereby enhancing the efficiency of root cause analysis.

LADRA~\cite{lu2019ladra} is a tool for detecting abnormal data analytic tasks and performing root cause analysis using Spark logs. LADRA employs a log parser to convert raw log files into structured data and extract relevant features. The detection method is then introduced to identify the time and location of the occurrence of abnormal tasks. Furthermore, predefined factors based on these features are extracted to facilitate root cause analysis. General Regression Neural Network (GRNN) is employed to determine the likelihood of reported root causes, weighted by the factors. LADRA is an offline tool that accurately analyzes abnormalities without additional monitoring overhead. It considers four potential root causes: CPU, memory, network, and disk I/O. Evaluated using three Spark benchmarks where root causes were artificially introduced, experimental findings demonstrate that LADRA's methodology is efficient in accurately identifying root causes.

The issue with the discussed methods is that they are capable of identifying known root causes: therefore, the DevOps teams are aware of the source of the failure. 
However, in complex IT systems, there may also be root causes that have not been previously identified.

\section{PU Learning}
\label{sec:related_work:pu_learning}
The last section of our related work is dedicated to PU learning, since this learning strategy is one of our main concepts in some of our methods.
PU-learning is a semi-supervised training technique that builds a binary classifier based only on positive and unlabeled examples, hence the origin of its name~\cite{bekker2020learning,liu2003building}.

Numerous PU learning techniques have been applied in diverse fields~\cite{liu2002partially,bekker2020learning,letouzey2000learning}. However, our focus here is on methods applicable to log analysis.

The authors in~\cite{liu2002partially} utilize the Expectation Maximization (EM) algorithm together with the Naive Bayes classification to investigate the problem of partially supervised classification, where the goal is to identify documents of a specific class within a larger set of mixed documents, without having labeled non-class documents. Traditional machine learning techniques are inapplicable due to the lack of labeled documents from both classes. To address this, the authors pose the problem as a constrained optimization problem and introduce a novel technique involving the Expectation-Maximization (EM) algorithm combined with naive Bayesian classification. The EM algorithm produces a sequence of Naive Bayesian classifiers, from which one is selected using a heuristic. This method is demonstrated to be effective through extensive experimentation.

A more conservative variant of PU learning is proposed in~\cite{fusilier2015detecting}. This method improves upon the original PU-learning approach by iteratively pruning the set of reliable negative instances using a binary classifier. In the context of detecting both positive and negative deceptive opinions—i.e., fictitious reviews promoting low-quality products or criticizing good ones—the proposed method selectively refines negative examples from the unlabeled set. This more conservative approach results in higher quality negative instances, which leads to significantly better prediction results. 

An ensemble learning method for PU learning is proposed in~\cite{mordelet2014bagging}. The authors motivate bagging SVM, i.e. the aggregation of multiple SVM classifiers in order to answer sources of instability often encountered in PU learning situations. 
This method addresses the challenges associated with PU learning—where only positive and unlabeled examples are available, and negative examples are absent—by iteratively training multiple binary classifiers to differentiate known positive examples from random subsamples of the unlabeled set. The predictions of these classifiers are then averaged to improve overall accuracy.
They find their approach to be comparable to existing works, while being often faster to train, especially for $|\mathcal{P}| \ll |\mathcal{U}|$.

In~\cite{WuCWWZW20}, the authors introduce a hybrid semi-supervised learning model, hPSD, for spammer detection. Unlike traditional methods, hPSD uses both user characteristics and user-product relations to detect deceptive review spammers. 
Thereby, they extract a set of reliable negative instances via a feature strength function using a greedy heuristic on sorted features until $|\mathcal{P}|\simeq|RN|$.
By iteratively injecting various positive samples and constructing classifiers in a semi-supervised framework, hPSD achieves superior performance over baseline methods. Experiments on movie datasets and real-life Amazon data confirm its effectiveness in identifying hidden spammers and their employers. 

In another work~\cite{LuoCLJ18}, the authors propose a web anomaly detection method that combines supervised learning with PU learning, specifically targeting HTTP payload data. Their approach involves training a base supervised XG-Boost model to detect known attack traffic. Subsequently, normal traffic is analyzed using a PU learning classifier to identify unknown malicious traffic.

\cleardoublepage
\chapter{Multi-Layered Architecture for Log Analysis}
\minitoc
\label{ch:log_analysis_layer}

Log analysis for AIOps is a huge research area~\cite{oliner2012advances,oliner2007datasets}. 
The industry's expectation was complete automation~\cite{andenmatten2019aiops}, to become more efficient and profitable. 
It became obvious that single methods were too system-specific to be generalized, and it became apparent that log analysis applications were too specialized for complete automation. 
Consequently, to address these issues, we present a layered architecture for log analysis in this chapter.

In the intricate ecosystem of software development and system management, understanding how an IT system or service functions in real-world operation is essential. 
Therefore, right from the start of a software project, log statements are embedded in the source code to record specific events of the software during runtime, establishing a path of insights that reveal how the system operates during execution. 
The generated logs, thoroughly gathered in files, act as critical assets for analysis after deployment, providing information on internal functions of the system and interactions with users.
Consequently, the log lines contain status messages or current values of variables and additional information, such as a timestamp or service IDs~\cite{gulenko2016system}.
However, obtaining practical insights from logs is a complex task. 
It involves traversing a labyrinth of log files, deciphering patterns, and extracting meaningful information. 
This process, essential for troubleshooting and improving system stability in subsequent iterations, demands substantial time and effort from DevOps teams.
Hence, our attention is directed exclusively towards recorded log data.

Therefore, we have developed methods that help DevOps teams gain important insights into the execution of the IT system or service, by analyzing the produced log data, to subsequently develop or correctly apply effective anomaly detection and root cause analysis methods to mitigate failures in subsequent updates.
To achieve this, we have divided the log analysis into three layers.
Initially, \textit{Log Investigation} serves as an initial layer that encompasses the processes of extracting insights from log data anomalies and, furthermore, preparing log data for model training and evaluation. 
In the second layer, \textit{Anomaly Detection}, we utilize the insights obtained from the first layer to implement the appropriate anomaly detection method, ensuring it is properly trained to detect system behaviors that deviate from the normal behavior of the system.
Third, we address the identification of the origin of failures. 
Since failures propagate through the system, pinpointing the exact log lines that show the propagation and therefore reveal the root cause is termed \textit{Root Cause Analysis}.


\section{Architecture Overview}
\label{sec:log_analysis_layer:architecture}


In this section, we provide the foundation for the methods that we will present in detail in the following chapters.
In this way, we provide detailed definitions for each task that must be accomplished by the methods in each layer and reveal the required input data for each layer and the advantageous results for the subsequent layer. 
In addition, we present an overview of how our methods can assist DevOps teams in the log analysis lifecycle for IT operations.

\begin{figure}[htbp]
\centering
\includegraphics[width=1.0\columnwidth]{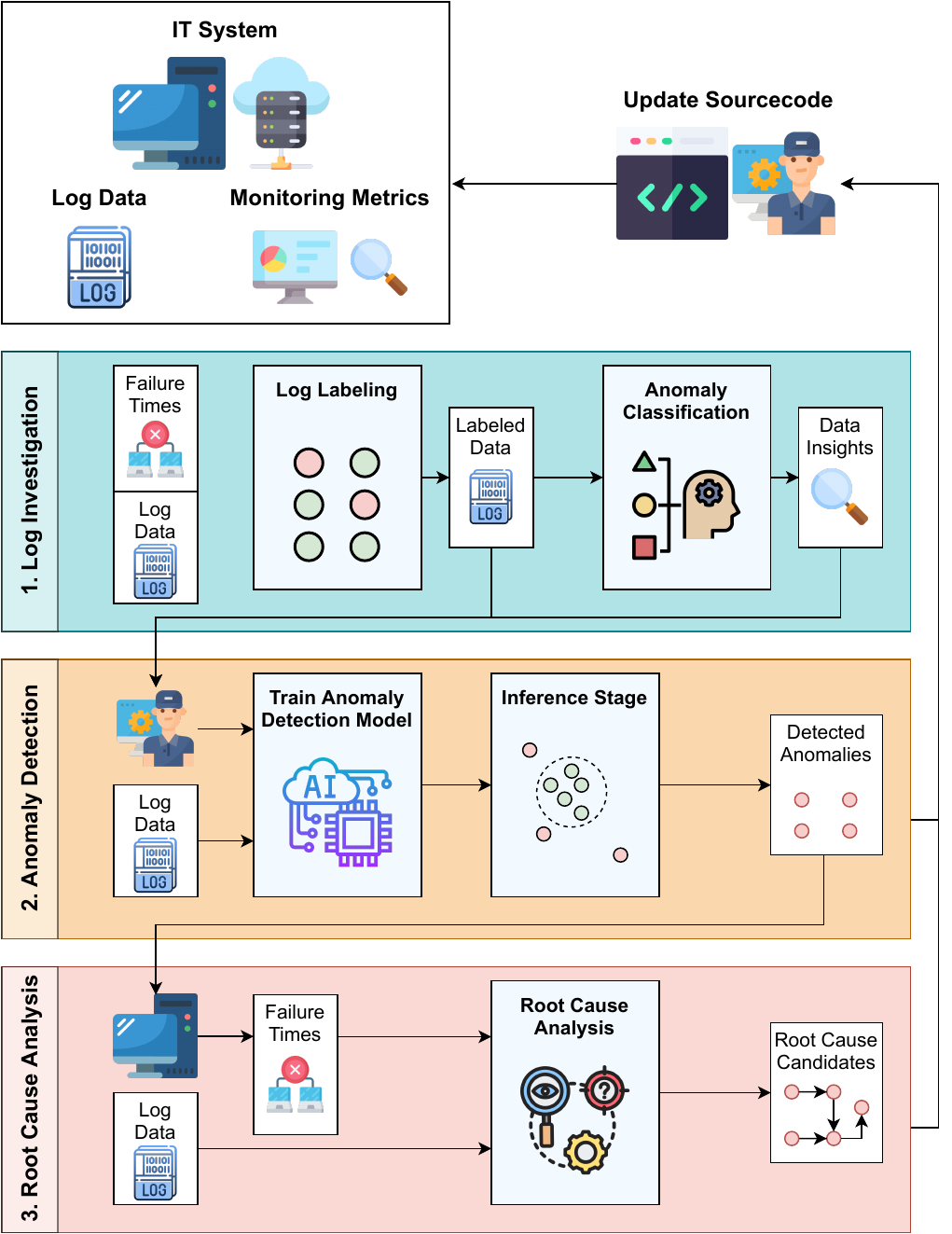}
\caption{Schematic representation of our layered architecture, illustrating its three layers: log investigation, anomaly detection, and root cause analysis}
\label{fig:log_analysis_layer:architecture:holistic_view}
\end{figure}

\autoref{fig:log_analysis_layer:architecture:holistic_view} illustrates the three layers of our architecture. 
In the upper left, there is an IT system or service that produces log data during its operation. 
The system metrics are also monitored, which helps to identify any operational problems within the system. 
However, we cannot determine where and what exactly is going wrong in the software, as the metrics are collected at the hardware level.

The initial layer is the \acf{LI}, which consists of anomaly labeling and anomaly classification. 
If there are no labeled logs, they must be labeled either by experts or using the appropriate techniques.
To label data, our method requires approximate failure times and logs from the IT system as initial input. 
Estimated failure times are derived from the monitoring of metric data. 
Using these approximate times, it can identify the precise log lines that contain anomalies and label them, thus creating a labeled data set.
The labeled data can then be used for further investigation of the log data or to evaluate trained models. 
Following the creation of a labeled data set, a detailed classification of each labeled anomaly is performed during the anomaly classification process. 
Here, anomalies are classified into various types.
This gives the AIOps team precise knowledge about the nature of the anomalies that occur primarily in the system.
Therefore, they can then apply or develop customized methods for detecting anomalies.

The next layer involves an \acf{AD} method, which consists of model training and inference. 
Insights from the initial layer are utilized to develop or choose an appropriate model, depending on the dominant type of anomalies.
The available log data determines the training strategy for the model, whether it is trained supervised, unsupervised, or weakly supervised. 
After obtaining a trained model, it is crucial to accurately interpret the results of the model during inference and to establish an appropriate decision boundary. 
This boundary is critical for defining the model's accuracy and managing the rate of false positives or negatives to achieve the desired outcomes for the application.

The final layer involves a \acf{RCA}. 
Considering the various anomalies that occur during operation and are detected through anomaly detection, it is essential to pinpoint and separate those that are directly related to the failures for precise \ac{RCA}. 
To achieve this, approximate failure times are required. 
These times can be sourced either from the monitoring system, derived from the anomaly frequency distribution, or other sources.
After performing \ac{RCA}, we identify potential root cause candidates. 
It is crucial to accurately identify these potential root cause candidates to support the DevOps team in comprehensively understanding the nature and origin of the failure.
Using these methods, the DevOps team can accurately identify and analyze system failures, allowing them to refine the software in future revisions and increase the robustness of the IT system or service.

It is evident that the layers are interconnected and dependent on one another. 
Hence, log investigation serves as foundational work for anomaly detection, which in turn acts as groundwork for root cause analysis.
Nevertheless, each method can be applied independently when the prerequisites are met and only its specific outcomes are required.

\subsection{Log Investigation}
\label{subsec:log_analysis_layer:log_investigation}
Log investigation is the first and essential layer of our log analysis architecture. 
Within our log analysis architecture, it serves as a crucial foundation for subsequent layers. 
Thus, we define log investigation as follows:

\begin{definition}[Log Investigation]
\label{def:log_investigation}
Log investigation involves systematic exploration and analysis to extract insights, patterns, and anomalies that contribute to the development of effective algorithms. This process allows human analysts to gain a deeper understanding of the log data, identify relevant features, and refine algorithmic models for improved anomaly detection or root cause analysis.
\end{definition}

Although there are additional areas of log investigation, this thesis focuses on specific aspects, particularly emphasizing the processes of labeling and subsequent classification of log data anomalies. 
Details of the methods for both areas are provided in~\autoref{ch:data}.  

The initial step is log labeling and serves multiple purposes: it is instrumental in training new supervised anomaly detection models by providing them with accurately labeled data. 
In addition, it is used to evaluate the performance and accuracy of existing models to ensure that they function as intended. 
Moreover, log labeling facilitates deeper investigations such as anomaly classification.

We define log labeling for this thesis to ensure a common understanding.

\begin{definition}[Log Labeling]
\label{def:log_labeling}
Log labeling is the assignment of a binary label to every single log line within a data set, designated as either \textbf{normal} or \textbf{anomalous}.

Let $\mathcal{L} = \{l_1, l_2, \ldots, l_n\}$ be a log data set, consisting of a sequence of log lines, where each $l_i$ is an individual log line. 
The process of log labeling involves assigning a label $y_i$ to each log line $l_i$, where $y_i \in \{0, 1\}$.

\begin{itemize}
    \item $y_i = 0$ indicates that $L_i$ is a normal log line.
    \item $y_i = 1$ indicates that $L_i$ is an anomalous log line.
\end{itemize}

Formally, the log labeling function can be defined as:
$f: \mathcal{L} \rightarrow \{0, 1\}$
where $f(l_i) = y_i$ assigns a binary label to each log line $l_i$.
\end{definition}

The primary issue is that expert data labeling is expensive and time-consuming, prompting numerous companies to seek automated or semi-automated solutions. 
Consequently, we introduce a method for autonomous log labeling in~\autoref{sec:data:autonomous_data_labeling}.

The subsequent step is anomaly classification, which is the process of categorizing anomalies. 
Thus, we utilize the log message and the contextual log messages to classify each anomaly.
Thereby, we define anomaly classification for this thesis to ensure a common understanding of this process.

\begin{definition}[Log Anomaly Classification]
\label{def:anomaly_classification}
Let $\mathcal{A} = \{a_1, a_2, \ldots, a_k\}$ be the set of predefined anomaly types, and let $\mathcal{L} = \{l_1, l_2, \ldots, l_n\}$ be a sequence of log lines, where each $l_i$ is an individual log line. 
Anomaly classification involves assigning an anomaly type $a_j \in \mathcal{A}$ to each abnormal log line $l_i \in \mathcal{L}_{\text{abnormal}}$, resulting in a classified data set $\{(l_i, a_i) \mid a_i \in \mathcal{A} \wedge l_i \in \mathcal{L}_{\text{abnormal}}\}$.

The classification function can be defined as:
$g: \mathcal{L}_{\text{abnormal}} \rightarrow \mathcal{A}$
where $\mathcal{L}_{\text{abnormal}} \subseteq \mathcal{L}$ represents the subset of log lines labeled abnormal.
\end{definition}

Classifying anomalies is crucial for a deeper understanding of the data and making informed decisions. 
For example, it is necessary to determine whether to detect contextual anomalies, point anomalies, or a combination of both.
Whether an anomaly detection model is needed for multiple anomaly types or just one, and whether to develop a new model or use an existing one are key questions. 
Anomaly classification aids in addressing these issues and provides guidance to select a suitable model for anomaly detection. 
Thus, we develop a classification method for different anomaly types and set formulas to compute these types in~\autoref{sec:data:classifying_anomalies}.

\subsection{Anomaly Detection}
\label{subsec:log_analysis_layer:anomaly_detection}
The detection of anomalies within log data is a critical aspect in supporting maintenance in IT systems and services and ensuring regulatory compliance~\cite{reuben2016automated}. 
It provides insights into system performance, assists in faulty execution during runtime, and enables proactive maintenance.

We have segmented log anomaly detection into model training and inference stage, which we present in detail in~\autoref{ch:anomaly_detection}. 
The objective of the model training is to accurately assign values to each log line, so that the decision boundary can divide the output into two classes.
Thereby we define log anomaly detection as follows:

\begin{definition}[Log Anomaly Detection Model]
\label{def:anomaly_detection}
A Log Anomaly Detection Model is designed to identify unusual or unexpected patterns within log data.

Formally, let $\mathcal{L} = \{l_1, l_2, \ldots, l_n\}$ be a sequence of log lines. 
Each log line $l_i$ represents a normal or abnormal log line: $y_i \in \{0, 1\}$, where $y_i = 0$ indicates a normal log line and $y_i = 1$ indicates an anomalous log line.
These 

The goal of the model $\Phi(l_i,\Theta)$ is to train its parameters $\Theta$ to output a numeric representation $\Phi(l_i) \in \mathbb{R}^k$ of the input log line $l_i$, so that a decision boundary function $b$ (see definition \ref{def:decision_boundary}) is capable of interpreting the output and assign a label of 0 or 1. 

Ideally, the subsequent equations should hold true:
$\{l_i \in \mathcal{L} \mid y_i = 1\}  = \{l_i \in \mathcal{L} \mid b(\Phi(l_i)) = 1\}$ 
and
$\{l_i \in \mathcal{L} \mid y_i = 0\} = \{l_i \in \mathcal{L} \mid b(\Phi(l_i)) = 0\}$
\end{definition}

For model training, various types of data can be utilized, including unlabeled, partially labeled, or labeled data, leading to different training strategies such as unsupervised, weakly supervised, or supervised (see~\autoref{sec:background:learning_strategies}).  

In the inference stage, the trained model evaluates the incoming log data. 
This process involves applying the learned patterns and statistical characteristics of the trained data to the new data to detect deviations that may indicate anomalies.

A crucial component of the inference stage is the decision boundary function, which interprets the output of the trained anomaly detection model and determines whether a given log message is classified as normal or abnormal. 
If the trained model is a neural network, the output is a neuron or a complete layer, which is a single scalar value or a vector.
The task of the decision boundary is to translate this output into a binary representation.
We define the decision boundary as follows:

\begin{definition}[Decision Boundary]
\label{def:decision_boundary}
Let $\Phi: \mathcal{L} \rightarrow \mathbb{R}^k$ be the trained anomaly detection model, where $\mathcal{L}$ is the set of log lines, and $\mathbb{R}^k$ is the output space of the model. 
The decision boundary function $b: \mathbb{R}^k \rightarrow \{0, 1\}$ translates the output of the model $\Phi$ into a binary representation.

\begin{equation}
b(\vec{x}) = 
\begin{cases} 
1 & \text{if } b_t(\vec{x}) \geq \tau \\
0 & \text{if } b_t(\vec{x}) < \tau 
\end{cases}
\end{equation}
where $\vec{x} \in \mathbb{R}^k$ is the output of the model, $b_t: \mathbb{R}^k \rightarrow \mathbb{R}$ is a function that maps the output vector to a scalar value, and $\tau$ is the threshold.
\end{definition}

Therefore, the decision boundary is crucial for the accuracy of the anomaly detection process. Its calculation is explained in detail in~\autoref{sec:anomaly_detection:inference}.
There are various techniques for defining the decision boundary function, depending on the type of anomaly detection model used. 
For example, in a simple threshold-based approach, the decision boundary function can compare the output score or probability assigned to each log line by the model to a predefined threshold. 
Log lines with scores exceeding the threshold are classified as anomalies, while those below are considered normal.

\subsection{Root Cause Analysis}
\label{subsec:log_analysis_layer:root_cause_analysis}
The final layer of our architecture for log analysis is \acf{RCA}.
Having successfully identified the anomalies in the second layer, the focus now shifts to isolating specific anomalies that assist the DevOps team in diagnosing the failure.
The task is to analyze the log anomalies and find these that pinpoint the primary cause of the failure, along with its cascading impacts during the execution of the software. 
Therefore, \ac{RCA} on log data is a essential procedure in the field of \ac{AIOps} that we define as follows:

\begin{definition}[Root Cause Analysis]
\label{def:root_cause_analysis}
Formally, let $\mathcal{L} = \{l_1, l_2, \ldots, l_n\}$ be a sequence of log lines. 
Let $\mathcal{L}_{\text{anomalous}} = \{l_i \in \mathcal{L} \mid b(\Phi(l_i)) = 1\}$ be the set of log lines identified as anomalous by the Log Anomaly Detection Model, where $\Phi$ is the trained model and $b$ is the decision boundary function.

The goal of RCA is to determine the root cause log lines 
$\mathcal{L}_{\text{root}} = \arg\max\mathcal{S} \subseteq \mathcal{L}_{\text{anomalous}} \sum_{l_i \in \mathcal{S}} \mathrm{Impact}(l_i)$
where $\mathrm{Impact}(l_i)$ is a measure of the significance of the log line $l_i$ in contributing to the system failure, and the maximization is over all subsets $\mathcal{S}$ of $\mathcal{L}_{\text{anomalous}}$ such that $\mathcal{L}_{\text{root}}$ contains the abnormal log lines that describe the primary cause of the observed system failure.
\end{definition}

The DevOps team needs to analyze the failure and its root cause to remedy it in the upcoming software update. 
Given the varied interpretations of log lines among individuals of DevOps teams, our method, which we explain in detail in~\autoref{ch:root_cause_analysis} seeks to identify the most accurate root cause candidates that effectively explain the failure.
Consequently, this enables the team to address these failures in future updates.

\subsubsection{Summary}
In this section, we discussed the interplay of our three-layered architecture for log analysis.
We explained how the DevOps team can derive insights from their log data and the anomalies that are in the log data.
Subsequently, we illustrated the process of creating or choosing a suitable anomaly detection model for IT systems or services to ensure reliable anomaly detection. 
In the final layer, we provided guidance on pinpointing the root causes of failures among numerous anomalies.

Although we did not explore the specific methods applicable in these layers, we highlighted a broad overview, defined each part of our architecture, and pointed to later chapters for more comprehensive details.
The details of the methods of the Log Investigation Layer are given in~\autoref{ch:data}, for the Anomaly Detection Layer in~\autoref{ch:anomaly_detection}, and for the Root Cause Analysis Layer in~\autoref{ch:root_cause_analysis}.
The evaluation of all methods is presented in~\autoref{ch:evaluation}.

\cleardoublepage

\chapter{Investigating Data: Empowering Log Analysis}
\minitoc
\label{ch:data}

In log analysis, automation plays a crucial role due to the vast size of log files and the continuous generation of new logs, making it impossible for humans to analyze them without the assistance of machines.
To minimize human effort, two methods are presented in this chapter. 
The first method is used for automatic labeling of log data, since especially in the domain of log analysis, the lack of labeled data remains a major obstacle~\cite{wittkopp2020decentralized,wen2020time}.
The objectives of autonomous log labeling are to improve the evaluation of methods or to train supervised methods through the use of labeled data. 
Furthermore, we present an additional method to go deeper into the identified anomalies.
This second method is used to classify anomalies in log data. 
Anomalies require careful categorization to allow efficient detection.
Having an understanding of the type of anomaly allows the development of more effective algorithms for its detection.
The objective of anomaly classification is to collect enough details about anomalies to determine the appropriate \ac{AD} method to employ or create.

Data labeling and anomaly classification are performed predominantly by humans, resulting in substantial effort and costs for companies due to the need for specialized skills and the corresponding training for the individual~\cite{sabharwal2022hands}.
Consequently, minimizing manual labor in these domains is crucial, as they build the foundation for effective detection of anomalies and root cause analysis~\cite{rijal2022aiops}.


This chapter is based on the following publications:
\begin{itemize}
    \item LogLAB: attention-based labeling of log data anomalies via weak supervision~\cite{wittkopp2021loglab}
    \item A taxonomy of anomalies in log data~\cite{wittkopp2021taxonomy}   
\end{itemize}


We first present our autonomous log labeling method in~\autoref{sec:data:autonomous_data_labeling} to then present our taxonomy with which we can classify anomalies in~\autoref{sec:data:classifying_anomalies}.
In addition, we provide a detailed explanation of how both methods interact with each other.

\section{Autonomous Data Labeling}
\label{sec:data:autonomous_data_labeling}

In the landscape of machine learning and data analysis, the scarcity of labeled data presents a significant challenge. 
Especially in the domain of log analysis, the lack of labeled data remains a major obstacle~\cite{wittkopp2021loglab,wen2020time}.
Labeling data is costly and time-consuming, as experts need to analyze every single log message and investigate which messages reflect their corresponding errors.
Hence, there are only a very limited number of openly accessible labeled data sets for log anomaly detection.
Moreover, businesses are hesitant to share their log data due to the potential exposure of confidential details regarding their systems, as well as information about customers or employees who engage with the systems.

Although log lines are assigned severity levels such as Debug, Info, Warning, and Error, these levels do not necessarily reflect whether a log line is normal or abnormal.
The final decision for each log line depends on the context in which it occurs~\cite{oliner2007datasets,bogatinovski2022qulog}. 
For example, an error that occurs during routine system maintenance may be less critical than the same error during peak business hours.
Furthermore, an investigation of a large number of git repositories revealed that a certain number of severity levels are incorrectly assigned~\cite{bogatinovski2022qulog}.
This demonstrates the importance of accurately labeling each individual log line according to its content.

Therefore, this section introduces methods for autonomous data labeling. 
Since labeled data serve as the foundation for training supervised models and evaluating existing models, recognizing this obstacle, the concept of autonomous data labeling has gained prominence, using automation to streamline the labeling process and make it more scalable and cost-efficient.
Supervised models that train on large volumes of normal and abnormal data show significant performance in log anomaly detection~\cite{zhang2019robust,yang2021semi}. 
Large volumes of labeled training data are especially important for deep learning methods and are therefore a strong accelerator for log anomaly detection \cite{ratner2016data}.

To address this problem, we propose an iterative and attention-based model for binary labeling of anomalies in log data.
It relies only on rough estimates of when an error has occurred, information that can often be derived from other monitoring systems~\cite{sukhwani2017monitoring}.
We utilize weak supervision to identify actual abnormal log messages in the provided estimates and assign labels correspondingly.
Furthermore, we apply an iterative training strategy that identify biases from a previous model and train a new model without those biases. 
With that strategy, we are able to improve the autonomous labeling.

The remainder of this section is structured as follows: \autoref{subsec:data:autonomous_data_labeling:problem_description} provides a description of the problem and provides general information. 
\autoref{subsec:data:autonomous_data_labeling:learning_from_pu} explains our approach in detail and gives a formal definition of learning from positive and unknown samples. 
\autoref{subsec:data:autonomous_data_labeling:log_processing_pipeline} demonstrates the log processing pipeline, detailing its steps and significance in data preprocessing. 
\autoref{subsec:data:autonomous_data_labeling:model_architecture} presents the architecture of our autonomous labeling model, discussing its design and integration of our method. 
\autoref{subsec:data:autonomous_data_labeling:objective_function} elaborates on the objective function used to optimize the precision of the labeling, discussing the selection criteria. 
\autoref{subsec:data:autonomous_data_labeling:iterative_training_strategy} outlines our iterative training strategy to improve model performance over time, describing its benefits and techniques.

\subsection{Problem Description}
\label{subsec:data:autonomous_data_labeling:problem_description}

To automatically label log data, it is essential to have details which log lines are considered as anomalies and which ones are normal. 
To achieve this, we depend on the monitoring system, which gives us estimated error timestamps of the IT system or service.

Modern monitoring systems provide alerts indicating issues, but they are unable to pinpoint a specific log line.
Modern monitoring solutions are equipped with advanced algorithms and technologies designed to detect and notify administrators when a system experiences deviations from expected behavior. 
These monitoring solutions carefully examine a wide range of system metrics, including both hardware and software components, keeping track of parameters such as CPU usage, memory consumption, disk I/O operations, network traffic, and application performance~\cite{sukhwani2017monitoring}.

However, we assume that failure times are roughly known through the monitoring solution.
We use this information in retrospect to label abnormal log lines in an automated way.
Therefore, we use the failure time windows of the monitoring system as an auxiliary source to train our approach without previously known labels.

\begin{figure}[htbp]
\centering
\includegraphics[width=1.0\columnwidth]{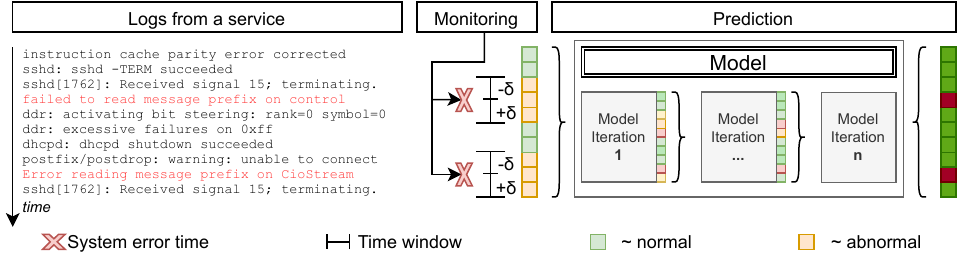}
\caption{We use rough estimates for failure times provided by monitoring systems in order to identify actually abnormal log messages via weak supervision.}
\label{fig:data:autonomous_data_labeling:problem_draw}
\end{figure}

\autoref{fig:data:autonomous_data_labeling:problem_draw} exemplifies the problem described.
It displays the log of a system with two abnormal log events colored red for illustration purposes.
As described above, we utilize monitoring information to estimate time windows of the length $2*\delta$ in which we suspect abnormal log events to be present. 
Thus, the task of the model is to determine the error of the preliminary labels to then assign the correct labels to the individual log lines.
By utilizing the knowledge from the time windows, two classes can be approximated. 
One class with presumably normal data and another class with unknown data that contains both abnormal and normal. 

All log messages that occur during the time windows represent the unknown class, whereas all other log messages represent the normal class. 
Since the log messages of the unknown class occur in a timely manner and are correlated with a failure, they are initial labeled abnormal and are colored orange in \autoref{fig:data:autonomous_data_labeling:problem_draw} in the monitoring phase. 
The other log messages are preliminary labeled normal and colored green. 
It becomes transparent that the preliminary labels derived from the assignment of the class do not have to correspond to the underlying ground truth, which is represented by the red colored log messages.

The goal of the method is to label these log messages as abnormal and to label all others as normal, regardless of whether they are included in the respective time windows.
To maximize precision, we train a model over several epochs and then use the output to train a new model. 
By incorporating the output predictions from the previous model as part of the training input for the next model iteration, the iterative approach enriches the training dataset, effectively guiding the subsequent model towards higher precision. 
This iterative refinement mechanism not only improves the quality of the training input but also mitigates the potential biases introduced by inaccurate preliminary labeled log messages from the time span of $2\delta$ or the output of previous iterations. 
As the training process iterates through multiple cycles, each successive model benefits from increasingly precise training input, thereby fostering continuous improvement in the accuracy and reliability of log message labeling.

Therefore, the monitoring system does not need to be able to precisely identify the actual log message, as a broad localization is sufficient. 
Through accurately pinning down real anomalous log messages with our approach, we effectively support reliability engineers.

\subsection{Learning from Positive and Unknown}
\label{subsec:data:autonomous_data_labeling:learning_from_pu}
As described above, we utilize monitoring information to estimate time windows of length $2*\delta$ in which we suspect abnormal log lines. 

We describe our log labeling problem as a weak supervision learning problem with inaccurate labels for a binary classification, since we cannot assign accurate labels due to imprecision of the failure time windows.
Therefore, we assign preliminary abnormal labels for all log lines in the estimated time windows and preliminary normal labels for all other log lines. 
Weak supervision with inaccurate labels is defined as a situation in which the supervision information does not always match the ground truth~\cite{zhou2018brief}.

It becomes transparent that the preliminary labels derived from the assignment do not have to correspond to the underlying ground truth. 
In this case, we use PU learning, one of our core concepts that we introduced in ~\autoref{subsec:background:concepts:pu_learning}, including a formal definition.

In particular, the underlying log data is divided into two classes, positive $\mathcal{P}$ and unknown $\mathcal{U}$, where $\mathcal{U}$ consists of all log lines that occur in the aforementioned time windows and $\mathcal{P}$ of the rest. 
The unknown class $\mathcal{U}$ includes all log lines that require labeling to complete the labeled data set.
This class contains both normal and abnormal log lines.
To train our network, we assign all log lines that are in $\mathcal{U}$ the preliminary label $1$, abnormal.
Whereas $\mathcal{P}$ contains presumably normal log lines, the label normal $1$ is assigned.
Hence, the autonomous labeling method must be able to learn a good representation of the normal log massages even if the same normal log lines appear in the unknown class with wrong labels to still label all normal log lines as normal and abnormal correspondingly.

\subsection{Log Processing Pipeline}
\label{subsec:data:autonomous_data_labeling:log_processing_pipeline}

After outlining and defining the problem and the concept with which we want to solve this, we present the individual steps of our method and identify which steps are included in the iterative training strategy.

For the autonomous labeling of log data, we design a processing and modeling pipeline illustrated in ~\autoref{fig:data:autonomous_data_labeling:high_level_pipeline}.
Individual steps are as follows:

\begin{figure}[htbp]
\centering
\includegraphics[width=0.8\columnwidth]{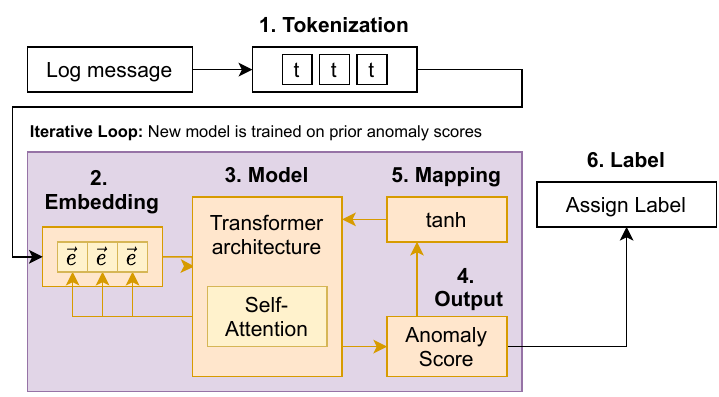}
\caption{6 steps to obtain labels for single log lines}
\label{fig:data:autonomous_data_labeling:high_level_pipeline}
\end{figure}

In step 1, we convert the content $c_i$ of each log line $l_i$ into a sequence of tokens $t_i$ using the splitting symbols \texttt{.,:/} and whitespaces as separators. 
This process is called tokenization and explained in detail in~\autoref{subsec:background:log_data:processing}.
Subsequently, we further clean the resulting sequence of tokens by replacing certain tokens with placeholders that adequately represent the original token without losing relevant information. 
We introduce placeholder tokens for hexadecimal values \texttt{'[HEX]'} or any number greater than or equal to 10 \texttt{'[NUM]'}, which are typical values in log data. 
Finally, we prefix the sequence of transformed tokens with a special placeholder token \texttt{'[CLS]'} which will be beneficial later on, as explained in~\autoref{subsec:background:concepts:transformer}. 
An exemplary log message
\begin{center}
    \texttt{time.c: Detected 3591.142 MHz.}
\end{center}
is thus transformed into a sequence of tokens
\begin{center}
    \texttt{['[CLS]', 'time', 'c', 'Detected', '[NUM]', '[NUM]', 'MHz']}.
\end{center}

Since these sequences can vary in length, we truncate them to a fixed size $s$ and fill up smaller sequences with padding tokens \texttt{'[PAD]'}.

For each token $w_j$ in the token sequence $t_i$, an embedding vector $\vec{e}_{i}(j)$ is obtained in step 2.
The truncated sequences of the embeddings $\vec{e_i}$ serve as input for the model in step 3.

The utilized model is indirectly responsible for training the aforementioned embeddings via backpropagation of prediction errors, thereby the embeddings will be optimized for preserving information.
The model computes an output embedding for each truncated input embedding sequence $\vec{e_i}$, which summarizes the log message by utilizing the embeddings of all tokens. 
The information in this output embedding is contained within the embedding of the \texttt{’[CLS]’} token and is adjusted through loss minimization during training.
During the training process, the model is supposed to learn the deviations of the log messages, thereby getting an intuition of what is normal and abnormal, respectively.

The output of the model $z_i$ is referred to as the anomaly score and is displayed in step 4.
Thus, the anomaly score for the label is calculated by the length of the output vector $\lVert z_i \rVert$.
Anomaly scores close to $0$ represent normal log messages, where large vectors indicate an abnormal log message. 

However, we employ an iterative training strategy in which the computed anomaly scores are converted and standardized using the $tanh$ function and then utilized as input labels for subsequent training models as shown in step 5.
Following the last iteration of the iterative training strategy, the final calculated anomaly score for each log line $l_i$ is then used to classify it with a label $\widehat{y_i}$, which can be normal or abnormal, depending on a threshold determined at step 6.


\subsection{Model Architecture}
\label{subsec:data:autonomous_data_labeling:model_architecture} 

This section shows the model architecture we use.
Therefore, we utilize one of our core concepts, the transformer encoder architecture, explained in~\autoref{subsec:background:concepts:transformer} and illustrated in~\autoref{fig:log_analysis_layer:concepts:transformer:transformer_architecture}.

This well-known architecture leverages the advantages of attention mechanisms and combines it with a straightforward architecture optimized to tackle different potential problems that are often encountered during training of neural networks.
Most importantly, the attention mechanism allows for attending over input embeddings and thus determining their overall importance, whereas the utilization of multiple attention mechanisms simultaneously (i.e. multi-head) stabilizes the learning process.
Since this architecture does not take into account the order of the input embeddings, we further enrich them with positional encoding.

We provide encoded embeddings of our processed input log lines in the neural network.
Then, all embeddings of the individual tokens are considered and the entire log message is embedded in the pre-panded \texttt{’[CLS]’} token.
The length of this embedding determines the anomaly score.
Lower scores signify normal log lines, whereas higher scores denote abnormal log lines.
As we solely use this vector representation in the subsequent steps, we therefore refer to it as $z_i$.

\subsection{Objective Function}
\label{subsec:data:autonomous_data_labeling:objective_function}

To label the log data, the transformer encoder model must be trained in such a way that it is capable to handle the problem of weak supervision with inaccurate labels for binary classification, as described in ~\autoref{subsec:data:autonomous_data_labeling:learning_from_pu}. 
To achieve this, the model needs to understand the semantics of various tokens, enabling it to differentiate between various log messages.
Thereby several log messages can be completely different but express the same state of the system and therefore have the same meaning. 
Since $\mathcal{P}$ consists only of normal log messages, the objective function must be modeled in such a way that the same meanings of log messages in class $\mathcal{P}$ and $\mathcal{U}$ have low anomaly scores. 
The semantics of the log messages that occur only in $\mathcal{U}$ are most likely abnormal and therefore must have higher anomaly scores.
Furthermore, the objective function must be able to handle large amounts of incorrectly labeled log messages, since the class $\mathcal{U}$ can increase rapidly for large $\delta$, as $\delta$ can only be estimated roughly from the monitoring systems.

Since the anomaly scores for each input sequence $\vec{e_i}$ are calculated by the length of the corresponding outcome vector $\lVert z_i\rVert$ of the model, the model must map normal log messages to small vectors, close to $0$ using the Euclidean distance.

Therefore, the objective function must consist of two parts, one part must minimize the errors of samples from class $\mathcal{P}$, from which the calculated anomaly scores should close to $0$. 
The second part of the objective must minimize the errors of the samples in class $\mathcal{U}$, pushing them away from $0$. 

The general structure of the objective function is shown in \autoref{eq:data:autonomous_data_labeling:objective_function_gen}, where $\tilde{y}_i$ is the inaccurate label, $z_i$ is the model output, and $m$ the number of samples per batch. 

\begin{equation}
    \label{eq:data:autonomous_data_labeling:objective_function_gen}
    \frac{1}{m}\sum\limits_{i=1}^{m}((1-\tilde{y}_i)*a(z_i) + (\tilde{y}_i)*b(z_i)
\end{equation}

The first part $a$ of the objective function becomes $0$ if the sample is in class $\mathcal{U}$, while the second part $b$ becomes $0$ if the sample is in class $\mathcal{P}$. 
For the first part, we choose the following \autoref{eq:data:autonomous_data_labeling:objective_function_a} to minimize the error for positive samples.

\begin{equation}
\label{eq:data:autonomous_data_labeling:objective_function_a}
    a(z_i) = \lVert z_i \rVert^2
\end{equation}

\autoref{eq:data:autonomous_data_labeling:objective_function_a} calculates the squared error of the output length. 
In contrast, we choose the following \autoref{eq:data:autonomous_data_labeling:objective_function_b} to increase the error for all anomaly scores when the log message is of class $\mathcal{U}$. 
Thus, $q$ is a numerator between 0 and 1 that represents the relation of the number of samples in $\mathcal{P}$ and $\mathcal{U}$, shown in ~\autoref{eq:data:autonomous_data_labeling:objective_function_pu_lim}.

\begin{equation}
\label{eq:data:autonomous_data_labeling:objective_function_b}
    b(z_i) = \frac{q^2}{\lVert z_i \rVert}
\end{equation}

To ensure that $q$ represents the relation of $\mathcal{P}$ and $\mathcal{U}$ and remains within the boundaries of 0 to 1, to not grow to fast, we model $q$ as a limited function, which is provided with the relation of $\mathcal{P}$ and $\mathcal{U}$.

\begin{align}
\label{eq:data:autonomous_data_labeling:objective_function_pu_lim}
    f(x) = \frac{x}{x+1}, \lim \limits_{x \to \infty} f(x) = 1, \nonumber \\
    q = f(\frac{|\mathcal{P}|}{|\mathcal{U}|}) = \frac{\frac{|\mathcal{P}|}{|\mathcal{U}|}}{(\frac{|\mathcal{P}|}{|\mathcal{U}|}+1)} = \frac{|\mathcal{P}|}{|\mathcal{P}|+|\mathcal{U}|}
\end{align}

Therefore, $f$ is the limited function with limes of 1 to ensure the requirements for $q$.
Thus, the total loss function can be expressed as

\begin{equation}
\label{eq:data:autonomous_data_labeling:objective_function}
    \frac{1}{m}\sum\limits_{i=1}^{n}\Big((1-y)*\lVert z_i \rVert^2 + (y)*\frac{(\frac{|\mathcal{P}|}{|\mathcal{P}|+|\mathcal{U}|})^2}{\lVert z_i \rVert}\Big).
\end{equation}

The final loss function of~\autoref{eq:data:autonomous_data_labeling:objective_function} enables the transformer model to train log messages with inaccurate labels by modifying the calculated error according to the relation of $\mathcal{P}$ and $\mathcal{U}$.

\subsection{Iterative Training Strategy}
\label{subsec:data:autonomous_data_labeling:iterative_training_strategy}

Due to numerous log lines being inaccurately labeled in the model's initial training data, particularly for high values of $\delta$, we utilize our iterative training strategy to improve the initial setup for a subsequent model.
This approach is more effective than running a model for more epochs, as more epochs are not able to mitigate the bias present in the training data.

Suppose that in the first iteration, we train our model on the available data and obtain an anomaly score $\lVert z_i\rVert$ for each original log event $l_i \in \mathcal{L}$.
Instead of mapping the anomaly scores directly to labels based on a threshold, we instantiate a new model and use them as training input. 
Prior to this, we refine the anomaly scores to remove biases that were previously learned.
We assume that at least half of all samples are normal, since anomalies are by definition~\cite{liang2007failure} comparably rare. 
Therefore, we calculate the median $m$ of all anomaly scores and subtract it from each anomaly score $\lVert z_i\rVert$ to obtain a smoothed anomaly score.
In the next step, we replace all negative scores with 0 to ensure that each score lives in the range $[0, \infty)$.
Lastly, we employ the hyperbolic tangent as a smoothing function, which effectively squeezes all anomaly scores into the range $[0, 1)$ and thus decreases the influence of very large anomaly scores during the next training iteration by using \autoref{eq:data:autonomous_data_labeling:smooth}. 

\begin{equation}
\label{eq:data:autonomous_data_labeling:smooth}
\forall l_i \in \mathcal{L}: \tanh(\max(0, \lVert z_i\rVert - m))
\end{equation}

This process causes at least half of all log lines to be labeled 0, and the other half of the log lines to have values between 0 and 1, depending on the extent to which the model has regarded these log lines as abnormal.
One more advantage is that the anomaly scores of the model are normalized between 0 and 1 to prevent us from knowing their potential magnitude.

Following our iterative training strategy, we then utilize the transformed anomaly scores as inaccurate labels of our log messages and train a new model.

\begin{figure}[h!]
    \centering
    \includegraphics[width=0.7\columnwidth]{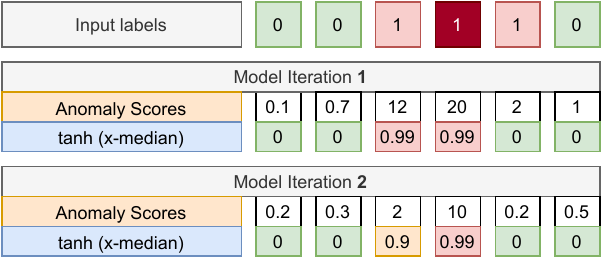}
    \caption{Anomaly scores are smoothed for training.}
    \label{fig:data:autonomous_data_labeling:iterativly_approach}
\end{figure}

\autoref{fig:data:autonomous_data_labeling:iterativly_approach} depicts how the training labels change from iteration to iteration. 
Therefore, the real anomaly is marked in dark red. 
The preliminary training label for this sample also denotes an anomaly, as the respective log message is, along with other log messages, within the failure time window. 
The anomaly scores of the first iteration are then smoothed by utilizing \autoref{eq:data:autonomous_data_labeling:smooth}. 
It can be observed that after the second iteration, the sample that was originally mislabeled receives a lower anomaly score than after the first iteration and is from then on treated as normal.
This exemplifies the optimal process and the potential of our iterative method.
After a configurable number of iterations, the final anomaly scores are eventually mapped to labels.

Using the transformer mechanism, our objective function, and an iterative training strategy, we can accurately label the log data on log line level of an IT system or service, provided we have approximate error times from the underlying system.

\section{Classifying different Types of Anomalies}
\label{sec:data:classifying_anomalies}
After demonstrating a method for autonomous labeling anomalies in log data, our next objective is to explore the various types of anomalies present in a subsequent phase.
In this section, we explore the terrain of anomaly classification and present a taxonomy specifically tailored for anomalies present in log data. 

In the context of anomaly detection, a fundamental distinction is drawn between normal $\mathcal{N}$ and abnormal samples $\mathcal{A}$. 
The anomaly classification further distinguishes $\mathcal{A}$ in more fine-grained classes~\cite{wittkopp2021taxonomy}.
Understanding and classifying abnormal samples in log data is a crucial step in log analysis.

The anomalies, which represent deviations from the expected behavior of a system, exhibit a wide range of characteristics.
Each sample is intricately characterized through a feature set custom-tailored to the specific domain of the underlying data.
For instance, in time series data, a sample might be delineated by its position in a multidimensional space coupled with a temporal component. 
On the other hand, areas like natural language processing could make use of feature sets that include word embedding.

Various academic publications~\cite{chandola2009anomaly,sebestyen2018taxonomy,blazquez2020review} have established a shared anomaly taxonomy. 
This taxonomy systematically categorizes anomalies into two primary classes: Point Anomalies and Contextual Anomalies.
This serves as the foundation for our taxonomy.

The outline of this section is structured as follows:
Firstly, we explore the concept of point anomalies in \autoref{subsec:data:classifying_anomalies:point_anomalies}.
Subsequently, we explore the context anomalies in \autoref{subsec:data:classifying_anomalies:context_anomalies}.
Lastly, we examine an anomaly taxonomy, providing a systematic classification of anomalies for better understanding and analysis in \autoref{subsec:data:classifying_anomalies:anomaly_taxonomy}.

\subsection{Point Anomalies}
\label{subsec:data:classifying_anomalies:point_anomalies}

Depending on the domain, the definitions are slightly different; in general, a point anomaly is a single data sample that can be considered anomalous compared to the rest of the data if its feature set is significantly different~\cite{chandola2009anomaly}.
Consequently, it has a set of characteristics, markedly different from the set of characteristics of normal samples $\mathcal{N}$.

\begin{figure}[htbp]
\centering
\includegraphics[width=0.5\columnwidth]{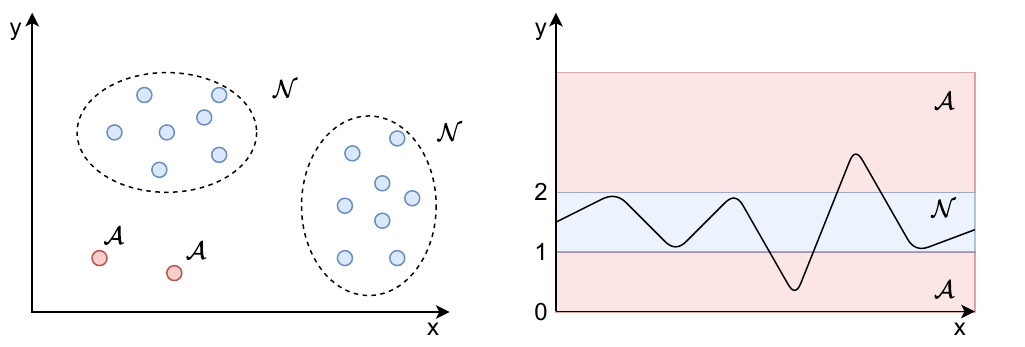}
\caption{Two examples for point anomalies. On the left side: Point anomalies in 2D space. On the right side: Point anomalies in a time series.}
\label{fig:data:classifying_anomalies:point_anomalies_examples}
\end{figure}

Two instances demonstrating point anomalies are depicted in \autoref{fig:data:classifying_anomalies:point_anomalies_examples}.
In the left example, two anomalies deviate from the specified region for normal samples, with their feature set represented by their position in a 2-dimensional space. 
Consequently, the anomalies' positions do not coincide with the designated normal area, rendering them point anomalies. 
In the second example, point anomalies manifest themselves in a time-series context. Given that the normal feature set is defined as $y \in [1,2]$, anomalies are characterized by displaying values of $y$ outside the range $[1,2]$.

\begin{figure}[htbp]
\centering
\includegraphics[width=0.4\columnwidth]{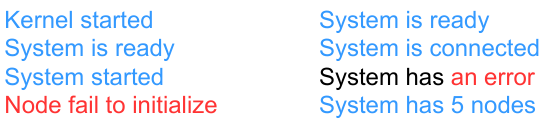}
\caption{Two examples for point anomalies in written text.}
\label{fig:data:classifying_anomalies:point_anomalies_examples_nlp}
\end{figure}

\autoref{fig:data:classifying_anomalies:point_anomalies_examples_nlp} portrays two instances of point anomalies within the written text. 
The anomaly on the left side is directly characterized by the words in the sentence that represent the feature set of this sample. 
The sentence \textit{"Node failed to initialize"} and is an anomalous sample that does not show an intersection with the characteristics of the other sentences, by analyzing the words and the semantics.
The second example is more fine-grained as only specific words are relevant to determine an abnormal sample. 
The presence of certain words within a sentence can indicate whether the sentence is anomalous or not. 
All sentences share a common or similar prefix, using the following descriptors such as "ready", "connected", "5 nodes" or "an error" determining the anomalous behavior. 

This section explored the concept of point anomalies in various domains, emphasizing the general definition of a point anomaly as a single data sample with a significantly different feature set compared to normal data. 
Illustrated through examples in 2D space, time series, and written text, deviations from the designated normal area were observed to characterize point anomalies.

\subsection{Context Anomalies}
\label{subsec:data:classifying_anomalies:context_anomalies}

Samples that exhibit unusual behavior only in a particular context are referred to as contextual anomalies~\cite{chandola2009anomaly}, alternatively known as conditional anomalies~\cite{song2007conditional}.
Samples that fall into this category of anomalies may exhibit behavioral characteristics similar to those of normal instances, but are still considered anomalous within a particular context determined by their contextual properties.

\begin{figure}[htbp]
\centering
\includegraphics[width=0.6\columnwidth]{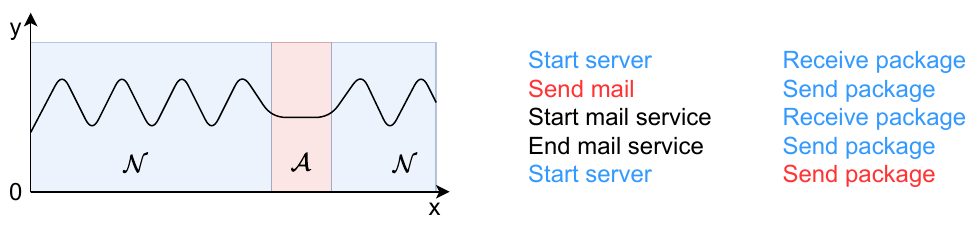}
\caption{One example of a contextual anomaly in time series and two examples for contextual anomalies in written text.}
\label{fig:data:classifying_anomalies:context_anomalies_examples_both}
\end{figure}

\autoref{fig:data:classifying_anomalies:context_anomalies_examples_both} illustrates this situation:
On the left-hand side, there is a time series that exhibits a repetitive oscillation pattern. 
However, this oscillation is disrupted within a specific section.
Despite the fact that the values in this section fall within the standard range, the abnormality is evident in the absence of any oscillation.
The examples on the right-hand side show samples that are typical but their order and context make them an anomaly.
The context of the abnormal samples defined by their contextual properties (y-values / sentence order) is different, as the normally observable strict pattern is interrupted. 
Furthermore, \autoref{fig:data:classifying_anomalies:context_anomalies_examples_both} illustrates two variants of a contextual anomaly in written text. 
The \textit{Send mail} statement in the left example is considered an anomaly due to its context. This is because the statements \textit{Start mail service} and \textit{End mail service} come after the \textit{Send mail} statement.
Since the mail service must be started before sending any mail, this set of statements exemplifies a contextual anomaly. 
The second example is similar to the time series example. 
The statements \textit{Receive package} and \textit{Send package} alternate constantly. 
The anomaly is described by the fact that this alternating pattern is interrupted by a second \textit{Send package} statement.

This section introduced contextual anomalies, characterized by instances that may display normal behavior in general but become anomalous within a specific context determined by their contextual properties. 
The anomalies become evident when the expected patterns are disrupted, as demonstrated by examples in time series and written text.

\subsection{Anomaly Taxonomy for Log Data}
\label{subsec:data:classifying_anomalies:anomaly_taxonomy}

This section discusses the classification of anomalies into two fundamental types, which is then applied to log data.
The taxonomy is further improved and equations are provided to determine the category of anomaly to which an anomalous log line belongs.
This taxonomy relies on the categorization into \textit{Point Anomalies} and \textit{Context Anomalies}. 
Furthermore, we differentiate the \textit{Point Anomalies} into two categories: \textit{Template Anomalies} and \textit{Attribute Anomalies}.

\begin{figure}[htbp]
\centering
\includegraphics[width=0.6\columnwidth]{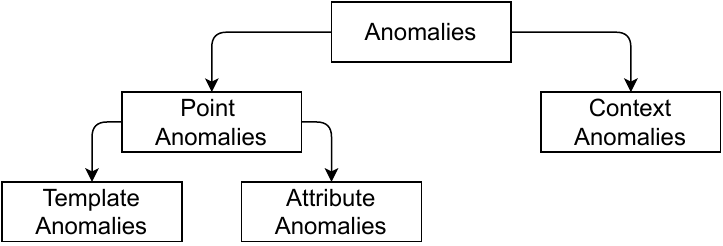}
\caption{Taxonomy for anomalies in log data.}
\label{fig:data:classifying_anomalies:anomaly_taxonomy}
\end{figure}

\autoref{fig:data:classifying_anomalies:anomaly_taxonomy} depicts our taxonomy. 
In the domain of log data, a \textit{Point Anomaly} refers to an abnormal log message that is characterized by its own content. 
The log line could be classified as anomalous only by investigating its content and without observing its context.
Hence, the unusual behavior of a log message is defined by either the corresponding template or a specific attribute, such as a word or number, in the log message.
Therefore, we define a \textit{Template Anomaly} as being characterized by the template of the corresponding log message. 
On the other hand, an \textit{Attribute Anomaly} is defined based on the attributes that are obtained during the template generation process.

Log data anomalies can be classified into a second type known as \textit{Context Anomalies}.
Anomalous behavior in this case is determined by the context, which refers to the surrounding log messages.
The significance of an anomaly is therefore only relevant in relation to the surrounding log messages.

Our taxonomy focuses exclusively on single-threaded event-logging scenarios for contextual anomalies. 
Currently, we have not addressed distributed logging or state log messages~\cite{NagarajKN12}, as well as the challenges associated with them.

\subsubsection{Classifying Log Data Anomalies}

\begin{figure}[htbp]
\centering
\includegraphics[width=0.9\columnwidth]{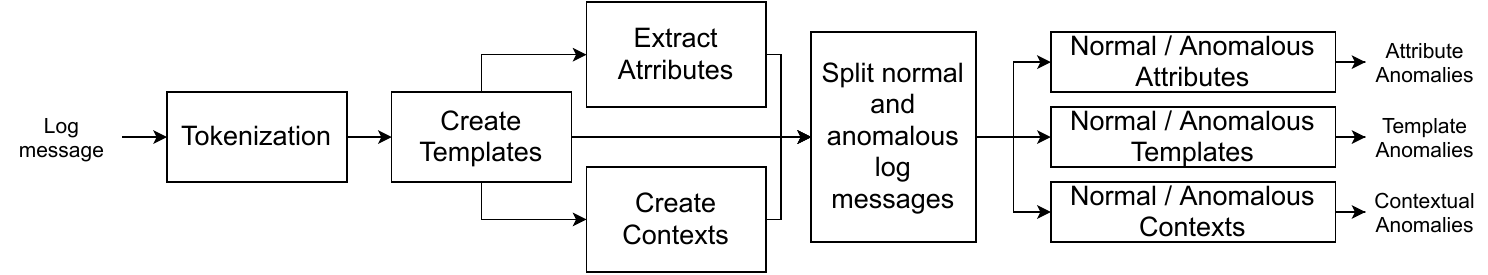}
\caption{Mining process of the different anomaly types.}
\label{fig:data:classifying_anomalies:classification_process}
\end{figure}

The process of classifying types of anomalies with respect to our taxonomy is illustrated in~\autoref{fig:data:classifying_anomalies:classification_process}. 

Each log message is first divided into sequences of tokens in order to extract the log template and a unique id for this template. 
After all log templates are generated, we extract the attributes of each log message and calculate the context for each log line. 

The context $cx_i$ for each log message $l_i$ is based on log template ids and is modeled as a set of template ids.
\begin{equation}
cx_i = \{ t^x_j : j = i-a,\ldots,i-1,i+1,\ldots,i+b] \},
\label{eq:context}
\end{equation}
where $a$ and $b$ are the boundaries of the context. 
For example, we calculate the context of the 10th log message with boundaries $a=2$ and $b=1$ as $c_{10}=\{l_8,l_9, l_{11}\}$.
The template of the log message whose context is created is not considered, as described in \autoref{eq:context}.  
After deriving templates, attributes, and contexts, we divide the data set into a set of normal $\mathcal{N}$ and anomalous log messages $\mathcal{A}$ based on labels determined by experts or automated processes.
Next, utilizing these two sets and previously calculated entities, we derive a score for each anomaly type for each abnormal log message. 
The scores represent how strongly the respective anomaly type is pronounced. 
Each score is in $[0,1]$, with $1$ referring to the strongest manifestation.
\\

\textbf{Template Anomalies.}
The template anomaly $\alpha$ is calculated for each template id $x$. To get all the templates for a specific template ID $x$, we write $t^x(\cdot)$.
\begin{equation}
\alpha(t^x) = \frac{|t^x(\mathcal{A})|}{|t^x(\mathcal{A})|+|t^x(\mathcal{N})|}
\label{eq:template_anomaly}
\end{equation}
\vspace{3mm}

\textbf{Attribute Anomalies.} 
The attribute anomaly $\beta$ is calculated for each log message. 
Since each log message can have multiple attributes, a score for each attribute is calculated, and the attribute anomaly is then represented by the maximal score.
Here, $a_j(\cdot)$ receives all the same tokens as $a_j$ from the corresponding normal or anomalous set. 
\begin{equation}
\beta(a_i) = \max{(s: \forall a_j \in a_i. s= \frac{|a_j(\mathcal{A})|}{|a_j(\mathcal{A})|+|a_j(\mathcal{N})|})}
\label{eq:attribute_anomaly}
\end{equation}
\vspace{3mm}

\textbf{Context Anomalies.} 
The context anomaly $\gamma$ is also calculated for each log message.
For this we calculate all contexts of all log messages from$\mathcal{A}$ and $\mathcal{N}$.
To obtain the context for a specific log message, we write $cx_i(\cdot)$. 

\begin{equation}
\gamma(cx_i) = \frac{|cx_i(\mathcal{A})|}{|cx_i(\mathcal{A})| + |cx_i(\mathcal{N})|}
\label{eq:context_anomaly}
\end{equation}

This formula calculates the probability that a given context is abnormal or normal for each log line in $\mathcal{A}$.

Thus, all scores can be calculated by dividing the occurrences of an event in the anomalous set by the occurrences of this event across both sets. 
As a result, $\alpha$, $\beta$, and $\gamma$ do not make an exact assignment to the anomaly types but create a score that indicates how strongly it behaves to a particular anomaly type. 
Hence, a log line can also have several types of anomaly.

As indicated in the literature, there exist anomaly detection methods that rely on single log lines, the patterns of single log lines, the surrounding context, or a combination of both.
Through the process of classifying the anomalies, we are able to choose or create suitable detection algorithms.

\subsubsection{Chapter Summary}
Log investigation is a challenging and time-consuming task in machine learning and log analysis. 
Therefore, this chapter explored a method for autonomous data labeling that is an essential task for supervised learning and the evaluation of existing models.
Furthermore, we presented an anomaly taxonomy to acquire a deeper understanding of the underlying anomalies for better understanding and analysis. 
Through these contributions, we aim to gain a comprehensive understanding of the log data, identify potential anomalies and their characteristics to enhance current algorithms or develop tailored algorithms for the IT system.

\cleardoublepage

\chapter{Anomaly Detection Model For Failure Recognition}
\minitoc
\label{ch:anomaly_detection}

In the previous chapter, we have presented a method that can autonomously label data with very high precision, and we described an approach for classifying anomalies. 
Using these two methods, we can effectively examine the nature of log data and anomalies to create a precise anomaly detection method.
This is necessary due to the complexity of IT systems, where the volume of log data is growing to the extent that it cannot be manually analyzed.
Therefore, this chapter presents an anomaly detection method that can be trained in different ways, through different objective functions.
Our method concentrates on point anomalies that include template and attribute anomalies. These types of anomaly are prevalent in most log data, as we show in our evaluation in~\autoref{ch:evaluation}.

Although we have the ability to label training data using our method presented in~\autoref{sec:data:autonomous_data_labeling}, there are times when we lack the required prerequisites, such as the observation of the system metrics and the alerts that follow.
In these cases, it is necessary to create unsupervised anomaly detection methods.
Moreover, there are more reasons to create an unsupervised method, because supervised methods are trained on known anomalies, they perform better on those specific anomalies. 
In contrast, unsupervised methods have the advantage of being able to identify new and unknown anomalies~\cite{baier2019cope,chandola2009anomaly}. 
This also offers the advantage of being more resilient in identifying novel anomalies that were also unknown in a supervised setting.
The second benefit of unsupervised methods is that they do not require labeled data, which is hard to obtain and cost intensive~\cite{wittkopp2020a2log}.
In contrast, supervised trained methods demonstrate higher accuracy in stable systems where anomalies remain consistent.
In addition to that, there are also cases where the labels contain inaccuracies~\cite{wittkopp2022pull}. 
In such scenarios, a weakly supervised method can be beneficial.

This presents the challenge of developing a method for various training paradigms, including supervised, weakly supervised, and unsupervised scenarios. 
Regardless of the training strategy employed, during the inference stage, the output of the trained anomaly detection model needs to be interpreted to classify each log line as either abnormal or normal. 
Interpreting the output is challenging, since the appearance of the output values is unknown in advance.
Even recent research on anomaly detection~\cite{guo2021logbert,du2017deeplog,nedelkoski2020logsy} assumes the existence of a sufficient amount of labeled validation data for parameter tuning for the inference stage. 
However, in production settings, where services evolve, such data is hard to obtain, volatile, and requires manual evaluation by experts. 
The issue here is that numerous techniques involve the use of a threshold or decision boundary that is manually adjusted to determine whether a log line is classified as anomaly or not.

In response to these challenges, this chapter presents a two-fold solution. 
First, we present a neural network based on the self-attention mechanism to perform an anomaly scoring for each log line that is capable of accommodating supervised, unsupervised, and weakly supervised training paradigms through different objective functions. 
Second, we perform data augmentation to generate deviations from the respective training data, to analyze the response of the model, and to calculate the final decision boundary for the inference stage of the model. 
The objective of this chapter is to present our anomaly detection method that can identify anomalies with high levels of precision and recall, regardless of whether the training data are unlabeled, contain errors, or are labeled.

This chapter is based on the following publications:
\begin{itemize}
    \item A2Log: Attentive Augmented Log Anomaly Detection~\cite{wittkopp2020a2log}
    \item PULL: Reactive Log Anomaly Detection Based on Iterative PU Learning~\cite{wittkopp2022pull}   
\end{itemize}




In~\autoref{sec:anomaly_detection:anomaly_detection_model:problem_description}, we present the different problems of training an anomaly detection model for logs in different fashions. 
In the following~\autoref{sec:anomaly_detection:anomaly_detection_model:overview}, we provide an overview of our method, including preprocessing data, model training, and the inference stage. 
Next, in~\autoref{sec:anomaly_detection:anomaly_detection_model:preprocessing}, we further examine and demonstrate the preprocessing steps for the log data used in both training and inference. 
Section~\ref{sec:anomaly_detection:anomaly_detection_model:model_training} focuses on model training, detailing the processes of selecting training data, choosing the network architecture, and selecting the objective function. 
In~\autoref{sec:anomaly_detection:inference}, we discuss model inference, starting with the problem description, followed by data augmentation strategies, and concluding with the selection of a decision boundary.

\section{Problem Description}
\label{sec:anomaly_detection:anomaly_detection_model:problem_description}
Systems in various domains, including network infrastructure, cloud computing platforms, and software applications, continuously produce large volumes of log data documenting events, transactions, and state of the system. 
In the contemporary landscape of digital infrastructure, the generation and management of log data have become integral components of operational monitoring and troubleshooting. 
Detecting anomalies in a timely and accurate manner is essential to maintain the integrity and availability of IT systems or services. 

Anomalies are patterns in the data that do not conform to a defined notion of normal behavior~\cite{chandola2009anomaly}.
Anything that deviates from normal behavior can be considered abnormal behavior. 
Steinwall et al.~\cite{steinwart2005classification} state that this can be considered as a binary classification task.
Consequently, anomaly detection on log data is defined as the problem of assigning a binary label to each log line. 
However, the sheer volume and complexity of log data, coupled with the dynamic nature of modern computing environments, pose significant challenges to anomaly detection models.
Depending on the system, environment, and expertise of the DevOps teams, different types of training data are available and can be used.

Supervised learning methods, which rely on labeled examples of normal and abnormal behavior for training, are highly effective but require significant human effort and domain expertise to annotate large volumes of data. 
Furthermore, labeled data may be insufficient or unavailable for certain types of anomalies or operational contexts, limiting the applicability of supervised methods.

In scenarios where roughly labeled data is available, weak-supervised methods can help reduce the load of manual annotation by leveraging partially labeled or noisy data sets. 
However, incorporating weak supervision into an anomaly detection method poses its own set of challenges, including the need to infer reliable labels from noisy data sources and effectively incorporate weak signals into the learning process.

In contrast, unsupervised anomaly detection methods offer a promising avenue for addressing the scalability and labeling challenges inherent in supervised approaches. 
Using the inherent structure and statistical properties of the log data, unsupervised methods aim to identify anomalies without the need for labeled examples. 
However, unsupervised techniques often struggle to distinguish between harmless fluctuations in system behavior and genuine anomalies, leading to high false positive rates and reduced detection accuracy.

Thus, the central problem lies in developing an effective solution capable of accommodating varying degrees of supervision, from fully supervised to weakly supervised and unsupervised learning strategies.

\section{Method Overview}
\label{sec:anomaly_detection:anomaly_detection_model:overview}

Using advances in machine learning, our method offers a flexible approach that seamlessly accommodates varying degrees of supervision, from fully supervised to weakly supervised and unsupervised learning paradigms.

With access to labeled training data, our method can employ supervised learning techniques to accurately identify anomalies in log lines. 
In scenarios where only partial labeling is available or the labels are noisy, our method is adapted to incorporate weak supervision. 
Using the inherent structure and statistical properties of unlabeled normal log data, the model autonomously identifies deviations from normal behavior patterns, effectively uncovering anomalies without the need for labeled examples.

Since our method is based on a neural network, we now describe the individual steps of our anomaly detection method.
The output of a neural network is commonly transformed in a way that can be interpreted as a probability or a probability distribution over one or multiple output neurons, from which further decisions can be inferred.
Therefore, an anomaly detection method requires a final decision on interpreting the neural network's output. 
From this we conclude that a general anomaly detection method, which includes a neural network, consists of two parts:
The \textit{Scoring} and the \textit{Inference}.

The neural network learns the patterns in the first part and calculates an anomaly score for each log line.
Therefore, \textit{Scoring} must transform each log line into a real-valued output. 
The second part is the final decision to detect anomalies on inference stage, based on the anomaly scoring, which transforms the scores into a binary classification.

\begin{figure}[htbp]
\centering
\includegraphics[width=1\columnwidth]{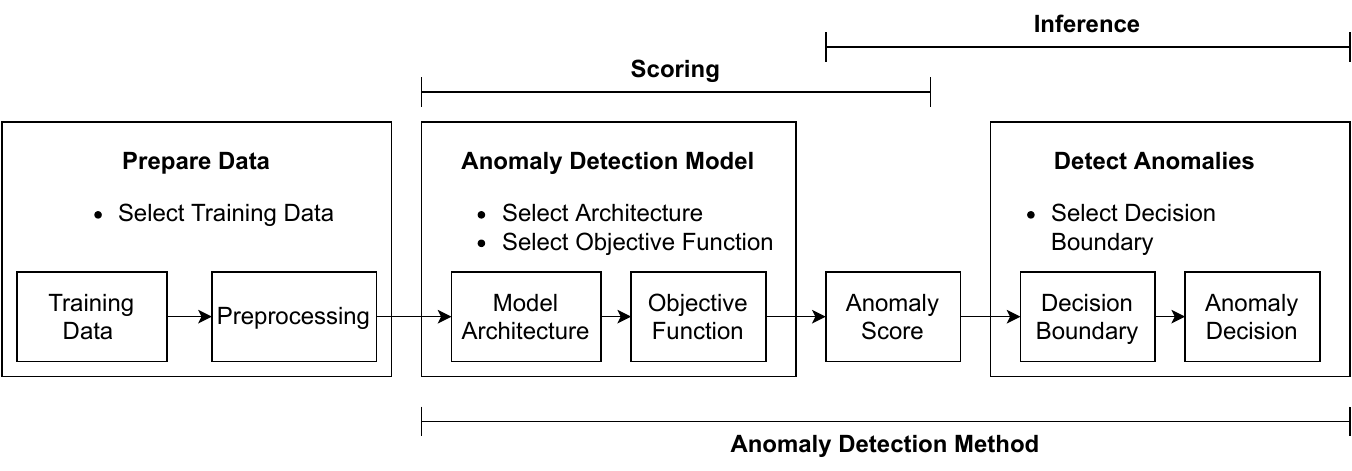}
\caption{Structure of our general anomaly detection method including inference and overview of adaptable components for different training paradigms.}
\label{fig:anomaly_detection:anomaly_detection_model:general_model}
\end{figure}

Formally, the anomaly detection model $\Phi$ for log data $L$ is described by two functions, $s$ and $i$, where $s: L \rightarrow \mathbb{R}$ is the \textit{Scoring} function, represented by a neural network and $i: \mathbb{R} \rightarrow \{0,1\}$ is the \textit{Inference}. 
Therefore, $\Phi = i(s(x))$, where $x$ is the input of preprocessed log data.
That leads us to the general visualization in~\autoref{fig:anomaly_detection:anomaly_detection_model:general_model}.

~\autoref{fig:anomaly_detection:anomaly_detection_model:general_model} illustrates the general anomaly detection method, which can be roughly divided into 3 phases.

In the first phase of the general anomaly detection method, the focus is on selecting and preparing the input data to facilitate the subsequent anomaly scoring phase. 
The log message must be transformed into a format that is suitable for machine learning algorithms.
This initial step plays a crucial role in determining the training paradigm adopted for the anomaly detection model, which could be unsupervised, supervised, or weakly supervised.

Once the training paradigm is established based on the selected or available training data, the second phase of the anomaly detection method begins. 
In this phase, anomaly scores are computed for each log line, indicating the likelihood that it is anomalous. 
The choice of neural network architecture, along with the appropriate objective function, plays a crucial role in determining the accuracy and effectiveness of the anomaly scoring process.
In our method, we employ an encoder structure from the transformer architecture that is tailored to the specific requirements of each training paradigm. 
However, alternative architectures may be considered based on the specific characteristics of the log data and the objectives of the anomaly detection task. 
The diversity of our approach is mainly attributed to various objective functions that are customized to the particular training paradigm employed. 

Following the computation of the anomaly scores, the third phase of the method involves determining a decision boundary from these scores during the inference stage. 
This decision boundary serves to differentiate between normal and anomalous log lines, allowing the model to classify incoming log entries in real-time. 
Further details on this final phase of the anomaly detection process are provided in~\autoref{sec:anomaly_detection:inference}.

Our method provides a unified solution that seamlessly integrates supervised, weakly supervised, and unsupervised learning strategies and can also compute the decision boundary, thus providing a significant advantage for real-world use cases.
Through extensive experimentation and real-world deployment, we demonstrate the efficacy and adaptability of our approach in~\autoref{sec:evaluation:anomaly_detection}.

\section{Log Preprocessing}
\label{sec:anomaly_detection:anomaly_detection_model:preprocessing}

Logging is commonly employed to investigate the faulty behavior of systems and services and to increase dependability, which results in information being written to a log file. 
It is crucial to preprocess the log files to convert them into a format that a neural network can understand.
Thus, systematic preprocessing of log lines improves feature extraction and accuracy of anomaly detection models.

As described in~\autoref{subsec:background:log_data:characteristics} the log file documents the executions of the software and is created by log instructions (e.g. \texttt{printf()} or \texttt{log.info()}). 
Each log instruction results in a single log message, such that the complete log is a sequence of messages $\mathcal{L} = (l_i : i = 1,2,3,\ldots)$. 
There is a commonly used separation in \textit{meta-information} and \textit{content}. 
To access the content of a log message $l_i$, we write $c_i$.

In the initial step of log preprocessing, the focus is on extracting the content $c_i$ from each log line $l_i$. 
We apply our anomaly detection method only to the extracted content that we also call log message.
To process the log message into a structure appropriate for machine learning models, a few preprocessing procedures are applied.

\begin{figure}[htbp]
\centering
\includegraphics[width=1\columnwidth]{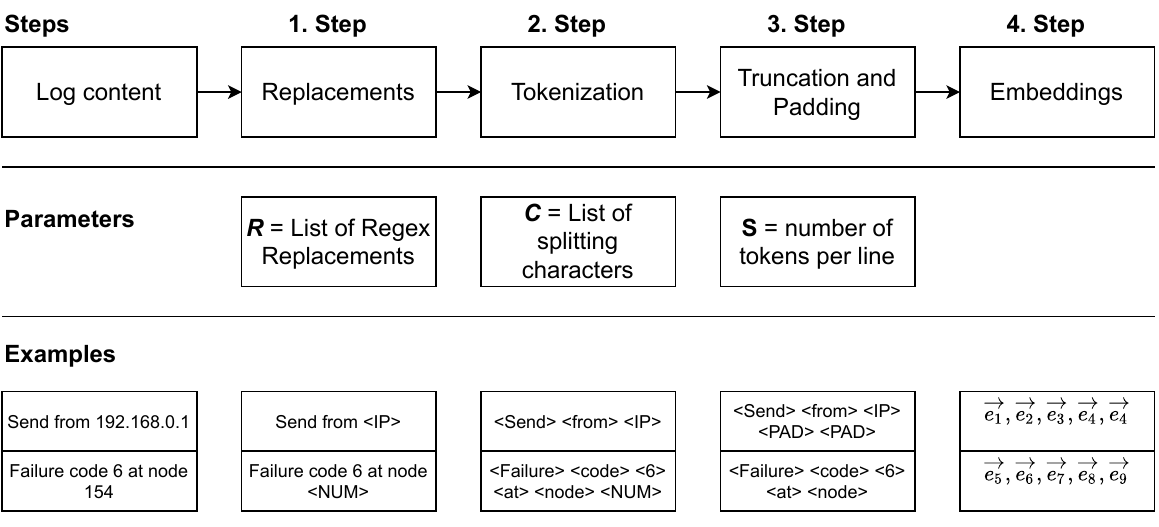}
\caption{Preprocessing steps for every single log line.}
\label{fig:anomaly_detection:anomaly_detection_model:preprocessing}
\end{figure}

~\autoref{fig:anomaly_detection:anomaly_detection_model:preprocessing} shows the 4 preprocessing steps with their possible parameterization.

The first step is to reduce variability, which is done through replacements with placeholder tokens. 
A critical step in promoting uniformity within log messages. 
By systematically replacing recurring patterns like IP addresses, hexadecimal values, or numerical values with specific tokens can effectively decrease variability in the log content.
This logic can also be more fine-grained, so that, for example, only numerical values that exceed a predefined threshold are replaced with a num-token.
This not only simplifies and reduces variability in subsequent processing, but also ensures consistency.

For example, there could be placeholder tokens for hexadecimal values \texttt{[HEX]} or any number greater than or equal to 10 \texttt{[NUM]}. 
Exemplary log messages:
\begin{center}
    \texttt{Memory bar 0 runs at 3591.142 MHz}\\
    \texttt{Clean cell 0x125789}
\end{center}
are thus transformed into:
\begin{center}
    \texttt{Memory bar 0 runs at [NUM].[NUM] MHz}\\
    \texttt{Clean cell [HEX]}
\end{center}

However, the number of replacement rules in $R$ depends on the nature of the log messages. 
If there are more template anomalies (see~\autoref{subsec:data:classifying_anomalies:point_anomalies}) in the underlying log data, several replacement rules can be developed. 
If there are more attribute anomalies (see~\autoref{subsec:data:classifying_anomalies:point_anomalies}), there should be fewer replacement rules, since attribute anomalies can be found in the more variable parts of the log messages and we are reducing the variability with these replacements.
Therefore, it is important that an expert carefully chooses the replacement rules. 
The guideline is to establish the minimum number of rules needed for the situation.
For example, if the log messages contain memory addresses, these would potentially increase the number of tokens by the number of memory addresses of the system, which leads to unnecessary waste of resources, as a memory address in a log line is not an anomaly.

Following the replacement step, the log message undergoes the second step: tokenization (see~\autoref{subsec:background:log_data:processing}).
A process essential for extracting meaningful features and enabling granular analysis by utilizing a customizable list of separators such as colons, commas, whitespace, periods, question marks, semicolons, and hyphens, the log content is segmented into individual tokens. 
During this step, the content $c_i$ is tokenized to a sequence of tokens $t_i$ using the symbols that are set in the parameter $C$, which contains the collection of characters used for separation.

For example, we are using the separators \texttt{.} and \texttt{whitespace}.
Exemplary log messages:
\begin{center}
    \texttt{Memory bar 0 runs at [NUM].[NUM] MHz}\\
    \texttt{Clean cell [HEX]}
\end{center}
are thus transformed into the sequences of tokens:
\begin{center}
    \texttt{[Memory, bar, 0, runs, at, [NUM], [NUM], MHz]}\\
    \texttt{[Clean, cell, [HEX]]}
\end{center}

The more separators are used, the more the log lines are split and the tokens are more granular.
Depending on the nature of the log data, the variability of the input for the neural network can then increase or decrease.
For example, when the hyphen serves as a separator, a compound term like \texttt{server-node} is separated into its constituent parts, \texttt{server} and \texttt{node}, both of which should already exist in other log lines, resulting in no new token. 
However, if the hyphen is not used as a separator, \texttt{server-node} is considered a single new token, which increases the variability.

The next step is the normalization of token counts for each log message to ensure uniformity and consistency across log messages that serve as input for the neural network after the next preprocessing step.
Since these token sequences $t_i$ can vary in length, we normalize them to a fixed length.
The quantity of tokens for all log lines can be chosen as desired and is regulated by the parameter $S$. 
Thereby we truncate longer sequences and fill up smaller sequences with padding tokens \texttt{'[PAD]'}.

For example, we set $S=6$:
\begin{center}
    \texttt{[Memory, bar, 0, runs, at, [NUM], [NUM], MHz]}\\
    \texttt{[Clean, cell, [HEX]]}
\end{center}
are thus truncated or padded to:
\begin{center}
    \texttt{[Memory, bar, 0, runs, at, [NUM]]}\\
    \texttt{[Clean, cell, [HEX], [PAD], [PAD], [PAD]]}
\end{center}

Identifying the appropriate parameter $S$ is a challenging task that necessitates expertise.
An option could be to configure the parameter to align with the length of tokens in the longest log message. 
However, if a log message significantly exceeds the length of other log messages, this could result in excessive overhead, as all other messages would need to be padded with a token to match this extended length. 
On the contrary, setting the parameter $S$ too low could result in a loss of valuable information from numerous log messages. 
As $s$ determines the maximum length of the token list, longer log messages are truncated while shorter ones are supplemented with pad tokens. 
Therefore, a practical guideline is to set $S$ to correspond to the 75th percentile of all token lists within the system.

The final step of the preprocessing pipeline involves token embedding, transforming each token into a continuous vector representation within a high-dimensional space, thereby generating a series of embeddings. 
For every token $w_j$ in the token sequence $t_i$, an embedding vector $\vec{e}_{i}(j)$ is derived. 
These embedding sequences $\vec{e_i}$ are then used as input to the neural network. 
A comprehensive explanation of embedding representations can be found in~\autoref{subsec:background:log_data:processing}.

In conclusion, the log preprocessing pipeline for anomaly detection is a carefully designed structure to enhance the performance and efficiency of the neural network. 
By systematically replacing, tokenizing, normalizing, and embedding log data, organizations can fine-tune the preprocessing process and improve the accuracy of anomaly detection. 


\section{Model Training}
\label{sec:anomaly_detection:anomaly_detection_model:model_training}
In this section, we describe the training procedure of our anomaly detection method, depending on the given training data. 
This involves selecting a suitable objective function and a suitable network architecture. 


\subsection{Selecting Training Data}
\label{subsec:anomaly_detection:anomaly_detection_model:selecting_training_data}

Before our anomaly detection can be trained, it is essential that we understand the available training data. 
As illustrated in~\autoref{fig:anomaly_detection:anomaly_detection_model:general_model}, it is necessary to first examine our available training data before we can choose the suitable elements for the anomaly detection task.
As we know from~\autoref{sec:background:learning_strategies}, there are three possible ways to train our model: Unsupervised, Weakly Supervised or Supervised. 
The training data and the quality of their labels determine which strategy we can apply.

Unsupervised training can be utilized when we are certain that our training data consists solely of normal samples and we lack any labels.
Provided that our data are accurately labeled and that we are confident that there are no inaccuracies in either the normal or abnormally labeled training data, we have the ability to conduct supervised training.
If we are uncertain about the accurate labeling of all instances within a class, such as the presence of anomalies in the normal training class or normal instances within the anomalous training class, our PU learning method (see~\autoref{subsec:data:autonomous_data_labeling:learning_from_pu}) allows us to perform weakly supervised training.

Based on these scenarios, we can divide the training data into 5 classes:

\begin{definition}[Normal Training Data]
\label{def:normal_training_data}
Normal training data refers to a data set consisting of logs of system behavior during normal expected operation. 
These logs capture typical patterns, behaviors, and events that occur when the system is operating normally. 
It must be ensured that these logs really do not contain any anomalies.
\end{definition}

Normal training data can be sourced from the operational system during periods of normal function. 
However, it is crucial that these data undergo an expert review to confirm the absence of any anomalies that may have gone unnoticed because of their failure to cause significant system errors. 
Alternatively, normal training data can be produced in a testing environment where the identical system operates but not with a normal user load.
In such instances, user behavior can be emulated to generate normal log data. 
The challenge lies in accurately and intricately simulating user behavior.

If we only have normal training data, we can apply an unsupervised training strategy, which comes with several advantages and disadvantages (see~\autoref{subsec:background:training_strategies:unsupervised}).

\begin{definition}[Abnormal Training Data]
\label{def:abnormal_training_data}
Abnormal training data refers to a data set that contains logs that deviate from the expected patterns of system behavior. 
Abnormal training data consist exclusively of anomalous behavior, such as system failures, errors, security breaches, or unusual events.
\end{definition}

Obtaining abnormal training data for log anomaly detection may be sourced from various sources, including historical logs of known errors or failures, security incidents, simulated attack scenarios, or manually curated examples of abnormal behavior.

Obtaining abnormal log data for training in log anomaly detection typically involves a multistep process. 
First, identify potential sources of abnormal behavior within your system, such as system failures or unusual user interactions.
Then, experts are required to meticulously scrutinize the log files from these sources and tag the respective anomalies. 
They must ensure that the labeled data cover a diverse range of abnormal scenarios to provide sufficient training examples for the model. 
Therefore, this procedure is labor intensive and costly.
Alternatively, if the appropriate criteria are satisfied, we can implement our automatic labeling method demonstrated in~\autoref{sec:data:autonomous_data_labeling}.

The purpose of using abnormal training data in log anomaly detection is to provide the model with examples of abnormal behavior that may indicate potential threats or issues within the system. 
By incorporating abnormal data during training, the model can learn to distinguish between normal and abnormal patterns, improving its ability to detect anomalies effectively.
In case we have normal training data and abnormal training data, we can apply supervised training (see~\autoref{subsec:background:training_strategies:supervised}). 

\begin{definition}[Positive Training Data]
\label{def:positive_training_data}
Positive training data denotes a data set in which all the logs belong exclusively to a single class. 
Thus, positive training data could be classified as normal training data or abnormal training data.
It is crucial to confirm that all instances in the positive training data truly belong to a single class.
\end{definition}

Consequently, the source of positive training data could be one of the two processes previously outlined for normal or abnormal training data.

\begin{definition}[Unknown Training Data]
\label{def:unknown_training_data}
Unknown training data is defined as a data set that contains log data from any class. 
This implies that this data includes both types; normal training data and abnormal training data. 
However, we are uncertain whether the sample in the unknown class is normal or anomalous.
Furthermore, we remain uninformed about the distribution of both classes in the unknown and lack any labels.
\end{definition}

Any component of the system that generates logs can provide the logs for the unknown training data. 
Ideally, these logs should originate from the production system to ensure the most accurate representation of the system.

We can subsequently utilize positive training data and unknown training data for the training of our weakly supervised anomaly detection model. 
It is crucial to note that the positive class is composed exclusively of either normal or abnormal samples, while the unknown class contains samples from both categories, according to our definition.

If we employ both of these training data classes, the training data contain partial labels or noisy annotations, which require the model to infer anomaly patterns from these imperfect signals. 
This paradigm offers a middle ground between the supervision extremes of fully supervised and unsupervised learning, enabling the model to learn from partially labeled or noisy datasets.

\begin{definition}[Auxiliary Training Data]
\label{def:auxiliary_training_data}
Auxiliary training data essentially refer to unknown training data but originates from a different system. 
The crucial aspect is that this type of training data has distinct features compared to the training data from the original system. 
\end{definition}

The logs could originate from any system that is not our original system. Regardless of whether they hold anomalies or merely normal data, it is irrelevant.
This data is used to ensure the stability of the training process under specific circumstances.

After we have defined and described the various classes of training data, in the next chapter, we will demonstrate how the logs need to be preprocessed before applying our anomaly detection method to them.

\subsection{Selecting Network Architecture}
\label{subsec:anomaly_detection:anomaly_detection_model:network}

As we have explained in~\autoref{sec:anomaly_detection:anomaly_detection_model:overview} and illustrated in~\autoref{fig:anomaly_detection:anomaly_detection_model:general_model}, the \textit{Scoring} phase consists of 2 parts. 
First, we select a suitable neural network and then a suitable objective function based on the training data. 
In this section, we focus on choosing the suitable neural network.

The design of our approach enables us to take advantage of any network architecture that eventually yields a single scalar value. 
When it comes to anomaly detection within our method, choosing the right neural network is crucial. 
Taking into account the wide range of neural networks at our disposal, this may seem like an intimidating endeavor. 
However, the nature of our data can guide us towards the right choice. 
We are working with log data, which falls under the subdomain of natural language. 
Although a variety of network structures can be utilized, research indicates that the transformer architecture has already performed well in the domain of natural language processing~\cite{yates2021pretrained}. 
This architecture, originally designed for translation tasks, has proven to be highly effective in handling textual data, making it ideal for log anomaly detection. 
For our specific task, we only require the encoding component of the transformer architecture.
The reason being that the downstream anomaly detection task is performed on the encoded embedding of the entire log line, providing us with the necessary information for accurate and efficient anomaly detection.
Therefore, we utilize the encoder of the transformer architecture~\cite{devlin2018bert} with self-attention~\cite{VaswaniSPUJGKP17}. 
We have elaborated on this network architecture and its mathematical foundations in ~\autoref{subsec:background:concepts:transformer} in detail.

The encoder of the transformer architecture is applied to map the token sequences of the log lines to a d-dimensional vector embedding (see~\autoref{subsec:background:log_data:processing}), which is represented through the \texttt{[CLS]} token and prefixed to all log messages.
The preprocessed token sequence $t_i$ (see~\autoref{sec:anomaly_detection:anomaly_detection_model:preprocessing}) serves as input for our model. 
Then the corresponding embeddings $\vec{e_i}$ are assigned to each token of the sequence $t_i$.
Hence, the architecture used does not consider the order of the tokens, the input sequence is enriched with positional encoding information (see~\autoref{subsec:background:concepts:transformer}). 
The model then calculates an output embedding for each positional encoded input sequence $\vec{e_i}$, which summarizes the log message using the embeddings of all tokens.
The utilized model is indirectly responsible for training the aforementioned embeddings via backpropagation of prediction errors; therefore, the embeddings will be optimized to preserve information of the corresponding tokens.
The information of each log message is embedded within the embedding of the \texttt{[CLS]} token and is adjusted by minimizing losses during training.
Given that our neural network generates an embedding, we assess it by determining the length of the embedding vector, which provides us with the anomaly score. 
Consequently, the anomaly score is determined by the length of the output vector $\lVert \vec{z_i} \rVert$ and is calculated for each log line.
We denote the calculated score of the model that is the output of the \textit{Scoring} phase as $\vec{z_i}$ and use it throughout the remaining steps.

We set the scoring target for normal data at $0$ as the absolute normal state, so that the likelihood of an anomaly in the log message is proportional to the increase in positive values.

\subsection{Selecting an Objective Function}
\label{subsec:anomaly_detection:anomaly_detection_model:objective_function}


In this section, we present the three separate objective functions that can be utilized to train our selected neural network for anomaly detection during the \textit{Scoring} phase.
Each function comes with its unique strengths and limitations, corresponding to the specific characteristics of the selected or available training data.


Before exploring the details of each objective function, we present the objective for our neural network.
To express it in a formal manner, we depict our model as $z_i=\Phi(\vec{e_i}, y_i, \Theta)$, with $z_i$ being the output. 
$\Phi$ is a trainable model, which needs to acquire the parameters $\Theta$ during the training process, to compute an anomaly score for the embeddings of the log lines $\vec{e_i}$ and its associated labels $y_i$.
Since the output of the model $\Phi$ is a vector, we use the length of the vector $\lVert z_i \rVert$ as anomaly scores for each log message.
Anomaly scores close to $0$ represent normal log messages, where large scores indicate an abnormal log message. 

Initially, we outline the objective function required for unsupervised training and highlight the essential training data utilized. 
Subsequently, we establish the objective function for the weakly supervised training scenario. 
Finally, we demonstrate the objective function for supervised training and also refer to the training data that are needed for that approach.

\subsubsection{Unsupervised Objective Function} 
When it comes to anomaly detection, unsupervised learning (see~\autoref{subsec:background:training_strategies:unsupervised}) implies that our method only has access to normal logs and does not have knowledge of any anomalies. 
In order to be able to train our model in an unsupervised fashion, we need two types of training data:

\begin{itemize}
  \item Normal training data (see definition \ref{def:normal_training_data} from~\autoref{subsec:anomaly_detection:anomaly_detection_model:selecting_training_data})
  \item Auxiliary training data (see definition \ref{def:auxiliary_training_data} from~\autoref{subsec:anomaly_detection:anomaly_detection_model:selecting_training_data})
\end{itemize}

The normal training data serves to educate the model so that it can grasp the structures and inherent semantics of the log data that are generated during run time without any faults and create a mathematical representation of the training data.
Given that our model $\Phi$ requires a label $y_i$ for each training data sample, we assign a label $y_i=0$ to each log line in the normal training data set.
Given that the data are solely made up of normal data, the model would assign the same label to every input, leading to a collapse. 
This is because it would not be required to "learn" the structures; instead, it would learn that the output should always be 0, irrespective of the input data. In our scenario: $\lVert z \rVert=0$.
To prevent this, we need a stabilization class as a second class that exists of auxiliary training data.
The auxiliary training data are labeled with $y_i=1$.
This gives the model an abstraction of variety in log messages, and, therefore, the model should be able to assign different anomaly scores to real anomalies based on log message characteristics. 

Hence, it is necessary to have an error term for normal training data which remains close to zero if the log line corresponds to the normal representation. 
The subsequent expression experiences a quadratic growth, if the outcome vector of the model is large for the input data that have the label $y_i=0$:

\begin{equation}
\label{eq:anomaly_detection:unsupervised:normal}
(1-y_i)\left\|\vec{z_i}\right\|^2
\end{equation}

\begin{figure}[htbp]
\centering
\includegraphics[width=0.5\columnwidth]{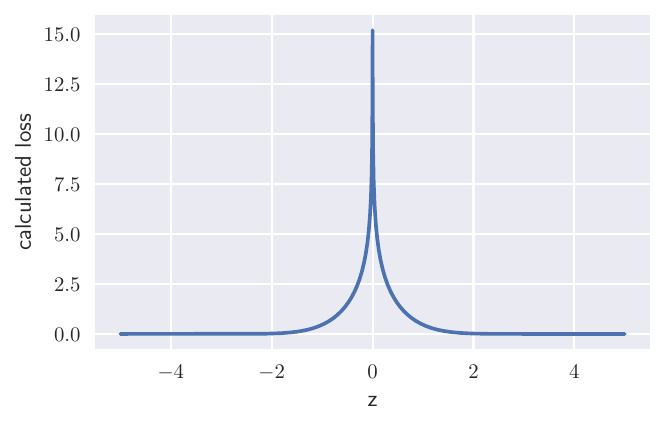}
\caption{Part of the objective function for unsupervised learning to calculate the loss for auxiliary data.}
\label{fig:anomaly_detection:anomaly_detection_model:objective_function:unsupervised}
\end{figure}

In contrast, our goal is to have greater outcome vectors for our auxiliary data, which is marked with $y=1$. 
These data could be normal, abnormal, or a combination of both, according to our definition \ref{def:auxiliary_training_data}. 
We continue to desire higher anomaly scores, but not excessively high, considering that any data could be part of the auxiliary data. 
As a result, we devised a function that does not incentivize the network to produce larger outcome vectors $z_i$, allowing us to have a concise representation of the auxiliary data and retains the model of implosion.
The function is depicted in~\autoref{fig:anomaly_detection:anomaly_detection_model:objective_function:unsupervised} ranging from $-5$ to $5$ and computes substantial losses, the closer the outcome vector is to $0$.

The displayed function, which is utilized to compute the loss for training samples where $y_i=1$, is as follows:

\begin{equation}
\label{eq:anomaly_detection:unsupervised:auxiliary}
- y_i \cdot log(1-exp(-\left\|\vec{z_i}\right\|^2))
\end{equation}

We now combine~\autoref{eq:anomaly_detection:unsupervised:normal} and~\autoref{eq:anomaly_detection:unsupervised:auxiliary} into a unified function, resulting in a composite objective function to calculate the loss in unsupervised training~\cite{nedelkoski2020logsy}. 

\begin{equation}
\label{eq:anomaly_detection:unsupervised:full_formular}
\frac{1}{n}\sum_{i=1}^n(1-y_i)\left\|\vec{z_i}\right\|^2 \\ - y_i \cdot log(1-exp(-\left\|\vec{z_i}\right\|^2))
\end{equation}

Since several training data samples are combined into one batch, we write the objective function as the average of the loss values in a batch of size $n$.

\subsubsection{Weak supervised Objective Function}
We now present the objective function for weakly supervised learning. 
Thus, we refer to our defined learning problem when learning positive and unknown samples is required, as explained in~\autoref{subsec:background:concepts:pu_learning}.
This is necessary when there is uncertainty about the labels within a class.
Hence, we utilize a positive class with confirmed labels and an unknown class with uncertain labels.
For this training approach, the required training data includes:

\begin{itemize}
  \item Positive training data (see definition \ref{def:positive_training_data} from~\autoref{subsec:anomaly_detection:anomaly_detection_model:selecting_training_data})
  \item Unknown training data (see definition \ref{def:unknown_training_data} from~\autoref{subsec:anomaly_detection:anomaly_detection_model:selecting_training_data})
\end{itemize}

In this scenario, we can be sure that only normal samples are found in the positive class. 
However, the unknown class can contain both anomalies and normal samples, but at least there must be some anomalies that are consistent with our definition \ref{def:unknown_training_data}.
This could be a result of failures during labeling or due to the pressure of time, since the process of identifying and tagging anomalies is quite expensive, a point we have previously discussed in~\autoref{sec:data:autonomous_data_labeling}.

In this case, we set the labels $y_i=0$ for the positive class $\mathcal{P}$ since it consists only of normal examples, while all samples of the unknown class receive the label $y_i=1$.

The objective function must be modeled so that the semantics of normal log messages in class $\mathcal{P}$ and $\mathcal{U}$ have low anomaly scores. 
The semantics of the log messages that occur only in $\mathcal{U}$ are most likely abnormal and therefore must have higher anomaly scores.
In addition to that, the objective function must be able to handle large amounts of incorrectly labeled log messages, since the class $\mathcal{U}$ can increase rapidly for large $\delta$, as $\delta$ can only be estimated roughly from the monitoring systems.

Since the anomaly scores for each input sequence $\vec{e_i}$ are calculated by the length of the corresponding outcome vector $\lVert z_i\rVert$ of the model, the model must map normal log messages with label $y_i=0$ to small vectors.

\begin{equation}
\label{eq:anomaly_detection:weak_supervised:positive}
    (1-y_i)\cdot\lVert z_i \rVert^x
\end{equation}

Where $x$ could be $2$ or $4$, depending on the sizes of $\mathcal{P}$ and $\mathcal{U}$.
In the event that the size of the unknown class surpasses that of the positive class, it becomes imperative to ensure that the samples from the positive class are closely approximated to $0$, irrespective of the presence of similar samples in the unknown class. 
Therefore, we increase the loss many times faster if $\mathcal{P} \leq \mathcal{U}$.

\begin{equation}
x=\begin{cases}
2 & \mathcal{P}>\mathcal{U} \\
4 & else
\end{cases}
\end{equation}

In contrast, we need to increase the loss for all anomaly scores when the log message is of class $\mathcal{U}$ and therefore has the label $y_i=1$.
However, we must ensure that growth is not too swift, given that the unknown class also encompasses normal samples. 
Therefore, the relationship of $\mathcal{P}$ and $\mathcal{U}$ should remain within the boundaries of $0$ to $1$, to not grow too fast, we model it as a limited function $f$, which provided the relation of $\mathcal{P}$ and $\mathcal{U}$.

\begin{align}
\label{eq:anomaly_detection:weak_supervised:objective_function_pu_lim}
    f(x) = \frac{x}{x+1}, \lim \limits_{x \to \infty} f(x) = 1, \nonumber \\
    f(\frac{|\mathcal{P}|}{|\mathcal{U}|}) = \frac{\frac{|\mathcal{P}|}{|\mathcal{U}|}}{(\frac{|\mathcal{P}|}{|\mathcal{U}|}+1)} = \frac{|\mathcal{P}|}{|\mathcal{P}|+|\mathcal{U}|}
\end{align}

Consequently, we employ the ratio of the positive $\mathcal{P}$ to the unknown $\mathcal{U}$ class as a numerator to the outcome vector of the neural network in~\autoref{eq:anomaly_detection:weak_supervised:unknown}.

\begin{equation}
\label{eq:anomaly_detection:weak_supervised:unknown}
    (y_i)\cdot\frac{(\frac{|\mathcal{P}|}{|\mathcal{P}|+|\mathcal{U}|})^2}{\lVert z_i \rVert}
\end{equation}

After combining both functions \ref{eq:anomaly_detection:weak_supervised:positive} and \ref{eq:anomaly_detection:weak_supervised:unknown} in a single objective function, we receive the objective~\autoref{eq:anomaly_detection:weak_supervised:objective_function} for training batches of size $m$.

\begin{equation}
\label{eq:anomaly_detection:weak_supervised:objective_function}
    \frac{1}{m}\sum\limits_{i=1}^{n}\Big((1-y_i)*\lVert z_i \rVert^x + (y_i)*\frac{(\frac{|\mathcal{P}|}{|\mathcal{P}|+|\mathcal{U}|})^2}{\lVert z_i \rVert}\Big).
\end{equation}

The final loss function of~\autoref{eq:anomaly_detection:weak_supervised:objective_function} enables the transformer model to train log messages with inaccurate labels of class $\mathcal{U}$ by modifying the calculated error according to the relation of $\mathcal{P}$ and $\mathcal{U}$.





\subsubsection{Supervised Objective Function}
We still require an objective function for the supervised training process.
The data required for this training approach are straightforward. 
We need normal training data on the one side, and on the other side we need abnormal training data.
The training data are defined as follows:

\begin{itemize}
  \item Normal training data (see definition \ref{def:normal_training_data} from~\autoref{subsec:anomaly_detection:anomaly_detection_model:selecting_training_data})
  \item Abnormal training data (see definition \ref{def:abnormal_training_data} from~\autoref{subsec:anomaly_detection:anomaly_detection_model:selecting_training_data})
\end{itemize}

The normal training data receive the label $y_i=0$, whereas the abnormal training data receive the label $y_i=1$.

Given that the anomaly scores for each input sequence $\vec{e'_i}$ are determined by the magnitude of the associated outcome vector $\lVert z_i\rVert$ from the model, it is necessary for the model to associate normal log messages with label $y_i=0$ with smaller vectors and therefore generate higher losses for normal log messages with large output vectors.

\begin{equation}
\label{eq:anomaly_detection:supervised:a}
    (1-y_i)\cdot\lVert z_i \rVert^x
\end{equation}

Abnormal log messages with the label $y_i=1$ should receive large vectors, and therefore high anomaly scores, which is ensured by~\autoref{eq:anomaly_detection:supervised:b}
\begin{equation}
\label{eq:anomaly_detection:supervised:b}
    (y_i)\cdot\lVert z_i \rVert^x
\end{equation}

Combining both functions to one loss function, we obtain the loss function for batches with the size $m$ in~\autoref{eq:anomaly_detection:supervised}.

\begin{equation}
\label{eq:anomaly_detection:supervised}
    \frac{1}{m}\sum\limits_{i=1}^{n}(1-y_i)*\lVert z_i \rVert^2 + (y_i)*\lVert z_i \rVert^2
\end{equation}

In conclusion, we have explored three distinct loss functions tailored for unsupervised, supervised, and weak supervised training scenarios. 
Each approach offers unique advantages and considerations in guiding the model towards optimal performance for the specific training data that is used. 
By understanding the nuances of these loss functions, we pave the way for more informed decision-making in selecting the most suitable training methodology for a given task. 

We are now in a position to accurately train our neural network using our training data and compute an anomaly score for each log message. 
However, this score initially provides us with no meaningful information; therefore, we need to determine a threshold value. 
This threshold will be used in the inference phase to decide if the anomaly score is high enough to be considered an anomaly.

\section{Model Inference}
\label{sec:anomaly_detection:inference}

Following our detailed discussion of the anomaly scoring phase, in this section we continue to explain the method for making a final decision based on the anomaly score during inference as shown in~\autoref{fig:anomaly_detection:anomaly_detection_model:general_model}.

Our model assigns anomaly scores close to $0$ to normal log data, while anomalies receive higher anomaly scores.
The challenge lies in automatically determining a sufficiently high decision boundary such that it is above the scores of normal log lines and below those of anomalies.

Recent research studies employing deep learning for log data analysis and anomaly detection~\cite{guo2021logbert, du2017deeplog, nedelkoski2020logsy, li2020swisslog} face the same challenge. 
These studies generally presuppose the availability of adequate labeled validation data to manually establish the threshold optimally. 
However, in production settings, where services evolve, such data are hard to obtain, volatile, and require manual evaluation by experts. 
In particular, when it comes to deriving an anomaly decision during inference for each log line, these methods lack the capability to obtain a decision boundary in an automated and unsupervised manner. 
Commonly employed methods need to be aware of the anomalies in the validation data to set this decision boundary optimally. 
This relatively strong requirement poses limitations. 

Therefore, in this section, we describe how the decision boundary for the inference stage can be obtained from the training data.

Thereby we first describe the problem in~\autoref{subsec:anomaly_detection:inference:problem_description} to then explain the data augmentation for log data in~\autoref{subsec:anomaly_detection:inference:data_augmentation}.
Finally, we provide the calculations of our decision boundary in an unsupervised scenario in~\autoref{subsec:anomaly_detection:inference:decision_boundary_unsupervised}, for a weakly supervised scenario in~\autoref{subsec:anomaly_detection:inference:decision_boundary_weak_supervised} and for a supervised scenario in~\autoref{subsec:anomaly_detection:inference:decision_boundary_supervised}.

\subsection{Problem Description}
\label{subsec:anomaly_detection:inference:problem_description}

In this subsection, we investigate the challenge of selecting an optimal decision boundary.
The primary challenge we face is the uncertainty of the anomaly scores produced by the neural network for log messages during inference. 
For example, after training, it is unclear whether scores such as $0.05$ or $0.5$ indicate a high or low score, since there are no labels associated with input log messages during inference. 
This makes it difficult to identify abnormal log messages as such, as the computed anomaly scores cannot be interpreted, and thus scores of real abnormal log messages can have an arbitrary form.
On the other hand, the decision boundary faces the problem that the neural network will not be able to set the scores in such a way that the log messages can be separated exactly into 2 classes and that these two classes then exactly reflect the normal and abnormal data. 
It is most likely that there will be a normal log message that has a higher score than an abnormal message, and vice versa, independently of the training mode.

\begin{figure}[htbp]
\centering
\includegraphics[width=0.8\columnwidth]{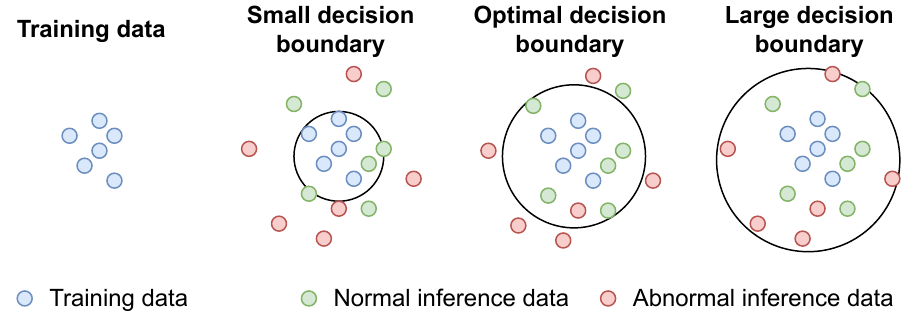}
\caption{Illustration of the various possibilities for setting a decision boundary in an unsupervised setting.}
\label{fig:anomaly_detection:inference:decision_boundary_unsupervised}
\end{figure}

\begin{figure}[htbp]
\centering
\includegraphics[width=0.8\columnwidth]{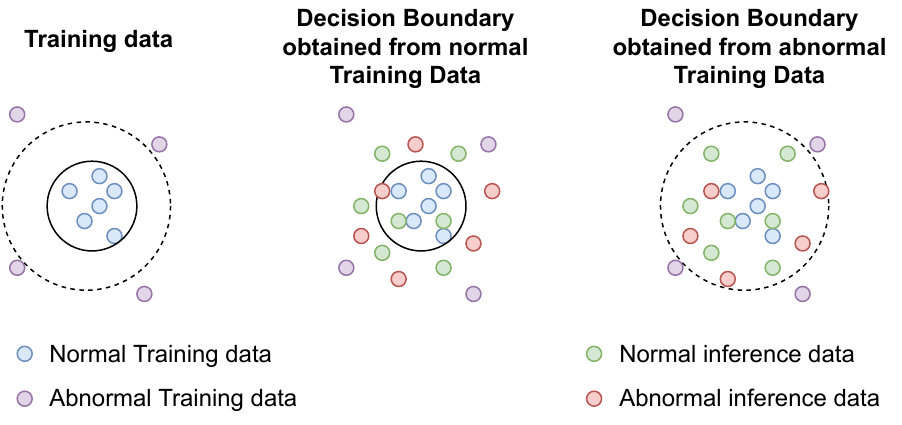}
\caption{Illustration of the various possibilities for setting a decision boundary in a supervised setting.}
\label{fig:anomaly_detection:inference:decision_boundary_supervised}
\end{figure}

The challenge presents itself differently for unsupervised and supervised methods. Therefore, we will illustrate the challenge for unsupervised in~\autoref{fig:anomaly_detection:inference:decision_boundary_unsupervised} and supervised in ~\autoref{fig:anomaly_detection:inference:decision_boundary_supervised}.

\autoref{fig:anomaly_detection:inference:decision_boundary_unsupervised} demonstrates the difficulty of setting a precise decision boundary using only normal data points from the training. 
The blue points represent normal data points that are part of the training data. 
The green and red points represent normal and abnormal data points during the inference phase. 
It can be observed that the green points are on average closer to the normal trained data points, yet considering them for configuration of the decision boundary would lead to a suboptimal separation. 
If the decision boundary is too small, too many log messages will be classified as false positive by the method. 
If the decision boundary is too large, too many log messages will be classified as false negatives. 
Both scenarios result in weaker precision, recall, and F1 scores. 
we can also transfer the problem description in the unsupervised scenario to the weakly supervised scenario, since we have no information about the unknown class, we also only have the normal data at our disposal for the decision boundary.

The second example in~\autoref{fig:anomaly_detection:inference:decision_boundary_supervised} demonstrates the difficulty of setting a precise decision boundary even if we have anomalies during training in a supervised setting.
While the model effectively distinguishes between normal and abnormal data during training, the inference phase may introduce new and unforeseen log messages. 
These new normal log messages might score higher than those seen during training, whereas newly encountered anomalies could score lower.
This results in an issue where, if the decision boundary is determined using either the normal or abnormal training data, it fails to effectively distinguish between new normal samples and anomalies during the inference phase, as illustrated in~\autoref{fig:anomaly_detection:inference:decision_boundary_supervised}.

These examples demonstrate that both components, the \textit{Anomaly Scoring} and calculating the decision boundary for \textit{Anomaly Inference}, have to be precise. 

Due to the various factors influencing the training of a neural network, each trained anomaly scoring model will most likely produce different anomaly scores for the same data, and therefore the final decision boundary must be set individually for each model.
Moreover, as the nature and relation of future normal and abnormal log messages cannot be known in advance, it is required that the decision boundary can cope with these uncertainties and still provide good decisions solely based on the training data that are available.

Therefore, each model needs its own anomaly decision function $i$, which is calculated before inference and applied in the inference stage.
Eventually, the \textit{Anomaly Decision} function $i$ assigns to each log message an explicit label $\in \{0,1\}$ during inference with

\begin{equation}
i(\left\|\vec{z_i}\right\|)= 
\begin{cases}
    1, & \left\|\vec{z_i}\right\| > \epsilon,\\
    0, & \left\|\vec{z_i}\right\| \leq \epsilon
\end{cases}
\end{equation}
where $\epsilon$ is a decision boundary that is used to perform a binary classification for log messages.

\subsection{Data Augmentation}
\label{subsec:anomaly_detection:inference:data_augmentation}
Data augmentation in general is a technique widely used in machine learning and deep learning to artificially expand the size and diversity of a training dataset. 
Its primary goal is to improve the generalization and robustness of machine learning models by exposing them to a wider range of variations and scenarios present in real-world data.

In our case, we use data augmentation to explore the model's behavior toward new and unknown patterns in log data.
As we do not know how the anomaly score will result for new log messages that appear in the future, we require a method that simulates deviations from the already trained log messages to understand how the model will react and determines the decision boundary $\epsilon$.
We simulate deviations from training data by applying data augmentation.

\begin{figure}[htbp]
\centering
\includegraphics[width=0.8\columnwidth]{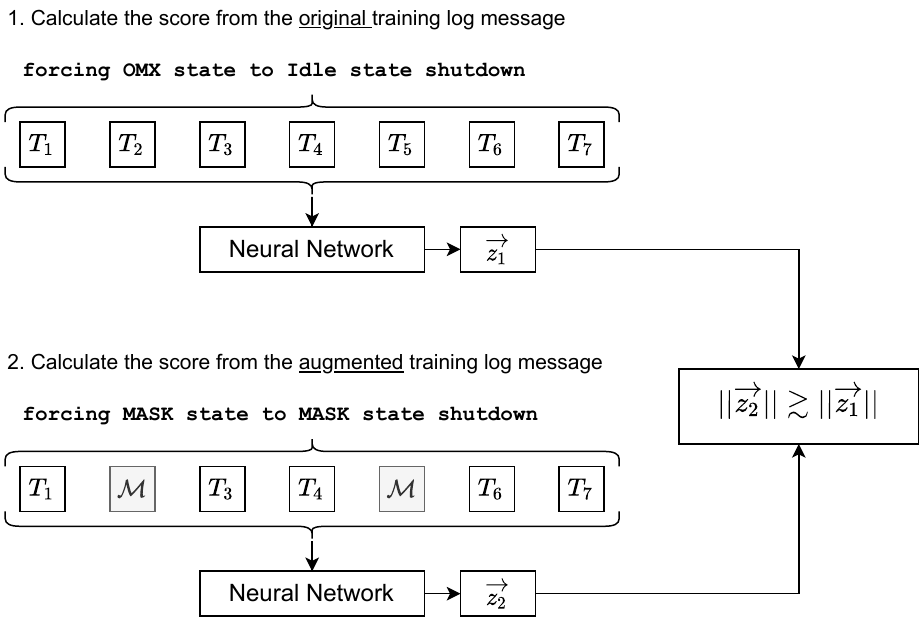}
\caption{Data augmentation for a log message with $\alpha=2$ replaced tokens.}
\label{fig:anomaly_detection:inference:data_augmentation}
\end{figure}

\autoref{fig:anomaly_detection:inference:data_augmentation} depicts our method for modifying log data to see how the model reacts.
First, the training data log messages are tokenized, followed by replacing the tokens $\alpha \in \mathbb{N}$ at random positions in each sequence of tokens $t_i$ with a masking token $\mathcal{M}$, as described in \autoref{eq:anomaly_detection:inference:data_augmentation}.

\begin{equation}
\label{eq:anomaly_detection:inference:data_augmentation}
t'_i = (w_j : w_j \in V \lor w_j = \mathcal{M}, j=1,2,\ldots,|t_i|)
\end{equation}

Second, we received the anomaly scores of the trained model for the augmented training data. 
As these masking tokens are excluded during training, the trained model calculates a score for each augmented token sequence $t'_i$ that should be greater than or equal to the score of the original token sequence $t_i$, because it contains $\alpha$ masking tokens $\mathcal{M}$ and therefore does not conform to established knowledge. 

This method allows us to emulate changes across all available training data sets, aiding in the computation of a decision boundary for the inference stage. 
The calculation of $\epsilon$ is then based on the training strategy chosen.

\subsection{Calculating a Decision Boundary for an Unsupervised Model}
\label{subsec:anomaly_detection:inference:decision_boundary_unsupervised}
First, we calculate a decision boundary for an unsupervised model. 
Therefore, we apply data augmentation to log messages from the normal training data set $l_n$ only, without applying the data augmentation to the stabilization class $S$.

All values computed by the model for the augmented token sequences are collected to form the distribution $D$.
Hence, we cannot put token sequences in our neural network, we utilize the embedding sequence $e'_i$, which corresponds to the token sequence $t'_i$, as input.

\begin{equation}
D = (z_i : e'_i \in l_n, 0 \leq i \leq |l_n).
\end{equation}

Here $z_i$ is the output of the model for each log message, and the short form for $z_i=\Phi(\vec{e'_i}, y_i, \Theta)$ as we described in chapter~\autoref{subsec:anomaly_detection:anomaly_detection_model:network}.
This distribution is the basis for the decision boundary $\epsilon$. 
We calculate the decision boundary by choosing the i-th percentile $p$ of the distribution $D$ and multiplying this value by a regulator variable $\beta$.

\begin{equation}
\epsilon(D, p,\beta) = D_p \cdot \beta
\end{equation}

This formula calculates the final decision boundary from the given distribution for the deviated token sequences. 
The parameter $p$ is a bias regulator because there is the possibility that the model calculates outliers for a few normal augmented log messages. 
These outliers can occur when rare normal log messages resemble the stabilization class from the training data and therefore have high anomaly scores. 
Hence, we use the i-th percentile as the base value for $\epsilon$.
With $\beta$, we can control whether we want to allow more deviations for normal data during the inference phase.

$p$, $\beta$, and $\alpha$ represent three general variables. 
However, our research across various data sets and models indicates that setting $\alpha=1$, $p=95$, corresponding to the 95th percentile, and $\beta=2$, achieves an almost optimal configuration. 
Setting $\alpha=1$ means that we only exchange one token per log message with a masking token $\mathcal{M}$.

\subsection{Calculating a Decision Boundary for a Weak-Supervised Model}
\label{subsec:anomaly_detection:inference:decision_boundary_weak_supervised}
The calculation of the decision boundary for a weakly supervised model is pretty similar to the calculation of the decision boundary for the unsupervised model. 
Data augmentation is applied to the positive data set, which is the normal data set in this learning strategy, designated as $l_n$, to create a distribution $D_n$ of anomaly scores.
For this purpose, the neural network is supplied with the embedding sequences $e'_i$ from the augmented token sequences, enabling it to generate scores for each log message in the positive training data set.

\begin{equation}
D_n = (z_i : e'_i \in l_n, 0 \leq i \leq |l_n).
\end{equation}

We calculate the base value $p_{95}^n$ for the decision boundary by choosing the 95th percentile of the distribution $D_n$.

The same process is applied to the unknown training data set $l_u$. 
It is known to contain both classes, yet the exact number of anomalies versus normal log messages remains unknown.

\begin{equation}
D_u = (z_i : e'_i \in l_u, 0 \leq i \leq |l_u).
\end{equation}

Typically, the values in the distribution $D_u$ are expected to be higher on average compared to the normal data distribution, since these data are associated with the label $1$ and include anomalies. 
Consequently, we calculate the 95th percentile $p_{95}^u$ for $D_u$ and use the average of this percentile and $p_{95}^n$ as the decision boundary.

\begin{equation}
\epsilon=\begin{cases}
\frac{p_{95}^n + p_{95}^u}{2} & p_{95}^n < p_{95}^u \\
p_{95}^n & else
\end{cases}
\end{equation}

In the uncommon scenario where $p_{95}^n \geq p_{95}^u$, $p_{95}^n$ is utilized as the decision boundary, ensuring that the decision boundary is large enough for normal data during inference.

\subsection{Calculating a Decision Boundary for a Supervised Model}
\label{subsec:anomaly_detection:inference:decision_boundary_supervised}
To determine the decision boundary for the supervised training model, data augmentation is carried out on both the normal training dataset $l_n$ and the abnormal training dataset $l_a$.

All values calculated by the trained model for the augmented token sequences in $l_n$ are collected to form the distribution $D_n$.
As in the previous section, we cannot input token sequences directly into our neural network and make use of the embedding sequence $e'_i$.

\begin{equation}
D_n = (z_i : e'_i \in l_n, 0 \leq i \leq |l_n).
\end{equation}

Similarly, the distribution $D_a$ is computed for every augmented token sequence within $l_a$ in the abnormal training data set.

\begin{equation}
D_a = (z_i : e'_i \in l_a, 0 \leq i \leq |l_a).
\end{equation}

In each of the equations, $z_i$ represents the model output for each log messages, and is the short form of $z_i=\Phi(\vec{e'_i}, y_i, \Theta)$.
Both distributions are the basis for the decision boundary $\epsilon$. 
We calculate the decision boundary by choosing the 95th percentile $p_{95}^n$ of the distribution $D_n$ and the 5-th percentile $p_5^a$ of the distribution $D_a$.

We use the 95th percentile for the distribution $D_n$ as in the calculation of the decision boundary for unsupervised training. 
This filters out possible outliers of the anomaly scores that may occur due to the data augmentation process.
For the distribution of anomaly scores from the augmented abnormal training data, we utilize the 5th percentile. 
Given that the anomalies were effectively distinguished and assigned high scores during training, our goal is to identify the lowest scores among these anomalies. 
We select the 5th percentile as it accounts for anomalies that may have been assigned lower values. 
Our aim is to exclude these outliers with excessively low scores.

From this we derive two base values, $p_{95}^n$ and $p_5^a$.
In case $p_{95}^n < p_5^a$ we chose the mean of both values as $\epsilon$ and otherwise we choose $p_{95}^n$ as $\epsilon$.

\begin{equation}
\epsilon=\begin{cases}
\frac{p_{95}^n + p_{5}^a}{2} & p_{95}^n < p_5^a \\
p_{95}^n & else
\end{cases}
\end{equation}

This guarantees that most of the normal data fall below the decision boundary. 
By examining the minimum anomaly scores of the abnormal data after augmentation, taking the mean value of $p_{95}^n$ and $p_5^a$ proves adequate for accurate differentiation during inference. 
Should the anomalies present lower values, we apply the value of the normal distribution, given the higher frequency of normal data occurrences compared to anomalies. 
However, such scenarios are expected to be infrequent in practice and should be considered as a fallback plan.

\paragraph{Summary}
Anomaly detection methods have become increasingly important in ensuring the reliable and stable operation of IT systems, including their serviceability.
However, existing anomaly detection methods are applied under constrained assumptions for the final anomaly decision that included the decision boundary.
Therefore, we proposed three ways to address the current limitations of anomaly detection methods.
Unlike other approaches, it calculates its decision boundary for the final decision by exploring the behavior of the model based on augmented training data. 
With data augmentation, we simulate deviations in log data that occur from service updates over time.

Even if we can simulate deviations of the normal log lines with the help of our data augmentation and thus detect anomalies efficiently on inference stage, we see as a limitation that a fundamental concept drift in the data pushes the method to its limits and thus leads to misclassifications.
However, we can keep up with the optimal decision boundary, which can only be calculated utilizing the anomaly examples available on the inference stage. 
\cleardoublepage

\chapter{Root Cause Analysis For Failure Investigation}
\minitoc
\label{ch:root_cause_analysis}

A key use case for AIOps is to support developers in detecting and resolving system failures by analyzing their root causes. 
Due to the heavy use of logging in modern systems, many works have thus focused on log anomaly detection~\cite{zawawy2010log,nedelkoski2020self,wittkopp2021loglab,wittkopp2022pull,hamooni2016logmine,korzeniowski2022landscape,lu2017log}. 
However, faults often propagate extensively within systems before actual failures occur.
This can result in large amounts of anomalies being detected across various different services.
Presenting DevOps teams with hundreds of potentially anomalous log lines is not a sensible approach when the main concern is to quickly understand the root cause of a failure and then mitigate it in further updates.

In complex IT service architectures, the root cause of a failure is often not determined by a single event.
Instead, it must be described by a set of faults and system states that are often distributed across many different services~\cite{wittkopp2024logrca}.
This set can describe the course of action~\cite{zawawy2010log}.
The key challenge in \emph{Root Cause Analysis} is to identify the minimal set of relevant information required to understand the core problem.
This is especially relevant for log data, which can contain large amounts of noise and irrelevant information.

Several recent works explore root cause analysis for log data~\cite{korzeniowski2022landscape}.
However, a significant drawback of these studies is that they were developed for specific systems, where the number of possible root causes was predetermined~\cite{lu2017log,lu2019ladra,sharp2016semi}. 
This limitation makes it difficult to apply these methods to complex, constantly evolving IT systems. 
In most real-world services, not all root causes are known in advance.

To facilitate a quick understanding of failures in distributed IT services, in this chapter, we propose a log-based root cause analysis method.
Our method determines a set of log lines that together describe the root cause of a failure.
We operate under the assumption that the failure time is already identified, but the exact failure does not have to be known.
For this, we rank all log lines that were generated within a \ac{ITW} prior to the failure by relevance.
DevOps teams can then dynamically investigate different numbers of \ac{RCC}, resulting in a desired set of \acp{RCC}.
These root cause candidates are expected to be temporarily ordered and causally related~\cite{9529498}.

This chapter is based on the following publications:
\begin{itemize}
    \item LogRCA: Log-based Root Cause Analysis for Distributed Services~\cite{wittkopp2024logrca}
    \item Progressing from Anomaly Detection to Automated Log Labeling and Pioneering Root Cause Analysis~\cite{wittkopp2023progressing}   
\end{itemize}


Section~\ref{sec:rca:from_ad_to_rca} explains the problem and challenges.
Section~\ref{sec:rca:pu_learning} describes how we formulate root cause analysis as a PU learning problem.
Section~\ref{sec:rca:balancing} proposes a data balancing approach to increase performance in rare root causes.
Section~\ref{sec:rca:method} presents our root cause analysis method in detail.
We will conduct the evaluation of our method in the next chapter in~\autoref{sec:evaluation:root_cause_analysis}.

\section{From Anomaly Detection to Root Cause Analysis}
\label{sec:rca:from_ad_to_rca}
Since a single failure can propagate through multiple services in complex distributed systems, traditional anomaly detection approaches can alert for very large numbers of anomalous log lines, limiting their usefulness for DevOps teams.

The main difference between log anomaly detection and log-based \ac{RCA} is that anomaly detection aims to identify anomalous behavior in IT services by selecting individual and often contextually unrelated log lines. 
Log-based root cause analysis, on the other hand, aims to select a minimal set of contextually related log lines to support development or operations teams in better understanding the actual root cause of a failure~\cite{wittkopp2023progressing}.
This is a nontrivial task, as the objective of root cause analysis is to provide information on the actual course of action~\cite{zawawy2010log} of the failure.
For example, there might be important log lines that help a team understand the root cause, which would not have been detected by traditional anomaly detection because they describe normal behavior.

\begin{figure}[h]
\centering
\includegraphics[width=0.9\columnwidth]{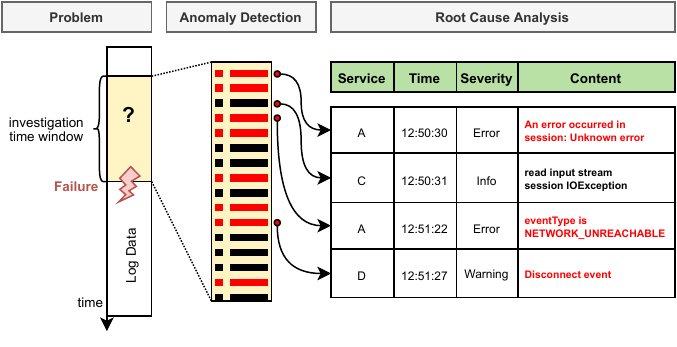}
\caption{Illustration of the problem and the desired solution for root cause analysis.}
\label{fig:rca:problem_description}
\end{figure}

The problem and the desired solution are illustrated in~\autoref{fig:rca:problem_description}, which includes four log lines. The initial log line indicates that an error occurred in a session of service A. This results in an IOException and eventually leads to a disconnect event, representing the actual failure.
Furthermore, it shows that the anomaly detection itself would show more anomalies than is necessary to understand the root cause.
Furthermore, the IOException does not need to be an anomaly in some cases but helps to understand the root cause.
Due to the lack of information about the characteristics of a root cause, the process of root cause analysis poses three main challenges:

\begin{enumerate}
    \item We do not know which log lines within an investigation time window represent the root cause of the failure. This means that we must treat all log lines prior to a failure as possible root cause candidates, which leads to a large number of incorrectly labeled data during training.
    \item Training data is often very unbalanced, which means that some failures have been common in the past, while others have rarely or never occurred.
    This can harm model performance due to training biases.
\end{enumerate}

\section{Root Cause Analysis as a PU Learning Problem}
\label{sec:rca:pu_learning}
As a prerequisite for our root cause analysis method, we require the actual detection of a failure.
This can be determined through specific logs (for example, a dedicated \emph{disconnect} event like in our evaluation data) or through other monitoring systems.
We do not need to know the exact failure. It is sufficient to recognize that a failure has occurred, which allows us to also analyze unknown failures.
DevOps then have to determine a reasonable investigation time window size, which describes the amount of time prior to the failure in which we expect the root cause log lines to occur. 

\begin{figure}[h]
    \centering
    \includegraphics[width=1\columnwidth]{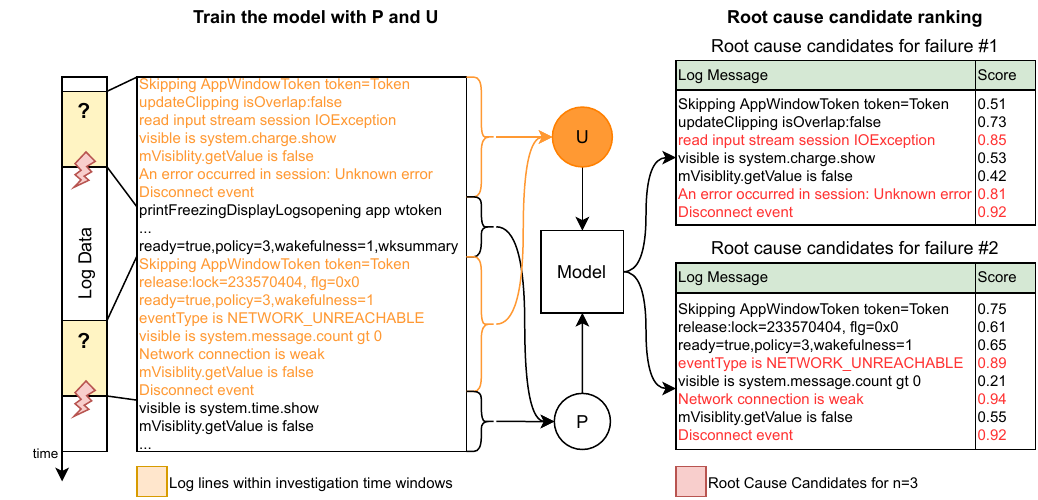}
    \caption{Illustrating the training process with false labeled data and the desired outcome for n=3.}
    \label{fig:rca:problem_example}
\end{figure}

The example in~\autoref{fig:rca:problem_example} contains two of these investigation time windows (in yellow) prior to their respective failures (red flash). Here, $n$ represents the quantity of log lines that the DevOps team wishes to display.

We formulate our training process as a PU learning problem as we have described in~\autoref{subsec:background:concepts:pu_learning}. 
Thus, our method learns the distribution of positive samples (in our case log lines outside the investigation time windows) to assess unknown samples (in our case log lines within the investigation time windows)~\cite{liu2002partially,zhu2009introduction,bekker2020learning,liu2003building}.
All orange log lines have been assigned to the unknown class $\mathcal{U}$, while black log lines are assigned to the normal class $\mathcal{P}$.

In other words, we train our model based on two classes: A \emph{positive} class $\mathcal{P}$ with normal data and an \emph{unknown} class $\mathcal{U}$, which contains both normal log lines and root cause log lines.
This leads to a training setup with a large number of inaccurate labels, as most samples in $\mathcal{U}$ are considered normal log lines.
As we do require information on when a failure occurred, as well as the size of the investigation time window to create the two classes, we can speak of a weakly supervised learning environment~\cite{zhou2018brief}.
Therefore, we can utilize our anomaly detection model in a weakly supervised training scenario, which we presented in~\autoref{sec:anomaly_detection:anomaly_detection_model:model_training} to calculate the relevance of the log line as \aclp{RCC}. 
However, we need to modify certain aspects of the training data and also analyze the model's output in a different manner.

As we do not previously know how many log lines are required to fully understand a particular and often entirely unknown root cause, our method does not decide how many log lines should be presented.
Therefore, DevOps teams are responsible to experiment with different numbers of \ac{RCC} to be displayed, which correspond to the $n$ log lines with the highest scores in each investigation time window.

\section{Balancing Data to Boost Performance on Rare Cases}
\label{sec:rca:balancing}

To improve our performance for rare or even unknown failures, we employ a training data balancing strategy based on automatic clustering.

\subsection{Imbalanced Training Data}
\label{sec:rca:balancing:imbalance_training_data}

In practical settings, a particular system failure can be caused by a variety of different root cases.
For example, unexpected disconnect events of clients in a mobile computing setting can be caused by a variety of reasons, ranging from weak wireless network connection, over actual software crashes, to power supply issues.
However, some of these causes may be much more common than others, resulting in highly imbalanced training data.
If there are only very few samples of a specific root cause available within $\mathcal{U}$, we cannot expect the machine learning model deployed to sufficiently learn the distribution of the corresponding root cause log lines.
The bias towards majority classes will result in a model that may struggle to distinguish actual root cause log lines from log lines in the normal class $\mathcal{P}$ and hence assign a relatively low root cause score.

Although handling class imbalance in deep learning is a well-researched problem~\cite{johnson2019class_imbalance_survey}, in our scenario, we operate in a semi-supervised setting.
Therefore, we neither know the class data distribution nor how many classes there are.
To deal with this, we estimate the number of different root causes by automatic clustering and then balance the data with the goal of improving performance in underrepresented root causes.

\subsection{Balancing Through Automatic Clustering}
\label{sec:rca:balancing:clustering}

To balance the training data, we use automatic clustering~\cite{jose2016automatic} to obtain an estimate of the number of root causes and their occurrences in the training data set.
Each cluster estimates a different type of root cause.
After the estimation, we balance the training data so that rare root causes are weighted more strongly in the training process, although not as strong as common root causes.

\begin{figure}[h]
\centering
\includegraphics[width=0.8\columnwidth]{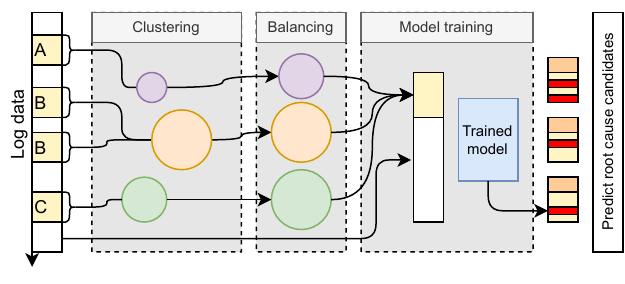}
\vspace{-5mm}
\caption{Balancing the training data.}
\label{fig:rca:balancing}
\end{figure}

Figure~\ref{fig:rca:balancing} illustrates the balancing procedure. 
As a first step, investigation time windows, depicted in yellow on the left, are encoded in a vector $\vec{w}$, by utilizing the meta-information of each log line in those windows. 
Specifically, the services of each log line are utilized, so each vector represents the services involved in each root cause.
Therefore, we create a vector $\vec{w}$, with a dimensionality equal to the number of different services in all investigation time windows, represented through $x_i$. 
This vector serves as input for the clustering:
$$dim(\vec{w}) = |\{\forall x_i \in \mathcal{L}: unique(x_i)\}|$$
Thereby $\mathcal{L}$ represents all log lines in all investigation time windows.

Each cluster within the automatic clustering output estimates a specific root cause.
This step is not accurate, but still gives a good estimation to balance the training data.
In our example, the root cause 'B' occurs two times and the root cause 'A' and 'C' occurs one time each.
The size of the circle illustrates the number of log lines within the corresponding investigation time windows of a specific cluster.
The log lines from the investigation windows of 'B' are therefore combined in one cluster.

During the balancing step, the number of log lines in each cluster is calculated: 
\begin{equation}
N = \left\{ \sum_{t \in T_{k_i}} \left| \{ l_i \in t \} \right| : k_i \in \mathcal{K} \right\} 
\end{equation}
Where $\mathcal{K} $ is the set of all clusters,
$T_{k_i}$ is the set of investigation time windows assigned to each cluster $k_i$.
Thus, $N$ becomes a set of the sizes of all clusters and $min(N)$ is the number of log lines in the smallest cluster and $max(N)$ is the number of log lines in the largest cluster.
The numbers of log lines for each cluster in $\mathcal{K}$ are then normalized between $\frac{max(N)}{2}$ and $max(N)$. 
This means that the smallest cluster will have half as many log lines as the largest cluster. 
We achieve this by replicating the log lines from the smaller clusters until the desired number is reached.
The target size $t(\cdot)$ of each cluster $k_i$ is then calculated by the following equation:

\begin{equation}
t_{|k_i|} = (\frac{|k_i| - min(N)}{max(N) - min(N)} \cdot \frac{max(N)}{2}) + \frac{max(N)}{2})
\end{equation}

With $|k_i|$ being the number of log lines in each cluster, we want to normalize between the desired range of $\frac{max(N)}{2}$ and $max(N)$.

Overall, balancing the training data is a critical step, as it allows the model to develop comprehensive training of each possible root cause but still train properly for the most occurring root causes, since they should also be the ones that occur very often in general. 
Subsequently, the class $\mathcal{U}$ is now balanced, whereas the class $\mathcal{P}$ is not touched during this step.

\section{Method for Investigating Root Causes}
\label{sec:rca:method}

In this section, we present our method for detecting a set of root cause log lines by examining an investigation time window prior to a failure through PU learning and finding the main involved services in for each root cause, to only keep root cause log lines and filter noise and irrelevant anomalies.
Our approach allows us even to identify the causes of rare or entirely new failures.

\subsection{Utilizing Anomaly Detection to Score the Log Lines}

We use the scoring of our anomaly detection model for weak supervised learning, including its preprocessing steps, presented in~\autoref{sec:anomaly_detection:anomaly_detection_model:preprocessing} and the objective function for weak supervised training presented in~\autoref{subsec:anomaly_detection:anomaly_detection_model:objective_function}.

\begin{figure}[htbp]
\centering
\includegraphics[width=0.8\columnwidth]{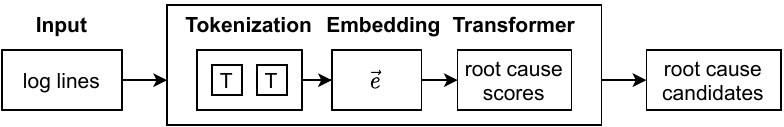}
\caption{Steps for selecting root cause candidates.}
\label{fig:rca:process}
\end{figure}

Figure~\ref{fig:rca:process} illustrates the different steps: After data preprocessing, the transformer model, trained using PU learning and our objective function, determines a root cause score for each log line.

As illustrated in~\autoref{fig:rca:process}, we apply tokenization (see~\autoref{subsec:background:log_data:processing}) to each log line using the seperators: \texttt{,} and \texttt{:} and \texttt{whitespace}. 
The resulting sequence of tokens is then further processed by replacing certain tokens with placeholders that adequately represent the original token while preserving relevant information. 
A placeholder token \texttt{[IP]} is introduced for IP address values, \texttt{[NUM]} is used for any number greater or equal to 10, \texttt{[HEX]} is utilized for hexadecimal numbers and \texttt{[ADDR]} for internal addresses of the application.
Lastly, we truncate the token sequences obtained from the log events to a fixed length and pad smaller sequences with the special token \texttt{[PAD]} to match the required input size.
Therefore, we set the parameters for the preprocessing pipeline of the anomaly detection model (seen~\autoref{sec:anomaly_detection:anomaly_detection_model:preprocessing}) as follows:
\begin{itemize}
    \item R = [IP, NUM>10, HEX, ADDR]
    \item C = [\texttt{,} and \texttt{:} and \texttt{whitespace}]
    \item S = 10
\end{itemize}

Using the embeddings of the individual tokens of the log message, we can train our model to generate a score for each log line. 
Instead of calculating and using the decision boundary for inference, we obtain the scores for each log line in each investigation time window. 
Consequently, our anomaly detection model serves as a foundation for \ac{RCA}.
With the scores for each log message within each investigation time window, we can further analyze them to identify the root cause and avoid displaying too many anomalies to DevOps teams.

\subsection{Analyze the Main Services for Root Cause Log Lines}
As described in~\autoref{sec:rca:from_ad_to_rca}, we cannot automatically determine how many log lines are required to fully understand a root cause.
Because of this, the DevOps teams must decide how many log lines should be returned for root cause analysis, setting the parameter $n$.
Based on the previously calculated root cause scores, we then return the $n$ log lines with the highest scores in chronological order, after finding the main services involved.

To find the main services involved, we focus on the log lines with the highest scores by selecting the top $n \cdot 10$ log lines of each designated investigation time window. 
Given that $n$ represents the number of log lines a DevOps team developer requires to understand the failure, we ensure that an adequate number of log lines are chosen for the filtering process by using $10$ times this number.
This targeted selection allows us to filter out the noise and concentrate on the most significant log lines with the highest scores. 
With these high-scored log lines, we perform clustering based on the associated service to each log line, similar to our approach during the training data balancing step. 
This leads to clusters where the log lines in the \ac{ITW} are generated by identical or extensively intersecting services. 
Because these clusters are derived solely from the top-ranked log lines, numerous anomalies appear within the clusters, with the root cause log lines being a part of them.
Therefore, the intersection of the services in a cluster should represent the main services involved in the root cause, and we can filter out log lines that are not generated by the main involved services.

\begin{figure}[htbp]
    \centering
    \includegraphics[width=0.9\columnwidth]{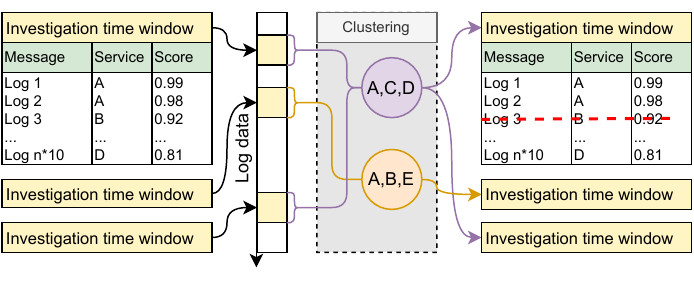}
    \caption{Find main involved services for each root cause.}
    \label{fig:rca:main_services}
\end{figure}

Figure~\ref{fig:rca:main_services} illustrates the clustering process of the investigation time windows. Each cluster consists of all the services involved in all the \acp{ITW} within this cluster. 
Thus, this example shows that service \textit{B} is not in the violet cluster because it does not occur in the third \ac{ITW}.
Thus, we remove all log lines in this cluster that originate from this service, no matter which \ac{ITW} they belong to.
This step ensures that we only keep logs from services that are involved in the root cause, because every cluster indicates a different root cause.
Even if a log line has a high score, it could be deleted through this process.
This log line could be an anomaly by high chance, but is not helpful for the root cause analysis, since it is probably a side effect of the failure and comes not from the main involved services that are important for the DevOps team to fix the failure in future updates.

\cleardoublepage

\chapter{Evaluation}
\minitoc
\label{ch:evaluation}

In this chapter, we evaluate the methods of our three layers, presented in~\autoref{ch:data}, \autoref{ch:anomaly_detection} and \autoref{ch:root_cause_analysis}.


First, we introduce the data sets employed for evaluation in~\autoref{sec:evaluation:datasets}. 
Following this, we evaluate the first layer log investigation and address RQ 1.1 in~\autoref{sec:evaluation:log_investigation}. 
Subsequently, we assess our log anomaly detection method and respond to RQ 1.2 in~\autoref{sec:evaluation:anomaly_detection}. 
Lastly, we analyze our root cause analysis method and answer RQ 1.3 in~\autoref{sec:evaluation:root_cause_analysis}. 
The main RQ is answered in the conclusion.

\section{Data Sets}
\label{sec:evaluation:datasets}
To assess the effectiveness of our proposed methods, we perform evaluations using three publicly accessible data sets and two proprietary industry data sets.
For anomaly classification, log labeling, and anomaly detection, we use the three publicly available data sets.
Furthermore, we use one of the two proprietary data sets exclusively for anomaly detection and the other for root cause analysis.

In subsequent subsections, we initially describe the data sets, their origins, and their contents. Following this, we illustrate the various characteristics of the data sets by detailing the distribution of log lines across different training and test splits.
Additionally, we present the number of unique log lines in the corresponding training and test split before and after preprocessing to highlight the complexity of the tasks for our models.

\subsection{Public Available Data Sets}
\label{subsec:evaluation:datasets:public}
The three publicly available data sets are well known in the log analysis area and are recorded on different large-scale computer systems~\cite{oliner2007datasets}.
These data sets were manually labeled by experts.
The names of the three data sets are as follows: BGL, Thunderbird, and Spirit.
In the following, we present the origin of the data sets.
    
\begin{itemize}
    \item The \emph{BGL} data set is collected from a BlueGene/L supercomputer system at Lawrence Livermore National Laboratory (LLNL) and contains 4,747,963 log messages of which 7.3 \% are abnormal.
    The data set contains a total of 348,460 anomalies.
    The log messages originate from a period of ~214 days, with on average 0.25 log messages per second.
    The logging period begins on 2005-06-03 at 22:42:50 and concludes on 2006-01-04 at 16:00:05.

    \item The \emph{Thunderbird} data set is collected from a Thunderbird supercomputer system at Sandia National Labs (SNL) in Albuquerque and contains 211,212,192 log messages. 
    For the evaluation we selected the first 5,000,000 log messages of which 4.5 \% are abnormal. 
    The data set contains a total of 226,287 anomalies. 
    They account for a period of ~9 days, with on average 6.4 log messages per second.
    The logging period begins on 2005-11-09 at 08:05:01 and ends on 2005-11-18 at 09:09:39.

    \item The \emph{Spirit} data set is collected from a Spirit supercomputer system in SNL and contains more than 272 million log messages. Again, we selected the first 5 million log messages, of which 15.3 \% are abnormal. 
    Thereby, 10 log lines had no content; therefore, we use only 4,999,90 log lines of this data set.
    The data set contains a total of 764,890 anomalies. 
    The extracted log messages cover a period of ~48 days, with on average 1.2 log messages per second. 
    The logging period begins on 2005-01-01 at 08:00:00 and ends on 2005-02-19 at 04:15:58.
\end{itemize}


\begin{table}[h!]
    \centering
    \small
        \begin{tabular}{c|cc|cc}
        \toprule
        \% Split & Train Normal & Train Abnormal & Test Normal & Test Abnormal \\
        \midrule
        \multicolumn{5}{c}{BGL} \\
        \midrule
        20 & 717,777 & 224,921 & 3,647,256 & 123,539 \\
        40 & 1,657,454 & 227,943 & 2,707,579 & 120,517 \\
        60 & 2,596,216 & 231,879 & 1,768,817 & 116,581 \\
        80 & 3,468,617 & 302,177 & 896,416 & 46,283 \\
        \midrule
        \multicolumn{5}{c}{Thunderbird} \\
        \midrule
        20 & 999,974 & 26 & 3,773,739 & 226,261 \\
        40 & 1,993,381 & 6619 & 2,780,332 & 219,668 \\
        60 & 2,905,590 & 94,410 & 1,868,123 & 131,877 \\
        80 & 3,836,865 & 163,135 & 936,848 & 63,152 \\
        \midrule
        \multicolumn{5}{c}{Spirit} \\
        \midrule
        20 & 710,093 & 289,905 & 3,525,007 & 474,985 \\
        40 & 1,538,199 & 461,797 & 2,696,901 & 303,093 \\
        60 & 2,275,122 & 724,872 & 1,959,978 & 40,018 \\
        80 & 3,237,143 & 762,849 & 997,957 & 2,041 \\
        \bottomrule
        \end{tabular}
    \caption{Overview of the 3 publicly available data sets and the distribution of normal and abnormal log lines in different training and evaluation scenarios.}
    \label{tab:evaluation:datasets:public:split}
\end{table}

~\autoref{tab:evaluation:datasets:public:split} shows the number of normal log lines and anomalies in the data sets, considering training evaluation splits ranging from 20\% to 80\%.
When 20\% of the data set is allocated for training, the remaining portion is utilized for evaluation.
However, because numerous log lines can appear multiple times, it is crucial to know the number of unique log lines present in the respective splits.

\begin{table}[h!]
    \centering
    \small
        \begin{tabular}{c|cc|cccc}
        \toprule
        \% Split & \makecell{Train \\ unique \\ normal}  & \makecell{Train \\ unique \\ abnormal} & \makecell{Test \\ unique \\ normal} & \makecell{Test unknown \\ unique \\ Normal} & \makecell{Test \\ unique \\ abnormal} & \makecell{Test unknown \\ unique \\ abnormal} \\
        \midrule
        \multicolumn{7}{c}{BGL} \\
        \midrule
        20 & 37,156 & 5,994 & 292,983 & 272,202 & 43,015 & 43,007 \\
        40 & 121,134 & 6,095 & 256,388 & 188,224 & 42,942 & 42,906 \\
        60 & 144,784 & 6,378 & 184,058 & 164,574 & 42,640 & 42,623 \\
        80 & 240,942 & 27,421 & 72,117 & 68,416 & 23,028 & 21,580 \\
        \midrule
        \multicolumn{7}{c}{Thunderbird} \\
        \midrule
        20 & 150,717 & 17 & 428,290 & 418,972 & 15 & 15 \\
        40 & 346,159 & 23 & 236,858 & 223,530 & 12 & 9 \\
        60 & 388,879 & 31 & 196,489 & 180,810 & 4 & 1 \\
        80 & 496,892 & 32 & 85,422 & 72,797 & 2 & 0 \\
        \midrule
        \multicolumn{7}{c}{Spirit} \\
        \midrule
        20 & 402,545 & 168 & 1,576,096 & 1,463,871 & 830 & 823 \\
        40 & 644,381 & 388 & 1,358,175 & 1,222,035 & 621 & 603 \\
        60 & 949,085 & 544 & 1,055,383 & 917,331 & 465 & 447 \\
        80 & 1,514,253 & 826 & 478,920 & 352,163 & 180 & 165 \\
        \bottomrule
        \end{tabular}
    \caption{Overview of the unique normal and abnormal log lines in the 3 publicly available data sets for different splits and number of unknown unique normal and abnormal log lines in the evaluation sets.}
    \label{tab:evaluation:datasets:public:split_unique}
\end{table}

~\autoref{tab:evaluation:datasets:public:split_unique} presents the number of unique log lines for each split, along with the number of unique log lines that are absent from the respective training data set.
Here, we observe that even with 80\% of the data allocated for training, the BGL data set still contains numerous entirely unknown anomalies, whereas the Thunderbird data set no longer has any unknown anomalies. 
These metrics are crucial for supervised training. 
Additionally, regardless of the split, there are always more new unique normal log lines. 
This is significant for unsupervised training, as it necessitates learning a comprehensive representation of the system's normal state. 
This highlights the challenge of unsupervised learning, where we must identify new, previously unseen log lines as normal, even if they were not part of the training data.

Given that log lines often include numerous numeric and hexadecimal values, we simplify the complexity through preprocessing as detailed in~\autoref{sec:anomaly_detection:anomaly_detection_model:preprocessing}. 
~\autoref{tab:evaluation:datasets:public:split_unique_preprocessed} presents the number of unique log lines for each split after prepreprocessing, along with the number of unknown normal and abnormal log lines in the evaluation data sets.
During preprocessing, each hex value is substituted with <HEX> and any number greater than 10 is replaced with <NUM>, which reduces the variability of different log lines.

\begin{table}[H]
    \centering
    \small
        \begin{tabular}{c|cc|cccc}
        \toprule
        \% Split & \makecell{Train\\unique\\normal\\/w pre.}  & \makecell{Train\\unique\\abnormal\\/w pre.} & \makecell{Test\\unique\\normal\\/w pre.} & \makecell{Test\\unknown\\unique normal\\/w pre.} & \makecell{Test\\unique\\abnormal\\/w pre.} & \makecell{Test\\unknown\\unique abnormal\\/w pre.} \\
        \midrule
        \multicolumn{7}{c}{BGL} \\
        \midrule
        20 & 1,541 & 12 & 47,944 & 47,267 & 2,299 & 2,291 \\
        40 & 3,959 & 50 & 46,155 & 44,849 & 2,293 & 2,253 \\
        60 & 4,521 & 84 & 45,732 & 44,287 & 2,236 & 2,219 \\
        80 & 38,821 & 2,212 & 11,705 & 9,987 & 1,069 & 91 \\
        \midrule
        \multicolumn{7}{c}{Thunderbird} \\
        \midrule
        20 & 29,063 & 7 & 15,081 & 13,608 & 14 & 14 \\
        40 & 31,916 & 13 & 12,032 & 10,755 & 11 & 8 \\
        60 & 35,995 & 21 & 7,910 & 6,676 & 4 & 0 \\
        80 & 39,696 & 21 & 4,043 & 2,975 & 2 & 0 \\
        \midrule
        \multicolumn{7}{c}{Spirit} \\
        \midrule
        20 & 72,863 & 13 & 815,952 & 813,381 & 45 & 36 \\
        40 & 122,783 & 27 & 768,180 & 763,461 & 39 & 22 \\
        60 & 256,403 & 31 & 635,114 & 629,841 & 32 & 18 \\
        80 & 679,558 & 39 & 211,676 & 206,686 & 25 & 10 \\
        \bottomrule
        \end{tabular}
    \caption{Overview of the unique normal and abnormal log lines after preprocessing in the 3 publicly available data sets for different splits and number of unknown unique normal and abnormal log lines in the evaluation sets.}
    \label{tab:evaluation:datasets:public:split_unique_preprocessed}
\end{table}


\subsection{Industry Data Set for Anomaly Detection}
\label{subsec:evaluation:datasets:indutry_anomaly}

In this section, we present the metrics from a proprietary industry data set.
The industry data set comes from the production environment of an IT service and cloud provider.
More specifically, the log data originate from different disk controllers that manage a variety of hard disks.
Thereby, the focus is to ensure a dependable storage service, by detecting anomalies in the log data from the underlying hardware. 
The task is to identify anomalies by only training normal log data from some hard disks with no recorded anomalies and apply the model to new disks without manual optimization.
Given that new anomalies may arise in this environment over time due to the continuous addition of new disks to the cloud service, the model needs to be capable of addressing this task unsupervised.
Hence, we present the metrics that are relevant for an unsupervised training scenario. 
The data set comprises a total of 7,121,274 log lines, of which 6,924,454 are normal and 196,820 are abnormal.

\begin{table}[h!]
    \centering
    \small
        \begin{tabular}{c|c|cc}
        \toprule
        \% Split & Train Normal & Test Normal & Test Abnormal \\
        \midrule
        20 & 1,423,953 & 5,500,501 & 196,519 \\
        40 & 2,797,788 & 4,126,666 & 146,099 \\
        60 & 4,176,771 & 2,747,683 & 100,827 \\
        80 & 5,544,522 & 1,379,932 & 4,4323 \\
        \bottomrule
        \end{tabular}
    \caption{Overview of the industry data sets and the anomaly distribution in different training and evaluation scenarios.}
    \label{tab:evaluation:datasets:industry:ad:split}
\end{table}

~\autoref{tab:evaluation:datasets:industry:ad:split} illustrates the various distributions of training data for each split, along with the corresponding normal and abnormal log lines in the evaluation data.

\begin{table}[h!]
    \centering
    \small
        \begin{tabular}{c|c|cc|cc}
        \toprule
        \% Split & \makecell{Train \\ unique \\ normal} & \makecell{Test \\ unique \\normal} & \makecell{Test \\ unique \\ abnormal}  & \makecell{Test \\ unique \\ normal /w pre.} & \makecell{Test \\ unique \\ abnormal /w pre.}\\
        \midrule
        20 & 465,666 & 1,060,920 & 27 & 493,786 & 21 \\
        40 & 536,947 & 987,394 & 23 & 450,475 & 18 \\
        60 & 895,520 & 614,071 & 18 & 207,040 & 14 \\
        80 & 1,178,875 & 311,169 & 11 & 101,358 & 9 \\
        \bottomrule
        \end{tabular}
    \caption{Overview of the unique log lines in the industry data set and the unique anomaly log line distribution in different training and evaluation scenarios.}
    \label{tab:evaluation:datasets:industry:ad:split_unique}
\end{table}

Given that the total number of log lines does not reflect the variability of the data, we present the distribution of unique normal and unique abnormal log lines for each split, as well as the distribution after preprocessing for both classes in~\autoref{tab:evaluation:datasets:industry:ad:split_unique}. 
We apply the same rules as those applied to publicly available data sets, replacing numbers greater than 10 and hexadecimal values with specific tokens.
The entire data set comprises 1,427,710 unique normal log lines and 28 unique abnormal log lines. After preprocessing, there are 668,287 unique normal log lines and still 28 unique abnormal log lines remaining in the entire data set.
It can be concluded that there are only very few different anomalies in the test data but they occur frequently. 
Furthermore, the anomaly rate for the data set is between 3 and 3.5 \%, in relation to the respective split.

\subsection{Industry Data Set for Root Cause Analysis}
\label{subsec:evaluation:datasets:industry_rca}

The log data utilized for our \acl{RCA} were gathered by an IT company that tests their Android devices and software during field tests.
Consequently, they collected logs from Android smartphones that encountered failures.
As the data originate from an industry production setting, they cannot be made available in this thesis, but we describe their characteristics in this section.

The log data was generated by 46666 different services running on those Android devices. 
The data collected comprises 44.3 million log lines, of which 7.7 million log lines are unique, whereby each log line consists of 8.8 tokens on average.
After replacing specific tokens with the placeholders listed in~\autoref{sec:rca:method}, we are left with 0.7 million unique log lines.
Specifically, we substituted IP addresses, hexadecimal values, numbers exceeding 10, and internal addresses with respective placeholder tokens.
In total, the data covers 398 failures of Android devices, of which for 80 failures the root cause log lines are provided by the company's DevOps team for evaluation. 
For each failure, we define an investigation time window of 3 seconds prior to the failure, during which we expect to find the relevant log lines representing the root cause.

\begin{figure}[h]
    \centering
    \includegraphics[width=0.7\linewidth]{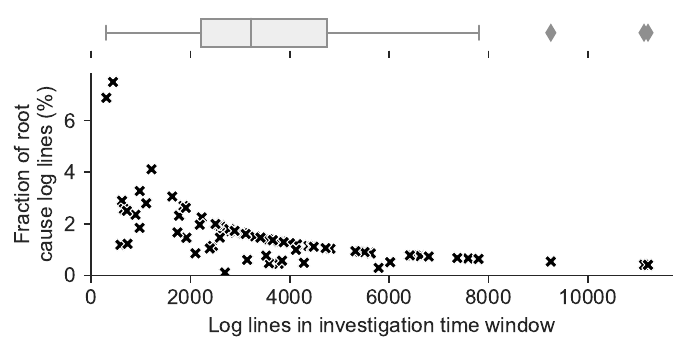} 
    \caption{Number of log lines in the \acl{ITW} and their fraction of root cause log lines.}
    \label{fig:evaluation:rca:dataset}
\end{figure}

Figure~\ref{fig:evaluation:rca:dataset} shows the distribution of the root cause log lines compared to all log lines in each of the 80 investigation time windows.
The minimal set of root cause log lines for each failure are labeled by industry experts.
Hence, the figure illustrates the fraction between root cause log lines and all log lines in the 80 investigation time windows.
Each window contains between 305 and 11209 log lines, each of which contains between 3 and 50 root cause log lines.
Thus, the total number of log lines in the investigation time windows is $\sim300,000$, where outside the windows there are still $\sim44$ millions.


\section{Log Investigation}
\label{sec:evaluation:log_investigation}
First, we provide an evaluation of the \acl{LI} layer within our log analysis architecture, described in~\autoref{sec:log_analysis_layer:architecture}. 
This layer includes labeling and classification. 

\subsection{Log Labeling}
\label{subsec:evaluation:log_investigation:labeling}
This section evaluates our labeling method (see~\autoref{sec:data:autonomous_data_labeling}) along with nine common methods for log anomaly detection and text classification.
We evaluate their effectiveness on three distinct data sets described in~\autoref{subsec:evaluation:datasets:public}, taking into account four different time windows $\delta$. 
These time windows emulate the monitoring system, which can approximately predict the failure time.

\subsubsection{Experimental Setup}
We first explain our experimental setup: data set preparation, benchmark methods we chose, and how we constructed our evaluation data sets.

\textbf{Data Set Preparation.} 
Initially, we utilize the BGL, Thunderbird, and Spirit data sets (see~\autoref{subsec:evaluation:datasets:public}).
We create our data sets by including all abnormal log events as well as their surrounding events within a time window $2*\delta$ in $\mathcal{U}$; all remaining log events are in $\mathcal{P}$. 
The classes of $\mathcal{P}$ and $\mathcal{U}$ are described in~\autoref{subsec:data:autonomous_data_labeling:learning_from_pu} for our autonomous labeling method.
As different systems with different quality of monitoring solutions can only provide failure time windows with varying degrees of accuracy, we investigate the performance at four different time windows $\delta$: $\pm 1000\,ms$ (2s), $\pm 5000\,ms$ (10s), $\pm 10000\,ms$ (20s), and $\pm 20000\,ms$ (40s). 

\begin{table*}[h]
    \footnotesize
	\centering
	\caption{Data sets with their total number of abnormal samples in the ground truth, and number of samples in $\mathcal{U}$ for different time windows ($\delta$).}
	\begin{tabular}{p{2.0cm}p{2.4cm}p{1.8cm}p{1.8cm}p{1.8cm}p{1.8cm}}
		\toprule
        \multirow{2}{*}{Data Set} & Abnormal & \multicolumn{4}{c}{$\mathcal{U}$ for different $\delta$} \\
        \cmidrule{3-6}
                & (ground truth) & $\pm 1,000 ms$ & $\pm 5,000,ms$ & $\pm 10,000 ms$ & $\pm 20,000 ms$ \\
        \midrule
        BGL        & 348,460 & 392,485 & 439,656 & 459,098 & 479,030\\
        Thunderbird& 226,287 & 1,423,207 & 2,368,016 & 2,853,171 & 2,924,896\\
        Spirit     & 764,890 & 1,006,419 & 2,333,800 & 3,251,475 & 3,280,803\\
        \bottomrule
	\end{tabular}
	\label{tab:evaluation:investigation:labeling:datasets}
\end{table*}

\autoref{tab:evaluation:investigation:labeling:datasets} presents the number of samples in $\mathcal{U}$ for each data set.
For small failure time windows, class $\mathcal{U}$ does not deviate too much from the ground truth at BGL and Spirit. 
Consequently, these evaluation data sets contain only very few incorrectly labeled normal log messages.
For Thunderbird, on the other hand, $\mathcal{U}$ is already about six times larger than the ground truth at $\delta=1000\,ms$, meaning it is the hardest data set for this task.

For larger $\delta$, $\mathcal{U}$ grows rapidly in all data sets until a certain limit.
At the largest observed windows, between $\delta=10000$ and $\delta=20000$, only relatively few new samples are added to $\mathcal{U}$.
This is reasonable, because systems often produce significantly more logs when an error occurs, placing them in close temporal proximity.

\textbf{Benchmark Methods.} To obtain a significant and wide benchmark, we compare our method to nine state-of-the-art text classification and anomaly detection methods presented in a recent text classification survey~\cite{kowsari2019text}.
These methods are also investigated in an established survey for anomaly detection in system logs~\cite{he2016experience}.
Namely, we choose Deeplog, Rocchio, Invariants Miner, SVM, Boosting, PCA, Logistic Regression, Decision Tree, and Random Forest as benchmarks.
These methods are explained in more detail in~\autoref{ch:related_work}.
DeepLog, Invariants Miner, and PCA are only able to train on samples in $\mathcal{P}$ to then detect deviations, while all others train on all samples in $\mathcal{P}$ and $\mathcal{U}$.

\textbf{Preprocessing and Model Setup.} The tokenization process as described in \autoref{subsec:data:autonomous_data_labeling:log_processing_pipeline} is applied for all methods to ensure a fair comparison.
Each sequence of tokens $t_i$ is truncated to have a length of 26 for \emph{Thunderbird}, 18 for \emph{Spirit}, and 12 for \emph{BGL}. 
The dimensionality $d$ of our embeddings is set to 128 and the hidden layer is set to a dimensionality of 256.
We initially use Xavier weight initialization for model weights and embeddings.
To train our model, we use a batch size of 1024, a total of 8 epochs, and a dropout rate of 10\%. 
For optimization, we use the Adam optimizer with a learning rate of $10^{-4}$ and a weight decay of $5 \cdot 10^{-5}$.

\subsubsection{Effect of Iterative Training}
Before comparing the performance of our autonomous labeling method to the benchmarks, we analyze the effect of iterative PU learning on all available methods to find out which methods benefit from this technique.
For this, we trained all methods three times on their respective output and observed the model performance after every iteration.
We chose three iterations to detect general trends for each method. 
In practice, the number of iterations would be determined by a suitable stopping condition or constraint.

\begin{figure}[h]
    \centering
    \includegraphics[width=0.6\columnwidth]{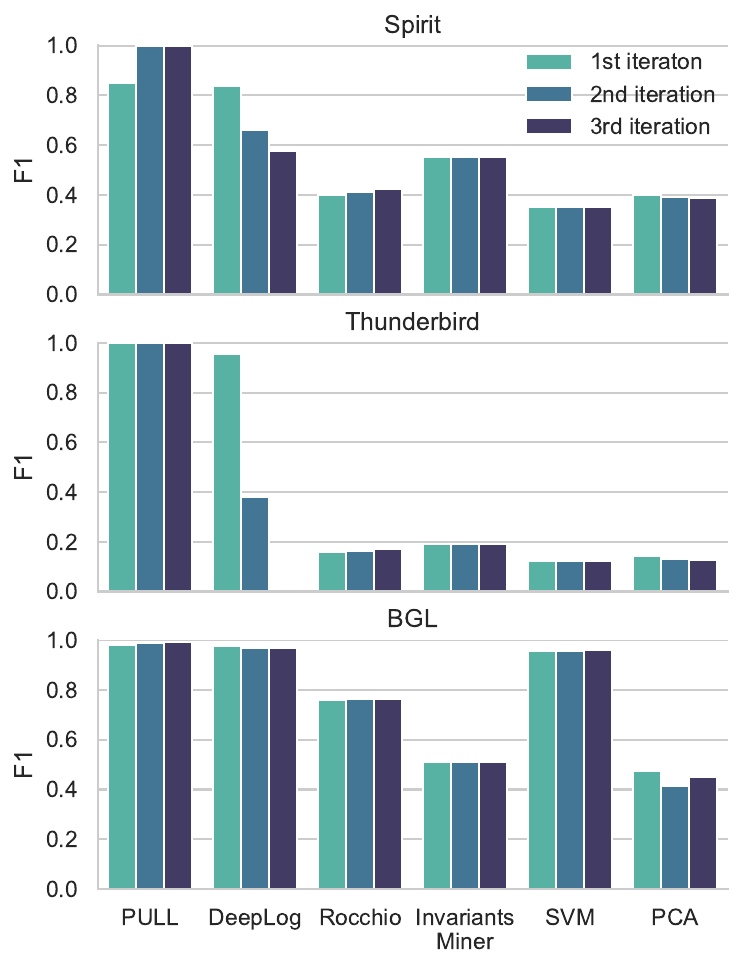}
    \caption{Performance after iterative training for $\delta=\pm 10000ms$. While our method and Rocchio improve on each iteration, other methods degrade (e.g. DeepLog and PCA) or stay the same (e.g. Invariant Miners and SVM).}
    \label{fig:evaluation:log_investigation:labeling:iterative_evaluation}
\end{figure}

\autoref{fig:evaluation:log_investigation:labeling:iterative_evaluation} displays the results of the methods that are influenced by the iterative training strategy.
Performance improvement or degradation was consistent across the three data sets for every method and can be classified into three categories: 

\begin{itemize}
    \item \emph{Performance Improvement.} Besides our method, also the Roccio algorithm's performance improved consistently and often significantly after every iteration.
    For both methods, the range of possible anomalies is becoming smaller in each iteration, as more normal log messages are now labeled as such. 
    Consequently, there are now more real anomalous messages in $\mathcal{U}$.
    Similarly, also the Random Forest performance increased on every iteration, but only at the 4th digit behind the comma.\\ \\ \\
    \item \emph{Performance Degradation.} DeepLog and PCA performance degraded after retraining, meaning that an increasing number of actual anomalies are classified as normal by these methods. 
    As a result, anomalies are now considered normal during further iterations, and thus the model tends to classify similar anomalies as normal as well. 
    \item \emph{No Change in Performance.} Decision Tree, Logistic Regression, and Invariants Miner did not change their performance at all between training iterations, as these methods already obtained their optimal result during the first iteration. 
    The boosting method and the SVM did not change their performance either, but show minor fluctuations in the precision and recall at 4 or 5 digits behind the comma.
\end{itemize}

\begin{adjustbox}{center,label={tab:evaluation:investigation:labeling:results},caption={F1-Scores for autonomous labeling methods for different $\delta$.},float=table}
	{\scriptsize
	
	\begin{tabular}[h!]{p{0.3cm} p{1.2cm} p{0.8cm} p{0.8cm} p{1.0cm}  p{0.8cm} p{0.8cm} p{0.8cm} p{0.8cm} p{0.8cm} p{0.8cm} p{0.8cm}}
	\toprule
 	& & \multicolumn{3}{c}{Learning $\mathcal{P}$} & \multicolumn{7}{c}{Learning $\mathcal{P}$ and $\mathcal{U}$} \\
 	\cmidrule(lr){3-5}
 	\cmidrule(lr){6-12}
	\rotatebox{90}{Dataset} & Metric & \rotatebox{45}{PCA} & \rotatebox{45}{Invariant Miners} & \rotatebox{45}{Deeplog} & \rotatebox{45}{Decision Tree} & \rotatebox{45}{Random Forest} & \rotatebox{45}{SVM} & \rotatebox{45}{Logistic Regr.} & \rotatebox{45}{Boost} & \rotatebox{45}{Rocchio} & \rotatebox{45}{\textbf{Our Method}} \\ [0.5ex]
	
 	\midrule
	\multicolumn{12}{c}{$\delta=\pm 1000ms$} \\
	\midrule
 
 	\multirow{3}{*}{\rotatebox{90}{BGL}} 
 	& Recall     & 1.0000 & 0.9999 & 0.9855 & 1.0000 & 1.0000 & 1.0000 & 0.9999 & 0.9898 & 0.6206 & 0.9954 \\
 	& Precision  & 0.4248 & 0.3425 & 0.9994 & 0.9948 & 0.9666 & 0.9685 & 0.9952 & 0.9918 & 0.8286 & 0.9999 \\
 	& F1-Sco.   & 0.5963 & 0.5102 & \cellcolor{blue!20}0.9924 & \cellcolor{blue!20}0.9974 & \cellcolor{cyan!20}0.9830 & \cellcolor{cyan!20}0.9840 & \cellcolor{blue!20}0.9976 & \cellcolor{blue!20}0.9908 & 0.7096 & \cellcolor{blue!20}0.9977 \\ \midrule
 
	\multirow{3}{*}{\rotatebox{90}{\parbox{1.1cm}{\centering\linespread{0.8}\selectfont thunder bird}}} 
	& Recall     & 1.0000 & 1.0000 & 0.9999 & 1.0000 & 1.0000 & 0.9998 & 0.9998 & 0.9990 & 0.9990 & 0.9990 \\
 	& Precision  & 0.1798 & 0.1003 & 0.9982 & 0.1934 & 0.1865 & 0.1930 & 0.1934 & 0.2020 & 0.2077 & 1.0000 \\
 	& F1-Sco.   & 0.3048 & 0.1824 & \cellcolor{blue!20}0.9990 & 0.3242 & 0.3144 & 0.3235 & 0.3242 & {0.3361} & 0.3440 & \cellcolor{blue!20}0.9995 \\ \midrule
 
 	\multirow{3}{*}{\rotatebox{90}{spirit}} 
 	& Recall     & 1.0000 & 1.0000 & 0.9999 & 1.0000 & 1.0000 & 0.9999 & 0.9994 & 0.9990 & 0.9943 & 0.9995 \\
 	& Precision  & 0.6726 & 0.4092 & 0.9855 & 0.9935 & 0.9238 & 0.9719 & 0.9930 & 0.9945 & 1.0000 & 0.9999 \\ 
 	& F1-Sco.   & 0.8043 & 0.5807 & \cellcolor{blue!20}0.9926 & \cellcolor{blue!20}0.9967 & 0.9604 & \cellcolor{cyan!20}0.9857 & \cellcolor{blue!20}0.9962 & \cellcolor{blue!20}0.9968 & \cellcolor{blue!20}0.9971 & \cellcolor{blue!20}0.9997 \\ 
 	
 	\midrule
	\multicolumn{12}{c}{$\delta=\pm 5000ms$} \\
	\midrule
 
 	\multirow{3}{*}{\rotatebox{90}{BGL}} 
 	& Recall     & 1.0000 & 0.9999 & 0.9898 & 0.9999 & 0.9999 & 0.9999 & 0.9999 & 0.9894 & 0.7614 & 0.9915 \\
 	& Precision  & 0.4214 & 0.3434 & 0.9916 & 0.9753 & 0.9317 & 0.9381 & 0.9755 & 0.9698 & 0.8549 & 0.9984 \\
 	& F1-Sco.   & 0.5930 & 0.5112 & \cellcolor{blue!20}0.9907 & \cellcolor{cyan!20}0.9874 & 0.9646 & 0.9680 & \cellcolor{cyan!20}0.9875 & 0.9795 & 0.8054 & \cellcolor{blue!20}0.9949 \\ \midrule
 
	\multirow{3}{*}{\rotatebox{90}{\parbox{1.1cm}{\centering\linespread{0.8}\selectfont thunder bird}}} 
	& Recall     & 1.0000 & 1.0000 & 0.9999 & 0.9999 & 1.0000 & 0.9999 & 0.9998 & 0.9991 & 0.9998 & 0.9990 \\
 	& Precision  & 0.1801 & 0.1072 & 0.9971 & 0.1540 & 0.1373 & 0.1389 & 0.1546 & 0.1675 & 0.1866 & 1.0000 \\
 	& F1-Sco.   & 0.3053 & 0.1936 & \cellcolor{blue!20}0.9985 & 0.2669 & 0.2415 & 0.2439 & 0.2678 & 0.2869 & 0.3146 & \cellcolor{blue!20}0.9995 \\ \midrule
 
 	\multirow{3}{*}{\rotatebox{90}{spirit}} 
 	& Recall     & 1.0000 & 1.0000 & 0.9998 & 0.9999 & 1.0000 & 1.0000 & 0.9999 & 1.0000 & 0.9943 & 0.9959 \\
 	& Precision  & 0.6249 & 0.4025 & 0.9862 & 0.4829 & 0.3749 & 0.3874 & 0.4829 & 0.4115 & 0.9949 & 0.9999 \\
 	& F1-Sco.   & 0.7691 & 0.5740 & \cellcolor{blue!20}0.9929 & 0.6513 & 0.5453 & 0.5584 & 0.6560 & 0.5830 & \cellcolor{blue!20}0.9946 & \cellcolor{blue!20}0.9980 \\
 	
 	\midrule
	\multicolumn{12}{c}{$\delta=\pm 10000ms$} \\
	\midrule
 	\multirow{3}{*}{\rotatebox{90}{BGL}} 
 	& Recall     & 1.0000 & 0.9999 & 0.9858 & 0.9999 & 0.9999 & 0.9999 & 0.9999 & 0.9851 & 0.7631 & 0.9922 \\
 	& Precision  & 0.4164 & 0.3450 & 0.9674 & 0.9518 & 0.9018 & 0.9090 & 0.9545 & 0.9674 & 0.8185 & 0.9883 \\
 	& F1-Sco.   & 0.5879 & 0.5130 & 0.9765 & 0.9753 & 0.9483 & 0.9523 & 0.9767 & 0.9762 & 0.7898 & \cellcolor{blue!20}0.9941 \\ \midrule
 
	\multirow{3}{*}{\rotatebox{90}{\parbox{1.1cm}{\centering\linespread{0.8}\selectfont thunder bird}}} 
	& Recall     & 1.0000 & 1.0000 & 0.9918 & 0.9999 & 1.0000 & 0.9999 & 0.9999 & 0.9991 & 0.9998 & 0.9990 \\
 	& Precision  & 0.1782 & 0.1070 & 0.9249 & 0.0719 & 0.0666 & 0.0723 & 0.0723 & 0.0797 & 0.1262 & 1.0000 \\
 	& F1-Sco.   & 0.3025 & 0.1933 & 0.9572 & 0.1341 & 0.1248 & 0.1348 & 0.1350 & 0.1476 & 0.2241 & \cellcolor{blue!20}0.9995 \\ \midrule
 
 	\multirow{3}{*}{\rotatebox{90}{spirit}} 
 	& Recall     & 1.0000 & 1.0000 & 0.9999 & 0.9999 & 1.0000 & 1.0000 & 1.0000 & 0.9999 & 0.9944 & 0.9579 \\
 	& Precision  & 0.4256 & 0.3984 & 0.7001 & 0.2220 & 0.2102 & 0.2222 & 0.2221 & 0.2063 & 0.3740 & 0.9240 \\ 
 	& F1-Sco.   & 0.4958 & 0.4254 & 0.8236 & 0.4909 & 0.4735 & 0.4836 & 0.4917 & 0.4887 & 0.5192 & \cellcolor{blue!20}0.9971 \\ 

 	\midrule
	\multicolumn{12}{c}{$\delta=\pm 20000ms$} \\
	\midrule
 	\multirow{3}{*}{\rotatebox{90}{BGL}} 
 	& Recall     & 0.7283 & 0.9999 & 0.9843 & 0.7816 & 0.9999 & 0.9999 & 0.7816 & 0.9926 & 0.6205 & 0.9683 \\
 	& Precision  & 0.3516 & 0.3450 & 0.9619 & 0.2805 & 0.9068 & 0.9066 & 0.2805 & 0.9456 & 0.9986 & 0.9916 \\
 	& F1-Sco.   & 0.4743 & 0.5130 & 0.9730  & 0.4128 & 0.9511 & 0.9510 & 0.4128 & 0.9685 & 0.7654 & 0.9798 \\ \midrule
 
	\multirow{3}{*}{\rotatebox{90}{\parbox{1.1cm}{\centering\linespread{0.8}\selectfont thunder bird}}} 
	& Recall     & 1.0000 & 1.0000 & 0.9454 & 1.0000 & 0.9999 & 0.9999 & 1.0000 & 0.9990 & 0.9990 & 0.9998 \\
 	& Precision  & 0.0774 & 0.1070 & 0.8816 & 0.0705 & 0.0663 & 0.0663 & 0.0705 & 0.0743 & 0.0932 & 0.9990 \\
 	& F1-Sco.   & 0.1437 & 0.1933 & 0.9124 & 0.1317 & 0.1243 & 0.1243 & 0.1317 & 0.1384 & 0.1705 & \cellcolor{blue!20}0.9994 \\ \midrule
 
 	\multirow{3}{*}{\rotatebox{90}{spirit}} 
 	& Recall     & 1.0000 & 1.0000 & 0.9956 & 1.0000 & 0.9999 & 1.0000 & 1.0000 & 0.9999 & 0.9964 & 0.7877 \\
 	& Precision  & 0.2490 & 0.3984 & 0.7157 & 0.2421 & 0.2115 & 0.2113 & 0.2421 & 0.2117 & 0.2692 & 0.9943 \\ 
 	& F1-Sco.   & 0.3987 & 0.4254 & 0.8328 & 0.3898 & 0.3492 & 0.3489 & 0.3898 & 0.3494 & 0.4239 & 0.8790 \\ 
 	\bottomrule
\end{tabular}}
\end{adjustbox}

\subsubsection{Comparison of Methods}

We now assess the performance of our method for autonomous labeling and all baselines in terms of precision, recall, and F1-Score metrics.
We show the best F1-Scores from the iterative training strategy for each method in \autoref{tab:evaluation:investigation:labeling:results}.
F1-Scores greater than 0.99 and 0.98 are colored blue and cyan, respectively. 
Our method outperforms other methods across all data sets, especially for large time windows ($\delta$).
Furthermore, as expected, with increasing $\delta$ and thus the increasing size of $\mathcal{U}$, the performance of all methods tends to decrease.



For $\delta = \pm 1000\,ms$, this task was apparently easy to achieve as simple methods such as logistic regression or decision trees also achieve similar performance. 
An exception is the Thunderbird data set, which is characterized by a large $\mathcal{U}$ class, even for $\delta \pm 1000$: No other method manages to achieve an F1-Score greater than 0.35, while our method reaches 0.999 recall with perfect precision.

For $\delta = \pm 5000\,ms$, we notice that the performance degradations previously observed for most methods on the Thunderbird data set now also start to manifest on the Spirit data set. 
We can see that simple methods tend to have close-to-perfect recall (i.e., they correctly identify all abnormal log messages) but suffer from poor precision, meaning that they fail to identify normal log messages in $\mathcal{U}$ and misclassify them as abnormal. 
Our method, on the other hand, maintains an F1-Score of more than 0.99 across all data sets.
However, in contrast to Thunderbird and Spirit, most methods that learn from $\mathcal{P}$ and $\mathcal{U}$ continue to achieve very good results on the BGL data set.

The greatest drop in performance becomes evident at the time window of $\delta = \pm 10000\,ms$. For the BGL data set, we notice a notable drop in F1-Scores of other methods to 0.97 or less, while only our method maintains a high performance of more than 0.994. 
On the Spirit data set, the F1-Score of most methods drops below 0.5 while our method still reaches 0.997. 
For Thunderbird, the precision of other methods drops to only 0.2 or less, meaning that only one out of five log events labeled as abnormal is actually abnormal. 
Our method, on the other hand, maintains the same performance for $\delta = \pm 10000\,ms$ as for smaller time windows, resulting in an F1-Score of 0.9995.

As shown in~\autoref{tab:evaluation:investigation:labeling:datasets}, the size of $\mathcal{U}$ does not change significantly when increasing the window size from $\delta = \pm 10000\,ms$ to $\delta = \pm 20000\,ms$, but we still notice some notable performance drops. 
Specifically, our method now only reaches an F1-Score of 0.87 on the Spirit data set, followed by DeepLog with an F1-Score of 0.83. 
On Thunderbird, DeepLog's performance decreases to 0.67 while our method nearly maintains its F1-Score of 0.9994 from the previous time window.

In general, it can be observed that our method and DeepLog, which are based on neural networks, perform significantly better than most traditional methods.
While traditional methods try to distinguish between black and white, some neural network based methods get a better intuition of what constitutes an anomaly. 

However, some threats to validity remain. On the one hand, we make the assumption that at least half of the data is normal. 
But in practice, it can be assumed that the majority of data are normal and only a few anomalies exist.
On the other hand, the failure times must be roughly known, which is not a strong precondition, since they can be deduced from commonly employed monitoring solutions.

\subsection{Anomaly Classification}
\label{subsec:evaluation:investigation:classification}

In the last section, we have demonstrated our capability to label data sets with exceptionally high precision and recall. 
Assuming that the monitoring system is sufficiently accurate, we encountered only a few mislabeled log lines. 
For this evaluation , we presume that these few log lines were corrected by experts. 
In this section, we evaluate our anomaly classification method (see~\autoref{sec:data:classifying_anomalies}) and calculate the types of anomalies present in the entire data sets used in the previous section and presented in~\autoref{subsec:evaluation:datasets:public}.

\autoref{tab:evaluation:investigation:classification:datasets} contains the number of normal and abnormal log messages in each data set.
Furthermore, we created log templates (see~\autoref{subsec:background:log_data:processing}) with Drain~\cite{he2017drain} and show the number of different templates for normal and abnormal log lines along with the number of intersecting templates.

\begin{table}[h]
	\centering
	\begin{tabular}{p{2.2cm}cc p{0.3cm} cc p{0.2cm} c}
		\toprule
        \multirow{2}{*}{Dataset}  & \multicolumn{2}{c}{Log messages} & & \multicolumn{2}{c}{Templates} && \\
        \cmidrule{2-3} \cmidrule{5-8}
                &  normal & anomalous && normal & anomalous &&  intersection \\
        \midrule
        Thunderbird   & 4\,773\,713 & 226\,287  && 969  & 17 && 3 \\
        Spirit        & 4\,235\,109 & 764\,891  && 1\,121 & 23 && 5 \\
        BGL           & 4\,399\,503 & 348\,460  && 802  & 58 && 10 \\
        \bottomrule
	\end{tabular}
	\caption{Data set statistics for anomaly classification.}
	\label{tab:evaluation:investigation:classification:datasets}
\end{table}

Since our classification method produces a probability value, we have calculated the number of different anomaly types for different thresholds.
Therefore, we applied our method to classify the types of anomalies in the data sets using threshold values of 0.6, 0.7, 0.8, 0.9, and 1.0. 
Because contextual anomalies are calculated by the templates of their surrounding log lines, we have set the following context boundaries: $a=10$ and $b=0$.

\begin{figure}[h]
\centering
\includegraphics[width=1.0\columnwidth]{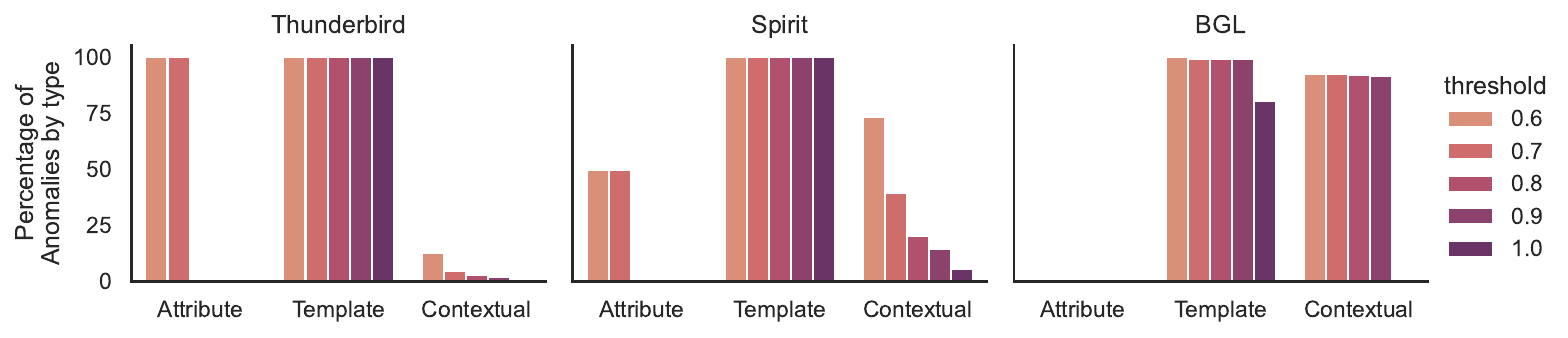}
\caption{Percentage of anomalies by type at different thresholds. Anomalies can be classified into multiple categories at the same time.}
\label{fig:evaluation:investigation:classification:results}
\end{figure}

\autoref{fig:evaluation:investigation:classification:results} displays the percentage of anomaly types present in each data set. 
We can observe that even at threshold 1.0, more than 99 \% of all anomalies in the Thunderbird and Spirit data sets are classified as template anomalies. 
This can be explained by the fact that the intersection of normal and abnormal templates is very small.
The case is very similar for BGL, although log templates are more heterogeneous in this data set. Only at a threshold of 1 the number of log messages classified as template anomalies drops to around 80 \%.
Furthermore, 226,071 of all anomalies in Thunderbird have the same template, that is, 99.9 \%. Similarly, 99.4 \% of all log anomalies in Spirit belong to one of the templates shown in Listing 1.

\begin{lstlisting}[label=listing,title=Listing 1: Most anomalies in Spirit belong to these two templates (380\,271 each).,frame=tb]
kernel: hda: drive not ready for command
kernel: hda: status error: status=<:HEX:> { }
\end{lstlisting}
\vspace{0.3cm}


For Spirit, we can observe that only one of the two most important log templates shown in Listing 1 contains an attribute, which explains why the number of attribute anomalies is around 50\,\%. 
At higher thresholds, the number of attribute anomalies drops to zero. 
Until a threshold of 0.7, almost 100 \% of all anomalies in Thunderbird are classified as attribute anomalies, which means that all template anomalies are also attribute anomalies in this case. 
As anomalous attributes are also contained in some normal log messages, the number of attribute anomalies drops to zero for higher thresholds.
For the BGL data set, no attribute anomalies were identified, not even at lower thresholds.
However, more than 91 \% of all anomalies in BGL are classified as contextual anomalies for thresholds between 0.6 and 0.9. 
This is significantly more than in the Thunderbird (13\,-\,2\,\%) and Spirit (63\,-\,14\,\%) data sets.

In conclusion, methods that focus on detecting template anomalies are expected to perform very well in all three benchmark data sets.
Additionally, identifying attribute anomalies may be helpful, but since most attribute anomalies are also template anomalies, the expected benefit is limited.
Approaches that identify anomalies by observing the context seem promising on data sets like BGL, but are not expected to perform well on Thunderbird and only to a certain degree on Spirit.

\subsubsection{Anomalies Outside the Taxonomy}
Our method does not guarantee that a given anomaly can be attributed to at least one of the classes in the proposed taxonomy, especially at high thresholds. 
The severity of this problem is evident to varying degrees in the data sets.
For Thunderbird, only 2 of 226\,287 messages could not be classified.
For Spirit, it is 28 out of 764\,891 for thresholds up to 0.9.
For a threshold of 1.0, we are unable to classify 1113 log messages, which is still only 0.15\%.
For the BGL data set, for thresholds of 0.6, 0.7, 0.8, and 0.9, we cannot classify 524, 831, and 2\,646 of 348\,460 messages, respectively, which is only a small fraction of all anomalies in BGL.
However, 68\,372 log messages, 19.6\% of all anomalous log messages, remained unlabeled at a threshold of 1.0.

\subsubsection{Research Question 1.1 Revisited}

Following the evaluation of our automated log labeling method and the anomaly classification method, we can address the research question for our first layer.

\textit{\textbf{RQ 1.1} How can an anomaly taxonomy and labeling support DevOps teams in the deployment and evaluation of appropriate anomaly detection methods?}



Given the diversity of anomaly detection models, which operate in different ways, understanding the nature of anomalies in the log data is crucial to select an appropriate detection method.
Our classification method can separate the majority of log anomalies into two primary types: point anomalies and contextual anomalies. 
Furthermore, we can divide point anomalies into template and attribute anomalies.
This knowledge allows us to determine an appropriate anomaly detection method required for the underlying IT system or service. 

To effectively evaluate the chosen or developed method, having a labeled data set is crucial. 
Our results demonstrated that we achieved F1-Scores of at least 0.997 in autonomously labeling all data sets. 
This accuracy of labeled data should be adequate to evaluate the method's performance, since only a really tiny amount of false positives and false negatives is present in the data for F1-Scores above 0.997.

In general, our \acl{LI} layer supports DevOps teams in selecting a suitable \ac{AD} method by knowing what type of anomalies are present in the log data and quickly evaluating the method through the automation of labeling data sets.

\section{Log Anomaly Detection}
\label{sec:evaluation:anomaly_detection}


In this section, we present an evaluation of our second layer of our log analysis architecture (see~\autoref{sec:log_analysis_layer:architecture}). 
Thereby we evaluate our anomaly detection method in different training scenarios on the three publicly available data sets and the industry data set for anomaly detection.
Details of the implementation are described in~\autoref{ch:anomaly_detection}.

As indicated in the previous~\autoref{subsec:evaluation:investigation:classification}, the publicly accessible data sets mainly include point anomalies, which are composed of template and attribute anomalies. 
Consequently, our anomaly detection method is specifically tailored for point anomalies.

\subsection{Preprocessing and Model Setup}
\label{subsec:evaluation:anomaly_detection:preprocessing_and_model}
This section details the preprocessing steps and model configuration applied across all training scenarios for all data sets.

For data preprocessing, we utilize the pipeline described in~\autoref{sec:anomaly_detection:anomaly_detection_model:preprocessing}. 
We establish the following rules for the replacement list $R$: \begin{itemize} 
    \item Substitute numbers exceeding $10$ with a \texttt{<NUM>} token \item Substitute Hexadecimal values with a \texttt{<HEX>} token. 
\end{itemize}

The set of split characters includes: $C=(\texttt{whitespace}, \texttt{;}, \texttt{:}, \texttt{.}, \texttt{-}, \texttt{,})$.

The number of tokens per log content is truncated to a length of $S=20$ for all data sets.

To set up our model, we initially use Xavier weight initialization for model weights and embeddings with the size 256.
For training our model, we use a hidden dimensionality of 256, a batch size of 1024, a total of 20 epochs, and a dropout rate of 10\%. 
Our output dimension $z$ is set to 128.
For optimization, we use the Adam optimizer with a learning rate of $10^{-4}$ and weight decay of $5 \cdot 10^{-5}$. 

For baselines built on neural networks, we use the same dimensionality for the layers and the same optimizer configuration.
For all statistical baselines, we employ TF-IDF (term frequency inverse document frequency) to encode tokens.

\subsection{Unsupervised Anomaly Detection}
\label{subsec:evaluation:anomaly_detection:unsupervised}
In this section we evaluate our log anomaly detection method in an unsupervised training scenario.
Since we evaluated which type of anomaly occurs how often in the last section, we first compare our method with four other methods concerning the different types of anomaly to determine which type is best detected. 
Next, we analyze the decision boundary during the inference stage for our model to identify the optimal decision boundary, using three publicly available data sets and an industry data set for anomaly detection, without focusing on specific anomaly types.
Finally, we benchmark our model against four other models on the three publicly available data sets.

Because our model requires a stabilization class for unsupervised training, we begin by showcasing the metrics of the auxiliary data.
The meaning of auxiliary data is explained in~\autoref{subsec:anomaly_detection:anomaly_detection_model:selecting_training_data} and defined in Definition~\ref{def:auxiliary_training_data}.

\subsubsection{Auxiliary Data}
Since we need a stabilization class to train our model in an unsupervised training scenario, we use 60,000 random log messages from other data sets. 
The corresponding mappings are shown in \autoref{tab:evaluation:anomaly_detection:unsupervised:stabilization}. 

\begin{table}[h!]
    \centering
    \begin{tabular}{|c|ccc|c|}
    \hline
    \multirow{2}{*}{Data set for evaluation} & \multicolumn{3}{|c|}{Auxiliary data from} & \multirow{2}{*}{Total} \\
    \cline{2-4}
     & BGL & Thunderbird & Spirit & \\
    \hline
    \hline
    BGL & / & 60,000 & 60,000 & 120,000 \\
    \hline
    Thunderbird & 60,000 & / & 60,000 & 120,000 \\
    \hline
    Spirit & 60,000 & 60,000 & / & 120,000 \\
    \hline
    Industry Data Set & 60,000 & 60,000 & 60,000 & 180,000 \\
    \hline
    \end{tabular}
    \caption{Log messages utilized for the stabilization class.}
\label{tab:evaluation:anomaly_detection:unsupervised:stabilization}
\end{table}

For the data sets BGL, Thunderbird, and Spirit, we utilize 120,000 log lines for each as a stabilization class. 
For the industry data set, we utilize a total of 180,000 log lines as the stabilization class.
Due to the disparity in the number of training samples of the normal log messages and the stabilization class, we use a weighted sampler for the training to balance the training.
Since the samples are randomly drawn for the auxiliary data, they can contain both abnormal and normal log lines.

\subsubsection{Evaluation of Different Anomaly Types}
To examine the capability of our method in identifying various anomaly types, we evaluate how accurately it detects anomalies that can be assigned to a class at a threshold of 0.7.
The number of each type of anomaly at this threshold is detailed in~\autoref{subsec:evaluation:investigation:classification}.
The goal is to identify which kinds of anomalies are easy or hard to detect for our method and also which baseline methods perform well on which anomalies.
As baselines, we chose a deep learning method called Deeplog~\cite{du2017deeplog}, and three data mining approaches PCA~\cite{jolliffe2005principal}, Invarant Miners~\cite{lou2010mining}, and Isolation Forest~\cite{liu2008isolation}.
We evaluate all methods in four different train/test splits of 0.2/0.8, 0.4/0.6, 0.6/0.4, and 0.2/0.8.

\begin{figure}[h]
\centering
\includegraphics[width=1.0\columnwidth]{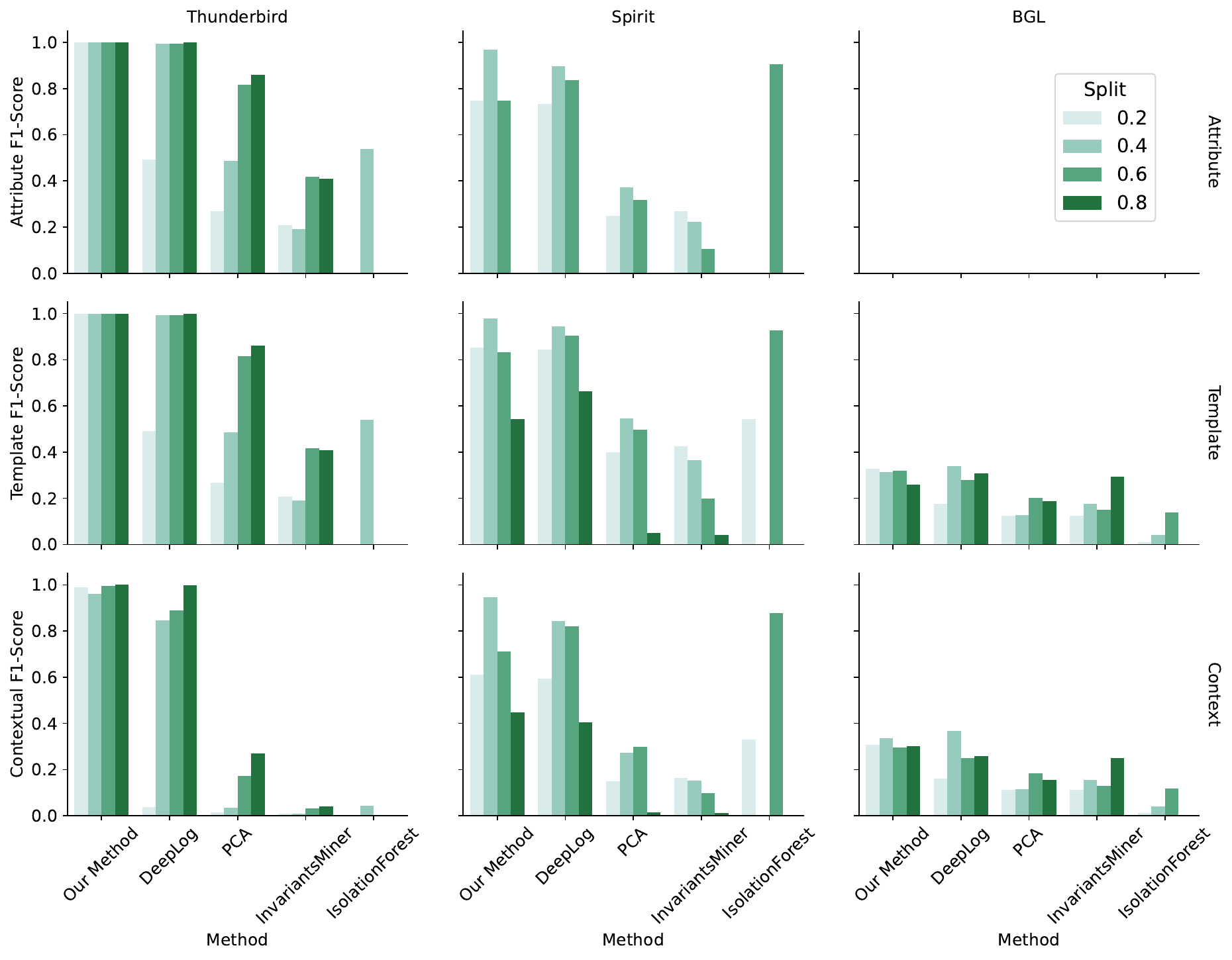}
\caption{F1-Scores for predicting attribute, template, and contextual anomalies at different train/test splits at threshold 0.7.
BGL contains no attribute anomalies.}
\label{fig:evaluation:investigation:classification:ml_results}
\end{figure}

All results are depicted in \autoref{fig:evaluation:investigation:classification:ml_results}.
We can observe that the two deep learning approaches outperform the data mining approaches in all experiments.
Isolation Forrest seems to be extremely sensitive to certain "sweet spots" in train/test splits but does not prove to be a robust method and is generally the worst-performing method. 
It is therefore excluded from any further discussion in the following paragraphs.
For the Thunderbird data set, DeepLog and our method manage to correctly classify almost all kinds of anomalies.
Our method performs generally better, even when only 20\,\% of the data are available as training data.
This might be attributed to the fact that DeepLog bases its predictions on templates only, while our method also takes attribute information into account.
The non-deep learning methods achieve worse F1-Scores on attribute and template anomalies and fail to predict most contextual anomalies.
On the Spirit data set, the performance of all methods is generally worse.
However, deep learning methods still achieve F1-Scores of around 0.75, 0.85, and 0.7 for attribute, template, and contextual anomalies, respectively.
On BGL, all methods obtain low F1-Scores for template and contextual anomalies: DeepLog around 0.27, our method around 0.35, PCA around 0.14, and Invariants Miner around 0.15. 
BGL does not contain any attribute anomalies.

However, the results do not represent the comprehensive results of our method, since we only examined anomalies with scores greater than 0.7 for the respective anomaly type. 
Anomalies with scores below 0.7 were excluded.
Consequently, increasing the training data might result in worse outcomes, as there were none or only a few anomalies of the corresponding type in the test set.

It can be concluded that for unsupervised methods template anomalies are the easiest to predict.
Furthermore, it can be suspected that attribute anomalies that are \emph{no} template anomalies are the hardest to predict.
However, this is hard to show, as attribute anomalies and template anomalies are highly correlated in all data sets.
Contextual anomalies were only predicted reliably by the two deep learning methods, but are generally harder to detect than template anomalies.

\subsubsection{Evaluation of the Decision Boundary}
In this section, we refrain from evaluating each anomaly type individually and instead assess the decision boundary calculation for all anomalies collectively.
Thus, the main focus is on the decision boundary function, explained in~\autoref{subsec:anomaly_detection:inference:decision_boundary_unsupervised} and is parameterized as shown in \autoref{tab:ad_param}.

\begin{table}[h!]
    \centering
    \begin{tabular}{|c||c|c|c|c|}
        \hline
        Parameter & BGL & Thunderbird & Spirit & Industry Data Set \\
        \hline
        $p$ & 0.95 & 0.95 & 0.95 & 0.95 \\
        $\alpha$ & 1 & 1 & 1 & 1 \\
        $\beta$ & 2.5 & 2.5 & 5.0 & 2.0 \\
        \hline
    \end{tabular}
    \caption{Decision boundary parameters to calculate the decision boundary.}
    \label{tab:ad_param}
\end{table}

We set $\alpha$ to $1$ in each experiment for each data set to replace only one token of the log message with the unknown token for data augmentation. 
Likewise, in each experiment, we set $p$ at 0.95 to filter out 5 \% of the augmented data with the highest anomaly scores. 
Furthermore, we adjust $\beta$ for each data set but keep it the same for each experiment on the respective data set.



\begin{figure*}[h]
\centering
\includegraphics[width=1\textwidth]{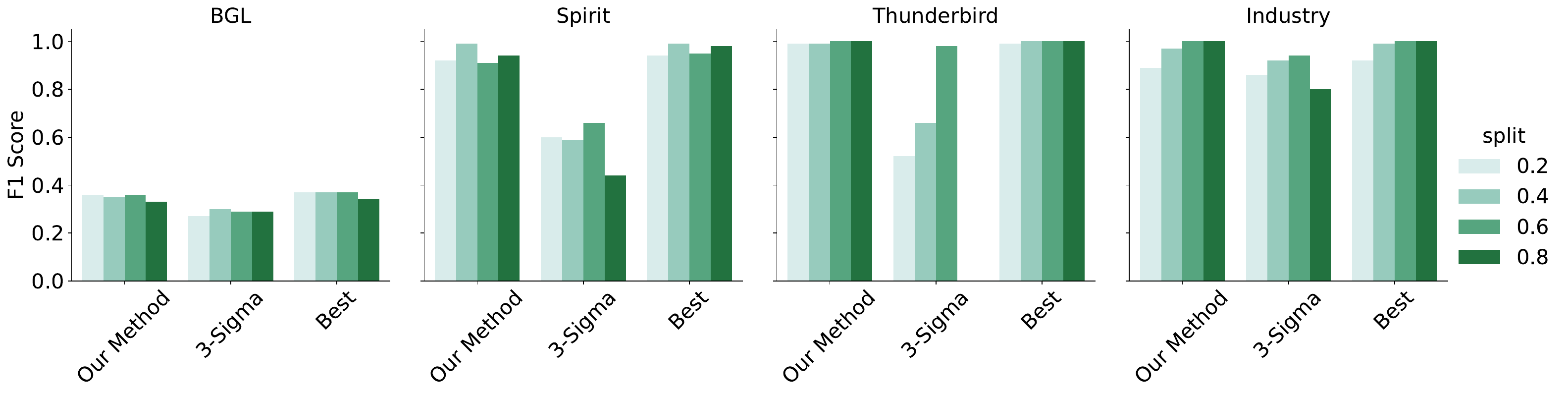}
\caption{Evaluation of our decision boundary against the best possible decision boundary and the 3-sigma method.}
\label{fig:evaluation:anomaly_detection:unsupervised:decision_boundary}
\end{figure*}

For all evaluations, we do every experiment three times and depict the best results in terms of the F1-Score in \autoref{fig:evaluation:anomaly_detection:unsupervised:decision_boundary}. 
In this evaluation, we employ our neural network and its configuration to compare our decision boundary calculation (\textit{Our Method}) with the optimal decision boundary (\textit{Best}) and the \textit{3-Sigma} method. The optimal decision boundary is determined retrospectively to explore the best achievable result for our network architecture. 
It shows the results for different amounts of training data of 20\%, 40\%, 60\%, 80\% of the respective data set. 
It can be seen that our method is superior to the baseline 3-Sigma in all experiments. 
In addition, our method achieves almost the same F1-Scores as \textit{Best} in all experiments. 

It is evident that our approach does not show an improvement in the BGL and Spirit data sets when additional training data is provided. 
However, on both data sets, the optimal decision boundary (Best) is only marginally better and follows the same pattern. This indicates that the model does not benefit from the increased training data in an unsupervised learning scenario.
The results in the BGL data set are not particularly high, as there is a fundamental concept drift in the data, which can be seen by the fact that even for high training splits of 60 \% there are still many normal unknown log lines after preprocessing present in the test data set that are not present in the training data set as shown in \autoref{tab:evaluation:datasets:public:split_unique_preprocessed}.

For the Thunderbird and industry data set, it is evident that our method shows strong performance right from the start, and the decision boundary calculation during the inference stage is highly precise, closely matching the optimal decision boundary. 
Furthermore, the two parameters $\alpha$ and $p$ for our decision boundary can be set to 1 and 0.95 independently of the data set. This shows that the method is robust in parameter choice and independent of the data set.

\subsubsection{Evaluation of all Anomalies in an Unsupervised Training Scenario}
After initially evaluating how our method and other methods perform on the different anomaly types, we now perform an evaluation of all anomalies. 
For our method, we employ the decision boundary function with the parameters of the preceding subsection.

\begin{figure*}[h]
\centering
\includegraphics[width=1\textwidth]{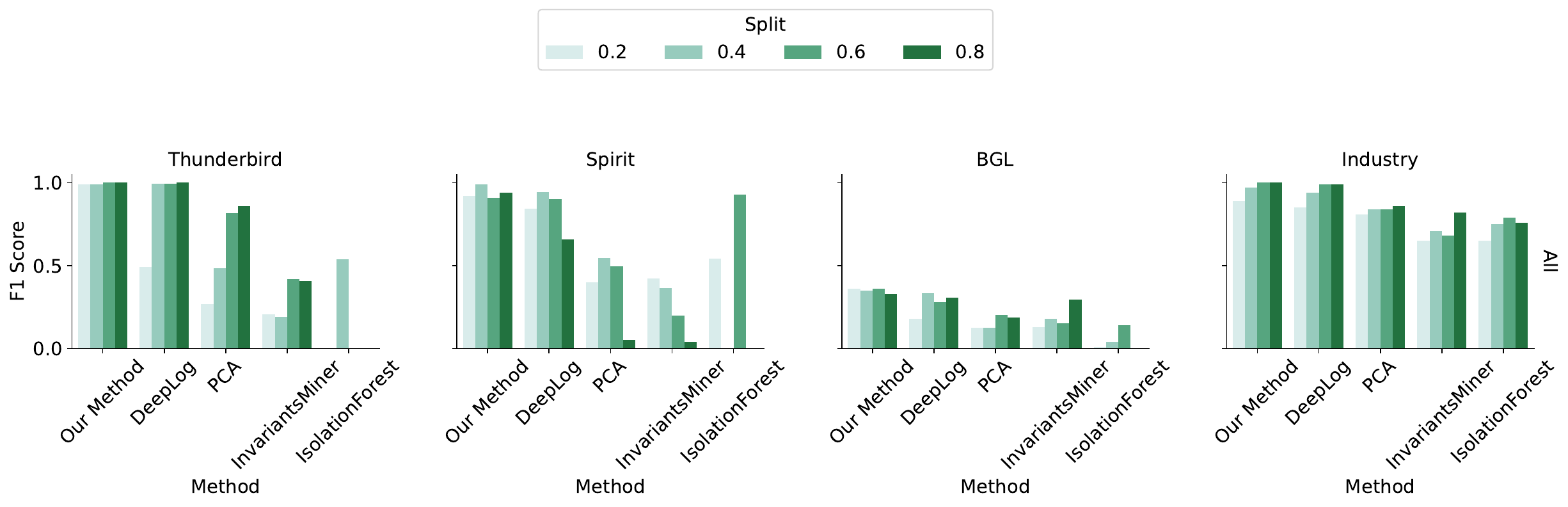}
\caption{F1-Scores of different methods in an unsupervised training scenario for different training splits.}
\label{fig:evaluation:anomaly_detection:unsupervised}
\end{figure*}

~\autoref{fig:evaluation:anomaly_detection:unsupervised} compares our method with Deeplog~\cite{du2017deeplog}, PCA~\cite{jolliffe2005principal}, Invariant Miners~\cite{lou2010mining}, and Isolation Forest~\cite{liu2008isolation}, which we have already used to evaluate their performance for different types of anomalies.

Our method outperforms others across all data sets, though it still underperforms on the BGL data set, where other methods also show poor performance. 
Additionally, the industry data set is the only one in which all methods benefit from more training data, while the effect of additional training data varies on the other data sets.

If 80\% of the BGL data set is utilized for training, only a small number of unknown normal log lines remain in the evaluation set. 
The fact that our unsupervised model still does not achieve high F1-Scores indicates that the anomalies closely resemble normal log lines. 
Therefore, a weakly or fully supervised training approach would be more effective and appropriate for this data set.

In summary, we could conclude that unsupervised methods demonstrate very good performance in some scenarios but not in all. 
Furthermore, our decision boundary function is suitable to calculate a precise decision boundary for the inference stage in an unsupervised scenario. 

\subsection{Weak-Supervised Anomaly Detection}
\label{subsec:evaluation:anomaly_detection:weak_supervised}

In this section, we evaluate our method in a weakly supervised scenario against eight other methods.
For our method, we use the objective function formulated in~\autoref{subsec:anomaly_detection:anomaly_detection_model:objective_function}.
For the inference stage, we use the decision boundary function formulated in~\autoref{subsec:anomaly_detection:inference:decision_boundary_weak_supervised}.

To evaluate the weakly supervised scenario, we simulate a monitoring environment that has an approximate knowledge of the error times. 
We achieve this by generating a failure time window around each anomaly in the training data set for the respective split. 
To explore various monitoring scenarios, we use time windows $\delta$ of $\pm$ 1 second, 5, 10, and 20 seconds. 
Consequently, for $\delta=20$, each failure time window is 40 seconds.

\begin{table}[h!]
    \centering
    \small
        \begin{tabular}{cc|ccccc}
        \toprule
        \% Split 
        & \makecell{Real Anomalies} 
        & \makecell{$\mathcal{U}$ for\\$\delta=\pm1 s$} 
        & \makecell{$\mathcal{U}$ for\\$\delta=\pm5 s$} 
        & \makecell{$\mathcal{U}$ for\\$\delta=\pm10 s$} 
        & \makecell{$\mathcal{U}$ for\\$\delta=\pm20 s$} \\
        \midrule
        \multicolumn{6}{c}{BGL} \\
        \midrule
        0.2 & 224,921 & 249,987 & 273,680 & 277,269 & 281,395 \\
        0.4 & 227,943 & 254,056 & 279,406 & 284,450 & 290,994 \\
        0.6 & 231,879 & 258,824 & 287,298 & 294,299 & 302,822 \\
        0.8 & 302,177 & 338,245 & 375,417 & 387,901 & 404,476 \\
        \midrule
        \multicolumn{6}{c}{Spirit} \\
        \midrule
        0.2 & 289,905 & 351,560 & 723,302 & 999,998 & 999,998 \\
        0.4 & 461,797 & 613,334 & 1,438,138 & 1,999,996 & 1,999,996 \\
        0.6 & 724,872 & 936,973 & 2,153,039 & 2,999,993 & 2,999,994 \\
        0.8 & 762,849 & 992,140 & 2,288,745 & 3,193,275 & 3,203,446 \\
        \midrule
        \multicolumn{6}{c}{Thunderbird} \\
        \midrule
        0.2 & 26 & 8,912 & 49,771 & 90199 & 14,4483 \\
        0.4 & 6,619 & 106,112 & 345,426 & 516,746 & 584,564 \\
        0.6 & 94,410 & 500,101 & 895,612 & 1,125,702 & 1,194,721 \\
        0.8 & 163,135 & 883,773 & 1,519,626 & 1,855,974 & 1,925,250 \\
        \bottomrule
        \end{tabular}
    \caption{Analysis of real anomalies count and log lines count in $\mathcal{U}$ for different training splits and failure time windows.}
    \label{tab:evaluation:anomaly_detection:weak_supervised:training_data}
\end{table}

We apply the PU learning scenario to each method. Where all log lines in the training data that are in the failure time windows are in $\mathcal{U}$ and all other in $\mathcal{P}$. 
The~\autoref{tab:evaluation:anomaly_detection:weak_supervised:training_data} provides a detailed analysis of the relationship between the number of real anomalies and the number of log lines in the class $\mathcal{U}$ for each split in three data sets: BGL, Spirit, and Thunderbird. 

In the BGL dataset, there is a relatively balanced increase in both real anomalies and the number of log lines in $\mathcal{U}$ as the percentage of training split and the sizes of the time window increase. Although the number of log lines in $\mathcal{U}$ is consistently larger than the real anomalies, the difference is not excessively large, indicating that the dataset is relatively easier to manage for anomaly detection tasks.

The Spirit data set exhibits a more substantial increase in the number of log lines in $\mathcal{U}$ compared to the real anomalies, particularly for longer time windows and higher percentages of training splits. Despite this greater increase, the ratio between $\mathcal{U}$ and real anomalies remains more manageable compared to Thunderbird, making Spirit moderately difficult for anomaly detection.

The Thunderbird data set is particularly difficult due to the significant difference between the count of actual anomalies and the number of log lines in $\mathcal{U}$. For every training split and time window size, the number of real anomalies is significantly smaller than the number of log lines in $\mathcal{U}$. This vast disparity shows that the Thunderbird data set has a high volume of incorrectly labeled training data in $\mathcal{U}$ relative to the number of anomalies, significantly increasing the difficulty of accurately detecting anomalies during inference.

\begin{figure*}[h]
    \centering
    \includegraphics[width=1.0\textwidth]{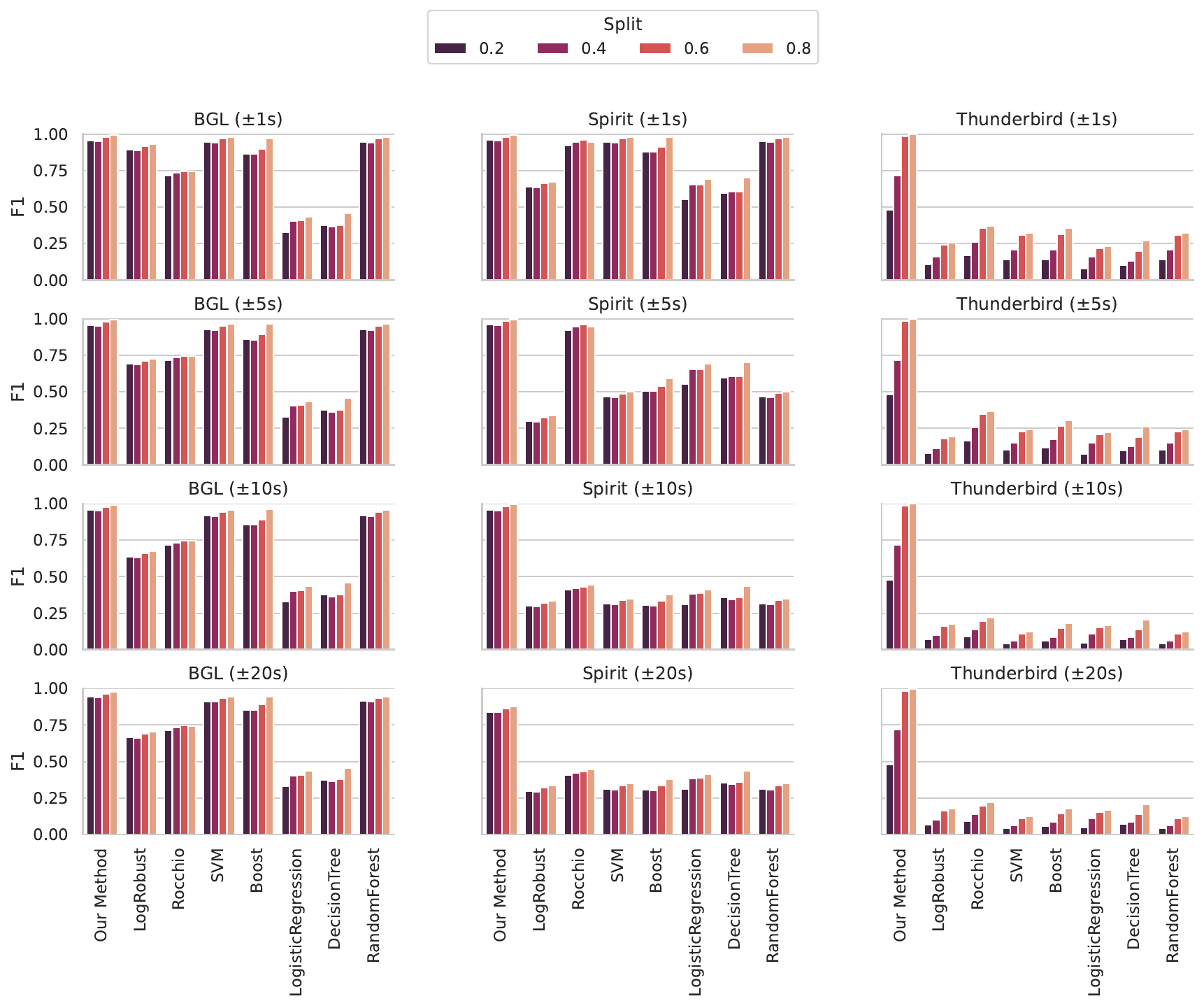}
    \caption{F1-Scores of different methods in a weak-supervised training scenario for different training splits and time windows.}
    \label{fig:evaluation:anomaly_detection:weak_supervised:results}
\end{figure*}


We show the performance of our method and the eight baselines in terms of the F1-Score for each data set, split, and failure time window size in~\autoref{fig:evaluation:anomaly_detection:weak_supervised:results}.
Our method achieves the highest F1-Scores in all experiments.
As expected, with increasing time window sizes and thus increasing size of $\mathcal{U}$, the performance of all methods tends to decrease.

\textbf{Thunderbird.}
For $\delta = \pm 1s$, most of the methods achieve good or acceptable performance. An exception is the Thunderbird dataset, which is characterized by a large class $\mathcal{U}$, even for $\delta \pm 1000$ in all splits. 
No methods but our method manage to achieve an F1-Score higher than 0.35, even for a split of $0.8$, while our method reaches an F1-Score of $0.99$.
Moreover, our method performs poorly for the splits of $0.2$ and $0.4$ in the Thunderbird data set. This is because, in the $0.2$ split, there are only 26 actual anomalies available, while nearly 9,000 log lines are in $\mathcal{U}$ (see~\autoref{tab:evaluation:anomaly_detection:weak_supervised:training_data}). 
Consequently, 99.7\% of the normal training data in $\mathcal{U}$ is mislabeled, a challenge that no method can handle.
With a training size of $0.4$, the performance of all methods shows improvement but remains unsatisfactory. Despite the fact that the ratio of true anomalies to mislabeled log lines in $\mathcal{U}$ becomes more positive, all methods other than ours still achieve an F1-Score below $0.19$.
Starting from a $0.6$ split, all methods show an improvement in performance across all sizes of the failure time windows. Certain methods show further enhancement with a $0.8$ split, whereas others, ours included, exhibit no change. 

\textbf{Spirit.}
In the Spirit data set, most methods (Rocchio, SVM, Boost, Random Forest, and our method) archive high F1-Scores above $0.9$ for $\delta=\pm1s$ for all training splits.
For $\delta = \pm 5s$, the performance of all methods except our method and Rocchio now also degrades in the Spirit data set. However, we see a more significant increase for each method if more training data are available.
Most methods tend to have close to perfect recall but suffer from poor precision, which means they do not identify normal log messages in $\mathcal{U}$, which is much larger for $\delta = \pm 5s$. 
The greatest drop in performance becomes evident at the time window of $\delta = \pm 10s$, now Rocchio and Random Forest archive approximately the same F1-Scores as the other methods.
With an error time window size of $\delta = \pm 20s$, the performance of our method experiences a more notable decrease, although it remains significantly better than all other methods.

\textbf{BGL.}
In contrast to Thunderbird and Spirit, most methods continue to achieve very good results on the BGL data set, even for large failure time windows.
For the BGL data set, we notice a drop in F1-Scores of other methods at $~0.96$ for training data sizes of $0.6$ and $0.8$, when $\delta = \pm 10s$, while only our method maintains a high performance of more than $0.99$.
In summary, the size of the training data appears to have a more significant impact on the methods for the BGL data set compared to the failure time window size, which is the main difference to the other data sets.

In general, it can be observed that our method, which is based on neural networks, performs significantly better than most traditional methods in all experiments.
However, neural network-based methods do not have to be better, since LogRobust, which is also based on a neural network, suffers from overfitting on $\mathcal{U}$.
It is not able to perform well if incorrectly labeled input data is given. 
Therefore, LogRobust is not applicable for a PU learning strategy, which demonstrates that neural networks are not generally superior.

Overall, it should be noted that all methods archive poor results in the Thunderbird data set, which was the simplest of all methods in the unsupervised training scenario. 
In contrast, the methods achieved high F1-Scores on the BGL data set, where all methods performed very poorly in the unsupervised scenario. 
This implies that anomaly detection methods can perform differently on the same log data from an IT system or service, depending on how they are trained.

\subsection{Supervised Anomaly Detection}
\label{subsec:evaluation:anomaly_detection:supervised}
In this section, we evaluate our method against four other methods in a supervised training scenario.
For this evaluation, we have chosen LogRobust~\cite{zhang2019robust}, which showed weak performance in the weakly supervised training scenario, but we anticipate it will perform good in the fully supervised scenario. Additionally, we chose Boost and SVM, as they demonstrated relatively good performance in the weakly supervised scenario compared to other baselines. Finally, we included Decision Tree, which had a rather poor performance in the weakly supervised scenario.


\begin{figure*}[h]
    \centering
    \includegraphics[width=1.0\textwidth]{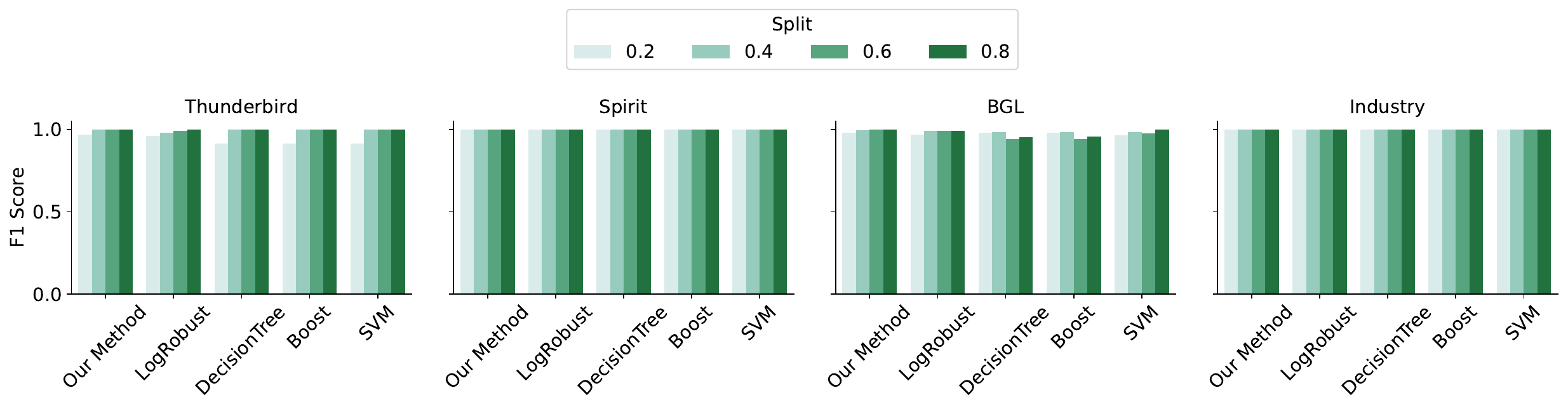}
    \caption{F1-Scores of different methods in a supervised training scenario for different training splits.}
    \label{fig:evaluation:anomaly_detection:supervised:results}
\end{figure*}

The~\autoref{fig:evaluation:anomaly_detection:supervised:results} shows the results for our four training splits: 0.2, 0.4, 0.6 and 0.8.
Firstly, it is evident that the performance of all methods across all data sets is excellent, with each achieving an F1-Scores exceeding 0.9 in every experiment. 
In addition, a score of f1 of 1.0 is frequently archived by several methods, indicating perfect anomaly detection without false positives or false negatives. 


Slight variations are noticeable in the Thunderbird data set when only 20\% of the data is used for training with traditional methods such as Decision Tree, Boost, and SVM. 
These methods perform marginally worse initially, but improve significantly when 40\% of the data is used for training.
All methods reach an F1-Score of 1.0 with a training split of 0.8 on the Thunderbird data set. 
LogRobust and Decision Tree achieve this score even with a split of 0.6, while our method achieves it with only 40\% training data.
This indicates that our method requires the smallest amount of training data to achieve perfect anomaly detection in a supervised training scenario for this data set.

For the Spirit data set, both LogRobust and our method achieve an F1-Score of 1.0. 
LogRobust requires 80 \% of the data for training, while our method requires only 60 \%. 
In addition, all methods perform excellent with F1-Scores of 0.99 and above.

In the BGL data set, no method achieves perfect anomaly detection, but all methods attain at least an F1-Score of 0.98.
Another observation is that the performance of the Decision Tree and Boost on the BGL data set degrades when trained on 60\% or more of the data set.

For the industry dataset, only our method achieves an F1-Score of 1.0 starting with 40\% training data. 
LogRobust achieves nearly perfect anomaly detection in all splits.

In general, this evaluation demonstrates that anomaly detection methods, regardless of whether they are based on neural networks or not, are extremely effective if they are trained in a supervised scenario. This further underscores the need for our automated labeling method for log data, since it is the prerequisite for supervised training.

\subsubsection{Research Question 1.2 Revisited}

Following the evaluation of our anomaly detection method, we can address the research question for our second layer.

\textit{\textbf{RQ 1.2} How should an anomaly detection method be designed to identify anomalies in any system without depending on specific training data?}

Our results show that our method can be utilized in a variety of situations involving log data from different systems and various types of training data.
This includes unlabeled training data consisting solely of normal data, imprecise labeled training data that contain incorrect labels, and labeled training data. 
Based on the type of training data, our method can be trained in an unsupervised, weakly supervised, or supervised manner.

In an unsupervised scenario, our method achieves F1-Scores ranging from 0.37 to 1.0, depending on the data set. 
In a weakly supervised training scenario, the results vary from F1-Score of 0.5 to 0.99 depending on the difficulty of the training parameter. 
However, it is possible to achieve F1-Scores of 0.99 and above on all data sets given sufficient training data. 
In a supervised scenario, all methods achieve F1-Scores ranging from 0.92 to 1.0, indicating that supervised training is the most efficient, but also the most costly scenario. 
Furthermore, we could demonstrate that we perform slightly better on all data sets as our baselines.

In every training scenario, there are baseline methods that can rival our method, but struggle to achieve comparable outcomes in different training scenarios.
This underscores the generality of our method.
Hence, we have demonstrated that we have developed a general method that can perform exceptionally well with different training data from different IT systems by simply modifying the objective function.

\section{Log Root Cause Analysis}
\label{sec:evaluation:root_cause_analysis}
In this section, we evaluate our \ac{RCA} method for our third log analysis layer presented in~\autoref{subsec:log_analysis_layer:root_cause_analysis} on the industry data set explored in~\autoref{subsec:evaluation:datasets:industry_rca}.

As described in~\autoref{ch:related_work}, all the approaches reviewed for root cause analysis use a previously defined set of root causes.
Our method, on the other hand, aims to help DevOps teams by analyzing previously unknown root causes.
Therefore, we assign a root cause score to each log line and present the most relevant log lines that explain the root cause to the DevOps team.

\subsection{Training Setup for all Methods}
\label{subsec:evaluation:root_cause_analysis:training_setup}
We train our method and four baselines on the described log data by applying our PU learning scenario. 
All log lines contained within the \acfp{ITW} ($\sim$300,000) are in the unknown class $\mathcal{U}$ and all remaining log lines ($\sim$44 million) in the positive class $\mathcal{P}$. 
We use three statistical models and one neural network-based method als baselines to compare our method.

We use the following statistical models als baselines.
First, a \emph{Decision Tree} with a maximum depth of 30, where the score describes the fraction of samples of the same class in a leaf.
Second, a \emph{Random Forest} with 100 trees with a maximum depth of 20, where the score describes the mean predicted class probabilities of all trees in the forest.
Third, a \emph{SVM}, where the score is determined by a 5-fold cross-validation.
To make SVM scores comparable, we calibrated its output probabilities using Platt scaling~\cite{Platt99}.
For all statistical baselines, we employ TF-IDF (term frequency inverse document frequency) to encode tokens.

As a baseline based on neural networks, we use a \ac{FFN}.
Our method, called LogRCA, and the \ac{FFN} are trained for 5 epochs, where both networks comprise an embedding layer with 128 units to encode tokens.
The \ac{FFN} baseline has two hidden layers with 256 hidden units each. 
The \acfp{RCC} scores are based on the calibrated output probability of its output layer~\cite{nn_calibration_2017}.
The attention layer of our method has two attention heads and connects to a fully connected layer with 256 units.
In these configurations, our method and FNN take roughly the same time to train.

To balance our training data, we use automatic clustering as described in~\autoref{sec:rca:balancing:clustering}.
For that, we use BIRCH~\cite{zhang1996birch} with branching factor $B=50$ and threshold $T=0.5$ to automatically determine the optimal number of clusters.
To analyze the main services involved in the root cause, we use BIRCH with a branching factor $B=50$ and a threshold $T=0.3$.

\subsection{Performance Analysis}
\label{subsec:evaluation:root_cause_analysis:performance_analysis}
\begin{figure}[h]
    \centering
    \includegraphics[width=0.95\columnwidth]{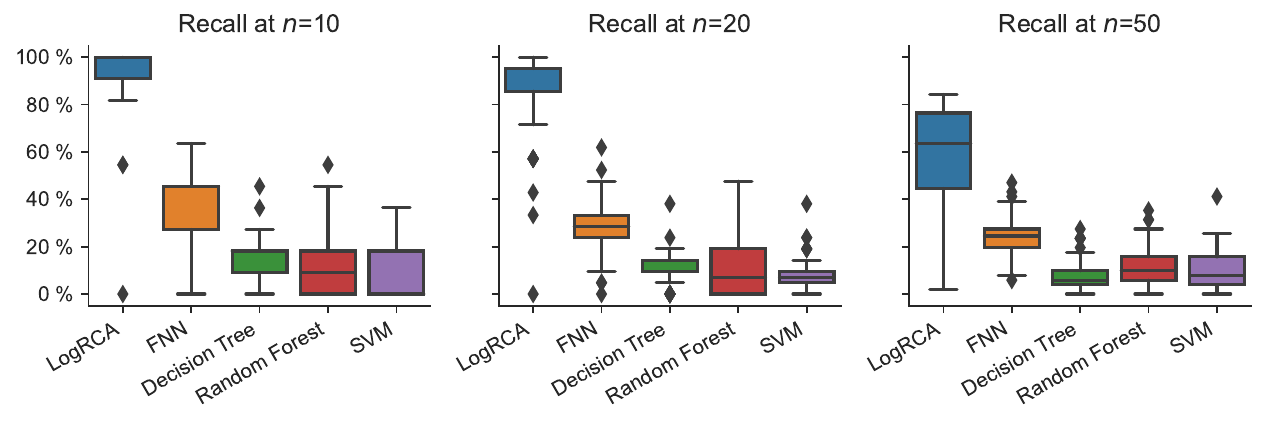}
    \vspace{-3mm}
    \caption{Fraction of root case log lines at 10/20/50 returned candidates.}
    \label{fig:rca:recall_at}
\end{figure}

We consider the recall of the returned root cause candidates to be the most relevant metric for DevOps teams to quickly understand a root cause. \emph{We therefore want to investigate how many of the n highest ranked root cause candidates are actually part of the root cause.}
Note, that our method does not automatically decide on a threshold~$n$, as we do not know how many log lines constitute the root cause in total.
Due to this, we analyze the recall at $n=10$, $n=20$, and $n=50$.
The results are presented in~\autoref{fig:rca:recall_at}.
DevOps teams have the flexibility to select the value of $n$ based on the number of candidates they wish to display.

We observe that our method clearly outperforms all baselines in all three experiments.
Within the 10 top-ranked root cause candidates, our method has an average recall of 93.5 \%.
In 75 of 80 cases, all log lines are part of the root cause, and in the remaining 5 cases the root cause is fully covered as it consists of $\le$10 root cause log lines. 
Due to this, additional log lines are shown since $n$ was configured to 10, which lowers the recall.
To compare, the baseline FNN has an average recall of 36.1 \% and the statistical methods only 15.8 \% (Decision Tree), 10.7 \% (Random Forest), down to 6.7 \% for SVN.
Similarly, for 20 and 50 returned \aclp{RCC}, our method maintains a high average recall of 86.6 \% and 57.7 \%, respectively, which means that the majority of log lines presented to the DevOps team are actually part of the root cause.

This advantage also translates to other related metrics, such as the question of whether we actually identified \emph{all} root cause log lines.
Since we have up to 50 root cause log lines in our ground truth, we report this metric for $n=50$: 
In this case, our method covered all root cause log lines in 65 out of 80 cases, compared to 57 for FFN, 29 for Decision Tree, 6 for Random Forest, and 11 for SVM.

\subsection{Impact of Balancing Training Data}
\label{subsec:evaluation:root_cause_analysis:impact_balancing}
\begin{figure}[h]
    \centering
    \includegraphics[width=0.9\columnwidth]{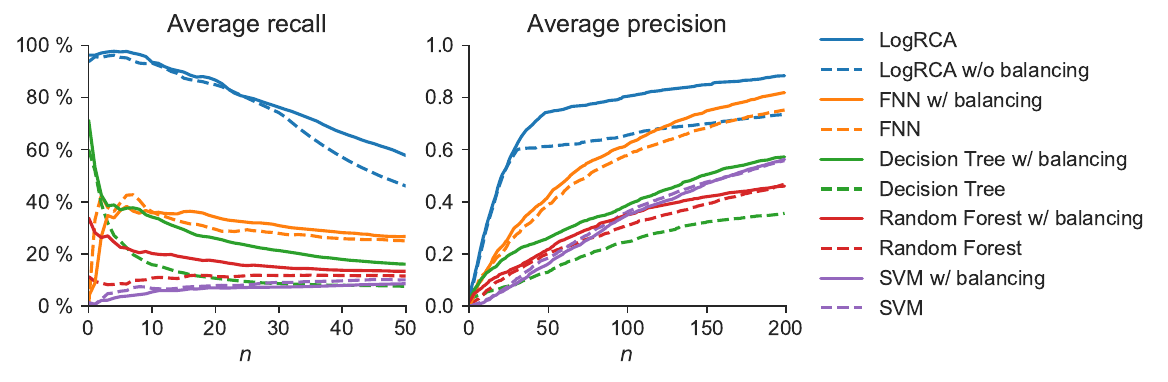}
    \caption{Average precision and recall of all approaches with and without balancing.}
    \label{fig:rca:precision_recall}
\end{figure}

To evaluate the impact of the proposed balancing of training data, we apply this strategy to all baselines.
In return, we also trained a version of our method without balancing.
Figure~\ref{fig:rca:precision_recall} presents the resulting precision and recall of all approaches, with and without balancing.
We report recall until $n=50$, since none of the \acf{ITW} contains more than 50 root cause log lines, and precision until $n=200$.

The recall indicates the proportion of actual root cause log lines present among the top $n$ log lines with the highest scores. 
Precision indicates the fraction of root cause log lines found within the top $n$ lines of all root cause log lines for this failure. 
Therefore, the precision increases with increasing $n$.

The intended effect of data balancing is to increase its performance in rare failures.
This is clearly visible: Our method average precision and recall remain the same until $n \le 30$ regardless of whether balancing is applied or not.
This suggests that many root causes were comparably easy for the transformer model to detect.
For $n > 30$, the effect of balancing considerably improves average performance, as there is less overfitting of frequently occurring failures.

Interestingly, balancing significantly improves the recall of the Decision Tree and Random Forest baselines, while there was no observable benefit to the SVM model.
For the FNN, balancing did not significantly improve or harm recall, but consistently improved precision with increasing root cause candidates.

\subsubsection{Research Question 1.3 Revisited}

Following the evaluation of our root cause analysis method, we can address the research question for our third layer.

\textit{\textbf{RQ 1.3} Is it possible to identify a set of log lines that describe the root cause leading to a failure, so that DevOps teams can remediate the failure?}

Our evaluation demonstrated that our anomaly detection method can be utilized in a weakly supervised training scenario to identify log lines that describe the root causes by adding additional techniques.
Therefore, we must balance the training data to also recognize root cause log lines in rare cases. 
Additionally, it is necessary to analyze the main services involved in each root cause to select the log lines produced by those services to exclude anomalies that are merely side effects, as they do not contribute to understanding the root cause or resolving the failure.

We performed our evaluation on a proprietary industry data set that contains 80 failures with unknown root causes. 
Our analysis showed that the log lines with the highest scores precisely described the root causes, allowing us to understand the failure.
Thus, we demonstrated that if we retrieve 10 log lines, they are relevant to the root cause with a probability of 98 \%. 
In the case where 50 log lines are retrieved, on average 60\% of them relate to the root cause, while other methods only achieve 40\%. 
However, not all root causes comprised 50 log lines; some had fewer, which degraded the values.

Since the top-ranked log lines generally describe the root cause, we were able to demonstrate that it is possible to identify a set of log lines that describe the root cause of a failure.

\cleardoublepage

\chapter{Conclusion}
\label{ch:conclusion}

It is important that failures in IT systems are resolved quickly, as they affect countless areas of our modern lives. 
Thus, we have focused on log analysis because logs document the execution of IT systems and offer insight into their behavior.
Therefore, this thesis presented a layered architecture for log analysis in complex IT systems. 
Specifically, we designed methods that provide sufficient detail and insights into the execution of the IT system to DevOps teams, enabling them to efficiently resolve failures. 
Our research identified three essential layers: Log investigations, anomaly detection, and root cause analysis.

In the first layer, we developed a method for autonomously labeling log data to evaluate or train anomaly detection methods. 
Our autonomous labeling method shows a strong performance of F1-Scores of 0.99, given that the failure time is approximately known.
This enables the DevOps team to quickly evaluate the anomaly detection method or train a supervised method without the need for manually labeled data.
This layer also provides a taxonomy for classifying anomalies into different types, which helps in selecting the appropriate anomaly detection method.
Our classification method proved highly effective, as we were able to classify a minimum of 99\% anomalies into classes in all data sets with a confidence level of 90\%. 

The second layer offers a general anomaly detection method that is adaptable to different training strategies: Unsupervised, weakly supervised, or supervised, depending on the available training data. 
We demonstrated the flexibility and robustness of our method by evaluating it on publicly available data and data from an industry company. 
We demonstrated that leveraging supervised and weakly supervised learning allows us to develop models that achieve F1-Scores between 0.99 and 1.0. 
Similar results were observed with unsupervised learning on certain data sets.
This layer ensures that anomalies are detected accurately and reliably, regardless of the type and source of the training data.

In the third layer, we implemented a root cause analysis method that filters out irrelevant log lines, focusing only on those that describe the origin and course of a failure. 
By balancing the training data and identifying the main services involved in the failure, we were able to effectively analyze and understand rare root causes. 
Our results demonstrated that the leading 10 root cause candidates consistently contained between 90\% and 98\% root cause log lines.
This method helps DevOps teams identify a set of log lines that describe the root causes of failures efficiently.

In general, our three-layer architecture for log analysis allows DevOps teams to gain insight into their IT systems and services by investigating the generated log data. 
By addressing the prerequisites of each layer and offering general methods that can handle different types of log data, our methods ensure that failures can be traced back to their root causes. 

Having answered the three research sub-questions in the corresponding evaluations of the three layers, we can finally answer our main research question:

\textit{\textbf{"How can log analysis methods be designed and optimized to help DevOps teams resolve failures efficiently?"}}


For DevOps teams to effectively resolve failures, understanding what and why something is failing is necessary. 
This can be accomplished by analyzing log files from the IT systems. 
To successfully analyze log data and find abnormal behavior, the methods must be independent of the system and not dependent on specific training data. 
Finally, they need to describe the failure and its cause to enable DevOps teams to understand and mitigate it.
Therefore, we have developed those methods within our three-layer architecture for log analysis. 

To choose and evaluate an appropriate anomaly detection method, we must know which type of anomalies are present in the log data and provide labeled data for evaluation.
Both prerequisites are provided by our methods in the Log Investigation layer.
Since different IT systems provide a variety of training and log data, we developed an anomaly detection method within our second layer that is capable of dealing with different types of training data of different systems.
To effectively analyze failures, it is crucial that the DevOps team understands them.
Therefore, in the Root Cause Analysis layer, we have introduced a method that identifies a set of log lines that describe the cause and origin of a failure.
This empowers DevOps teams to resolve these failures in future software updates by knowing what went wrong, and thus improving the reliability and performance of their IT systems.




\section{Limitations and Future Work}

Despite the effectiveness of the proposed three-layer architecture for log analysis, some limitations should be acknowledged.
In general, all log analysis methods are highly dependent on the completeness of the log data. 
Incomplete log data can negatively impact the performance of the methods, potentially leading to missed anomalies, incorrect classification, or missing root cause log lines.
Future work should investigate at what level of missing log lines our methods start to degrade.

There may be more types of anomalies than we identified, indicating that our classification may not cover all possible anomalies, since we could not classify all anomalies with high thresholds.
It is necessary to explore whether there are additional categories or if certain anomalies are just more subtle.

Moreover, some of our methods rely on the alerts of monitoring systems to have approximated failure times.
This is an important requirement for our log labeling method.
However, this requirement is less stringent for root cause analysis, as the failure time is already known since we want to investigate the failure.
Future research should explore alternative methods for estimating or determining approximate failure times.

Another limitation is that our RCA method returns a pre-defined number of log lines. 
If the DevOps team sets this number too low, not all relevant RCA log lines may be displayed, potentially omitting critical information. 
In contrast, setting the number too high could result in an excess of noise, making it harder to analyze the root cause. 
Therefore, it should be investigated if the number of root cause log lines can be dynamically determined for each root cause.



\cleardoublepage
\phantomsection

\addcontentsline{toc}{chapter}{Bibliography}
\newrefcontext[sorting=none]
\printbibliography[resetnumbers=true]

@inproceedings{wittkopp2020a2log,
  author       = {Thorsten Wittkopp and Alexander Acker and Sasho Nedelkoski and Jasmin Bogatinovski and Dominik Scheinert and Wu Fan and Odej Kao},
  title        = {A2Log: Attentive Augmented Log Anomaly Detection},
  booktitle    = {55th Hawaii International Conference on System Sciences, {HICSS} 2022,
                  Virtual Event / Maui, Hawaii, USA, January 4-7, 2022},
  pages        = {1853--1862},
  publisher    = {ScholarSpace},
  year         = {2022},
  url          = {http://hdl.handle.net/10125/79566},
  bibsource    = {dblp computer science bibliography, https://dblp.org}

}

@inproceedings{wittkopp2020decentralized,
  author       = {Thorsten Wittkopp and
                  Alexander Acker},
  editor       = {Hakim Hacid and
                  Fatma Outay and
                  Hye{-}young Paik and
                  Amira Alloum and
                  Marinella Petrocchi and
                  Mohamed Reda Bouadjenek and
                  Amin Beheshti and
                  Xumin Liu and
                  Abderrahmane Maaradji},
  title        = {Decentralized Federated Learning Preserves Model and Data Privacy},
  booktitle    = {Service-Oriented Computing - {ICSOC} 2020 Workshops - AIOps, CFTIC, STRAPS, AI-PA, AI-IOTS, and Satellite Events, Dubai, United Arab Emirates, December 14-17, 2020, Proceedings},
  series       = {Lecture Notes in Computer Science},
  volume       = {12632},
  pages        = {176--187},
  publisher    = {Springer},
  year         = {2020},
  url          = {https://doi.org/10.1007/978-3-030-76352-7\_20},
  doi          = {10.1007/978-3-030-76352-7\_20},
  bibsource    = {dblp computer science bibliography, https://dblp.org}
}

@inproceedings{wittkopp2021loglab,
  author       = {Thorsten Wittkopp and
                  Philipp Wiesner and
                  Dominik Scheinert and
                  Alexander Acker},
  editor       = {Hakim Hacid and
                  Odej Kao and
                  Massimo Mecella and
                  Naouel Moha and
                  Hye{-}young Paik},
  title        = {LogLAB: Attention-Based Labeling of Log Data Anomalies via Weak Supervision},
  booktitle    = {Service-Oriented Computing - 19th International Conference, {ICSOC}
                  2021, Virtual Event, November 22-25, 2021, Proceedings},
  series       = {Lecture Notes in Computer Science},
  volume       = {13121},
  pages        = {700--707},
  publisher    = {Springer},
  year         = {2021},
  url          = {https://doi.org/10.1007/978-3-030-91431-8\_46},
  doi          = {10.1007/978-3-030-91431-8\_46},
  bibsource    = {dblp computer science bibliography, https://dblp.org}
}

@inproceedings{wittkopp2022pull,
  author       = {Thorsten Wittkopp and
                  Dominik Scheinert and
                  Philipp Wiesner and
                  Alexander Acker and
                  Odej Kao},
  editor       = {Tung X. Bui},
  title        = {{PULL:} Reactive Log Anomaly Detection Based On Iterative {PU} Learning},
  booktitle    = {56th Hawaii International Conference on System Sciences, {HICSS} 2023,
                  Maui, Hawaii, USA, January 3-6, 2023},
  pages        = {1376--1385},
  publisher    = {ScholarSpace},
  year         = {2023},
  url          = {https://hdl.handle.net/10125/102802},
  bibsource    = {dblp computer science bibliography, https://dblp.org}
}

@inproceedings{wittkopp2021taxonomy,
  title={A taxonomy of anomalies in log data},
  author={Wittkopp, Thorsten and Wiesner, Philipp and Scheinert, Dominik and Kao, Odej},
  booktitle={International Conference on Service-Oriented Computing},
  pages={153--164},
  year={2021},
  organization={Springer}
}

@inproceedings{wittkopp2024logrca,
  title={LogRCA: Log-based Root Cause Analysis for Distributed Services},
  author={Wittkopp, Thorsten and Wiesner, Philipp and Kao, Odej},
  booktitle={European Conference on Parallel Processing},
  year={2024},
  organization={Springer}
}

@inproceedings{wittkopp2020superiority,
  author       = {Thorsten Wittkopp and
                  Alexander Acker and
                  Sasho Nedelkoski and
                  Jasmin Bogatinovski and
                  Odej Kao},
  editor       = {Maria Ganzha and
                  Leszek A. Maciaszek and
                  Marcin Paprzycki},
  title        = {Superiority of Simplicity: {A} Lightweight Model for Network Device
                  Workload Prediction},
  booktitle    = {Proceedings of the 2020 Federated Conference on Computer Science and
                  Information Systems, FedCSIS 2020, Sofia, Bulgaria, September 6-9,
                  2020},
  series       = {Annals of Computer Science and Information Systems},
  volume       = {21},
  pages        = {7--10},
  year         = {2020},
  url          = {https://doi.org/10.15439/2020F149},
  doi          = {10.15439/2020F149}
}

@inproceedings{wittkopp2023progressing,
  author       = {Thorsten Wittkopp and
                  Alexander Acker and
                  Odej Kao},
  editor       = {Jihe Wang and
                  Yi He and
                  Thang N. Dinh and
                  Christan Grant and
                  Meikang Qiu and
                  Witold Pedrycz},
  title        = {Progressing from Anomaly Detection to Automated Log Labeling and Pioneering
                  Root Cause Analysis},
  booktitle    = {{IEEE} International Conference on Data Mining, {ICDM} 2023 - Workshops,
                  Shanghai, China, December 4, 2023},
  pages        = {1231--1239},
  publisher    = {{IEEE}},
  year         = {2023},
  url          = {https://doi.org/10.1109/ICDMW60847.2023.00160},
  doi          = {10.1109/ICDMW60847.2023.00160},
  bibsource    = {dblp computer science bibliography, https://dblp.org}
}

@inproceedings{DBLP:conf/cluster/ScheinertTZWAWK21,
  author       = {Dominik Scheinert and
                  Lauritz Thamsen and
                  Houkun Zhu and
                  Jonathan Will and
                  Alexander Acker and
                  Thorsten Wittkopp and
                  Odej Kao},
  title        = {Bellamy: Reusing Performance Models for Distributed Dataflow Jobs
                  Across Contexts},
  booktitle    = {{IEEE} International Conference on Cluster Computing, {CLUSTER} 2021,
                  Portland, OR, USA, September 7-10, 2021},
  pages        = {261--270},
  publisher    = {{IEEE}},
  year         = {2021},
  url          = {https://doi.org/10.1109/Cluster48925.2021.00052},
  doi          = {10.1109/CLUSTER48925.2021.00052},
  bibsource    = {dblp computer science bibliography, https://dblp.org}
}

@inproceedings{DBLP:conf/bigdataconf/ScheinertABWWT21,
  author       = {Dominik Scheinert and
                  Alireza Alamgiralem and
                  Jonathan Bader and
                  Jonathan Will and
                  Thorsten Wittkopp and
                  Lauritz Thamsen},
  editor       = {Yixin Chen and
                  Heiko Ludwig and
                  Yicheng Tu and
                  Usama M. Fayyad and
                  Xingquan Zhu and
                  Xiaohua Hu and
                  Suren Byna and
                  Xiong Liu and
                  Jianping Zhang and
                  Shirui Pan and
                  Vagelis Papalexakis and
                  Jianwu Wang and
                  Alfredo Cuzzocrea and
                  Carlos Ordonez},
  title        = {On the Potential of Execution Traces for Batch Processing Workload
                  Optimization in Public Clouds},
  booktitle    = {2021 {IEEE} International Conference on Big Data (Big Data), Orlando,
                  FL, USA, December 15-18, 2021},
  pages        = {3113--3118},
  publisher    = {{IEEE}},
  year         = {2021},
  doi          = {10.1109/BIGDATA52589.2021.9671275}
}

@inproceedings{wiesner2022cucumber,
  title={Cucumber: Renewable-aware admission control for delay-tolerant cloud and edge workloads},
  author={Wiesner, Philipp and Scheinert, Dominik and Wittkopp, Thorsten and Thamsen, Lauritz and Kao, Odej},
  booktitle={European Conference on Parallel Processing},
  pages={218--232},
  year={2022},
  organization={Springer}
}

@inproceedings{scheinert2023karasu,
  title={Karasu: A collaborative approach to efficient cluster configuration for big data analytics},
  author={Scheinert, Dominik and Wiesner, Philipp and Wittkopp, Thorsten and Thamsen, Lauritz and Will, Jona-than and Kao, Odej},
  booktitle={2023 IEEE International Performance, Computing, and Communications Conference (IPCCC)},
  pages={403--412},
  year={2023},
  organization={IEEE}
}

@inproceedings{notaro2020systematic,
  title={A systematic mapping study in AIOps},
  author={Notaro, Paolo and Cardoso, Jorge and Gerndt, Michael},
  booktitle={International Conference on Service-Oriented Computing},
  pages={110--123},
  year={2020},
  organization={Springer}
}

@article{notaro2021survey,
  title={A survey of aiops methods for failure management},
  author={Notaro, Paolo and Cardoso, Jorge and Gerndt, Michael},
  journal={ACM Transactions on Intelligent Systems and Technology (TIST)},
  volume={12},
  number={6},
  pages={1--45},
  year={2021},
  publisher={ACM New York, NY}
}

@inproceedings{VaswaniSPUJGKP17,
  author    = {Ashish Vaswani and Noam Shazeer and Niki Parmar and Jakob Uszkoreit and Llion Jones and Aidan N. Gomez and Lukasz Kaiser and Illia Polosukhin},
  editor    = {Isabelle Guyon and Ulrike von Luxburg and Samy Bengio and Hanna M. Wallach and Rob Fergus and S. V. N. Vishwanathan and Roman Garnett},
  title     = {Attention is All you Need},
  booktitle = {NeurIPS},
  year      = {2017}
}

@article{devlin2018bert,
  title={Bert: Pre-training of deep bidirectional transformers for language understanding},
  author={Devlin, Jacob and Chang, Ming-Wei and Lee, Kenton and Toutanova, Kristina},
  journal={arXiv preprint arXiv:1810.04805},
  year={2018}
}

@inproceedings{sukhwani2017monitoring,
  title={Monitoring and mitigating software aging on IBM cloud controller system},
  author={Sukhwani, Harish and Matias, Rivalino and Trivedi, Kishor S and Rindos, Andy},
  booktitle={ISSREW},
  year={2017},
  organization={IEEE}
}

@inproceedings{DevlinCLT19,
  author    = {Jacob Devlin and Ming{-}Wei Chang and Kenton Lee and Kristina Toutanova},
  editor    = {Jill Burstein and Christy Doran and Thamar Solorio},
  title     = {{BERT:} Pre-training of Deep Bidirectional Transformers for Language Understanding},
  booktitle = {NAACL-HLT},
  publisher = {Association for Computational Linguistics},
  year      = {2019}
}

@inproceedings{GehringAGYD17,
  author    = {Jonas Gehring and Michael Auli and David Grangier and Denis Yarats and Yann N. Dauphin},
  editor    = {Doina Precup and Yee Whye Teh},
  title     = {Convolutional Sequence to Sequence Learning},
  booktitle = {ICML},
  publisher = {{PMLR}},
  year      = {2017}
}

@inproceedings{oliner2007datasets,
  author={Oliner, Adam and Stearley, Jon},
  booktitle={DSN}, 
  title={What Supercomputers Say: A Study of Five System Logs}, 
  year={2007}
}

@article{kowsari2019text,
  title={Text classification algorithms: A survey},
  author={Kowsari, Kamran and Jafari Meimandi, Kiana and Heidarysafa, Mojtaba and Mendu, Sanjana and Barnes, Laura and Brown, Donald},
  journal={Information},
  volume={10},
  number={4},
  year={2019},
  publisher={Multidisciplinary Digital Publishing Institute}
}

@inproceedings{he2016experience,
  author       = {Shilin He and
                  Jieming Zhu and
                  Pinjia He and
                  Michael R. Lyu},
  title        = {Experience Report: System Log Analysis for Anomaly Detection},
  booktitle    = {27th {IEEE} International Symposium on Software Reliability Engineering,
                  {ISSRE} 2016, Ottawa, ON, Canada, October 23-27, 2016},
  pages        = {207--218},
  publisher    = {{IEEE} Computer Society},
  year         = {2016},
  url          = {https://doi.org/10.1109/ISSRE.2016.21},
  doi          = {10.1109/ISSRE.2016.21},
  bibsource    = {dblp computer science bibliography, https://dblp.org}
}

@inproceedings{glorot2010understanding,
  title={Understanding the difficulty of training deep feedforward neural networks},
  author={Glorot, Xavier and Bengio, Yoshua},
  booktitle={AISTATS},
  year={2010},
  organization={JMLR Workshop and Conference Proceedings}
}

@inproceedings{baier2019cope,
  title={How to cope with change?-preserving validity of predictive services over time},
  author={Baier, Lucas and K{\"u}hl, Niklas and Satzger, Gerhard},
  booktitle={Proceedings of the 52nd Hawaii International Conference on System Sciences},
  year={2019}
}

@inproceedings{liu2020survey,
  title={A Survey of Text Data Augmentation},
  author={Liu, Pei and Wang, Xuemin and Xiang, Chao and Meng, Weiye},
  booktitle={2020 International Conference on Computer Communication and Network Security (CCNS)},
  pages={191--195},
  year={2020},
  organization={IEEE}
}

@article{shorten2019survey,
  title={A survey on image data augmentation for deep learning},
  author={Shorten, Connor and Khoshgoftaar, Taghi M},
  journal={Journal of Big Data},
  volume={6},
  number={1},
  pages={1--48},
  year={2019},
  publisher={Springer}
}

@article{steinwart2005classification,
  title={A Classification Framework for Anomaly Detection.},
  author={Steinwart, Ingo and Hush, Don and Scovel, Clint},
  journal={Journal of Machine Learning Research},
  volume={6},
  number={2},
  year={2005}
}

@article{ramponi2020neural,
  title={Neural Unsupervised Domain Adaptation in NLP - A Survey},
  author={Ramponi, Alan and Plank, Barbara},
  journal={arXiv preprint arXiv:2006.00632},
  year={2020}
}

@article{qi2020small,
  title={Small data challenges in big data era: A survey of recent progress on unsupervised and semi-supervised methods},
  author={Qi, Guo-Jun and Luo, Jiebo},
  journal={IEEE Transactions on Pattern Analysis and Machine Intelligence},
  year={2020},
  publisher={IEEE}
}

@article{naseer2018enhanced,
  title={Enhanced network anomaly detection based on deep neural networks},
  author={Naseer, Sheraz and Saleem, Yasir and Khalid, Shehzad and Bashir, Muhammad Khawar and Han, Jihun and Iqbal, Muhammad Munwar and Han, Kijun},
  journal={IEEE access},
  volume={6},
  pages={48231--48246},
  year={2018},
  publisher={IEEE}
}

@inproceedings{nedelkoski2020self,
  title={Self-supervised log parsing},
  author={Nedelkoski, Sasho and Bogatinovski, Jasmin and Acker, Alexander and Cardoso, Jorge and Kao, Odej},
  booktitle={Machine Learning and Knowledge Discovery in Databases: Applied Data Science Track: European Conference, ECML PKDD 2020, Ghent, Belgium, September 14--18, 2020, Proceedings, Part IV},
  pages={122--138},
  year={2021},
  organization={Sprin-ger}
}

@inproceedings{wen2021time,
  author    = {Qingsong Wen and Liang Sun and Fan Yang and Xiaomin Song and Jingkun Gao and Xue Wang and Huan Xu},
  title     = {Time Series Data Augmentation for Deep Learning: {A} Survey},
  booktitle = {IJCAI},
  publisher = {ijcai.org},
  year      = {2021}
}

@inproceedings{liang2007failure,
  title={Failure prediction in ibm bluegene/l event logs},
  author={Liang, Yinglung and Zhang, Yanyong and Xiong, Hui and Sahoo, Ramendra},
  booktitle={ICDM},
  year={2007},
  organization={IEEE}
}

@article{hearst1998support,
  title={Support vector machines},
  author={Hearst, Marti A. and Dumais, Susan T and Osuna, Edgar and Platt, John and Scholkopf, Bernhard},
  journal={IEEE Intelligent Systems and their applications},
  volume={13},
  number={4},
  pages={18--28},
  year={1998},
  publisher={IEEE}
}

@article{FarzadG20,
  author    = {Amir Farzad and T. Aaron Gulliver},
  title     = {Unsupervised log message anomaly detection},
  journal   = {{ICT} Express},
  year      = {2020},
}

@inproceedings{YuanALYL020,
  author    = {Yali Yuan and Sripriya Srikant Adhatarao and Mingkai Lin and Yachao Yuan and Zheli Liu and Xiaoming Fu},
  title     = {{ADA:} Adaptive Deep Log Anomaly Detector},
  booktitle = {IEEE International Conference on Computer Communications (INFOCOM)},
  year      = {2020}
}

@inproceedings{hamooni2016logmine,
  title={Logmine: Fast pattern recognition for log analytics},
  author={Hamooni, Hossein and Debnath, Biplob and Xu, Jianwu and Zhang, Hui and Jiang, Guofei and Mueen, Abdullah},
  booktitle={ACM International on Conference on Information and Knowledge Management (CIKM)},
  year={2016}
}

@inproceedings{meng2018device,
  title={Device-agnostic log anomaly classification with partial labels},
  author={Meng, Weibin and Liu, Ying and Zhang, Shenglin and Pei, Dan and Dong, Hui and Song, Lei and Luo, Xulong},
  booktitle={2018 IEEE/ACM 26th International Symposium on Quality of Service (IWQoS)},
  year={2018},
  organization={IEEE}
}

@inproceedings{gulenko2020ai,
  title={{AI}-Governance and Levels of Automation for AIOps-supported System Administration},
  author={Gulenko, Anton and Acker, Alexander and Kao, Odej and Liu, Feng},
  booktitle={IEEE International Conference on Computer Communications and Networks (ICCCN)},
  year={2020}
}

@inproceedings{rosendo2018improve,
  title={How to improve cloud services availability? Investigating the impact of power and it subsystems failures},
  author={Rosendo, Daniel and Leoni, Guto and Gomes, Demis and Moreira, Andr{\'e} and Gon{\c{c}}alves, Glauco and Endo, Patricia and Kelner, Judith and Sadok, Djamel and Mahloo, Mozhgan},
  booktitle={Hawaii International Conference on System Sciences (HICSS)},
  year={2018}
}

@book{sridharan2018distributed,
  title={Distributed systems observability: a guide to building robust systems},
  author={Sridharan, Cindy},
  year={2018},
  publisher={O'Reilly Media}
}

@article{korzeniowski2022landscape,
  title={Landscape of Automated Log Analysis: A Systematic Literature Review and Mapping Study},
  author={Korzeniowski, {\L}ukasz and Goczy{\l}a, Krzysztof},
  journal={IEEE Access},
  year={2022},
  publisher={IEEE}
}

@inproceedings{bogatinovski2022qulog,
  title={Qulog: Data-driven approach for log instruction quality assessment},
  author={Bogatinovski, Jasmin and Nedelkoski, Sasho and Acker, Alexander and Cardoso, Jorge and Kao, Odej},
  booktitle={Proceedings of the 30th IEEE/ACM International Conference on Program Comprehension},
  pages={275--286},
  year={2022}
}

@inproceedings{yates2021pretrained,
  title={Pretrained transformers for text ranking: BERT and beyond},
  author={Yates, Andrew and Nogueira, Rodrigo and Lin, Jimmy},
  booktitle={Proceedings of the 14th ACM International Conference on web search and data mining},
  pages={1154--1156},
  year={2021}
}

@article{gulenko2016system,
  title={A system architecture for real-time anomaly detection in large-scale nfv systems},
  author={Gulenko, Anton and Wallschl{\"a}ger, Marcel and Schmidt, Florian and Kao, Odej and Liu, Feng},
  journal={Procedia Computer Science},
  volume={94},
  pages={491--496},
  year={2016},
  publisher={Elsevier}
}

@article{yang2019nlsalog,
  author       = {Ruipeng Yang and
                  Dan Qu and
                  Ying Gao and
                  Yekui Qian and
                  Yongwang Tang},
  title        = {nLSALog: An Anomaly Detection Framework for Log Sequence in Security
                  Management},
  journal      = {{IEEE} Access},
  volume       = {7},
  pages        = {181152--181164},
  year         = {2019},
  url          = {https://doi.org/10.1109/ACCESS.2019.2953981},
  doi          = {10.1109/ACCESS.2019.2953981},
  bibsource    = {dblp computer science bibliography, https://dblp.org}
}

@article{krizhevsky2012imagenet,
  author       = {Alex Krizhevsky and
                  Ilya Sutskever and
                  Geoffrey E. Hinton},
  editor       = {Peter L. Bartlett and
                  Fernando C. N. Pereira and
                  Christopher J. C. Burges and
                  L{\'{e}}on Bottou and
                  Kilian Q. Weinberger},
  title        = {ImageNet Classification with Deep Convolutional Neural Networks},
  booktitle    = {Advances in Neural Information Processing Systems 25: 26th Annual
                  Conference on Neural Information Processing Systems 2012. Proceedings
                  of a meeting held December 3-6, 2012, Lake Tahoe, Nevada, United States},
  pages        = {1106--1114},
  year         = {2012},
  url          = {https://proceedings.neurips.cc/paper/2012/hash/c399862d3b9d6b76c8436e924a68c45b-Abstract.html},
  bibsource    = {dblp computer science bibliography, https://dblp.org}
}

@article{oliner2012advances,
  author       = {Adam J. Oliner and
                  Archana Ganapathi and
                  Wei Xu},
  title        = {Advances and challenges in log analysis},
  journal      = {Commun. {ACM}},
  volume       = {55},
  number       = {2},
  pages        = {55--61},
  year         = {2012},
  url          = {https://doi.org/10.1145/2076450.2076466},
  doi          = {10.1145/2076450.2076466},
  bibsource    = {dblp computer science bibliography, https://dblp.org}
}

@inproceedings{elkan2008learning,
  author       = {Charles Elkan and
                  Keith Noto},
  editor       = {Ying Li and
                  Bing Liu and
                  Sunita Sarawagi},
  title        = {Learning classifiers from only positive and unlabeled data},
  booktitle    = {Proceedings of the 14th {ACM} {SIGKDD} International Conference on
                  Knowledge Discovery and Data Mining, Las Vegas, Nevada, USA, August
                  24-27, 2008},
  pages        = {213--220},
  publisher    = {{ACM}},
  year         = {2008},
  url          = {https://doi.org/10.1145/1401890.1401920},
  doi          = {10.1145/1401890.1401920},
  bibsource    = {dblp computer science bibliography, https://dblp.org}
}

@article{ratner2020snorkel,
  author       = {Alexander Ratner and
                  Stephen H. Bach and
                  Henry R. Ehrenberg and
                  Jason A. Fries and
                  Sen Wu and
                  Christopher R{\'{e}}},
  title        = {Snorkel: rapid training data creation with weak supervision},
  journal      = {{VLDB} J.},
  volume       = {29},
  number       = {2-3},
  pages        = {709--730},
  year         = {2020},
  url          = {https://doi.org/10.1007/s00778-019-00552-1},
  doi          = {10.1007/S00778-019-00552-1},
  bibsource    = {dblp computer science bibliography, https://dblp.org}
}

@inproceedings{sun2017revisiting,
  author       = {Chen Sun and
                  Abhinav Shrivastava and
                  Saurabh Singh and
                  Abhinav Gupta},
  title        = {Revisiting Unreasonable Effectiveness of Data in Deep Learning Era},
  booktitle    = {{IEEE} International Conference on Computer Vision, {ICCV} 2017, Venice,
                  Italy, October 22-29, 2017},
  pages        = {843--852},
  publisher    = {{IEEE} Computer Society},
  year         = {2017},
  url          = {https://doi.org/10.1109/ICCV.2017.97},
  doi          = {10.1109/ICCV.2017.97},
  bibsource    = {dblp computer science bibliography, https://dblp.org}
}

@inproceedings{he2017drain,
  title={Drain: An online log parsing approach with fixed depth tree},
  author={He, Pinjia and Zhu, Jieming and Zheng, Zibin and Lyu, Michael R},
  booktitle={IEEE International Conference on Web Services (ICWS)},
  year={2017}
}

@inproceedings{zhu2019tools,
  author       = {Jieming Zhu and
                  Shilin He and
                  Jinyang Liu and
                  Pinjia He and
                  Qi Xie and
                  Zibin Zheng and
                  Michael R. Lyu},
  editor       = {Helen Sharp and
                  Mike Whalen},
  title        = {Tools and benchmarks for automated log parsing},
  booktitle    = {Proceedings of the 41st International Conference on Software Engineering:
                  Software Engineering in Practice, {ICSE} {(SEIP)} 2019, Montreal,
                  QC, Canada, May 25-31, 2019},
  pages        = {121--130},
  publisher    = {{IEEE} / {ACM}},
  year         = {2019},
  url          = {https://doi.org/10.1109/ICSE-SEIP.2019.00021},
  doi          = {10.1109/ICSE-SEIP.2019.00021},
  bibsource    = {dblp computer science bibliography, https://dblp.org}
}

@inproceedings{zhang2019robust,
  title={Robust log-based anomaly detection on unstable log data},
  author={Zhang, Xu and Xu, Yong and Lin, Qingwei and Qiao, Bo and Zhang, Hongyu and Dang, Yingnong and Xie, Chunyu and Yang, Xinsheng and Cheng, Qian and Li, Ze and others},
  booktitle={ESEC/FSE},
  year={2019}
}

@inproceedings{yang2021semi,
  title={Semi-supervised log-based anomaly detection via probabilistic label estimation},
  author={Yang, Lin and Chen, Junjie and Wang, Zan and Wang, Weijing and Jiang, Jiajun and Dong, Xuyuan and Zhang, Wenbin},
  booktitle={2021 IEEE/ACM 43rd International Conference on Software Engineering (ICSE)},
  pages={1448--1460},
  year={2021},
  organization={IEEE}
}

@inproceedings{guo2021logbert,
  author       = {Haixuan Guo and
                  Shuhan Yuan and
                  Xintao Wu},
  title        = {LogBERT: Log Anomaly Detection via {BERT}},
  booktitle    = {International Joint Conference on Neural Networks, {IJCNN} 2021, Shenzhen,
                  China, July 18-22, 2021},
  pages        = {1--8},
  publisher    = {{IEEE}},
  year         = {2021},
  url          = {https://doi.org/10.1109/IJCNN52387.2021.9534113},
  doi          = {10.1109/IJCNN52387.2021.9534113},
  bibsource    = {dblp computer science bibliography, https://dblp.org}
}

@inproceedings{du2017deeplog,
  title={Deeplog: Anomaly detection and diagnosis from system logs through deep learning},
  author={Du, Min and Li, Feifei and Zheng, Guineng and Srikumar, Vivek},
  booktitle={Proceedings of the 2017 ACM SIGSAC Conference on Computer and Communications Security},
  pages={1285--1298},
  year={2017}
}

@inproceedings{nedelkoski2020logsy,
  author={S. {Nedelkoski} and J. {Bogatinovski} and A. {Acker} and J. {Cardoso} and O. {Kao}},
  booktitle={2020 IEEE International Conference on Data Mining (ICDM)}, 
  title={Self-Attentive Classification-Based Anomaly Detection in Unstructured Logs}, 
  year={2020},
  volume={},
  number={},
  pages={1196-1201},
  doi={10.1109/ICDM50108.2020.00148}
}

@inproceedings{meng2019loganomaly,
  author       = {Weibin Meng and
                  Ying Liu and
                  Yichen Zhu and
                  Shenglin Zhang and
                  Dan Pei and
                  Yuqing Liu and
                  Yihao Chen and
                  Ruizhi Zhang and
                  Shimin Tao and
                  Pei Sun and
                  Rong Zhou},
  editor       = {Sarit Kraus},
  title        = {LogAnomaly: Unsupervised Detection of Sequential and Quantitative
                  Anomalies in Unstructured Logs},
  booktitle    = {Proceedings of the Twenty-Eighth International Joint Conference on
                  Artificial Intelligence, {IJCAI} 2019, Macao, China, August 10-16,
                  2019},
  pages        = {4739--4745},
  publisher    = {ijcai.org},
  year         = {2019},
  url          = {https://doi.org/10.24963/ijcai.2019/658},
  doi          = {10.24963/IJCAI.2019/658},
  bibsource    = {dblp computer science bibliography, https://dblp.org}
}

@inproceedings{li2020swisslog,
  author       = {Xiaoyun Li and
                  Pengfei Chen and
                  Linxiao Jing and
                  Zilong He and
                  Guangba Yu},
  editor       = {Marco Vieira and
                  Henrique Madeira and
                  Nuno Antunes and
                  Zheng Zheng},
  title        = {SwissLog: Robust and Unified Deep Learning Based Log Anomaly Detection
                  for Diverse Faults},
  booktitle    = {31st {IEEE} International Symposium on Software Reliability Engineering,
                  {ISSRE} 2020, Coimbra, Portugal, October 12-15, 2020},
  pages        = {92--103},
  publisher    = {{IEEE}},
  year         = {2020},
  url          = {https://doi.org/10.1109/ISSRE5003.2020.00018},
  doi          = {10.1109/ISSRE5003.2020.00018},
  bibsource    = {dblp computer science bibliography, https://dblp.org}
}

@inproceedings{yin2020improving,
  author       = {Kun Yin and
                  Meng Yan and
                  Ling Xu and
                  Zhou Xu and
                  Zhao Li and
                  Dan Yang and
                  Xiaohong Zhang},
  title        = {Improving Log-Based Anomaly Detection with Component-Aware Analysis},
  booktitle    = {{IEEE} International Conference on Software Maintenance and Evolution,
                  {ICSME} 2020, Adelaide, Australia, September 28 - October 2, 2020},
  pages        = {667--671},
  publisher    = {{IEEE}},
  year         = {2020},
  url          = {https://doi.org/10.1109/ICSME46990.2020.00069},
  doi          = {10.1109/ICSME46990.2020.00069},
  bibsource    = {dblp computer science bibliography, https://dblp.org}
}

@inproceedings{nicolau2016hybrid,
  author       = {Van Loi Cao and
                  Miguel Nicolau and
                  James McDermott},
  editor       = {Julia Handl and
                  Emma Hart and
                  Peter R. Lewis and
                  Manuel L{\'{o}}pez{-}Ib{\'{a}}{\~{n}}ez and
                  Gabriela Ochoa and
                  Ben Paechter},
  title        = {A Hybrid Autoencoder and Density Estimation Model for Anomaly Detection},
  booktitle    = {Parallel Problem Solving from Nature - {PPSN} {XIV} - 14th International
                  Conference, Edinburgh, UK, September 17-21, 2016, Proceedings},
  series       = {Lecture Notes in Computer Science},
  volume       = {9921},
  pages        = {717--726},
  publisher    = {Springer},
  year         = {2016},
  url          = {https://doi.org/10.1007/978-3-319-45823-6\_67},
  doi          = {10.1007/978-3-319-45823-6\_67},
  bibsource    = {dblp computer science bibliography, https://dblp.org}
}

@inproceedings{sakurada2014anomaly,
  author       = {Mayu Sakurada and
                  Takehisa Yairi},
  editor       = {Ashfaqur Rahman and
                  Jeremiah D. Deng and
                  Jiuyong Li},
  title        = {Anomaly Detection Using Autoencoders with Nonlinear Dimensionality
                  Reduction},
  booktitle    = {Proceedings of the {MLSDA} 2014 2nd Workshop on Machine Learning for
                  Sensory Data Analysis, Gold Coast, Australia, QLD, Australia, December
                  2, 2014},
  pages        = {4},
  publisher    = {{ACM}},
  year         = {2014},
  url          = {https://doi.org/10.1145/2689746.2689747},
  doi          = {10.1145/2689746.2689747},
  bibsource    = {dblp computer science bibliography, https://dblp.org}
}

@article{chandola2009anomaly,
  title={Anomaly detection: A survey},
  author={Chandola, Varun and Banerjee, Arindam and Kumar, Vipin},
  journal={ACM computing surveys (CSUR)},
  volume={41},
  number={3},
  year={2009},
  publisher={ACM New York, NY, USA}
}

@article{blazquez2020review,
  title={A review on outlier/anomaly detection in time series data},
  author={Blazquez-Garcia, Ane and Conde, Angel and Mori, Usue and Lozano, Jose A},
  journal={arXiv preprint arXiv:2002.04236},
  year={2020}
}

@inproceedings{sebestyen2018taxonomy,
  title={A taxonomy and platform for anomaly detection},
  author={Sebestyen, Gheorghe and Hangan, Anca and Czako, Zoltan and Kovacs, Gyorgy},
  booktitle={AQTR},
  year={2018},
  organization={IEEE}
}

@inproceedings{NagarajKN12,
  author    = {Karthik Nagaraj and Charles Edwin Killian and Jennifer Neville},
  title     = {Structured Comparative Analysis of Systems Logs to Diagnose Performance Problems},
  booktitle = {NSDI},
  publisher = {{USENIX} Association},
  year      = {2012}
 }

@article{song2007conditional,
  author    = {Xiuyao Song and Mingxi Wu and Christopher M. Jermaine and Sanjay Ranka},
  title     = {Conditional Anomaly Detection},
  journal   = {{IEEE} Trans. Knowl. Data Eng.},
  volume    = {19},
  number    = {5},
  year      = {2007},
}

@article{zhou2018brief,
  title={A brief introduction to weakly supervised learning},
  author={Zhou, Zhi-Hua},
  journal={National science review},
  volume={5},
  number={1},
  year={2018},
  publisher={Oxford University Press}
}

@inproceedings{liu2003building,
  title={Building text classifiers using positive and unlabeled examples},
  author={Liu, Bing and Dai, Yang and Li, Xiaoli and Lee, Wee Sun and Yu, Philip S},
  booktitle={ICDM},
  year={2003},
  organization={IEEE}
}

@inproceedings{liu2002partially,
  title={Partially supervised classification of text documents},
  author={Liu, Bing and Lee, Wee Sun and Yu, Philip S and Li, Xiaoli},
  booktitle={ICML},
  year={2002},
  organization={Sydney, NSW}
}

@article{zhu2009introduction,
  author       = {Xiaojin Zhu and
                  Andrew B. Goldberg},
  title        = {Introduction to Semi-Supervised Learning},
  series       = {Synthesis Lectures on Artificial Intelligence and Machine Learning},
  publisher    = {Morgan {\&} Claypool Publishers},
  year         = {2009},
  url          = {https://doi.org/10.2200/S00196ED1V01Y200906AIM006},
  doi          = {10.2200/S00196ED1V01Y200906AIM006},
  isbn         = {978-3-031-00420-9},
  bibsource    = {dblp computer science bibliography, https://dblp.org}
}

@article{bekker2020learning,
  title={Learning from positive and unlabeled data: A survey},
  author={Bekker, Jessa and Davis, Jesse},
  journal={Machine Learning},
  volume={109},
  number={4},
  year={2020},
  publisher={Springer}
}

@article{fusilier2015detecting,
  author       = {Donato Hern{\'{a}}ndez Fusilier and
                  Manuel Montes{-}y{-}G{\'{o}}mez and
                  Paolo Rosso and
                  Rafael Guzm{\'{a}}n{-}Cabrera},
  title        = {Detecting positive and negative deceptive opinions using PU-learning},
  journal      = {Information Process Management},
  volume       = {51},
  number       = {4},
  pages        = {433--443},
  year         = {2015},
  url          = {https://doi.org/10.1016/j.ipm.2014.11.001},
  doi          = {10.1016/J.IPM.2014.11.001},
  bibsource    = {dblp computer science bibliography, https://dblp.org}
}

@article{mordelet2014bagging,
  author       = {Fantine Mordelet and
                  Jean{-}Philippe Vert},
  title        = {A bagging {SVM} to learn from positive and unlabeled examples},
  journal      = {Pattern Recognition Letters},
  volume       = {37},
  pages        = {201--209},
  year         = {2014},
  url          = {https://doi.org/10.1016/j.patrec.2013.06.010},
  doi          = {10.1016/J.PATREC.2013.06.010},
  bibsource    = {dblp computer science bibliography, https://dblp.org}
}

@inproceedings{LuoCLJ18,
  author       = {Yuxuan Luo and
                  Shaoyin Cheng and
                  Chong Liu and
                  Fan Jiang},
  title        = {PU Learning in Payload-based Web Anomaly Detection},
  booktitle    = {Third International Conference on Security of Smart Cities, Industrial
                  Control System and Communications, {SSIC} 2018, Shanghai, China, October
                  18-19, 2018},
  pages        = {1--5},
  publisher    = {{IEEE}},
  year         = {2018},
  url          = {https://doi.org/10.1109/SSIC.2018.8556662},
  doi          = {10.1109/SSIC.2018.8556662},
  bibsource    = {dblp computer science bibliography, https://dblp.org}
}

@article{WuCWWZW20,
  author       = {Zhiang Wu and
                  Jie Cao and
                  Yaqiong Wang and
                  Youquan Wang and
                  Lu Zhang and
                  Junjie Wu},
  title        = {hPSD: A Hybrid PU-Learning-Based Spammer Detection Model for Product
                  Reviews},
  journal      = {{IEEE} Trans. Cybern.},
  volume       = {50},
  number       = {4},
  pages        = {1595--1606},
  year         = {2020},
  url          = {https://doi.org/10.1109/TCYB.2018.2877161},
  doi          = {10.1109/TCYB.2018.2877161},
  bibsource    = {dblp computer science bibliography, https://dblp.org}
}

@article{wen2020time,
  title={Time series data augmentation for deep learning: A survey},
  author={Wen, Qingsong and Sun, Liang and Yang, Fan and Song, Xiaomin and Gao, Jingkun and Wang, Xue and Xu, Huan},
  journal={arXiv preprint arXiv:2002.12478},
  year={2020}
}

@article{ratner2016data,
  title={Data programming: Creating large training sets, quickly},
  author={Ratner, Alexander J and De Sa, Christopher M and Wu, Sen and Selsam, Daniel and Re, Christopher},
  journal={Advances in neural information processing systems},
  volume={29},
  year={2016}
}

@inproceedings{LinZLZC16,
  author    = {Qingwei Lin and Hongyu Zhang and Jian{-}Guang Lou and Yu Zhang and Xuewei Chen},
  title     = {Log clustering based problem identification for online service systems},
  booktitle = {International Conference on Software Engineering (ICSE)},
  publisher = {{ACM}},
  year      = {2016}
}

@article{notaro2023logrule,
  title={LogRule: Efficient Structured Log Mining for Root Cause Analysis},
  author={Notaro, Paolo and Haeri, Soroush and Cardoso, Jorge and Gerndt, Michael},
  journal={IEEE Transactions on Network and Service Management},
  year={2023},
  publisher={IEEE}
}

@inproceedings{lu2017log,
  title={Log-based abnormal task detection and root cause analysis for spark},
  author={Lu, Siyang and Rao, BingBing and Wei, Xiang and Tak, Byungchul and Wang, Long and Wang, Liqiang},
  booktitle={IEEE International Conference on Web Services (ICWS)},
  year={2017}
}

@article{lu2019ladra,
  title={LADRA: Log-based abnormal task detection and root-cause analysis in big data processing with Spark},
  author={Lu, Siyang and Wei, Xiang and Rao, Bingbing and Tak, Byungchul and Wang, Long and Wang, Liqiang},
  journal={Future Generation Computer Systems},
  volume={95},
  year={2019},
  publisher={Elsevier}
}

@inproceedings{zawawy2010log,
  title={Log filtering and interpretation for root cause analysis},
  author={Zawawy, Hamzeh and Kontogiannis, Kostas and Mylopoulos, John},
  booktitle={IEEE International Conference on Software Maintenance (ICSM)},
  year={2010}
}

@inproceedings{9529498,
  author       = {Richard Jarry and
                  Satoru Kobayashi and
                  Kensuke Fukuda},
  title        = {A Quantitative Causal Analysis for Network Log Data},
  booktitle    = {{IEEE} 45th Annual Computers, Software, and Applications Conference,
                  {COMPSAC} 2021, Madrid, Spain, July 12-16, 2021},
  pages        = {1437--1442},
  publisher    = {{IEEE}},
  year         = {2021},
  url          = {https://doi.org/10.1109/COMPSAC51774.2021.00213},
  doi          = {10.1109/COMPSAC51774.2021.00213},
  bibsource    = {dblp computer science bibliography, https://dblp.org}
}

@book{jolliffe2005principal,
  author       = {Ian T. Jolliffe},
  title        = {Principal Component Analysis},
  series       = {Springer Series in Statistics},
  publisher    = {Springer},
  year         = {1986},
  url          = {https://doi.org/10.1007/978-1-4757-1904-8},
  doi          = {10.1007/978-1-4757-1904-8},
  isbn         = {978-1-4757-1906-2},
  bibsource    = {dblp computer science bibliography, https://dblp.org}
}

@incollection{breier2015anomaly,
  title={Anomaly detection from log files using data mining techniques},
  author={Breier, Jakub and Brani{\v{s}}ov{\'a}, Jana},
  booktitle={Information Science and Applications},
  pages={449--457},
  year={2015},
  publisher={Springer}
}

@inproceedings{chen2004failure,
  author       = {Mike Y. Chen and
                  Alice X. Zheng and
                  Jim Lloyd and
                  Michael I. Jordan and
                  Eric A. Brewer},
  title        = {Failure Diagnosis Using Decision Trees},
  booktitle    = {1st International Conference on Autonomic Computing {(ICAC} 2004),
                  17-19 May 2004, New York, NY, {USA}},
  pages        = {36--43},
  publisher    = {{IEEE} Computer Society},
  year         = {2004},
  url          = {https://doi.ieeecomputersociety.org/10.1109/ICAC.2004.31},
  doi          = {10.1109/ICAC.2004.31},
  bibsource    = {dblp computer science bibliography, https://dblp.org}
}

@inproceedings{lou2010mining,
  author       = {Jian{-}Guang Lou and
                  Qiang Fu and
                  Shengqi Yang and
                  Ye Xu and
                  Jiang Li},
  editor       = {Paul Barham and
                  Timothy Roscoe},
  title        = {Mining Invariants from Console Logs for System Problem Detection},
  booktitle    = {2010 {USENIX} Annual Technical Conference, Boston, MA, USA, June 23-25,
                  2010},
  publisher    = {{USENIX} Association},
  year         = {2010},
  url          = {https://www.usenix.org/conference/usenix-atc-10/mining-invariants-console-logs-system-problem-detection},
  bibsource    = {dblp computer science bibliography, https://dblp.org},
  pages        = {1--14}
}

@article{cinque2012event,
  author       = {Marcello Cinque and
                  Domenico Cotroneo and
                  Antonio Pecchia},
  title        = {Event Logs for the Analysis of Software Failures: {A} Rule-Based Approach},
  journal      = {IEEE Transactions on Software Engineering},
  publisher    = {IEEE},
  volume       = {39},
  number       = {6},
  pages        = {806--821},
  year         = {2013},
  url          = {https://doi.org/10.1109/TSE.2012.67},
  doi          = {10.1109/TSE.2012.67},
  bibsource    = {dblp computer science bibliography, https://dblp.org}
}

@inproceedings{xu2009detecting,
  author       = {Wei Xu and
                  Ling Huang and
                  Armando Fox and
                  David A. Patterson and
                  Michael I. Jordan},
  editor       = {Johannes F{\"{u}}rnkranz and
                  Thorsten Joachims},
  title        = {Detecting Large-Scale System Problems by Mining Console Logs},
  booktitle    = {Proceedings of the 27th International Conference on Machine Learning
                  (ICML-10), June 21-24, 2010, Haifa, Israel},
  pages        = {37--46},
  publisher    = {Omnipress},
  year         = {2010},
  url          = {https://icml.cc/Conferences/2010/papers/902.pdf},
  bibsource    = {dblp computer science bibliography, https://dblp.org}
}

@inproceedings{baseman2016relational,
  author       = {Elisabeth Baseman and
                  Sean Blanchard and
                  Zongze Li and
                  Song Fu},
  title        = {Relational Synthesis of Text and Numeric Data for Anomaly Detection
                  on Computing System Logs},
  booktitle    = {15th {IEEE} International Conference on Machine Learning and Applications,
                  {ICMLA} 2016, Anaheim, CA, USA, December 18-20, 2016},
  pages        = {882--885},
  publisher    = {{IEEE} Computer Society},
  year         = {2016},
  url          = {https://doi.org/10.1109/ICMLA.2016.0158},
  doi          = {10.1109/ICMLA.2016.0158},
  bibsource    = {dblp computer science bibliography, https://dblp.org}
}

@article{LANDAUER201894,
    title = {Dynamic log file analysis: An unsupervised cluster evolution approach for anomaly detection},
    journal = {Computers \& Security},
    volume = {79},
    pages = {94-116},
    year = {2018},
    issn = {0167-4048},
    doi = {https://doi.org/10.1016/j.cose.2018.08.009},
    url = {https://www.sciencedirect.com/science/article/pii/S0167404818306333},
    author = {Max Landauer and Markus Wurzenberger and Florian Skopik and Giuseppe Settanni and Peter Filzmoser}
}

@article{schapire2000boostexter,
  title={BoosTexter: A boosting-based system for text categorization},
  author={Schapire, Robert E and Singer, Yoram},
  journal={Machine learning},
  volume={39},
  number={2},
  year={2000},
  publisher={Springer}
}

@article{quinlan1986induction,
  title={Induction of decision trees},
  author={Quinlan, J. Ross},
  journal={Machine learning},
  volume={1},
  number={1},
  year={1986},
  publisher={Springer}
}

@article{safavian1991survey,
  title={A survey of decision tree classifier methodology},
  author={Safavian, S Rasoul and Landgrebe, David},
  journal={IEEE transactions on systems, man, and cybernetics},
  volume={21},
  number={3},
  year={1991},
  publisher={IEEE}
}

@article{manevitz2001one,
  title={One-class SVMs for document classification},
  author={Manevitz, Larry M and Yousef, Malik},
  journal={Journal of machine Learning research},
  volume={2},
  number={Dec},
  year={2001}
}

@inproceedings{ho1995random,
  author       = {Tin Kam Ho},
  title        = {Random decision forests},
  booktitle    = {Third International Conference on Document Analysis and Recognition,
                  {ICDAR} 1995, August 14 - 15, 1995, Montreal, Canada. Volume {I}},
  pages        = {278--282},
  publisher    = {{IEEE} Computer Society},
  year         = {1995},
  url          = {https://doi.org/10.1109/ICDAR.1995.598994},
  doi          = {10.1109/ICDAR.1995.598994},
  bibsource    = {dblp computer science bibliography, https://dblp.org}
}

@book{hosmer2013applied,
  title={Applied logistic regression},
  author={Hosmer Jr, David W and Lemeshow, Stanley and Sturdivant, Rodney X},
  volume={398},
  year={2013},
  publisher={John Wiley \& Sons}
}

@article{dou2018comparative,
  title={A comparative study of the Binary Logistic Regression (BLR) and Artificial Neural Network (ANN) models for GIS-based spatial predicting landslides at a regional scale},
  author={Dou, J and Yamagishi, H and Zhu, Z and Yunus, AP and Chen, CW},
  journal={Landslide Dynamics: ISDR-ICL Landslide Interactive Teaching Tools},
  volume={1},
  year={2018}
}

@article{genkin2007large,
  author       = {Alexander Genkin and
                  David D. Lewis and
                  David Madigan},
  title        = {Large-Scale Bayesian Logistic Regression for Text Categorization},
  journal      = {Technometrics},
  volume       = {49},
  number       = {3},
  pages        = {291--304},
  year         = {2007},
  url          = {https://doi.org/10.1198/004017007000000245},
  doi          = {10.1198/004017007000000245},
  bibsource    = {dblp computer science bibliography, https://dblp.org}
}

@inproceedings{sowmya2016large,
  title={Large scale multi-label text classification of a hierarchical dataset using Rocchio algorithm},
  author={Sowmya, BJ and Srinivasa, KG and others},
  booktitle={CSITSS},
  year={2016},
  organization={IEEE}
}

@inproceedings{selvi2017text,
  title={Text categorization using Rocchio algorithm and random forest algorithm},
  author={Selvi, S Thamarai and Karthikeyan, P and Vincent, A and Abinaya, V and Neeraja, G and Deepika, R},
  booktitle={ICoAC},
  year={2017},
  organization={IEEE}
}

@inproceedings{liu2008isolation,
  title={Isolation forest},
  author={Liu, Fei Tony and Ting, Kai Ming and Zhou, Zhi-Hua},
  booktitle={2008 eighth ieee international conference on data mining},
  year={2008},
  organization={IEEE}
}

@inproceedings{al2021isolation,
  title={Isolation forest based anomaly detection: A systematic literature review},
  author={Al Farizi, Wahid Salman and Hidayah, Indriana and Rizal, Muhammad Nur},
  booktitle={2021 8th International Conference on Information Technology, Computer and Electrical Engineering (ICITACEE)},
  pages={118--122},
  year={2021},
  organization={IEEE}
}

@article{kabinna2018examining,
  title={Examining the stability of logging statements},
  author={Kabinna, Suhas and Bezemer, Cor-Paul and Shang, Weiyi and Syer, Mark D and Hassan, Ahmed E},
  journal={Empirical Software Engineering},
  volume={23},
  pages={290--333},
  year={2018},
  publisher={Springer}
}

@article{cvach2012monitor,
  title={Monitor alarm fatigue: an integrative review},
  author={Cvach, Maria},
  journal={Biomedical instrumentation \& technology},
  volume={46},
  number={4},
  pages={268--277},
  year={2012}
}

@article{hays1982unix,
  title={Unix-based multiple-process system, for real-time data acquisition and control},
  author={Hays Jr, AV and Richmond, BJ and Optican, LM},
  year={1982},
  publisher={Electron Conventions, El Segundo, CA}
}

@article{khurana23natural,
  author       = {Diksha Khurana and
                  Aditya Koli and
                  Kiran Khatter and
                  Sukhdev Singh},
  title        = {Natural language processing: state of the art, current trends and
                  challenges},
  journal      = {Multim. Tools Appl.},
  volume       = {82},
  number       = {3},
  pages        = {3713--3744},
  year         = {2023},
  url          = {https://doi.org/10.1007/s11042-022-13428-4},
  doi          = {10.1007/S11042-022-13428-4},
  bibsource    = {dblp computer science bibliography, https://dblp.org}
}

@article{andenmatten2019aiops,
  title={AIOps-Artificial Intelligence for IT Operations: Todays Challenges of new Technologies and new Methodologies in IT Operations},
  author={Andenmatten, Martin},
  journal={HMD Praxis der Wirtschaftsinformatik},
  volume={56},
  pages={332--344},
  year={2019},
  publisher={Springer}
}

@incollection{beschastnikh2011mining,
  title={Mining temporal invariants from partially ordered logs},
  author={Beschastnikh, Ivan and Brun, Yuriy and Ernst, Michael D and Krishnamurthy, Arvind and Anderson, Thomas E},
  booktitle={Managing Large-scale Systems via the Analysis of System Logs and the Application of Machine Learning Techniques},
  pages={1--10},
  year={2011}
}

@article{jiang2011efficient,
  title={Efficient fault detection and diagnosis in complex software systems with informa-tion-theoretic monitoring},
  author={Jiang, Miao and Munawar, Mohammad A and Reidemeister, Thomas and Ward, Paul AS},
  journal={IEEE Transactions on Dependable and Secure Computing},
  volume={8},
  number={4},
  pages={510--522},
  year={2011},
  publisher={IEEE}
}

@inproceedings{shengqi2010mining,
  title={Mining program workflow from interleaved logs},
  author={Shengqi, J-GLQF and LI, YANG Jiang},
  booktitle={Proceedings of the 16th ACM SIGKDD international conference on Knowledge discovery and data mining (KDD'10)},
  year={2010}
}

@inproceedings{fu2009execution,
  title={Execution anomaly detection in distributed systems through unstructured log analysis},
  author={Fu, Qiang and Lou, Jian-Guang and Wang, Yi and Li, Jiang},
  booktitle={2009 ninth IEEE international conference on data mining},
  pages={149--158},
  year={2009},
  organization={IEEE}
}

@inproceedings{kreps2011kafka,
  title={Kafka: A distributed messaging system for log processing},
  author={Kreps, Jay and Narkhede, Neha and Rao, Jun and others},
  booktitle={Proceedings of the NetDB},
  volume={11},
  number={2011},
  pages={1--7},
  year={2011},
  organization={Athens, Greece}
}

@inproceedings{yen2013beehive,
  title={Beehive: Large-scale log analysis for detecting suspicious activity in enterprise networks},
  author={Yen, Ting-Fang and Oprea, Alina and Onarlioglu, Kaan and Leetham, Todd and Robertson, William and Juels, Ari and Kirda, Engin},
  booktitle={Proceedings of the 29th annual computer security applications conference},
  pages={199--208},
  year={2013}
}

@inproceedings{ko2008debugging,
  title={Debugging reinvented: asking and answering why and why not questions about program behavior},
  author={Ko, Amy J and Myers, Brad A},
  booktitle={Proceedings of the 30th international conference on Software engineering},
  pages={301--310},
  year={2008}
}

@inproceedings{fu2012logmaster,
  title={Logmaster: Mining event correlations in logs of large-scale cluster systems},
  author={Fu, Xiaoyu and Ren, Rui and Zhan, Jianfeng and Zhou, Wei and Jia, Zhen and Lu, Gang},
  booktitle={2012 IEEE 31st Symposium on Reliable Distributed Systems},
  pages={71--80},
  year={2012},
  organization={IEEE}
}

@article{santos2017analyzing,
  title={Analyzing the use of concept maps in computer science: A systematic mapping study},
  author={Santos, Vinicius dos and Souza, Erica F de and Felizardo, Katia R and Vijaykumar, Nandamudi L},
  journal={Informatics in Education},
  volume={16},
  number={2},
  pages={257--288},
  year={2017},
  publisher={Vilnius University Institute of Data Science and Digital Technologies}
}

@inproceedings{nedelkoski2020multi,
  title={Multi-source distributed system data for ai-powered analytics},
  author={Nedelkoski, Sasho and Bogatinovski, Jasmin and Mandapati, Ajay Kumar and Becker, Soeren and Cardoso, Jorge and Kao, Odej},
  booktitle={Service-Oriented and Cloud Computing: 8th IFIP WG 2.14 European Conference, ESOCC 2020, Heraklion, Crete, Greece, September 28--30, 2020, Proceedings 8},
  pages={161--176},
  year={2020},
  organization={Springer}
}

@inproceedings{bogatinovski2020multi,
  title={Multi-source anomaly detection in distributed {IT} systems},
  author={Bogatinovski, Jasmin and Nedelkoski, Sasho},
  booktitle={ICSOC},
  year={2020}
}

@article{johnson2019class_imbalance_survey,
    author = {Johnson, Justin and Khoshgoftaar, Taghi},
    year = {2019},
    month = {03},
    title = {Survey on deep learning with class imbalance},
    volume = {6},
    journal = {Journal of Big Data}
}

@inproceedings{sharp2016semi,
  title={Semi-Autonomous Labeling of Unstructured Maintenance Log Data for Diagnostic Root Cause Analysis},
  author={Sharp, Michael E and Sexton, Thurston B and Brundage, Michael P},
  year={2016},
  booktitle={International Conference Advances in Production Management Systems (APMS)}
}

@article{zio2009reliability,
  title={Reliability engineering: Old problems and new challenges},
  author={Zio, Enrico},
  journal={Reliability engineering \& system safety},
  volume={94},
  number={2},
  pages={125--141},
  year={2009},
  publisher={Elsevier}
}

@article{avizienis2004basic,
  title={Basic concepts and taxonomy of dependable and secure computing},
  author={Avizienis, Algirdas and Laprie, J-C and Randell, Brian and Landwehr, Carl},
  journal={IEEE transactions on dependable and secure computing},
  volume={1},
  number={1},
  pages={11--33},
  year={2004},
  publisher={IEEE}
}

@inproceedings{gray1986computers,
  title={Why do computers stop and what can be done about it?},
  author={Gray, Jim},
  booktitle={Symposium on reliability in distributed software and database systems},
  pages={3--12},
  year={1986},
  organization={Los Angeles, CA, USA}
}

@article{atzori2010internet,
  title={The internet of things: A survey},
  author={Atzori, Luigi and Iera, Antonio and Morabito, Giacomo},
  journal={Computer networks},
  volume={54},
  number={15},
  pages={2787--2805},
  year={2010},
  publisher={Elsevier}
}

@article{del2023big,
  title={Big Data Analytics in Healthcare: exploring the role of Machine Learning in Predicting patient outcomes and improving Healthcare Delivery},
  author={Del Giorgio Solfa, Federico and Simonato, Fernando Rogelio},
  journal={International Journal of Computations, Information and Manufacturing (IJCIM)},
  volume={3},
  year={2023}
}

@article{akter2022transforming,
  title={Transforming business using digital innovations: The application of AI, blockchain, cloud and data analytics},
  author={Akter, Shahriar and Michael, Katina and Uddin, Muhammad Rajib and McCarthy, Grace and Rahman, Mahfuzur},
  journal={Annals of Operations Research},
  pages={1--33},
  year={2022},
  publisher={Springer}
}

@article{kitsios2021digital,
  title={Digital transformation and strategy in the banking sector: Evaluating the acceptance rate of e-services},
  author={Kitsios, Fotis and Giatsidis, Ioannis and Kamariotou, Maria},
  journal={Journal of Open Innovation: Technology, Market, and Complexity},
  volume={7},
  number={3},
  pages={204},
  year={2021},
  publisher={MDPI}
}

@article{azzaakiyyah2023impact,
  title={The Impact of Social Media Use on Social Interaction in Contemporary Society},
  author={Azzaakiyyah, Hizbul Khootimah},
  journal={Technology and Society Perspectives (TACIT)},
  volume={1},
  number={1},
  pages={1--9},
  year={2023}
}

@article{zhang2010cloud,
  title={Cloud computing: state-of-the-art and research challenges},
  author={Zhang, Qi and Cheng, Lu and Boutaba, Raouf},
  journal={Journal of internet services and applications},
  volume={1},
  pages={7--18},
  year={2010},
  publisher={Springer}
}

@article{newkirk2008rapid,
  title={Rapid business and IT change: drivers for strategic information systems planning?},
  author={Newkirk, Henry E and Lederer, Albert L and Johnson, Alice M},
  journal={European Journal of Information Systems},
  volume={17},
  pages={198--218},
  year={2008},
  publisher={Springer}
}

@article{simon1977organization,
  title={The organization of complex systems},
  author={Simon, Herbert A and Simon, Herbert A},
  journal={Models of discovery: And other topics in the methods of science},
  pages={245--261},
  year={1977},
  publisher={Springer}
}

@inproceedings{dang2019aiops,
  title={Aiops: real-world challenges and research innovations},
  author={Dang, Yingnong and Lin, Qingwei and Huang, Peng},
  booktitle={2019 IEEE/ACM 41st International Conference on Software Engineering: Companion Proceedings (ICSE-Companion)},
  pages={4--5},
  year={2019},
  organization={IEEE}
}

@book{bass2015devops,
  title={DevOps: A software architect's perspective},
  author={Bass, Len and Weber, Ingo and Zhu, Liming},
  year={2015},
  publisher={Addison-Wesley Professional}
}

@inproceedings{senapathi2018devops,
  title={DevOps capabilities, practices, and challenges: Insights from a case study},
  author={Senapathi, Mali and Buchan, Jim and Osman, Hady},
  booktitle={Proceedings of the 22nd International Conference on Evaluation and Assessment in Software Engineering 2018},
  pages={57--67},
  year={2018}
}

@article{bogatinovski2021artificial,
  title={Artificial intelligence for it operations (aiops) workshop white paper},
  author={Bogatinovski, Jasmin and Nedelkoski, Sasho and Acker, Alexander and Schmidt, Florian and Wittkopp, Thors-ten and Becker, Soeren and Cardoso, Jorge and Kao, Odej},
  journal={arXiv preprint arXiv:2101.-06054},
  year={2021}
}

@inproceedings{shen2020evolving,
  title={Evolving from traditional systems to AIOps: design, implementation and measurements},
  author={Shen, Shijun and Zhang, Jiuling and Huang, Daochao and Xiao, Jun},
  booktitle={2020 IEEE International Conference on Advances in Electrical Engineering and Computer Applications (AEECA)},
  pages={276--280},
  year={2020},
  organization={IEEE}
}

@inproceedings{becker2020towards,
  title={Towards aiops in edge computing environments},
  author={Becker, Soeren and Schmidt, Florian and Gulenko, Anton and Acker, Alexander and Kao, Odej},
  booktitle={2020 IEEE International Conference on Big Data (Big Data)},
  pages={3470--3475},
  year={2020},
  organization={IEEE}
}

@article{zhaoxue2021survey,
  title={A survey on log research of aiops: Methods and trends},
  author={Zhaoxue, Jiang and Tong, Li and Zhenguo, Zhang and Jingguo, Ge and Junling, You and Liangxiong, Li},
  journal={Mobile Networks and Applications},
  volume={26},
  number={6},
  pages={2353--2364},
  year={2021},
  publisher={Springer}
}

@article{miranskyy2016operational,
  title={Operational-log analysis for big data systems: Challenges and solutions},
  author={Miranskyy, Andriy and Hamou-Lhadj, Abdelwahab and Cialini, Enzo and Larsson, Alf},
  journal={IEEE Software},
  volume={33},
  number={2},
  pages={52--59},
  year={2016},
  publisher={IEEE}
}

@inproceedings{sipos2014log,
  title={Log-based predictive maintenance},
  author={Sipos, Ruben and Fradkin, Dmitriy and Moerchen, Fabian and Wang, Zhuang},
  booktitle={Proceedings of the 20th ACM SIGKDD international conference on knowledge discovery and data mining},
  pages={1867--1876},
  year={2014}
}

@inproceedings{abbad2018toward,
  title={Toward an automated labeling of event log attributes},
  author={Abbad Andaloussi, Amine and Burattin, Andrea and Weber, Barbara},
  booktitle={Enterprise, Business-Process and Information Systems Modeling: 19th International Conference, BPMDS 2018, 23rd International Conference, EMMSAD 2018, Held at CAiSE 2018, Tallinn, Estonia, June 11-12, 2018, Proceedings 19},
  pages={82--96},
  year={2018},
  organization={Springer}
}

@article{samariya2023comprehensive,
  title={A comprehensive survey of anomaly detection algorithms},
  author={Samariya, Durgesh and Thakkar, Amit},
  journal={Annals of Data Science},
  volume={10},
  number={3},
  pages={829--850},
  year={2023},
  publisher={Springer}
}

@inproceedings{bogatinovski2023auto,
  title={Auto-Logging: AI-centred Logging Instrumentation},
  author={Bogatinovski, Jasmin and Kao, Odej},
  booktitle={2023 IEEE/ACM 45th International Conference on Software Engineering: New Ideas and Emerging Results (ICSE-NIER)},
  pages={95--100},
  year={2023},
  organization={IEEE}
}

@book{schmidt2012logging,
  title={Logging and log management: the authoritative guide to understanding the concepts surrounding logging and log management},
  author={Schmidt, Kevin and Phillips, Chris and Chuvakin, Anton},
  year={2012},
  publisher={Newnes}
}

@article{reuben2016automated,
  title={Automated log audits for privacy compliance validation: a literature survey},
  author={Reuben, Jenni and Martucci, Leonardo A and Fischer-H{\"u}bner, Simone},
  journal={Privacy and Identity Management. Time for a Revolution? 10th IFIP WG 9.2, 9.5, 9.6/11.7, 11.4, 11.6/SIG 9.2. 2 International Summer School, Edinburgh, UK, August 16-21, 2015, Revised Selected Papers 10},
  pages={312--326},
  year={2016},
  publisher={Springer}
}

@article{jose2016automatic,
  title={Automatic clustering using nature-inspired metaheuristics: A survey},
  author={Jose-Garcia, Adan and Gomez-Flores, Wilfrido},
  journal={Applied Soft Computing},
  volume={41},
  pages={192--213},
  year={2016},
  publisher={Elsevier}
}

@article{zhang1996birch,
  title={BIRCH: an efficient data clustering method for very large databases},
  author={Zhang, Tian and Ramakrishnan, Raghu and Livny, Miron},
  journal={ACM sigmod record},
  volume={25},
  number={2},
  pages={103--114},
  year={1996},
  publisher={ACM New York, NY, USA}
}

@article{Platt99,
author = {Platt, John},
year = {1999},
title = {Probabilistic Outputs for Support Vector Machines and Comparisons to Regularized Likelihood Methods},
volume = {10},
journal = {Advances in Large-Margin Classifiers}
}

@inproceedings{nn_calibration_2017,
author = {Guo, Chuan and Pleiss, Geoff and Sun, Yu and Weinberger, Kilian Q.},
title = {On Calibration of Modern Neural Networks},
year = {2017},
booktitle = {International Conference on Machine Learning (ICML)}
}

@inproceedings{lu2018detecting,
  title={Detecting anomaly in big data system logs using convolutional neural network},
  author={Lu, Siyang and Wei, Xiang and Li, Yandong and Wang, Liqiang},
  booktitle={2018 IEEE 16th Intl Conf on Dependable, Autonomic and Secure Computing, 16th Intl Conf on Pervasive Intelligence and Computing, 4th Intl Conf on Big Data Intelligence and Computing and Cyber Science and Technology Congress (DASC/PiCom/DataCom/CyberSciTech)},
  pages={151--158},
  year={2018},
  organization={IEEE}
}

@inproceedings{guo2023loglg,
  title={Loglg: Weakly supervised log anomaly detection via log-event graph construction},
  author={Guo, Hongcheng and Guo, Yuhui and Yang, Jian and Liu, Jiaheng and Li, Zhoujun and Zheng, Tieqiao and Zheng, Liangfan and Hou, Weichao and Zhang, Bo},
  booktitle={International Conference on Database Systems for Advanced Applications},
  pages={490--501},
  year={2023},
  organization={Springer}
}

@article{xie2023weakly,
  title={A weakly supervised anomaly detection method based on deep anomaly scoring network},
  author={Xie, Xin and Li, Zixi and Huang, Yuhui and Wu, Dengquan},
  journal={Signal, Image and Video Processing},
  volume={17},
  number={8},
  pages={3903--3911},
  year={2023},
  publisher={Springer}
}

@inproceedings{zhao2024weakly,
  title={Weakly Supervised Anomaly Detection via Knowledge-Data Alignment},
  author={Zhao, Haihong and Zi, Chenyi and Liu, Yang and Zhang, Chen and Zhou, Yan and Li, Jia},
  booktitle={Proceedings of the ACM on Web Conference 2024},
  pages={4083--4094},
  year={2024}
}

@inproceedings{pang2023deep,
  title={Deep weakly-supervised anomaly detection},
  author={Pang, Guansong and Shen, Chunhua and Jin, Huidong and van den Hengel, Anton},
  booktitle={Proceedings of the 29th ACM SIGKDD Conference on Knowledge Discovery and Data Mining},
  pages={1795--1807},
  year={2023}
}

@article{liu2023logbd,
  title={LogBD: A Log Anomaly Detection Method Based on Pretrained Models and Domain Adaptation},
  author={Liu, Shuxian and Deng, Le and Xu, Huan and Wang, Wei},
  journal={Applied Sciences},
  volume={13},
  number={13},
  pages={7739},
  year={2023},
  publisher={MDPI}
}

@article{minaee2021deep,
  title={Deep learning--based text classification: a comprehensive review},
  author={Minaee, Shervin and Kalchbrenner, Nal and Cambria, Erik and Nikzad, Narjes and Chen-aghlu, Meysam and Gao, Jianfeng},
  journal={ACM computing surveys (CSUR)},
  volume={54},
  number={3},
  pages={1--40},
  year={2021},
  publisher={ACM New York, NY, USA}
}

@article{chapelle2009semi,
  title={Semi-supervised learning},
  author={Chapelle, Olivier and Scholkopf, Bernhard and Zien, Alexander},
  journal={IEEE Transactions on Neural Networks},
  volume={20},
  number={3},
  pages={542--542},
  year={2009},
  publisher={IEEE}
}

@article{alexey2016discriminative,
  title={Discriminative unsupervised feature learning with exemplar convolutional neural networks},
  author={Alexey, Dosovitskiy and Fischer, Philipp and Tobias, Jost and Springenberg, Martin Riedmiller and Brox, Thomas},
  journal={IEEE TPAMI},
  volume={38},
  number={9},
  pages={1734--1747},
  year={2016}
}

@inproceedings{letouzey2000learning,
  title={Learning from positive and unlabeled examples},
  author={Letouzey, Fabien and Denis, Fran{\c{c}}ois and Gilleron, R{\'e}mi},
  booktitle={International Conference on Algorithmic Learning Theory},
  pages={71--85},
  year={2000},
  organization={Springer}
}

@inproceedings{dike2018unsupervised,
  title={Unsupervised learning based on artificial neural network: A review},
  author={Dike, Happiness Ugochi and Zhou, Yimin and Deveerasetty, Kranthi Kumar and Wu, Qingtian},
  booktitle={2018 IEEE International Conference on Cyborg and Bionic Systems (CBS)},
  pages={322--327},
  year={2018},
  organization={IEEE}
}

@article{sinaga2020unsupervised,
  title={Unsupervised K-means clustering algorithm},
  author={Sinaga, Kristina P and Yang, Miin-Shen},
  journal={IEEE access},
  volume={8},
  pages={80716--80727},
  year={2020},
  publisher={IEEE}
}

@inproceedings{arora2018analysis,
  title={An analysis of the t-sne algorithm for data visualization},
  author={Arora, Sanjeev and Hu, Wei and Kothari, Pravesh K},
  booktitle={Conference on learning theory},
  pages={1455--1462},
  year={2018},
  organization={PMLR}
}

@article{carbonneau2018multiple,
  title={Multiple instance learning: A survey of problem characteristics and applications},
  author={Carbonneau, Marc-Andr{\'e} and Cheplygina, Veronika and Granger, Eric and Gag-non, Ghyslain},
  journal={Pattern Recognition},
  volume={77},
  pages={329--353},
  year={2018},
  publisher={Elsevier}
}

@article{liu2021self,
  title={Self-supervised learning: Generative or contrastive},
  author={Liu, Xiao and Zhang, Fanjin and Hou, Zhenyu and Mian, Li and Wang, Zhaoyu and Zhang, Jing and Tang, Jie},
  journal={IEEE transactions on knowledge and data engineering},
  volume={35},
  number={1},
  pages={857--876},
  year={2021},
  publisher={IEEE}
}

@article{ren2020not,
  title={Not all unlabeled data are equal: Learning to weight data in semi-supervised learning},
  author={Ren, Zhongzheng and Yeh, Raymond and Schwing, Alexander},
  journal={Advances in Neural Information Processing Systems},
  volume={33},
  pages={21786--21797},
  year={2020}
}

@book{mohri2018foundations,
  title={Foundations of machine learning},
  author={Mohri, Mehryar and Rostamizadeh, Afshin and Talwalkar, Ameet},
  year={2018},
  publisher={MIT press}
}

@book{bishop2006pattern,
  title={Pattern recognition and machine learning},
  author={Bishop, Christopher M and Nasrabadi, Nasser M},
  volume={4},
  number={4},
  year={2006},
  publisher={Springer}
}

@inproceedings{brown2018recurrent,
  title={Recurrent neural network attention mechanisms for interpretable system log anomaly detection},
  author={Brown, Andy and Tuor, Aaron and Hutchinson, Brian and Nichols, Nicole},
  booktitle={Proceedings of the first workshop on machine learning for computing systems},
  pages={1--8},
  year={2018}
}

@article{chalapathy2019deep,
  title={Deep learning for anomaly detection: A survey},
  author={Chalapathy, Raghavendra and Chawla, Sanjay},
  journal={arXiv preprint arXiv:1901.03407},
  year={2019}
}

@book{hastie2009elements,
  title={The elements of statistical learning: data mining, inference, and prediction},
  author={Hastie, Trevor and Tibshirani, Robert and Friedman, Jerome H and Friedman, Jerome H},
  volume={2},
  year={2009},
  publisher={Springer}
}

@article{sabharwal2022hands,
  title={Hands-on AIOps},
  author={Sabharwal, Navin and Bhardwaj, G},
  journal={Apress eBooks, Springer},
  year={2022},
  publisher={Springer}
}

@article{rijal2022aiops,
  title={Aiops: A multivocal literature review},
  author={Rijal, Laxmi and Colomo-Palacios, Ricardo and S{\'a}nchez-Gord{\'o}n, Mary},
  journal={Artificial Intelligence for Cloud and Edge Computing},
  pages={31--50},
  year={2022},
  publisher={Springer}
}

@article{natella2016assessing,
  title={Assessing dependability with software fault injection: A survey},
  author={Natella, Roberto and Cotroneo, Domenico and Madeira, Henrique S},
  journal={ACM Computing Surveys (CSUR)},
  volume={48},
  number={3},
  pages={1--55},
  year={2016},
  publisher={ACM New York, NY, USA}
}

@inproceedings{bermbach2021future,
  title={On the future of cloud engineering},
  author={Bermbach, David and Chandra, Abhishek and Krintz, Chandra and Gokhale, Aniruddha and Slominski, Aleksander and Thamsen, Lauritz and Cavalcante, Everton and Guo, Tian and Brandic, Ivona and Wolski, Rich},
  booktitle={2021 IEEE International conference on cloud engineering (IC2E)},
  pages={264--275},
  year={2021},
  organization={IEEE}
}

@article{he2021survey,
  title={A survey on automated log analysis for reliability engineering},
  author={He, Shilin and He, Pinjia and Chen, Zhuangbin and Yang, Tianyi and Su, Yuxin and Lyu, Michael R},
  journal={ACM computing surveys (CSUR)},
  volume={54},
  number={6},
  pages={1--37},
  year={2021},
  publisher={ACM New York, NY, USA}
}

@inproceedings{nishu2022anomaly,
author="Bansal, Nishu and Pahuja, Swimpy",
editor="Tomar, Anuradha and Malik, Hasmat and Kumar, Pramod and Iqbal, Atif",
title="A Generic Review on Anomaly Detection",
booktitle="Proceedings of 3rd International Conference on Machine Learning, Advances in Computing, Renewable Energy and Communication",
year="2022",
publisher="Springer Nature Singapore",
address="Singapore",
pages="495--506"
}

@article{al2021review,
  title={A review of machine learning and deep learning techniques for anomaly detection in IoT data},
  author={Al-amri, Redhwan and Murugesan, Raja Kumar and Man, Mustafa and Abdulateef, Alaa Fareed and Al-Sharafi, Mohammed A and Alkahtani, Ammar Ahmed},
  journal={Applied Sciences},
  volume={11},
  number={12},
  pages={5320},
  year={2021},
  publisher={MDPI}
}

@article{diaz2023joint,
  title={A joint study of the challenges, opportunities, and roadmap of mlops and aiops: A systematic survey},
  author={Diaz-De-Arcaya, Josu and Torre-Bastida, Ana I and Z{\'a}rate, Gorka and Mi{\~n}{\'o}n, Ra{\'u}l and Almeida, Aitor},
  journal={ACM Computing Surveys},
  volume={56},
  number={4},
  pages={1--30},
  year={2023},
  publisher={ACM New York, NY, USA}
}

@article{franke2011optimal,
  title={Optimal IT service availability: Shorter outages, or fewer?},
  author={Franke, Ulrik},
  journal={IEEE Transactions on Network and Service Management},
  volume={9},
  number={1},
  pages={22--33},
  year={2011},
  publisher={IEEE}
}

@inproceedings{zhang2021cloudrca,
  title={CloudRCA: A root cause analysis framework for cloud computing platforms},
  author={Zhang, Yingying and Guan, Zhengxiong and Qian, Huajie and Xu, Leili and Liu, Hengbo and Wen, Qingsong and Sun, Liang and Jiang, Junwei and Fan, Lunting and Ke, Min},
  booktitle={Proceedings of the 30th ACM International Conference on Information \& Knowledge Management},
  pages={4373--4382},
  year={2021}
}

@inproceedings{larsson2007real,
  title={Real-time root cause analysis for complex technical systems},
  author={Larsson, Jan Eric and DeBor, Joseph},
  booktitle={2007 IEEE 8th Human Factors and Power Plants and HPRCT 13th Annual Meeting},
  pages={156--163},
  year={2007},
  organization={IEEE}
}

@inproceedings{wu2020microrca,
  title={Microrca: Root cause localization of performance issues in microservices},
  author={Wu, Li and Tordsson, Johan and Elmroth, Erik and Kao, Odej},
  booktitle={NOMS 2020-2020 IEEE/IFIP Network Operations and Management Symposium},
  pages={1--9},
  year={2020},
  organization={IEEE}
}

@article{a2018fast,
  title={Fast, consistent tokenization of natural language text},
  author={A. Mullen, Lincoln and Benoit, Kenneth and Keyes, Os and Selivanov, Dmitry and Arnold, Jeffrey},
  journal={Journal of Open Source Software},
  volume={3},
  number={23},
  pages={655},
  year={2018},
  publisher={The Open Journal}
}


\end{document}